\documentclass[10pt,journal,onecolumn,commsoc]{IEEEtran}
\IEEEoverridecommandlockouts

\ifCLASSOPTIONcompsoc
  \usepackage[nocompress]{cite}
\else
  \usepackage{cite}
\fi

\usepackage{xcolor}
\usepackage{amssymb}
\usepackage{amsmath}
\allowdisplaybreaks[4]
\usepackage{multicol}
\usepackage{amsfonts}
\usepackage{amsthm}
\usepackage{array}
\usepackage{bbm}
\usepackage[caption=false,font=normalsize,labelfont=sf,textfont=sf]{subfig}
\usepackage{textcomp}
\usepackage{stfloats}
\usepackage{url}
\usepackage{verbatim}
\usepackage{graphicx}
\usepackage{multirow}
\usepackage{textcase}
\usepackage{hyperref} 
\usepackage[lined,linesnumbered,ruled]{algorithm2e}
\usepackage[bordercolor=white, linecolor=orange, backgroundcolor=orange, shadow]{todonotes}

\usepackage{tikz}
\newcommand*\circled[1]{\tikz[baseline=(char.base)]{
            \node[shape=circle,draw,inner sep=1pt] (char) {#1};}}

\newtheorem{lemma}{Lemma}
\newtheorem{fact}{Fact}

\newtheorem{theorem}{Theorem}

\newtheorem{definition}{Definition}
\newtheorem{assumption}{Assumption}
\newtheorem{corollary}{Corollary}

\def\BibTeX{{\rm B\kern-.05em{\sc i\kern-.025em b}\kern-.08em
    T\kern-.1667em\lower.7ex\hbox{E}\kern-.125emX}}

\begin{document}

\title{ROSS: RObust decentralized Stochastic learning based on Shapley values}

\author{Lina~Wang, 
        Yunsheng~Yuan,
        Feng~Li,
        Lingjie~Duan
        
\IEEEcompsocitemizethanks{\IEEEcompsocthanksitem L. Wang, Y. Yuan and F. Li are with School of Computer Science and Technology, Shandong University, Qingdao 266237, China. 
E-mail: linawang425@mail.sdu.edu.cn, yuanys1028@gmail.com, fli@sdu.edu.cn
%
%
%
\IEEEcompsocthanksitem L. Duan is with both Internet of Things Thrust and Artificial Intelligence Thrust at Hong Kong University of Science and Technology (Guangzhou), Guangzhou 511453, China. 
E-mail: lingjieduan@hkust-gz.edu.cn
}
%
}



\IEEEtitleabstractindextext{
\begin{abstract}
  In the paradigm of decentralized learning, a group of agents collaborate to learn a global model using a distributed dataset without a central server; nevertheless, it is severely challenged by the heterogeneity of the data distribution across the agents. For example, the data may be distributed non-independently and identically, and even be noised or poisoned. To address these data challenges, we propose ROSS, a novel robust decentralized stochastic learning algorithm based on Shapley values, in this paper. Specifically, in each round, each agent leverages its local gradient and the cross-gradients from its neighbors (i.e., the derivative of its local loss function and the derivatives of its neighbors' local loss functions evaluated at its local model) to update its local model in a momentum-like manner, while we innovate in weighting the derivatives according to their contributions measured by Shapley values. We perform solid theoretical analysis to reveal the linear convergence speedup of our ROSS algorithm. We also verify the efficacy of our algorithm through extensive experiments on public datasets. Our results demonstrate that, in face of the above variety of data challenges, our ROSS algorithm has significant advantages over existing state-of-the-art proposals in terms of both convergence and prediction accuracy.
\end{abstract}

\begin{IEEEkeywords}
  Decentralized learning, Shapley values, data heterogeneity.
\end{IEEEkeywords}}

\maketitle

\IEEEdisplaynontitleabstractindextext
\IEEEpeerreviewmaketitle

\section{Introduction} \label{sec:introduction}
  The goal of distributed machine learning is to learn from data distributed across multiple agents~\cite{McMahanMRHA-AISTATS17,LianZZHZL-NIPS17,VerbraekenWKK-CSUR21,KairouzMABBN-FTML21}. The distributed learning algorithms can be classified into two categories according to communication topology: \textit{centralized} learning and \textit{decentralized} learning. In centralized learning, e.g., \textit{Federated Learning} (FL)~\cite{McMahanMRHA-AISTATS17,GaoTW-IEEENet23}, data are distributed across a group of agents, and a global model is learned by aggregating the updates computed locally at the agents through a central server. Nevertheless, in such a centralized architecture, all of the agents are coordinated by the central server; hence, the usability of the FL framework may suffer severe limitations, e.g., when the central server suffers cyber-attacks or it has to persistently communicate with the agents and thus becomes a significant bottleneck. To address these concerns, the paradigm of decentralized learning is proposed. There is no central server entailed in decentralized learning, and each agent interacts only with its neighbors.

  Unfortunately, decentralized learning is severely challenged by the complicated data distribution across the agents, especially considering there is no central infrastructure for global coordination. For example, the data of different agents naturally tend to be non-independently and identically distributed (non-IID)~\cite{LiDCB-ICDE2022}, such that the data of the agents may not align with the global objective of the decentralized learning. Furthermore, the data at the agents are usually noisy due to the measurement errors~\cite{GhoshKS-AAAI17,TuorWKLL-ICPR20}. Additionally, some of the agents may be corrupted such that their data are poisoned~\cite{LiHBS-ICML21}. Such noises or errors also diminish the data relevance at the agents and aggravate the heterogeneity of the data distribution, potentially inducing the failure of the global model training.

  Although these data challenges have been extensively investigated for centralized distributed learning (e.g., FL~\cite{McMahanMRHA-AISTATS17,TangYLZL-ICML19,WangKNL-INFOCOM20,ZhaoWHQDP-TWC24,LuPDSZ-IOTJ24}), there are only quite few studies for the decentralized counterpart. Recently, \cite{ VogelsHKKLSJ-NIPS21,VosFGKPS-NIPS23} propose to alleviate the impact of non-IID data through improving communication topology towards uniform data sampling. \cite{LinKSJ-ICML21} proposes a quasi-global momentum buffer to effectively learn from heterogeneous data distribution. \cite{PuA-MP21,ZhangFLYLZ-MobiHoc22,TakezawaBNSY-TMLR23,AketiHR-NIPS24} use gradient or momentum tracking mechanisms to address the problem of data heterogeneity. \cite{EsfandiariTJBHHS-ICML21, AketiKR-TMLR23} adopt cross-gradients to mitigate data heterogeneity, by enabling each agent to leverage its neighbors’ data to update its local model. Nevertheless, they cannot maintain effectiveness (e.g., in terms of convergence rate and prediction accuracy), as they cannot elaborately evaluate the influence of each agent, especially in face of a variety of data challenges.

  To address the above issues, we propose ROSS, a novel robust decentralized stochastic learning algorithm based on Shapley values. In each round of our ROSS algorithm, each agent leverages its local gradient and cross-gradients (i.e., the derivative of its local loss function and the ones of its neighbors' local loss functions evaluated at its model) to update its local model in a momentum-like manner. Specifically, to properly weight the gradients in the aggregation, each agent adopts Shapley values to measure the different contributions of its neighbors (i.e., the different potentials of the gradients to appropriately update its local model towards the global training objective). In summary, our main contributions are as follows:
  \begin{itemize}
    \item We study a very fundamental problem in decentralized learning: how to improve the robustness to a wide spectrum of data challenges, e.g., non-IID data and data poisoning attacks.
    \item By using the notion of Shapley value, we propose a robust decentralized stochastic learning algorithm. Specifically, each agent adopts Shapley values to elaborately weight its local gradient and cross-gradients and thus to properly evaluate the contributions of itself and its neighbors. 
    \item Through our solid theoretical analysis, we reveal the convergence of our ROSS algorithm. We also perform extensive experiments to verify the efficacy of our ROSS algorithm in a range of vulnerable scenarios on different real datasets.
  \end{itemize}

  The remainder of this paper is organized as follows. Related work is presented in Sec.~\ref{sec:relwork}. Then, we introduce some key preliminaries in Sec.~\ref{sec:pre}. In Sec.~\ref{sec:ross}, we present the details of our ROSS algorithm. Furthermore, we theoretically show the convergence rate of ROSS in Sec.~\ref{sec:analysis}. We also perform extensive experiments to verify the efficacy of ROSS algorithm in Sec.~\ref{sec:exp}. Finally, we conclude this paper and give an outlook on future research directions in Sec.~\ref{sec:con}.

\section{Related Work} \label{sec:relwork}
  In the context of distributed machine learning, decentralized learning algorithms have been attracting increasing attentions. \cite{LianZZHZL-NIPS17} proposes \textit{Decentralized Parallel Stochastic Gradient Descent} (D-PSGD) by combining the gossip-averaging \cite{BoydGPS-TIT06} with SGD. In \cite{JiangBHS-NIPS17}, a new consensus-based distributed SGD algorithm is proposed, enabling both data parallelization as well as decentralized computation. In \cite{AssranLBR-ICML19}, a \textit{Stochastic Gradient Push} (SGP) algorithm is proposed to extend D-PSGD to generic communication topologies that may be directed (asymmetric), sparse, and time-varying. In \cite{NadiradzeSASMA-arXiv19}, random interactions between agents are utilized to achieve consensus. \cite{KoloskovaLBJS-ICML2020} presents a unified framework for analyzing decentralized SGD algorithms. Moreover, \cite{BaluJTHLS-ICASSP21} proposes \textit{Decentralized Momentum Stochastic Gradient Descent} algorithm which introduces a momentum term to D-PSGD. \cite{YingYCHPY-NIPS21} reveals that applying decentralized SGD algorithms over exponential graphs has both fast communication and effective averaging simultaneously. \cite{YuanCHZPXY-ICCV21} finds that the momentum term can amplify the inconsistency bias in decentralized momentum SGD, and such bias grows large and thus results in considerable performance degradation, especially when batch-size grows. Therefore, \cite{YuanCHZPXY-ICCV21} proposes a decentralized large-batch momentum SGD to remove the momentum-incurred bias.

  The primary assumption of the above decentralized learning algorithms is that the data is independently and identically distributed across the agents. However, the assumption may not always hold in practice, and the robustness of the learning process thus cannot be ensured due to the ``abnormal'' data distribution~\cite{HsiehPMG-ICML20}. Although the problem of data heterogeneity has been extensively studied in centralized distributed machine learning (e.g., FL)~\cite{McMahanMRHA-AISTATS17,TangYLZL-ICML19,WangKNL-INFOCOM20,ZhaoWHQDP-TWC24,LuPDSZ-IOTJ24,YinCKB-ICML18}, there are only a handful of proposals investigating decentralized learning algorithms on non-IID data. In \cite{TangLYZL-ICML18}, the D-PSGD algorithm is improved with variance reduction technique to tackle large data variance across agents. Specifically, each agent linearly combines the stochastic gradient and its local model in last iteration with the current ones. It is found in \cite{LinKSJ-ICML21} that heterogeneous data distribution hinders the local momentum acceleration in decentralized SGD algorithm. Hence, \cite{LinKSJ-ICML21} proposes to integrate quasi-global momentum with local stochastic gradients to alleviate the drift in the local optimization caused by the heterogeneous data distribution. \cite{VogelsHKKLSJ-NIPS21} proposes RelaySUM, an alternative mechanism to gossip averaging, in order to handle the difference between the agents' local data distribution. RelaySUM uses spanning trees to distribute information exactly uniformly across all agents. In \cite{El-MhamdiFGGHR-NIPS21}, the agents exchange both local model updates and local models with their neighbor, and adopt Trimmean aggregation rule~\cite{YinCKB-ICML18} to combine the received local model updates and local models. In each round, each agent aggregates local model updates received from its neighbors multiple times, and then exchanges local models with its neighbors once. \cite{FangZHKLLLG-CCS24} proposes a novel similarity-based aggregation mechanism to effectively enhance the robustness of decentralized learning, particularly against adversarial model poisoning attacks. In \cite{VosFGKPS-NIPS23}, a new decentralized learning algorithm, \textit{Epidemic Learning} (EL), is proposed. It adopts a dynamically changing, randomized communication topology. Additionally, tracking mechanisms, e.g., gradient tracking~\cite{PuA-MP21} and momentum tracking~\cite{TakezawaBNSY-TMLR23} have been proposed to address the problem of heterogeneous data in decentralized learning. \cite{AketiHR-NIPS24} proposes to let each agent store a copy of its neighbors' model parameters and then track the model updates instead of gradients. \cite{ZhangFLYLZ-MobiHoc22} enhances the local update scheme by incorporating a recursive gradient correction technique to address the problem of data heterogeneity. Moreover, \cite{EsfandiariTJBHHS-ICML21, AketiKR-TMLR23} leverage the notion of cross-gradients to mitigate the data heterogeneity in decentralized learning algorithms. In particular, in \cite{EsfandiariTJBHHS-ICML21}, each agent collects the cross-gradient information from its neighbors, i.e., the derivatives of its neighbors' local loss functions evaluated at its local model, and then updates its model using the projected gradients based on quadratic programming. In \cite{AketiKR-TMLR23}, each agent aggregates more cross-gradient information. Except for the derivatives of its neighbors' local loss functions evaluated at its model, the gradients computed locally based on the received neighbors' models are taken into account in gradient aggregation. Although our algorithm also utilizes the cross-gradient information, each agent adopts  Shapley values to weight the cross-gradient information such that the contributions (or data relevance) of its neighbors can be characterized elaborately.

  Shapley value is a widely used concept in game theory. It is used to measure each player's marginal contribution in a cooperative game. Recently, to address the heterogeneity of data distribution, many proposals adopt Shapley values to measure the different contributions of agents (or clients) in FL~\cite{FanFZPFLZ-ICDE22, LiuCYLC-TIST22, SunLZXLLQR-KDD23}. In particular, the Shapley value of each agent is calculated by the central server based on the received local models. According to the Shapley values, the local models can be properly aggregated in the central server. Unfortunately, there is no central server for global orchestration in decentralized learning, and each agent has to estimate Shapley value independently; hence, a consensus on measuring the contributions of the agents cannot be achieved. Therefore, it is highly non-trivial to leverage the notion of Shapley value to design robust decentralized learning algorithms.

\section{Preliminaries} \label{sec:pre}
 We first give the basics of decentralized learning in Sec.~\ref{ssec:dsl}, and then introduce the notion of Shapley value in Sec.~\ref{ssec:shapley}.
  \subsection{Decentralized Stochastic Learning} \label{ssec:dsl}
    We consider the following stochastic learning problem
    \begin{align}  \label{eq:stochprob}
      \min_{x \in \mathbb{R}^d} & \mathcal{F}(x) = \mathbb{E}_{\xi \sim \mathcal{D}} F(x; \xi)
    \end{align}
    where $\mathcal{D}$ is a predefined probability distribution, $x \in \mathbb{R}^d$ represents model parameter, $\xi$ denotes a random data sample from the probability distribution $\mathcal{D}$, and $F(x; \xi)$ represents a real-valued loss function on data sample $\xi$ under the model parameterized by $x$. In general, this formulation summarizes many popular machine learning models including deep learning, linear regression, and logistic regression.

    Let $\mathcal{G} = \{ \mathcal{N}, \mathbf{W} \}$ denote a decentralized network, where $\mathcal{N} = \{1,\cdots,N\}$ represents a group of $N$ agents and $\mathbf{W} \in [0,1]^{N \times N}$ denotes a weighted adjacent matrix. Without loss of generality, we assume that $\mathbf{W}$ is a symmetric doubly stochastic matrix such that i) $\omega_{i,i'} \in [0,1]$ and $\omega_{i,i'} = \omega_{i',i}$ for $\forall i, i' \in \mathcal{N}$, and ii) $\sum_{i'\in\mathcal{N}} \omega_{i,i'} = 1$ for $\forall i \in \mathcal{N}$. Note that, for any agents $i, i' \in \mathcal{N}$, we have $\omega_{i,i'} = 0$ if and only if there is no edge between $i$ and $i'$ such that they cannot communicate with each other. Let $\mathcal{N}_i \subseteq \mathcal{N}$ be the set of neighbors of agent $i \in \mathcal{N}$, i.e., $\mathcal{N}_i = \{ i'\in\mathcal{N} \mid \omega_{i,i'} > 0 \}$. We also suppose that each agent $i$ is aware of $\omega_{i,i'}$ for $\forall i' \in \mathcal{N}_i$.

    Let $\mathcal{D}_i$ denote the local data distribution of agent $i \in \mathcal{N}$. Assume $\mathcal{D}_i$ is private for agent $i$ and will not be shared with others. To design distributed algorithms on a decentralized network, the problem defined in (\ref{eq:stochprob}) can be re-written as
    \begin{align}  \label{eq:decentstochprob}
      \min_{x \in \mathbb{R}^d} & \mathcal{F}(x) := \frac{1}{N} \sum^N_{i=1} f_i(x)
    \end{align}
    where
    \begin{equation} \label{eq:localloss}
      f_i (x) = \mathbb{E}_{\xi_i \sim \mathcal{D}_i} \left[ F_i (x; \xi_i) \right]
    \end{equation}
    denotes the expectation of the local loss function $F_i$ of agent $i$ on its local data distribution $\mathcal{D}_i$. Without loss of generality, $F_i (\cdot ; \cdot) = F(\cdot; \cdot)$ for $\forall i \in \mathcal{N}$.

    Standard decentralized learning aims to train a shared global model by minimizing the expected loss over each agent’s local data distribution. This requires that all agents’ datasets be relevant to the same underlying task. However, as discussed in Sec.~\ref{sec:introduction}, data are often not identically distributed across agents. In such heterogeneous settings, agents may contribute very differently to the global objective. In particular, each agent updates its local model using its neighbors’ updates, and when data distributions vary widely, these updates can become biased or misaligned. To address the above issue, inspired by prior work~\cite{WangRZJS-FL20,NagalapattiN-AAAI21,LiPH-KDD22,SunLZXLLQR-KDD23}, we assume each agent holds a small validation dataset $\mathcal Q$ sampled from the combined global distribution $\mathcal{D}$ to fairly evaluate neighbors’ contributions.  As will be shown in Sec.~\ref{sec:exp}, only a small fraction of data samples is sufficient for the validation dataset to enable our algorithm to achieve satisfactory performance. Therefore, the overhead of reserving such a small portion of data is negligible in practice.

  \subsection{Cooperative Game and Shapley Value} \label{ssec:shapley}
    In this section, we review the basics of cooperative game and Shapley value, which will be helpful to our algorithm design and analysis later.
    \begin{definition}[Cooperative Game]
      Let $\mathcal{Z}=\{1,\cdots,Z\}$ be a finite set of $Z$ players. A cooperative game is represented by a tuple $(\mathcal{Z}, v)$, where $v: 2^{\mathcal{Z}} \rightarrow \mathbb{R}$ is a characteristic function of the game, assigning a real number to each coalition $\mathcal{Z}' \subseteq \mathcal{Z}$ and satisfying $v(\emptyset)=0$.
    \end{definition}
    The characteristic function $v(\mathcal{Z}')$ denotes the payoff obtained by  $\mathcal{Z}' \subseteq \mathcal{Z}$. The distribution of the surplus of the players actually plays the essential role for cooperative game. As demonstrated in \cite{Shapley-CTG53}, the distribution of the payoff over the players should satisfy the following requirements, i.e., balance, symmetry, additivity and zero element. Particularly,
    \begin{itemize}
    \item {Balance:} The payoffs to all players must add up to $v(\mathcal{Z})$.
    \item {Symmetry:} Two players are considered substitutes if they contribute equally to each coalition, and the solution should treat them the same.
    \item {Additivity:} The solution to the sum of two cooperative games should equal the sum of the individual solution awarded to each of the two games.
    \item {Zero element:} A player should not be rewarded if the player contributes nothing to each coalition.
    \end{itemize}

    To develop a payoff allocation solution that meets the aforementioned requirements, \cite{shapley1971cores} allocated the payoff to each player based on its marginal contributions to all possible coalitions this player could participate in. Specifically, the Shapley value of a player is a weighted sum of the player's marginal contributions across all possible coalitions in the game. We formally define the Shapley value as follows.
    \begin{definition}[Shapley Value] \label{def:shapley}
      Let ${\mathcal{S}} \in \Omega(\mathcal{Z})$ be a permutation of players in $\mathcal{Z}$ and $\mathcal{Z}_i ({\mathcal{S}})$ be a coalition consisting of all predecessors of player $i$ in ${\mathcal{S}}$. Assuming ${\mathcal{S}}(i)$ denotes the location of $i$ in $\mathcal{S}$, $\mathcal{Z}_i ({\mathcal{S}})$ can be represented as
      \begin{equation} \label{eq:prede}
        \mathcal{Z}_i ({\mathcal{S}}) = \left\{ i' \in {\mathcal{S}} \mid {\mathcal{S}}(i') < {\mathcal{S}}(i) \right\}
      \end{equation}
      Given a cooperative game $(\mathcal{Z}, v)$, the Shapley value for player $i\in\mathcal{Z}$ can be defined by 
      \begin{equation} \label{eq:shapley-1}
        \varphi_i(v) =  \frac{1}{Z !} \sum_{\mathcal{S} \in \Omega(\mathcal{Z})} \left( v \left( \mathcal{Z}_i (\mathcal{S}) \bigcup \{i\} \right) - v \left( \mathcal{Z}_i (\mathcal{S}) \right) \right)
      \end{equation}
    \end{definition}
    In another word, the payoff allocated to $i$ in a coalition game is the average marginal contribution of $i$ to ${\mathcal{Z}}_i (\mathcal{S})$ over all $\mathcal{S} \in \Omega(\mathcal{Z})$. Therefore, (\ref{eq:shapley-1}) can be re-written as:
    \begin{equation} \label{eq:shapley-2}
      \varphi(i) = \sum_{\mathcal{Z}' \subseteq \mathcal{Z} \setminus \{i\}} \left(Z \dbinom{Z-1}{Z'} \right)^{-1}   \left(v \left( \mathcal{Z}' \bigcup \{i\} \right) - v \left( \mathcal{Z}' \right)\right) 
    \end{equation}
    where $Z = |\mathcal{Z}|$ and $Z' = |\mathcal{Z}'|$.

    In decentralized learning, agents aggregate gradient information from their neighbors (e.g., the cross-gradients in our case), and their contributions must be weighted fairly. Shapley values, derived from cooperative game theory, provide a principled way to do this by averaging marginal gains across all possible coalitions. This coalition-aware property is crucial, since an agent’s update can indirectly influence multiple others and complementary data distributions may yield joint contributions larger than individual ones. Although there is another potential method, i.e., influence functions, which has been used to evaluate contributions by measuring the local effect of individual data points in FL~\cite{LiZWHL-TPDS22}, they cannot capture the coalition effects among the agents. Hence, Shapley values are better suited for ROSS, where fair weighting of neighbor contributions across all coalitions is essential.

\section{Our ROSS Algorithm} \label{sec:ross}
  The basic idea of our algorithm is to let each agent leverage the notion of Shapley value to measure the contribution of its peers to the global objective such that the gradients are aggregated appropriately. Specifically, in each round $t$, each agent $i \in \mathcal{N}$ first calculates its local stochastic gradient $g^{[t]}_{i,i}$ according to its local dataset $\mathcal{D}_i$ and its current local model $x^{[t-1]}_i$, and then sends its local model $x^{[t-1]}_i$ to its neighbors. Once receiving the local model $x^{[t-1]}_j$ from each of its neighbors, agent $i$ calculates stochastic cross-gradient $g^{[t]}_{i,j}$ according to its local dataset and the local model $x^{[t-1]}_j$, and then sends $g^{[t]}_{i,j}$ back to agent $j$. Agent $i$ then computes its local model update $x^{[t]}_{i,j}$ according to $g^{[t]}_{j,i}$ for $\forall j \in \mathcal{N}_i$. Based on the updates, agent $i$ calculates Shapley values $\{\varphi^{[t]}_{i,j}\}_{j \in \mathcal{N}_i}$ to measure the data relevance of the neighbors (or the contributions of the neighbors to the global training objective). The Shapley values are then utilized as weight parameters for the aggregation of the received stochastic gradients, and finally the local model $x^{[t]}_i$ is updated in a momentum-like manner.
  \begin{algorithm}[htb!]
    \KwIn{Initial point $x^{[0]}_i$, learning rate $\gamma$, momentum buffer $u^{[0]}_i = 0$, momentum coefficient $\alpha$, doubly stochastic matrix $\mathbf{W}$, the number of iterations $T$.}
    \KwOut{Local model $x^{[T]}_i$ of agent $i$}
    \ForEach{$t=1,2,\cdots,T$}{
      Compute local stochastic gradient $g^{[t]}_{i,i}$ according to (\ref{eq:selfgrad}); \label{ln:selfgrad} \\
      Send $x^{[t-1]}_i$ to each neighbor $j \in \mathcal{N}_i \setminus \{i\}$; \label{ln:sendmodel01}\\
      Receive $x^{[t-1]}_j$ from each $j \in \mathcal{N}_i \setminus \{i\}$; \label{ln:recemodel01}\\
      \ForEach{$j \in \mathcal{N}_i \setminus \{i\}$}{
        Compute stochastic gradient $g^{[t]}_{i,j}$ according to (\ref{eq:crossgrad}); \label{ln:crossgrad} \\
        Send $g^{[t]}_{i,j}$ back to agent $j$; \label{ln:sendcrossgrad}\\
      }\label{ln:sendcrossgradend}
      Receive $g^{[t]}_{j,i}$ from each $j \in \mathcal{N}_i \setminus \{i\}$; \label{ln:rececrossgrad}\\
      Compute $x^{[t]}_{i,j} = x^{[t-1]}_{i} - \gamma \cdot g^{[t]}_{j,i}$ for $\forall j \in \mathcal{N}_i$ according to (\ref{eq:up4neighbor}) and (\ref{eq:up4self}); \label{ln:model4shapley}\\
      Compute Shapley value $\varphi^{[t]}_{i,j}$ for $\forall j\in\mathcal{N}_i$ according to (\ref{eq:compshapley});   \label{ln:compshapley}\\
      Compute normalized Shapley value $\hat{\varphi}^{[t]}_{i,j}$ for $\forall j\in\mathcal{N}_i$ according to (\ref{eq:compnormshapley}); \label{ln:compnormshapley} \\
      Compute weight $\pi^{[t]}_{i,j}$ for $\forall j \in \mathcal{N}_i$ according to (\ref{eq:weight}); \label{ln:compweight}\\
      $\bar{g}^{[t]}_i = \sum_{j\in \mathcal{N}_i} \pi^{[t]}_{i,j} g^{[t]}_{j, i}$;  \label{ln:avggrad}\\
      $\hat{u}^{[t]}_i = \alpha u^{[t-1]}_i + \bar{g}^{[t]}_i$;  \label{ln:uplocmoment01} \\
      $\hat{x}^{[t]}_i = x^{[t-1]}_i - \gamma \hat{u}^{[t]}_i$;  \label{ln:uplocpara01} \\
      Send $\hat{u}^{[t]}_i$ and $\hat{x}^{[t]}_i$ to each neighbor $j \in \mathcal{N}_i \setminus \{i\}$; \label{ln:sendmomentlocpara}\\
      Receive $\hat{u}^{[t]}_j$ and  $\hat{x}^{[t]}_j$ from $\forall j \in \mathcal{N}_i \setminus \{i\}$; \label{ln:recemomentlocpara}\\
      $u^{[t]}_i = \sum_{j\in \mathcal{N}_i} \omega_{i,j} \hat{u}^{[t]}_j$; \label{ln:uplocmoment02}  \\
      $x^{[t]}_i =  \sum_{j\in \mathcal{N}_i} \omega_{i,j} \hat{x}^{[t]}_j$;  \label{ln:uplocpara02}  
    }
  \caption{Our ROSS algorithm (on each agent $i$).} 
  \label{alg:ross}
  \end{algorithm}

  The pseudo-code of our ROSS algorithm is given in \textbf{Algorithm}~\ref{alg:ross}. In each round $t$, each agent $i \in \mathcal{N}$ first uniformly takes a sample $\xi_{i,t}$ from its local dataset $\mathcal{D}_i$ and computes its local stochastic gradient
  \begin{equation} \label{eq:selfgrad}
    g^{[t]}_{i,i} = \nabla F_i \left( x^{[t-1]}_i; \xi_{i,t} \right)
  \end{equation}
  as shown in Line~\ref{ln:selfgrad}. It then shares its current local model $x^{[t-1]}_i$ with its neighbors $\mathcal{N}_i \setminus \{i\}$ (see Line~\ref{ln:sendmodel01}). Once receiving the local model $x^{[t-1]}_j$ from its neighbor $j \in \mathcal{N}_i \setminus \{i\}$, agent $i$ computes stochastic cross-gradient 
  \begin{equation} \label{eq:crossgrad}
    g^{[t]}_{i,j} = \nabla F_i \left( x^{[t-1]}_j; \xi_{i,t} \right)
  \end{equation}
  on its local dataset, and sends $g^{[t]}_{i,j}$ back to agent $j$ (see Lines~\ref{ln:recemodel01}$\sim$\ref{ln:sendcrossgradend}). After receiving $g^{[t]}_{j,i}$ from each neighbor $j \in \mathcal{N}_i \setminus\{i\}$, agent $i$ adopts the gradient to update its local model
  \begin{equation} \label{eq:up4neighbor}
    x^{[t]}_{i,j} = x^{[t-1]}_{i} - \gamma \cdot g^{[t]}_{j,i}, ~\forall j \in \mathcal{N}_i \setminus\{i\}
  \end{equation}
  and it also computes local model update $x^{[t]}_{i,i}$ according to $g^{[t]}_{i,i}$
  \begin{equation} \label{eq:up4self}
    x^{[t]}_{i,i} = x^{[t-1]}_{i} - \gamma \cdot g^{[t]}_{i,i}
  \end{equation}
  as shown in Lines~\ref{ln:rececrossgrad}$\sim$\ref{ln:model4shapley}. These local model updates are then used to calculate the Shapley values (see Line~\ref{ln:compshapley}). Particularly, for any subset $\mathcal{N}^\prime \subseteq \mathcal{N}_i$, we let
  \begin{equation}
    x^{[t]}_{i, \mathcal{N}^\prime} = \frac{1}{|\mathcal{N}^\prime|} \sum_{j \in \mathcal{N}^\prime} x^{[t]}_{i, j}
  \end{equation}
  denote the averaged model update across $\mathcal{N}^\prime$, and $J \left( \xi, x^{[t]}_{i, \mathcal{N}^\prime} \right) \in \{0,1\}$ indicate if model $x^{[t]}_{i, \mathcal{N}^\prime}$ can make a correct prediction for any data sample $\xi \in \mathcal{Q}$. Then, we define a characteristic function $v(\mathcal{N}^\prime; \mathcal{Q})$ as the prediction accuracy of model $x^{[t]}_{i, \mathcal{N}^\prime}$ on the validation dataset $\mathcal{Q}$, i.e., 
  \begin{equation} \label{eq:charfunc}
    v(\mathcal{N}^\prime; \mathcal{Q}) = \frac{1}{|\mathcal{Q}|} \sum_{\xi \in \mathcal{Q}} J \left( \xi, x^{[t]}_{i, \mathcal{N}^\prime} \right), ~\forall \mathcal{N}^\prime \subseteq \mathcal{N}_i.
  \end{equation}
  The Shapley value for $\forall j \in \mathcal{N}_i$ then is defined as
  \begin{equation} \label{eq:compshapley}
    \varphi^{[t]}_{i,j} = \sum_{\mathcal{N}' \subseteq \mathcal{N}_i \setminus \{j\}}  \frac{ v \left( \mathcal{N}' \bigcup \{j\}; {\mathcal{Q}} \right) - v \left( \mathcal{N}'; {\mathcal{Q}} \right) }{ |\mathcal{N}_i| \binom{ |\mathcal{N}_i| - 1}{|\mathcal{N}'|} }
  \end{equation}
  Next, as shown in Line~\ref{ln:compnormshapley}, we employ the min-max normalization in each round to eliminate the influence of the unequal ranges of the Shapley value, the normalized Shapley value of $\forall j \in \mathcal{N}_i$ is calculated as 
  \begin{equation} \label{eq:compnormshapley}
    \hat{\varphi}^{[t]}_{i,j} = \frac{\varphi^{[t]}_{i,j} - \min_{k \in \mathcal{N}_i} \varphi^{[t]}_{i,k}}{\max_{k \in \mathcal{N}_i} \varphi^{[t]}_{i,k} - \min_{k \in \mathcal{N}_i} \varphi^{[t]}_{i,k}} 
  \end{equation}
  According to the normalized Shapley value of each $j \in \mathcal{N}_i$, agent $i$ then calculates weight
  \begin{equation} \label{eq:weight}
    \pi^{[t]}_{i,j} = \frac{\hat{\varphi}^{[t]}_{i,j}}{\omega_{i,j} \sum_{k \in \mathcal{N}_i} \hat{\varphi}^{[t]}_{i,k}}
  \end{equation}
  for $\forall j \in \mathcal{N}_i$ (see Line~\ref{ln:compweight}), based on which, agent $i$ calculates a weighted average across the received stochastic gradients
  \begin{equation}  \label{eq:avggrad}
    \bar{g}^{[t]}_i = \sum_{j\in \mathcal{N}_i} \pi^{[t]}_{i,j} g^{[t]}_{j, i}
  \end{equation}
  As shown in Lines~\ref{ln:uplocmoment01}$\sim$\ref{ln:uplocpara02}, the local model $x^{[t]}_i$ is finally updated in a momentum-like manner. Specifically, agent $i$ first calculates
  \begin{equation} \label{eq:imupmomentum}
    \hat{u}^{[t]}_i = \alpha {u}^{[t-1]}_i + \bar{g}^{[t]}_i
  \end{equation}
  and
  \begin{equation} \label{eq:imupmodel}
    \hat{x}^{[t]}_i = {x}^{[t-1]}_i - \gamma \hat{u}^{[t]}_i
  \end{equation}
  It then sends $\hat{u}^{[t]}_i$ and $\hat{x}^{[t]}_i$ to its neighbors (see Line~\ref{ln:sendmomentlocpara}). After receiving $\hat{u}^{[t]}_j$ and $\hat{x}^{[t]}_j$ from $\forall j \in \mathcal{N}_i \setminus \{ i \}$, agent $i$ updates its momentum and local model by
  \begin{equation} \label{eq:updatemomentum}
    u^{[t]}_i = \sum_{j\in \mathcal{N}_i} \omega_{i,j} \hat{u}^{[t]}_j
  \end{equation}
  and
  \begin{equation} \label{eq:updatemodel}
    x^{[t]}_i =  \sum_{j\in \mathcal{N}_i} \omega_{i,j} \hat{x}^{[t]}_j
  \end{equation}
  respectively.

  As demonstrated in Sec.~\ref{sec:pre}, for each agent $i$, calculating standard Shapley values according to (\ref{eq:compshapley}) mainly relies on the permutations of $\mathcal{N}_i$. This results in significant computational complexity, especially when agent $i$ has a large number of neighbors. To address this problem, the Monte Carlo method is employed to calculate the Shapley values with lower complexity~\cite{CastroGT-COR09, FatimaWJ-AI08}. The pseudo-code of our Monte Carlo-based estimation algorithm (for agent $i$ in round $t$) is given in \textbf{Algorithm}~\ref{alg:shapeyest}. Specifically, we initialize $\varphi^{[t]}_{i,j} = 0$ for $\forall j \in \mathcal{N}_i$ (see Line~\ref{ln:shapleyest-init}). In each iteration $ r = 1,2,\cdots,R$ of our algorithm, we first produce a random permutation $\phi_r$ of $\mathcal{N}_i$ (see Line~\ref{ln:shapleyest-perm}). Then, for each agent $j \in \mathcal{N}_i$, we construct $\mathcal{Z}_j(\phi_r)$, i.e., the set of predecessors of agent $j$ in $\phi_r$, as illustrated in Line~\ref{ln:shapleyest-prede}. We finally calculate $v\left(\mathcal{Z}_j\left(\phi_r\right) \bigcup \{j\}; \mathcal{Q} \right)$ and $v\left(\mathcal{Z}_j\left(\phi_r\right); \mathcal{Q} \right)$ according to (\ref{eq:charfunc}), and update $\varphi^{[t]}_{i,j}$ by
  \begin{equation} \label{eq:updateshap}
    \varphi^{[t]}_{i,j} \leftarrow \varphi^{[t]}_{i,j} + \frac{v\left(\mathcal{Z}_j\left(\phi_r\right) \bigcup \{j\}; \mathcal{Q} \right) - v\left(\mathcal{Z}_j\left(\phi_r\right); \mathcal{Q} \right)} {R}
  \end{equation}
  as demonstrated in Line~\ref{ln:shapleyest-shapley}. It is shown that, the cost of this Monte Carlo-based estimation mainly depends on the maximum number of iterations $R$. Larger $R$ yields higher accuracy at the expense of additional computation. As will be shown in Sec.~\ref{sec:exp}, setting $R=10$ is sufficient for ROSS to deliver superior performance, significantly better than the state-of-the-art baselines.
  \begin{algorithm}[htb!]
    \KwIn{Neighbors $\mathcal{N}_i$, local models $\left\{ x^{[t]}_{i,j} \right\}_{j \in \mathcal{N}_{i}}$ and validation dataset $\mathcal{Q}$, and the maximum number of iterations $R$.}
    \KwOut {Shapley value $\varphi^{[t]}_{i,j}$ of each agent $j \in \mathcal{N}_i$ in round $t$.}
    %
    %
    %
    $\varphi^{[t]}_{i,j} = 0$ for $\forall j \in \mathcal{N}_i$;  \label{ln:shapleyest-init}\\
    \ForEach{$r = 1, 2, \cdots, R$}{ 
      Let $\phi_r$ be a random permutation of  $\mathcal{N}_i$;  \label{ln:shapleyest-perm}\\
      \ForEach{$j \in \mathcal{N}_i $}{
        $\mathcal{Z}_j \left( \phi_r \right) = \left\{ j' \in \mathcal{N}_i ~\big{|}~ \phi_r(j') < \phi_r(j) \right\}$;  \label{ln:shapleyest-prede}\\
        $\varphi^{[t]}_{i,j} \leftarrow \varphi^{[t]}_{i,j} + \frac{v\left(\mathcal{Z}_j\left(\phi_r\right) \bigcup \{j\}; \mathcal{Q} \right) - v\left(\mathcal{Z}_j\left(\phi_r\right); \mathcal{Q} \right)} {R}$
            \label{ln:shapleyest-shapley} \\
        }
    }
  \caption{Shapley value estimation for agent $i$ in round $t$.} 
  \label{alg:shapeyest}
  \end{algorithm}

\section{Convergence Analysis} \label{sec:analysis}
  We reveal the convergence of our ROSS algorithm in this section. In particular, we first present our main result in Sec.~\ref{ssec:mainresult} and then demonstrate the detailed proof in Sec.~\ref{ssec:proof}.

  \subsection{Main Results} \label{ssec:mainresult}
    Before giving our main result about the convergence of our ROSS algorithm, we first introduce \textbf{Assumptions}~\ref{ass:L-smooth}$\sim$\ref{ass:ds_mat} which are usually used in convergence analysis in existing studies\cite{LianZZHZL-NIPS17,YuJY-ICML19,BaluJTHLS-ICASSP21,EsfandiariTJBHHS-ICML21,ZhangFLYLZ-MobiHoc22}. It is worth noting that, when these assumptions do not hold, our algorithm still works but the convergence may not be ensured. \textbf{Assumption}~\ref{ass:L-smooth} implies that $\mathcal{F}(x)$ is also $L$-smooth. Moreover, we assume the variances induced by the data heterogeneity are bounded in \textbf{Assumption}~\ref{ass:bd-varia}, and suppose the adjacent matrix of the communication graph is doubly stochastic in \textbf{Assumption}~\ref{ass:ds_mat}.
    \begin{assumption}  \label{ass:L-smooth}
      Each function $f_i(x)$ is $L$-smooth for $\forall i\in\mathcal{N}$.
    \end{assumption}
    \begin{assumption} \label{ass:bd-varia}
      There exist $\sigma > 0$, $\varsigma > 0$ such that
      \begin{equation} \label{eq:bd-varia0}
        \mathbb{E}_{\xi \sim \mathcal{D}_i} \Big\| \nabla F_i(x; \xi) - \nabla f_i(x) \Big\|^2 \leq \sigma^2, ~~\forall i \in \mathcal{N}
      \end{equation}
      and
      \begin{equation} \label{eq:bd-varia1}
        \Big\| \nabla f_i(x) - \nabla \mathcal F(x) \Big\|^2 \leq \varsigma^2,~~\forall i \in \mathcal{N}
      \end{equation}
    \end{assumption}
    \begin{assumption} \label{ass:ds_mat}
      Given the symmetric doubly stochastic matrix $\mathbf{W}$, let $\lambda_i(\mathbf{W})$ denote the $i$-th largest eigenvalue of $\mathbf{W}$. we have $\lambda_1(\mathbf{W}) = 1$ and there exists constant $\rho <1$ such that $ \max \{ |\lambda_2(\mathbf{W})|, |\lambda_N(\mathbf{W})| \} \leq \sqrt{\rho}$.
    \end{assumption}

    Our main results about the convergence of our ROSS algorithm is given in \textbf{Theorem}~\ref{thm:main}. It is revealed by \textbf{Theorem}~\ref{thm:main} that, the average gradient magnitude achieved by our ROSS algorithm is mainly upper-bounded by the difference between the initial value of the objective function and the optimal value.
    \begin{theorem} \label{thm:main}
      Suppose \textbf{Assumptions}~\ref{ass:L-smooth}-\ref{ass:ds_mat} hold, and learning rate $\gamma$ satisfies the following condition
      \begin{align} \label{eq:main-rate}
        \gamma \leq \min \left\{ \begin{aligned} & \frac{\alpha L}{(1-\alpha)^2},~~ \frac{(1-\alpha)(1-\sqrt{\rho})}{8L\sqrt{\frac{\hat{\varphi}^2_{\max}}{\omega^4_{\min}} + 2}}, \\& \frac{(1-\alpha)^2 + \sqrt{(1-\alpha)^4 - 8L^2\alpha \left(12+\frac{64\hat{\varphi}^2_{\max}}{\omega^4_{\min}}\right)^2}}{16L\left(3+\frac{16\hat{\varphi}^2_{\max}}{\omega^4_{\min}}\right)} \end{aligned} \right\}
      \end{align}
      For any $T \geq 1$, we have
      \begin{align} \label{eq:main-bound}
        & \frac{1}{T} \sum^{T}_{t=1} \mathbb{E} \left[ \left\| \nabla \mathcal{F} \left( \bar{x}^{[t-1]} \right) \right\|^2 \right]  \nonumber\\
        \leq & \frac{1}{m_1 T} \left( \mathcal{F} \left( \bar{x}^{[0]} \right) - \mathcal{F}^* \right) + \bigg( m_2 + \frac{ m_3 \gamma^2 \alpha^2}{(1-\alpha)^4} + m_4 \bigg) \left( \frac{\left(4 \sigma^2 + 16 \varsigma^2\right)\hat{\varphi}^2_{\max} }{\omega^4_{\min}} + \frac{6 \sigma^2}{N}\right) \nonumber\\
        & + \frac{8 \gamma^2 \left( \hat{\varphi}^2_{\max} + 2\omega^4_{\min} \right) \left( \sigma^2 + 4 \varsigma^2 \right)}{\omega^4_{\min} (1-\alpha)^2 \left( 1-\sqrt{\rho} \right)^2} \left( \frac{32L^2 \hat{\varphi}^2_{\max}}{\omega^4_{\min}} \left( m_2 + \frac{ m_3 \gamma^2 \alpha^2}{(1-\alpha)^4} \right) + \frac{16m_4 L^2 \hat{\varphi}^2_{\max}}{\omega^4_{\min}} + m_5 \right)  
      \end{align}
      for \textbf{Algorithm}~\ref{alg:ross}, where $\mathcal{F}^*$ denotes the optimal value of the objective function, and
      \begin{align} \label{eq:main-constants}
      \begin{cases}
        \bar{x}^{[t]} = \frac{1}{N} \sum^N_{i=1} x^{[t]}_i \vspace{1ex}\\
        \omega_{\min} = \min_{i \in \mathcal{N}, j \in \mathcal{N}_i} \omega_{i,j} \vspace{1ex}\\
        \hat{\varphi}_{\max} = \max_{t\in \{1,2,\cdots, T\}} \hat{\varphi}^{[t]}_{\max} \vspace{1ex}\\
        \hat{\varphi}^{[t]}_{\max} =\max_{i\in\mathcal{N},j\in\mathcal{N}_i}\frac{\hat{\varphi}^{[t]}_{i,j}} {\sum_{j\in\mathcal{N}_i} \hat{\varphi}^{[t]}_{i,j}}  \vspace{1ex}\\
        m_1 = \frac{\gamma}{2(1-\alpha)} - \frac{(1-\alpha)\gamma^2}{2\alpha L} \vspace{1ex} \\
        m_2 = \frac{1}{m_1} \left( \frac{\alpha L \gamma^2}{2(1-\alpha)^3} + \frac{L \gamma^2}{2(1-\alpha)^2} \right) \vspace{1ex}\\
        m_3 = \frac{L(1-\alpha)}{2 m_1 \alpha}  \vspace{1ex}\\
        m_4 = \frac{L \alpha }{2 m_1 (1-\alpha)^3}   \vspace{1ex}\\
        m_5 = \frac{L^2 \gamma}{2 m_1 (1-\alpha)}
      \end{cases}
      \end{align}
    \end{theorem}

    We then characterize the convergence of our algorithm more explicitly in \textbf{Corollary}~\ref{cor:main}.
    \begin{corollary} \label{cor:main}
      Let learning rate $\gamma = O\left(\sqrt{\frac{N}{T}}\right)$ and $\sigma = \varsigma = O\left(\frac{1}{\sqrt{NT}}\right)$. When $T$ is sufficiently large such that
      \begin{align} \label{eq:cormain-t}
        T \geq \max \left\{ \begin{aligned} & \frac{N(1-\alpha)^4} {\alpha^2 L^2}, ~~ \frac{64NL^2 \left( \frac{\hat{\varphi}^2_{\max}}{\omega^4_{\min}} + 2 \right)}{(1-\alpha)^2(1-\sqrt{\rho})^2}, \\ & 256 N L^2 \left( 3 \omega^4_{\min} + 16\hat{\varphi}^2_{\max} \right)^2 \Bigg( (1-\alpha)^2 \omega^4_{\min} +  \sqrt{(1-\alpha)^4 \omega^8_{\min} - 128 \alpha L^2 \bigg( \begin{aligned} & 3 \omega^4_{\min} \\ +& 16\hat{\varphi}^2_{\max} \end{aligned} \bigg)^2} \Bigg)^{-2} \end{aligned} \right\}
      \end{align}
      we have
      \begin{align} \label{eq:cor}
        \frac{1}{T} \sum^{T}_{t=1} \mathbb{E} \left[ \left\| \nabla \mathcal{F} \left( \bar{x}^{[t-1]} \right) \right\|^2 \right] \leq   M \left( \frac{1}{\sqrt{NT}} + \frac{1}{\sqrt{N^3T}} + \frac{1}{\sqrt{NT^3}} + \frac{\sqrt{N}}{\sqrt{T^5}} + \frac{1}{\sqrt{N^5T}}  + \frac{1}{\sqrt{N^3T^3}} + \frac{1}{T^2} \right) 
      \end{align}
      for some constant $M > 0$.
    \end{corollary}

    This corollary implies that when $T$ is sufficiently large, the convergence rate of our algorithm mainly depends on $\mathcal{O} \left( \frac{1}{\sqrt{NT}} \right)$, and hence, it is revealed that our algorithm has a linear speedup. Although various studies (including ours) propose algorithms to address the challenges of non-IID data in decentralized learning, they all maintain this rate~\cite{EsfandiariTJBHHS-ICML21, TakezawaBNSY-TMLR23, AketiHR-NIPS24}, which to the best of our knowledge remains the optimal bound even under IID data~\cite{LianZZHZL-NIPS17, AssranLBR-ICML19, YuJY-ICML19}. ROSS attains this state-of-the-art rate while leveraging Shapley values to fairly weight local and cross-gradients, thereby accounting for heterogeneous neighbor contributions without sacrificing convergence. As will be shown in Sec.~\ref{sec:exp}, although ROSS only achieves the same linear speedup as the state-of-the-art algorithms, our extensive experiments demonstrate that ROSS consistently outperforms them in practice.

  \subsection{Proof of Our Main Results} \label{ssec:proof}
    We present the detailed proof of \textbf{Theorem}~\ref{thm:main}. To maintain the fluency of our presentation, all proofs of the lemmas are given in the supplementary material (see Sec. I$\sim$VI). The key to our proof is to reveal the influence of Shapley values on the convergence process of our algorithm (especially in \textbf{Lemmas}~\ref{lem:diffhatgrad}, ~\ref{lem:bdavggrad}, \ref{lem:barG-G}, and \ref{lem:barx-x}).

    Recall that $\bar{g}^{[t]}_i$ denotes the average of the stochastic gradients $\Big\{ g^{[t]}_{j,i} \Big\}_{j\in\mathcal{N}_i}$ weighted by the Shapley values (see (\ref{eq:avggrad})). We first show in \textbf{Lemma}~\ref{lem:diffhatgrad} that the (expected) difference between the weighted average gradient and the gradients calculated by the agents based on their local datasets is bounded. In particular, \textbf{Lemma}~\ref{lem:diffhatgrad} reveals that, $\mathbb{E} \Big[ \Big\| \frac{1}{N} \sum^N_{i=1} \big( \bar{g}^{[t]}_i - g^{[t]}_{i,i} \big) \Big\|^2 \Big]$ is bounded by $\mathbb{E} \Big[ \frac{1}{N} \sum^N_{i=1}  \Big\| x^{[t-1]}_i - \bar{x}^{[t-1]} \Big\|^2 \Big]$ and $\mathbb{E} \Big[ \Big\| \frac{1}{N} \sum^N_{i=1} \nabla f_i \big(x^{[t-1]}_i \big) \Big\|^2 \Big]$
    \begin{lemma} \label{lem:diffhatgrad}
      When \textbf{Assumptions}~\ref{ass:L-smooth}-\ref{ass:ds_mat} hold, in each round $t \geq 1$, we have 
      \begin{align} \label{eq:diffhatgrad}
        & \mathbb{E} \left[ \left\| \frac{1}{N} \sum^N_{i=1} \bigg( \bar{g}^{[t]}_i - g^{[t]}_{i,i} \bigg) \right\|^2 \right]  \nonumber\\
        \leq & \frac{\left(2\sigma^2 + 8\varsigma^2\right)\hat{\varphi}^2_{\max}}{\omega^4_{\min}} + \frac{2\sigma^2}{N} 
        + \frac{16L^2\hat{\varphi}^2_{\max}} {N \omega^4_{\min}} \sum^N_{i=1} \mathbb{E} \bigg[ \bigg\| x^{[t-1]}_i - \bar{x}^{[t-1]} \bigg\|^2 \bigg]
        + \bigg( \frac{8\hat{\varphi}^2_{\max}}{\omega^4_{\min}} + 2 \bigg) \mathbb{E} \bigg[ \bigg\| \frac{1}{N} \sum^N_{i=1} \nabla f_i \bigg(x^{[t-1]}_i \bigg)  \bigg\|^2 \bigg]
      \end{align}
    \end{lemma}

    To prove \textbf{Theorem}~\ref{thm:main}, we first introduce an auxiliary sequence $\left\{ \bar{S}^{[t]} \right\}_{t \geq 0}$ such that
    \begin{align} \label{eq:auxseq}
      \bar{S}^{[t]} = 
      \begin{cases}
        \bar{x}^{[t]} & \text{with}~t = 0  \\
        \frac{1}{1-\alpha} \bar{x}^{[t]} - \frac{\alpha}{1-\alpha} \bar{x}^{[t-1]} & \text{with}~t \geq 1
      \end{cases}
    \end{align}
    where $\bar{x}^{[t]} = \frac{1}{N} \sum^N_{i=1} x^{[t]}_i$ (see (\ref{eq:main-constants})). $\bar{S}^{[t]}$ indicates the change of the average of the local models weighted by momentum coefficient $\alpha$. In the following, the initial value will be set to $0$.
    According to \textbf{Algorithm}~\ref{alg:ross}, we show in \textbf{Lemma}~\ref{lem:seqdiff} that $\bar{S}^{[t]} - \bar{S}^{[t-1]}$ can be represented by $\left\{ \bar{g}^{[t]}_{i} \right\}_{i\in\mathcal{N}}$ and demonstrate in \textbf{Lemma}~\ref{lem:sandx} that the bound of $\sum^{T}_{t=1} \left\| \bar{S}^{[t]} - \bar{x}^{[t]} \right\|^2 = \big(\frac{\alpha}{1-\alpha}\big)^2 \sum^T_{t=1}  \big\| \bar{x}^{[t]} - \bar{x}^{[t-1]}\big\|^2$ mainly depends on $\sum^{T}_{t=1} \left\| \frac{1}{N}\sum^{N}_{i=1} \bar{g}_i^{[t]} \right\|^2 $.
    \begin{lemma} \label{lem:seqdiff}
      According to \textbf{Algorithm}~\ref{alg:ross}, for any $t \geq 1$, we have 
      \begin{equation}  \label{eq:seqdiff}
        \bar{S}^{[t]} - \bar{S}^{[t-1]} = \frac{-\gamma}{N(1-\alpha)} \sum^N_{i=1} \bar{g}_i^{[t]}
      \end{equation}
    \end{lemma}
    \begin{lemma} \label{lem:sandx}
      For any $T \geq 1$, we have
      \begin{equation}
        \sum^{T}_{t=1} \left\| \bar{S}^{[t]} - \bar{x}^{[t]} \right\|^2 \leq \frac{\gamma^2 \alpha^2}{(1-\alpha)^4} \sum^{T}_{t=1} \bigg\| \frac{1}{N}\sum^{N}_{i=1} \bar{g}_i^{[t]} \bigg\|^2 
      \end{equation}
    \end{lemma}
    Also, based on \textbf{Lemma}~\ref{lem:diffhatgrad}, we illustrate the bound of $\mathbb{E} \left[ \left\| \frac{1}{N}\sum^{N}_{i=1} \bar{g}^{[t]}_i \right\|^2 \right]$ in \textbf{Lemma}~\ref{lem:bdavggrad}.

    \begin{lemma} \label{lem:bdavggrad}
      When \textbf{Assumptions}~\ref{ass:L-smooth}-\ref{ass:ds_mat} hold, for any $t \geq 1$, we have
      \begin{align} \label{eq:avggrad01}
        \mathbb{E} \left[ \left\| \frac{1}{N}\sum^{N}_{i=1} \bar{g}^{[t]}_i \right\|^2 \right] 
        \leq & \frac{\left(4\sigma^2 + 16\varsigma^2\right)\hat{\varphi}^2_{\max}}{\omega^4_{\min}} + \frac{6\sigma^2}{N} 
        + \frac{32L^2\hat{\varphi}^2_{\max}} {N \omega^4_{\min}} \sum^N_{i=1} \mathbb{E} \left[ \left\| x^{[t-1]}_i - \bar{x}^{[t-1]} \right\|^2 \right] \nonumber\\
        &+ \left( \frac{16\hat{\varphi}^2_{\max}}{\omega^4_{\min}} + 6 \right) \mathbb{E} \left[ \left\| \frac{1}{N} \sum^N_{i=1} \nabla f_i \left(x^{[t-1]}_i \right)  \right\|^2 \right]      
      \end{align}
    \end{lemma}

    Before continuing our proof, we first introduce some notations (see (\ref{eq:notation})) and facts (i.e., \textbf{Fact}~\ref{fact:fact-1} and \textbf{Fact}~\ref{fact:fact-2}) which will be useful later. Similar facts are used in \cite{YuJY-ICML19, wang2021cooperative}.
    \begin{align} \label{eq:notation}
    \begin{cases}
      \bar{\mathbf{G}}^{[t]} &\triangleq \left[ \bar{g}_1^{[t]}, \bar{g}_2^{[t]}, \cdots, \bar{g}_N^{[t]} \right]  \vspace{1ex}\\
      \mathbf{U}^{[t]} &\triangleq \left[ u^{[t]}_1, u^{[t]}_2, \cdots,  u^{[t]}_N \right]  \vspace{1ex}\\
      \hat{\mathbf{U}}^{[t]} &\triangleq \left[ \hat{u}^{[t]}_1, \hat{u}^{[t]}_2, \cdots,  \hat{u}^{[t]}_N \right]  \vspace{1ex}\\
      \mathbf{X}^{[t]} &\triangleq \left[ x^{[t]}_1, x^{[t]}_2, \cdots, x^{[t]}_N \right]   \vspace{1ex}\\
      \mathbf{G}^{[t]} &\triangleq \left[ g^{[t]}_{1,1}, g^{[t]}_{2,2}, \cdots, g^{[t]}_{N,N} \right]  \vspace{1ex}\\
      \mathbf{H}^{[t]} &\triangleq \left[ \nabla f_1\left(x^{[t-1]}_1\right),  \cdots, \nabla f_N\left(x^{[t-1]}_N\right)  \right]  \vspace{1ex}\\   
      \mathbf{J}^{[t]} &\triangleq \left[ \nabla f_1\left(\bar{x}^{[t-1]}\right), \cdots, \nabla f_N\left(\bar{x}^{[t-1]}\right)  \right] 
    \end{cases}
    \end{align}

    \begin{fact} \label{fact:fact-1}
      Let $\mathbf{Q} = \frac{1}{N} \mathbf{1}\mathbf{1}^{\mathrm T}$ where $\mathbf{1}$ denotes a $N$-dimensional column vector with all entries being $1$. For any doubly stochastic matrix $\mathbf{W}$, we have the following properties:
      \begin{itemize}
        \item $\mathbf{Q} \mathbf{W} = \mathbf{W} \mathbf{Q}$;
        \item $(\mathbf{I}- \mathbf{Q}) \mathbf{W} = \mathbf{W} (\mathbf{I}- \mathbf{Q})$;
        \item For any integer $t \geq 1$, $\left\| (\mathbf{I}- \mathbf{Q}) \mathbf{W}^t \right\|_{\mathfrak{s}} \leq (\sqrt{\rho})^t$, where $\|\cdot\|_{\mathfrak{s}}$ denotes the spectrum norm.
      \end{itemize}
    \end{fact}

    \begin{fact} \label{fact:fact-2}
      Let $\mathbf{B}_1, \mathbf{B}_2, \cdots, \mathbf{B}_N$ be $N$ arbitrary real-valued matrices such that $\|\mathbf{B}_i\|^2_{\mathfrak{F}} = \sum^N_{k=1} \| \mathbf{b}_{i,k} \|^2$, where $\|\cdot\|_{\mathfrak{F}}$ denotes the Frobenius norm, and $ \mathbf{b}_{i,k}$ is the $k$-th column of $\mathbf{B}_i$. We have 
      \begin{align}
        \left\| \sum^N_{i=1} \mathbf{B}_i \right\|^2_{\mathfrak{F}} \leq \sum^N_{i=1} \sum^N_{j=1} \left\| \mathbf{B}_i \right\|_{\mathfrak{F}} \left\| \mathbf{B}_j \right\|_{\mathfrak{F}}
      \end{align}
    \end{fact}

    In the following, we first reveal the bound of $\mathbb{E} \Big[ \left\| \bar{\mathbf{G}}^{[t]} - \mathbf{G}^{[t]} \right\|^2_{\mathfrak{F}} \Big]$ in \textbf{Lemma}~\ref{lem:barG-G}, based on which, we show in \textbf{Lemma}~\ref{lem:barx-x} that $\sum^{T}_{t=1} \mathbb{E} \Big[ \frac{1}{N} \sum^N_{i=1} \Big\| \bar{x}^{[t]} - x^{[t]}_i \Big\|^2 \Big]$ is bounded by $\sum^{T}_{t=1} \mathbb{E} \Big[ \left\| \frac{1}{N} \sum^N_{i=1} \nabla f_i \big( x^{[t-1]}_i \big) \right\|^2 \Big]$.
    \begin{lemma} \label{lem:barG-G} 
      When all \textbf{Assumptions}~\ref{ass:L-smooth}-\ref{ass:ds_mat} hold, for any $t \geq 1$, we have
      \begin{align} \label{eq:barG-G}
        \mathbb{E} \left[ \left\| \bar{\mathbf{G}}^{[t]} - \mathbf{G}^{[t]} \right\|^2_{\mathfrak{F}} \right]  
        \leq & \frac{\left(2N\sigma^2 + 8N\varsigma^2\right)\hat{\varphi}^2_{\max}}{\omega^4_{\min}} + 2N\sigma^2 +  8N\varsigma^2   \nonumber\\
        & + \left( \frac{16L^2\hat{\varphi}^2_{\max}}{\omega^4_{\min}} + 16L^2 \right) \sum^N_{i=1} \mathbb{E} \left[ \left\| x^{[t-1]}_i - \bar{x}^{[t-1]} \right\|^2 \right]  \nonumber\\
        & +  \left( \frac{8N\hat{\varphi}^2_{\max}}{\omega^4_{\min}} + 8N \right) \mathbb{E} \left[ \left\| \frac{1}{N} \sum^N_{i=1} \nabla f_i \left(x^{[t-1]}_i \right) \right\|^2 \right]
      \end{align}
    \end{lemma}
    \begin{lemma} \label{lem:barx-x} 
      Let \textbf{Assumptions}~\ref{ass:L-smooth}-\ref{ass:ds_mat} hold and  $\left\{ \bar{x}^{[t]} \right\}_{t \geq 1}$ be the sequence obtained by the iterations of \textbf{Algorithm}~\ref{alg:ross}. If momentum coefficient $\alpha \in [0,1)$ and learning rate 
      \begin{equation}
        \gamma \leq \frac{(1-\alpha)(1-\sqrt{\rho})}{8L\sqrt{\frac{\hat{\varphi}^2_{\max}}{\omega^4_{\min}} + 2}}
      \end{equation}
      then for any $T \geq 1$, we have
      \begin{align} \label{eq:barx-x}
        \sum^{T}_{t=1} \frac{1}{N} \sum^N_{i=1} \mathbb{E} \left[ \left\| \bar{x}^{[t]} - x^{[t]}_i \right\|^2 \right] 
        \leq 8 T \gamma^2 \frac{ \left( \frac{\hat{\varphi}^2_{\max}}{\omega^4_{\min}} +2 \right) \left( \sigma^2 + 4\varsigma^2 \right)}{(1-\alpha)^2 (1-\sqrt{\rho})^2} 
        + \frac{1}{2L^2} \sum^{T}_{t=1} \mathbb{E} \left[ \Big\| \frac{1}{N} \sum^N_{i=1} \nabla f_i \left( x^{[t-1]}_i \right) \Big\|^2 \right]   
      \end{align}
    \end{lemma}

    With the above lemmas, we present the main proof of \textbf{Theorem}~\ref{thm:main}.
    Based on the $L$-smoothness of $\mathcal{F}$, we have
    \begin{align} \label{eq:fbars}
      \mathbb{E} \Big[ \mathcal{F} \Big(\bar{S}^{[t]} \Big) \Big]
      \leq & \mathbb{E} \Big[ \mathcal{F} \Big(\bar{S}^{[t-1]} \Big) \Big] + \frac{L}{2} \mathbb{E} \Big[ \Big\| \bar{S}^{[t]} - \bar{S}^{[t-1]} \Big\|^2 \Big] \nonumber\\
      & + \mathbb{E} \left[ \left\langle \nabla \mathcal{F} \left(\bar{S}^{[t-1]} \right), \bar{S}^{[t]} - \bar{S}^{[t-1]} \right\rangle \right] 
    \end{align}
    According to \textbf{Lemma}~\ref{lem:seqdiff}, we have
    \begin{align} \label{eq:bars-bars}
      & \mathbb{E} \Big[ \Big\langle \nabla \mathcal{F} \Big( \bar{S}^{[t-1]} \Big), \bar{S}^{[t]} - \bar{S}^{[t-1]} \Big\rangle \Big] \nonumber\\
      = & \frac{-\gamma}{1-\alpha} \mathbb{E} \Big[ \Big\langle \nabla \mathcal{F} \Big( \bar{S}^{[t-1]} \Big),\frac{1}{N}\sum^N_{i=1} \bar{g}_i^{[t]} \Big\rangle \Big]  \nonumber\\
      = & \frac{-\gamma}{1-\alpha} \mathbb{E} \Big[ \Big\langle \nabla \mathcal{F} \Big( \bar{S}^{[t-1]} \Big) - \nabla \mathcal{F} \Big( \bar{x}^{[t-1]} \Big) ,\frac{1}{N}\sum^N_{i=1} \bar{g}_i^{[t]} \Big\rangle \Big]  \nonumber\\
      & + \frac{-\gamma}{1-\alpha} \mathbb{E} \Big[ \Big\langle \nabla \mathcal{F} \Big( \bar{x}^{[t-1]} \Big), \frac{1}{N}\sum^N_{i=1} \bar{g}_i^{[t]} \Big\rangle \Big]  \nonumber\\
      = & \underbrace{ \frac{-\gamma}{1-\alpha} \mathbb{E} \Big[ \Big\langle \nabla \mathcal{F} \Big( \bar{S}^{[t-1]} \Big) - \nabla \mathcal{F} \Big( \bar{x}^{[t-1]} \Big) ,\frac{1}{N}\sum^N_{i=1} \bar{g}_i^{[t]} \Big\rangle \Big] }_{\mathrm{\MakeTextUppercase{\romannumeral 3}}}  \nonumber\\
      & + \underbrace{ \frac{-\gamma}{1-\alpha} \mathbb{E} \Big[ \Big\langle \nabla \mathcal{F} \Big(\bar{x}^{[t-1]} \Big), \frac{1}{N}\sum^N_{i=1} \Big( \bar{g}_i^{[t]} - g^{[t]}_{i,i} \Big) \Big\rangle \Big] }_{\mathsf{\MakeTextUppercase{\romannumeral 4}}} \nonumber\\
      & + \underbrace{ \frac{-\gamma}{1-\alpha} \mathbb{E} \Big[ \Big\langle \nabla \mathcal{F} \Big(\bar{x}^{[t-1]} \Big), \frac{1}{N}\sum^N_{i=1} g^{[t]}_{i,i} \Big\rangle \Big] }_{\mathrm{\MakeTextUppercase{\romannumeral 5}}}
    \end{align}
    which can be bounded by bounding $\mathrm{\MakeTextUppercase{\romannumeral 3}}$, $\mathrm{\MakeTextUppercase{\romannumeral 4}}$ and $\mathrm{\MakeTextUppercase{\romannumeral 5}}$. For $\mathrm{\MakeTextUppercase{\romannumeral 3}}$, we have
    \begin{align} \label{eq:term-iii}
      & \frac{-\gamma}{1-\alpha} \mathbb{E} \Big[ \Big\langle \nabla \mathcal{F} \Big( \bar{S}^{[t-1]} \Big) - \nabla \mathcal{F} \Big(\bar{x}^{[t-1]} \Big) ,\frac{1}{N}\sum^N_{i=1} \bar{g}_i^{[t]} \Big\rangle \Big] \nonumber\\
      \stackrel{\scriptsize{\circled{1}}} \leq& \frac{(1-\alpha)L}{2{\alpha}} \mathbb{E} \Big[ \Big\| \bar{S}^{[t-1]} - \bar{x}^{[t-1]} \Big\|^2 \Big] \
      + \frac{\alpha L \gamma^2}{2(1-\alpha)^3} \mathbb{E} \Big[ \Big\| \frac{1}{N}\sum^N_{i=1} \bar{g}_i^{[t]} \Big\|^2 \Big]
    \end{align}
    where we have $\scriptsize{\circled{1}}$ by applying the basic inequality $\langle a,b \rangle \leq \frac{1}{2} \|a\|^2 + \frac{1}{2} \|b\|^2$ with $a = \frac{\sqrt{1-\alpha}}{\sqrt{\alpha L}} \left( \nabla \mathcal{F} \left( \bar{S}^{[t-1]} \right) - \nabla \mathcal{F} \left( \bar{x}^{[t-1]} \right) \right)$ and $b = \frac{-\gamma \sqrt{\alpha L}}{(1-\alpha)^{\frac{3}{2}}} \left( \frac{1}{N} \sum^N_{i=1} \bar{g}_i^{[t]} \right)$, and using the smoothness of function $\mathcal{F}(\cdot)$.
    For $\mathrm{\MakeTextUppercase{\romannumeral 4}}$, we have
    \begin{align} \label{eq:term-iv}
      & \frac{-\gamma}{1-\alpha} \mathbb{E} \Big[ \Big\langle \nabla \mathcal{F} \Big( \bar{x}^{[t-1]} \Big), \frac{1}{N}\sum^N_{i=1} \Big( \bar{g}_i^{[t]} - g^{[t]}_{i,i} \Big) \Big\rangle \Big]  \nonumber\\
      \leq & \frac{(1-\alpha)\gamma^2}{2\alpha L} \mathbb{E} \Big[ \Big\| \nabla \mathcal{F} \Big(\bar{x}^{[t-1]} \Big) \Big\|^2 \Big]
      + \frac{\alpha L}{2(1-\alpha)^3} \mathbb{E} \Big[ \Big\| \frac{1}{N}\sum^N_{i=1} \Big( \bar{g}_i^{[t]} - g^{[t]}_{i,i} \Big) \Big\|^2 \Big]  
    \end{align}
    by applying $\langle a,b \rangle \hspace{-1ex} \leq \frac{1}{2} \|a\|^2 + \frac{1}{2} \|b\|^2$ with $a = \frac{-\gamma \sqrt{1-\alpha}}{\sqrt{\alpha L}} \nabla \mathcal{F}\left( \bar{x}^{[t-1]} \right) $ and $b = \frac{\sqrt{\alpha L }}{(1-\alpha)^\frac{3}{2}} \left(\frac{1}{N}\sum^N_{i=1} \left( \bar{g}_i^{[t]} - g^{[t]}_{i,i} \right) \right)$.
    Then, for $\mathrm{\MakeTextUppercase{\romannumeral 5}}$, we have
    \begin{align} \label{eq:term-v-00}
      \mathbb{E} \left[ \Big\langle \nabla \mathcal{F} \Big(\bar{x}^{[t-1]} \Big), \frac{1}{N}\sum^N_{i=1} g^{[t]}_{i,i} \Big\rangle \right]
      = \mathbb{E} \Big[ \Big\langle \nabla \mathcal{F} \Big(\bar{x}^{[t-1]} \Big), \frac{1}{N}\sum^N_{i=1}  \nabla f_i \Big( x^{[t-1]}_i \Big) \Big\rangle \Big]
    \end{align}
    Therein,
    \begin{align} \label{eq:term-v-01}
      & \bigg\langle \nabla \mathcal{F} \left( \bar{x}^{[t-1]} \right), \frac{1}{N}\sum^N_{i=1} \nabla f_i \left( x^{[t-1]}_i \right) \bigg\rangle   \nonumber\\
      \stackrel{\scriptsize{\circled{1}}} = & \frac{1}{2} \left\| 
       \nabla \mathcal{F} \left( \bar{x}^{[t-1]} \right) \right\|^2 + \frac{1}{2} \Big\| \frac{1}{N}\sum^N_{i=1} \nabla f_i \Big( x^{[t-1]}_i \Big) \Big\|^2 
      - \frac{1}{2} \Big\| \nabla \mathcal{F} \Big(\bar{x}^{[t-1]} \Big) - \frac{1}{N}\sum^N_{i=1}  \nabla f_i \Big( x^{[t-1]}_i \Big) \Big\|^2   \nonumber\\
      \stackrel{\scriptsize{\circled{2}}} \geq & \frac{1}{2} \left\| \nabla \mathcal{F} \left( \bar{x}^{[t-1]} \right) \right\|^2 + \frac{1}{2} \Big\| \frac{1}{N}\sum^N_{i=1} \nabla f_i \Big( x^{[t-1]}_i \Big) \Big\|^2 
      - \frac{L^2}{2N} \sum^N_{i=1} \Big\| \bar{x}^{[t-1]} - x^{[t-1]}_i \Big\|^2  
    \end{align}
    where we have $\scriptsize{\circled{1}}$ according to $\langle a, b \rangle = \frac{1}{2} \left( \|a\|^2 + \|b\|^2 - \|a-b\|^2 \right)$ with $a=\nabla \mathcal{F}(\bar{x}^{[t-1]})$ and $b=\frac{1}{N}\sum^N_{i=1} \nabla f_i( x^{[t-1]}_i)$; $\scriptsize{\circled{2}}$ since
    \begin{align*}
      & \Big\| \nabla \mathcal{F}(\bar{x}^{[t-1]}) - \frac{1}{N}\sum^N_{i=1} \nabla f_i(x^{[t-1]}_i) \Big\|^2  \nonumber\\
      =& \Big\| \frac{1}{N} \sum^N_{i=1} \nabla f_i(\bar{x}^{[t-1]}) - \frac{1}{N}\sum^N_{i=1} \nabla f_i(x^{[t-1]}_i) \Big\|^2 \nonumber\\
      \stackrel{\scriptsize{\circled{1}}} \leq & \frac{1}{N} \sum^N_{i=1} \left\| \nabla f_i(\bar{x}^{[t-1]}) -  \nabla f_i(x^{[t-1]}_i) \right\|^2 \nonumber\\
      \stackrel{\scriptsize{\circled{2}}} \leq & \frac{1}{N} \sum^N_{i=1} L^2 \left\| \bar{x}^{[t-1]} - x^{[t-1]}_i \right\|^2
    \end{align*}
    where we have ${\scriptsize{\circled{1}}}$ due to the convexity of $\|\cdot\|^2$ and \textit{Jensen's Inequality}; and ${\scriptsize{\circled{2}}}$ according to the smoothness of $f_i(\cdot)$.
    By substituting (\ref{eq:term-v-01}) into (\ref{eq:term-v-00}), we continue to bound $\mathrm{\MakeTextUppercase{\romannumeral 5}}$ as follows
    \begin{align} \label{eq:term-v-02}
      \frac{-\gamma}{1-\alpha} \mathbb{E} \Big[ \Big\langle \nabla \mathcal{F} \Big(\bar{x}^{[t-1]} \Big), \frac{1}{N}\sum^N_{i=1} g^{[t]}_{i,i} \Big\rangle \Big]
      \leq & \frac{-\gamma}{2(1-\alpha)} \mathbb{E} \Big[ \Big\| \nabla \mathcal{F} \Big(\bar{x}^{[t-1]} \Big) \Big\|^2 \Big]   \nonumber\\
      & + \frac{-\gamma}{2(1-\alpha)} \mathbb{E} \Big[ \Big\| \frac{1}{N}\sum^N_{i=1} \nabla f_i \Big(x^{[t-1]}_i \Big) \Big\|^2 \Big]  \nonumber\\
      & - \frac{-\gamma L^2}{2N(1-\alpha)} \sum^N_{i=1} \mathbb{E} \Big[ \Big\| \bar{x}^{[t-1]} - x^{[t-1]}_i \Big\|^2 \Big]
    \end{align}
    By combining the bounds of $\mathrm{\MakeTextUppercase{\romannumeral 3}}$, $\mathrm{\MakeTextUppercase{\romannumeral 4}}$, and $\mathrm{\MakeTextUppercase{\romannumeral 5}}$ (see (\ref{eq:term-iii}), (\ref{eq:term-iv}) and (\ref{eq:term-v-02}), respectively), we obtain the bound of (\ref{eq:bars-bars}), and we thus continue to bound $\mathbb{E} \left[ \mathcal{F} \left(\bar{S}^{[t]} \right) \right]$ according to (\ref{eq:fbars}) as follows
    \begin{align} \label{eq:fbarx}
      \mathbb{E} \Big[ \mathcal{F} \Big(\bar{S}^{[t]} \Big) \Big] 
      \leq & \mathbb{E} \Big[ \mathcal{F} \Big(\bar{S}^{[t-1]} \Big) \Big] + \frac{(1-\alpha)L}{2{\alpha}} \mathbb{E} \Big[ \Big\| \bar{S}^{[t-1]} - \bar{x}^{[t-1]} \Big\|^2 \Big] \nonumber\\
      & + \left(\frac{\alpha L \gamma^2}{2(1-\alpha)^3} + \frac{L\gamma^2}{2(1-\alpha)^2}   \right) \mathbb{E} \Big[ \Big\| \frac{1}{N}\sum^N_{i=1} \bar{g_i}^{[t]} \Big\|^2 \Big]     \nonumber\\
      & + \bigg( \frac{(1-\alpha) \gamma^2}{2\alpha L} - \frac{\gamma}{2(1-\alpha)} \Big) \mathbb{E} \Big[ \Big\| \nabla \mathcal{F} \Big(\bar{x}^{[t-1]} \bigg) \Big\|^2 \Big]   \nonumber\\
      & + \frac{\alpha L}{2(1-\alpha)^3} \mathbb{E} \Big[ \Big\| \frac{1}{N}\sum^N_{i=1} \Big( \bar{g_i}^{[t]} - g^{[t]}_{i,i} \Big) \Big\|^2 \Big]  \nonumber\\
      & - \frac{\gamma}{2(1-\alpha)} \mathbb{E} \Big[ \Big\| \frac{1}{N}\sum^N_{i=1}  \nabla f_i \Big(x^{[t-1]}_i \Big) \Big\|^2 \Big] \nonumber\\
      & + \frac{\gamma L^2}{2(1-\alpha)} \frac{1}{N} \sum^N_{i=1} \mathbb{E} \left[ \left\| \bar{x}^{[t-1]} - x^{[t-1]}_i \right\|^2 \right]  
    \end{align}
    By rearranging (\ref{eq:fbarx}), we have
    \begin{align} \label{eq:fbarx-01}
      \mathbb{E} \Big[ \Big\| \nabla \mathcal{F} \Big( \bar{x}^{[t-1]} \Big) \Big\|^2 \Big] 
      \leq & \frac{1}{m_1} \Big( \mathbb{E} \Big[ \mathcal{F}\Big(\bar{S}^{[t-1]}\Big) \Big] - \mathbb{E} \Big[ \mathcal{F}\Big(\bar{S}^{[t]}\Big) \Big] \Big) \nonumber\\
      & + m_2 \mathbb{E} \Big[ \Big\| \frac{1}{N}\sum^N_{i=1} \bar{g_i}^{[t]} \Big\|^2 \Big] + m_3 \mathbb{E} \Big[ \Big\| \bar{S}^{[t-1]} - \bar{x}^{[t-1]} \Big\|^2 \Big]  \nonumber\\
      & + m_4  \mathbb{E} \Big[ \Big\| \frac{1}{N}\sum^N_{i=1} \Big( \bar{g_i}^{[t]} - g^{[t]}_{i,i} \Big) \Big\|^2 \Big]  \nonumber\\
      & + \frac{m_5}{N} \sum^N_{i=1} \mathbb{E} \Big[ \Big\| \bar{x}^{[t-1]} - x^{[t-1]}_i \Big\|^2  \Big] \nonumber\\
      & - m_6 \mathbb{E} \Big[ \Big\| \frac{1}{N}\sum^N_{i=1} \nabla f_i \Big( x^{[t-1]}_i \Big) \Big\|^2 \Big] 
    \end{align}
    where $m_1, \cdots, m_5$ are defined in (\ref{eq:main-constants}) and $m_6 = \frac{\gamma}{2 m_1 (1-\alpha)}$.
    Summing over $t \in \{1, \cdots, T\}$, we have
    \begin{align} \label{eq:fbarx-003}
      \sum^{T}_{t=1} \mathbb{E} \Big[ \Big\| \nabla \mathcal{F}(\bar{x}^{[t-1]}) \Big\|^2 \Big] 
      \leq& \frac{1}{m_1} \Big( \mathbb{E} \Big[ \mathcal{F} \Big( \bar{S}^{[0]} \Big) \Big] - \mathbb{E} \Big[ \mathcal{F}(\bar{S}^{[T]}) \Big] \Big) \nonumber\\
      & + m_2 \sum^{T}_{t=1} \mathbb{E} \Big[ \Big\| \frac{1}{N}\sum^N_{i=1} \bar{g_i}^{[t]} \Big\|^2 \Big] \nonumber\\
      & + m_3 \sum^{T}_{t=1} \mathbb{E} \Big[ \Big\| \bar{S}^{[t-1]} - \bar{x}^{[t-1]} \Big\|^2 \Big] \nonumber\\
      & + m_4 \sum^{T}_{t=1} \mathbb{E} \Big[ \Big\| \frac{1}{N}\sum^N_{i=1} \Big( \bar{g}_i^{[t]} - g^{[t]}_{i,i} \Big) \Big\|^2 \Big] \nonumber\\
      & + m_5 \sum^{T}_{t=1} \frac{1}{N} \sum^N_{i=1} \mathbb{E} \Big[ \Big\| \bar{x}^{[t-1]} - x^{[t-1]}_i \Big\|^2  \Big] \nonumber\\
      & - m_6 \sum^{T}_{t=1} \mathbb{E} \Big[ \Big\| \frac{1}{N}\sum^N_{i=1} \nabla f_i \Big( x^{[t-1]}_i \Big) \Big\|^2 \Big]
    \end{align}
    By combining $\textbf{Lemma}~\ref{lem:diffhatgrad}$, $\textbf{Lemma}~\ref{lem:sandx}$, $\textbf{Lemma}~\ref{lem:bdavggrad}$ and $\textbf{Lemma}~\ref{lem:barx-x}$ in (\ref{eq:fbarx-003}), we have:
    \begin{align} \label{eq:sumfbarx}
      &\sum^{T}_{t=1} \mathbb{E} \Big[ \Big\| \nabla \mathcal{F}(\bar{x}^{[t-1]}) \Big\|^2 \Big]   \nonumber\\
      \leq & \frac{1}{m_1} \Big( \mathbb{E} \Big[ \mathcal{F} \Big( \bar{S}^{[0]} \Big) \Big] - \mathbb{E} \Big[ \mathcal{F} \Big( \bar{S}^{[T]} \Big) \Big] \Big)  \nonumber\\
      & + T \bigg( m_2 + \frac{ m_3 \gamma^2 \alpha^2}{(1-\alpha)^4} + m_4 \bigg) \left( \frac{\left(4 \sigma^2 + 16 \varsigma^2\right)\hat{\varphi}^2_{\max} }{\omega^4_{\min}} + \frac{6 \sigma^2}{N}\right)  \nonumber\\
      & + T \frac{8 \gamma^2 \left( \hat{\varphi}^2_{\max} + 2\omega^4_{\min} \right) \left( \sigma^2 + 4 \varsigma^2 \right)}{\omega^4_{\min} (1-\alpha)^2 \left( 1-\sqrt{\rho} \right)^2} \left( \frac{32L^2 \hat{\varphi}^2_{\max}}{\omega^4_{\min}} \left( m_2 + \frac{ m_3 \gamma^2 \alpha^2}{(1-\alpha)^4} \right) + \frac{16m_4 L^2 \hat{\varphi}^2_{\max}}{\omega^4_{\min}} + m_5 \right)  \nonumber\\
      & - \left( m_6 - \bigg( \frac{32\hat{\varphi}^2_{\max}}{\omega^4_{\min}} + 6 \bigg) \bigg( m_2 + \frac{m_3 \gamma^2 \alpha^2}{(1-\alpha)^4} \bigg) - \frac{m_5}{2L^2} - \left( \frac{16\hat{\varphi}^2_{\max}}{\omega^4_{\min}}+2 \right) m_4 \right) \sum^{T}_{t=1} \mathbb{E} \Big[ \Big\| \frac{1}{N}\sum^N_{i=1} \nabla f_i \Big( x^{[t-1]}_i \Big) \Big\|^2 \Big] 
    \end{align}
    Dividing both sides of (\ref{eq:sumfbarx}) by $T$ and employing the fact that $\bar{S}^{[0]} = \bar{x}^{[0]}$, we have
    \begin{align} \label{eq:finalbound}
      & \frac{1}{T} \sum^{T}_{t=1} \mathbb{E} \Big[ \Big\| \nabla \mathcal{F} \Big( \bar{x}^{[t-1]} \Big) \Big\|^2 \Big] \nonumber\\
      \leq & \frac{1}{m_1 T} \Big( \mathcal{F} \Big( \bar{x}^{[0]} \Big) -  \mathcal{F}^* \Big) + \bigg( m_2 + \frac{ m_3 \gamma^2 \alpha^2}{(1-\alpha)^4} + m_4 \bigg) \left( \frac{\left(4 \sigma^2 + 16 \varsigma^2\right)\hat{\varphi}^2_{\max} }{\omega^4_{\min}} + \frac{6 \sigma^2}{N}\right) \nonumber\\
      & + \frac{8 \gamma^2 \left( \hat{\varphi}^2_{\max} + 2\omega^4_{\min} \right) \left( \sigma^2 + 4 \varsigma^2 \right)}{\omega^4_{\min} (1-\alpha)^2 \left( 1-\sqrt{\rho} \right)^2} \left( \frac{32L^2 \hat{\varphi}^2_{\max}}{\omega^4_{\min}} \left( m_2 + \frac{ m_3 \gamma^2 \alpha^2}{(1-\alpha)^4} \right) + \frac{16m_4 L^2 \hat{\varphi}^2_{\max}}{\omega^4_{\min}} + m_5 \right)  
    \end{align}
    when 
    \begin{align} \label{eq:condition-gamma}
     m_6 - \left( \frac{32\hat{\varphi}^2_{\max}}{\omega^4_{\min}}+6 \right) \left( m_2 +  \frac{m_3 \gamma^2 \alpha^2}{(1-\alpha)^4} \right) - \frac{m_5}{2L^2} - \left( \frac{16\hat{\varphi}^2_{\max}}{\omega^4_{\min}}+2 \right) m_4 \geq 0
    \end{align}
    To ensure the inequality (\ref{eq:condition-gamma}), we need to have
    \begin{align}
      m_1 = \frac{\gamma}{2(1-\alpha)} - \frac{(1-\alpha) \gamma^2}{2\alpha L} > 0 
    \end{align}
    and
    \begin{align}
      \frac{\gamma}{2} - \left( \frac{32\hat{\varphi}^2_{\max}}{\omega^4_{\min}}+6 \right) \left( \frac{2 L\gamma^2}{(1-\alpha)^2} + \frac{L\alpha}{(1-\alpha)^2} \right) \geq 0 
    \end{align}
    hold; therefore, the step size $\gamma$ is supposed to satisfy
    \begin{align}
      \gamma \leq \min \left\{ \begin{aligned} & \frac{\alpha L}{(1-\alpha)^2}, ~~ \frac{(1-\alpha)(1-\sqrt{\rho})}{8L\sqrt{\frac{\hat{\varphi}^2_{\max}}{\omega^4_{\min}} + 2}}\\
      & \frac{(1-\alpha)^2 + \sqrt{(1-\alpha)^4 - 8L^2\alpha \left(12+\frac{64\hat{\varphi}^2_{\max}}{\omega^4_{\min}}\right)^2}}{16L\left(3+\frac{16\hat{\varphi}^2_{\max}}{\omega^4_{\min}}\right)} \end{aligned} \right\}
    \end{align}

    We continue to present the proof of \textbf{Corollary}~\ref{cor:main}. If letting $\gamma = \mathcal{O}\Big(\sqrt{\frac{N}{T}}\Big)$ and $\sigma = \varsigma = \mathcal{O}\Big(\frac{1}{\sqrt{NT}}\Big)$, we have the constants in (\ref{eq:main-constants}) represented by
    \begin{align}
      m_1 = \mathcal{O}\Big(\sqrt{\frac{N}{T}}\Big),~
      m_2 = \mathcal{O}\Big(\sqrt{\frac{N}{T}}\Big),~ 
      m_3 = \mathcal{O}\Big(\sqrt{\frac{T}{N}}\Big),
      m_4 = \mathcal{O}\Big(\sqrt{\frac{T}{N}}\Big),~ 
      m_5 = \mathcal{O}\Big(1\Big),~
      \text{and}~m_6 = \mathcal{O}\Big(1\Big)
    \end{align}
    Then, as $\hat{\varphi}_{\max} \leq 1$, for each term in the right hand side of (\ref{eq:finalbound}), we have
    \begin{align}
      \frac{1}{m_1 T} \Big( \mathcal{F} \Big( \bar{x}^{[0]} \Big) -  \mathcal{F}^* \Big) = \mathcal{O}\Big(\frac{1}{\sqrt{NT}}\Big),
    \end{align}
    \begin{align}
      \bigg( m_2 + \frac{ m_3 \gamma^2 \alpha^2}{(1-\alpha)^4} + m_4 \bigg) \left( \frac{\left(4 \sigma^2 + 16 \varsigma^2\right)\hat{\varphi}^2_{\max} }{\omega^4_{\min}} + \frac{6 \sigma^2}{N}\right) 
      = \mathcal{O} \left( \frac{1}{\sqrt{NT^3}} + \frac{1}{\sqrt{N^3T^3}} + \frac{1}{\sqrt{N^3T}} + \frac{1}{\sqrt{N^5T}} \right) 
    \end{align}
    and
    \begin{align}
      \frac{8 \gamma^2 \left( \hat{\varphi}^2_{\max} + 2\omega^4_{\min} \right) \left( \sigma^2 + 4 \varsigma^2 \right)}{\omega^4_{\min} (1-\alpha)^2 \left( 1-\sqrt{\rho} \right)^2} 
      \left( \frac{32L^2 \hat{\varphi}^2_{\max}}{\omega^4_{\min}} \left( m_2 + \frac{ m_3 \gamma^2 \alpha^2}{(1-\alpha)^4} \right) + \frac{16m_4 L^2 \hat{\varphi}^2_{\max}}{\omega^4_{\min}} + m_5 \right) 
      = \mathcal{O} \left( \frac{\sqrt{N}}{\sqrt{T^5}} + \frac{1}{T^2} + \frac{1}{\sqrt{NT^3}} \right)
    \end{align}
    by combining which, we conclude that there exists constant $M$ such that
    \begin{align}
      \frac{1}{T} \sum^{T}_{t=1} \mathbb{E} \left[ \left\| \nabla \mathcal{F} \left( \bar{x}^{[t-1]} \right) \right\|^2 \right]
      \leq M \left( \frac{1}{\sqrt{NT}} + \frac{1}{\sqrt{N^3T}} + \frac{1}{\sqrt{NT^3}} + \frac{\sqrt{N}}{\sqrt{T^5}} +   \frac{1}{\sqrt{N^5T}} + \frac{1}{\sqrt{N^3T^3}} + \frac{1}{T^2}  \right) 
    \end{align}
    If we fix $N$ and let $T$ be sufficiently large such that $\gamma = \mathcal{O} \left( \sqrt{\frac{N}{T}} \right)$ (and thus (\ref{eq:cormain-t}) holds), the convergence rate of \textbf{Algorithm}~\ref{alg:ross} can be represented by $\mathcal{O}\left(\frac{1}{\sqrt{NT}}\right)$.

\section{Experiments} \label{sec:exp}
   In this section, we experimentally analyze the performance of our ROSS algorithm. In Sec.~\ref{ssec:setup}, we briefly introduce the two datasets used in our evaluation and the setup of our experiments. In Sec.~\ref{ssec:results}, we present and thoroughly discuss the experiment results.

  \subsection{Datasets and Experiment Setup} \label{ssec:setup}
    We conduct extensive experiments on MNIST and CIFAR-10 datasets. The MNIST dataset consists of $70,000$ images of handwritten digits in ten classes, the training dataset has $60,000$ images, and the test dataset has $10,000$ images. The CIFAR-10 dataset consists of ten classes of images, including airplane, automobile, bird, cat, deer, dog, frog, horse, ship, and truck, $50,000$ of the images are used for training and the rest $10,000$ ones are adopted for testing. For each dataset, we construct a validation set by uniformly sampling $2,000$ images ($20\%$) at random from the corresponding test dataset.

    In our experiments, we adopt \textit{Convolutional Neural Network} (CNN) models to evaluate our ROSS algorithm, considering they are widely used in many applications (e.g., image classification etc.). Specifically, for MNIST dataset, the CNN model is composed of two $3 \times 3$ convolution layers, each followed by a $2 \times 2$ max pooling module, then a fully connected layer, and ReLU activation is utilized in the convolutional layers. The CNN model for CIFAR-10 dataset consists of two $5 \times 5$ convolution layers, each followed by a $2 \times 2$ max pooling module, then two fully connected layers. ReLU activation is utilized again in the convolutional layers. We set the mini-batch size to be $260$, and let $\alpha=0.5, 0.7$ and $\gamma=0.001, 0.01$ for the above two datasets, respectively. In addition, we let $R=10$ for our Monte Carlo-based Shapley value estimation algorithm. Note that we still have the performance of ROSS ensured if choosing different values for these hyperparameters.

    We consider the following three types of graphs: fully connected, bipartite, and ring, with the latter two being sparser. For the data setting, we first use the Dirichlet distribution to simulate non-IID data across different agents~\cite{YurochkinAGGHK-ICML19,LinKSJ-ICML21}. To better reflect real-world data challenges, we further consider the following four scenarios based on the non-IID setting: long-tailed distribution~\cite{CaoWGAM-NIPS19}, data noise~\cite{SunLZXLLQR-KDD23}, label noise~\cite{TolpeginTGL-ESORICS20}, and gradient poisoning~\cite{BaruchBG-NIPS19}. Under these conditions, we compare our ROSS algorithm against seven state-of-the-art baselines: DMSGD~\cite{YuJY-ICML19}, CGA~\cite{EsfandiariTJBHHS-ICML21}, NET-FLEET~\cite{ZhangFLYLZ-MobiHoc22}, MEDIAN~\cite{YinCKB-ICML18}, TRIM-MEAN~\cite{YinCKB-ICML18}, LEARN~\cite{El-MhamdiFGGHR-NIPS21}, and BALANCE~\cite{FangZHKLLLG-CCS24}. Due to space limitations, we present only the experiment results on fully connected and bipartite graphs. More details on the experimental setup and the results for ring graphs are provided in the supplementary material.

  \subsection{Experiment Results}  \label{ssec:results}
    \subsubsection{Experiment Results on MNIST Dataset}  \label{ssec:res-mnist}
      \begin{figure*}[htb!]
      \begin{center}
        \parbox{.24\textwidth}{\center\includegraphics[width=.24\textwidth]{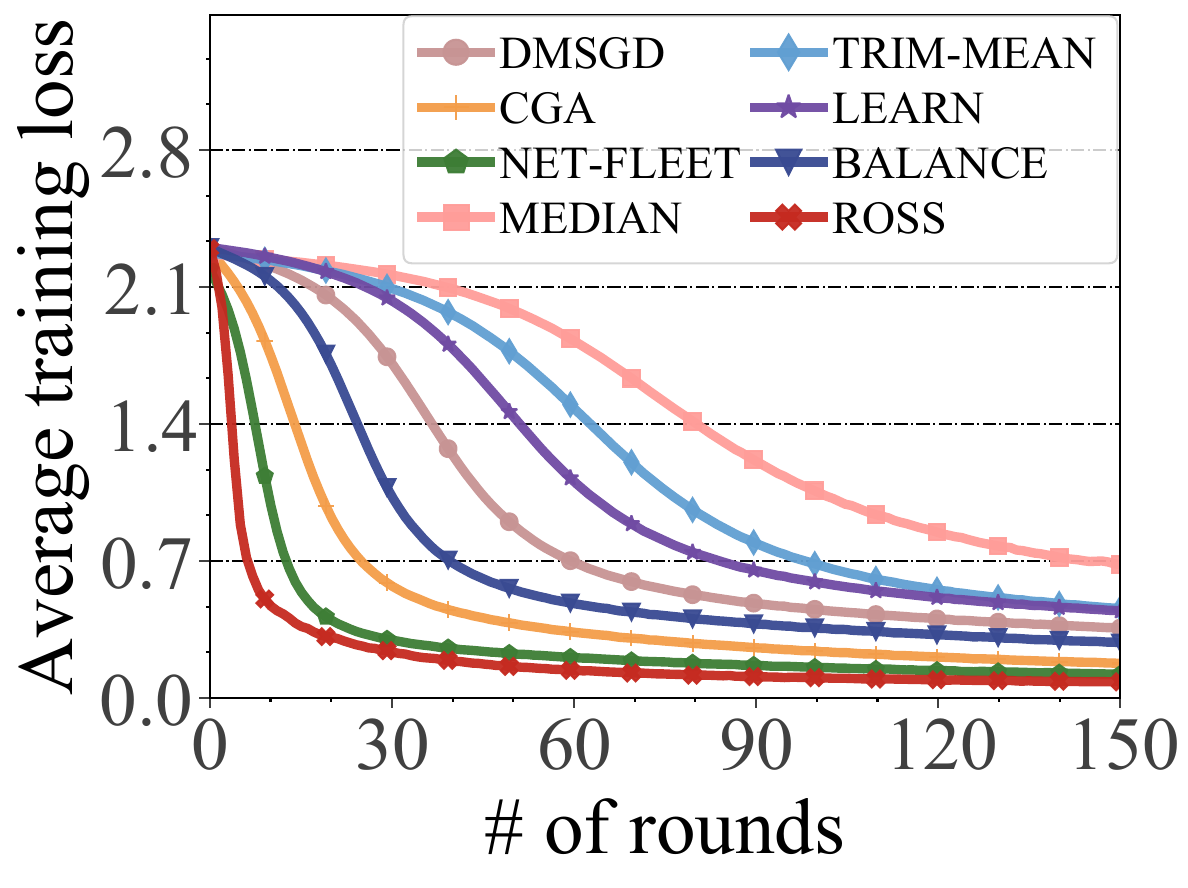}}
        \parbox{.24\textwidth}{\center\includegraphics[width=.24\textwidth]{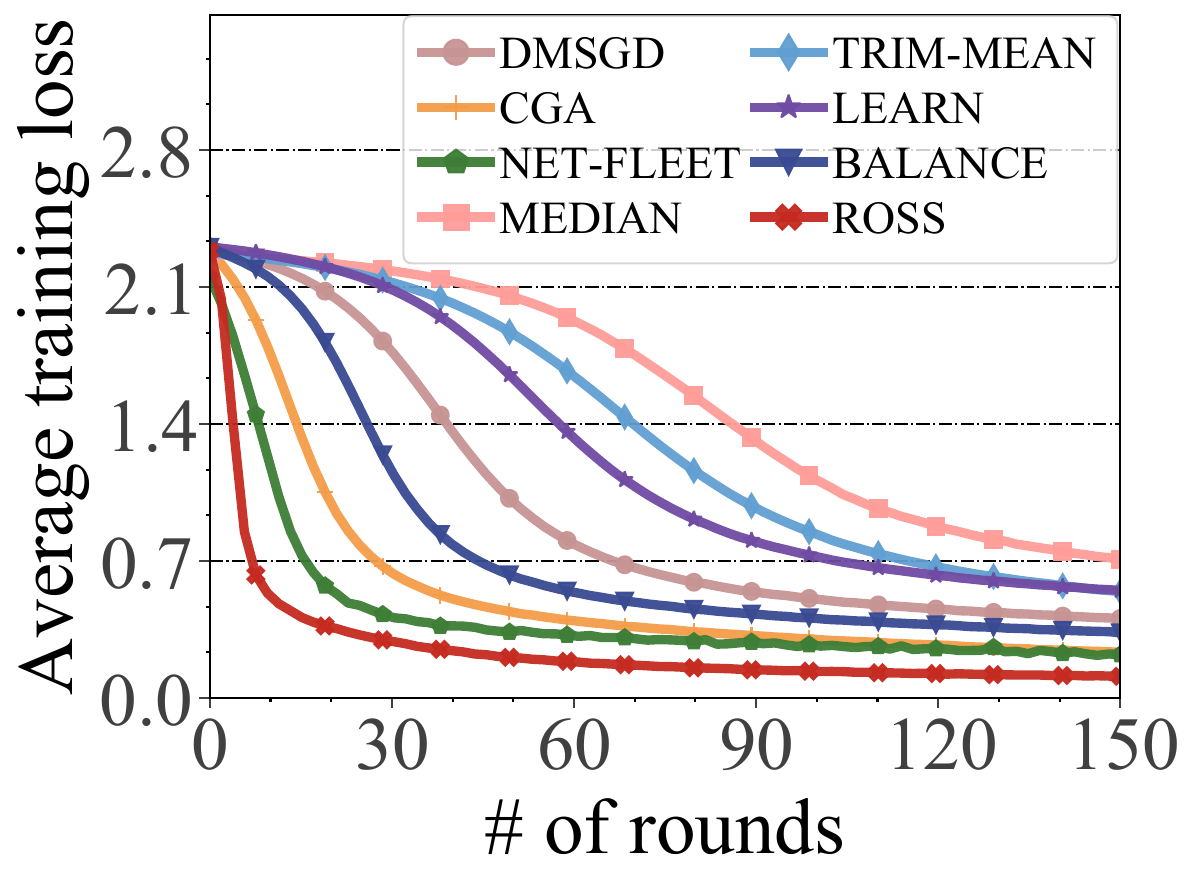}}
        \parbox{.24\textwidth}{\center\includegraphics[width=.24\textwidth]{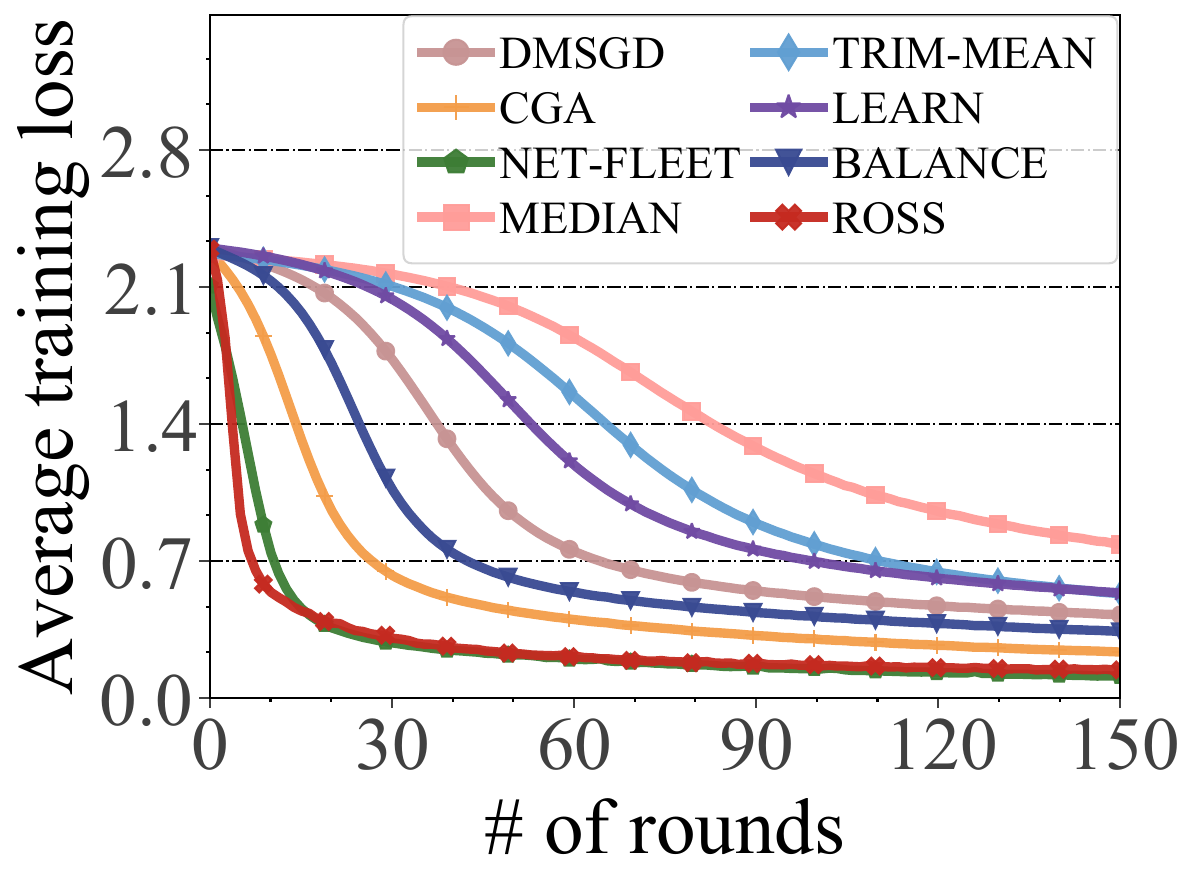}}
        \parbox{.24\textwidth}{\center\includegraphics[width=.24\textwidth]{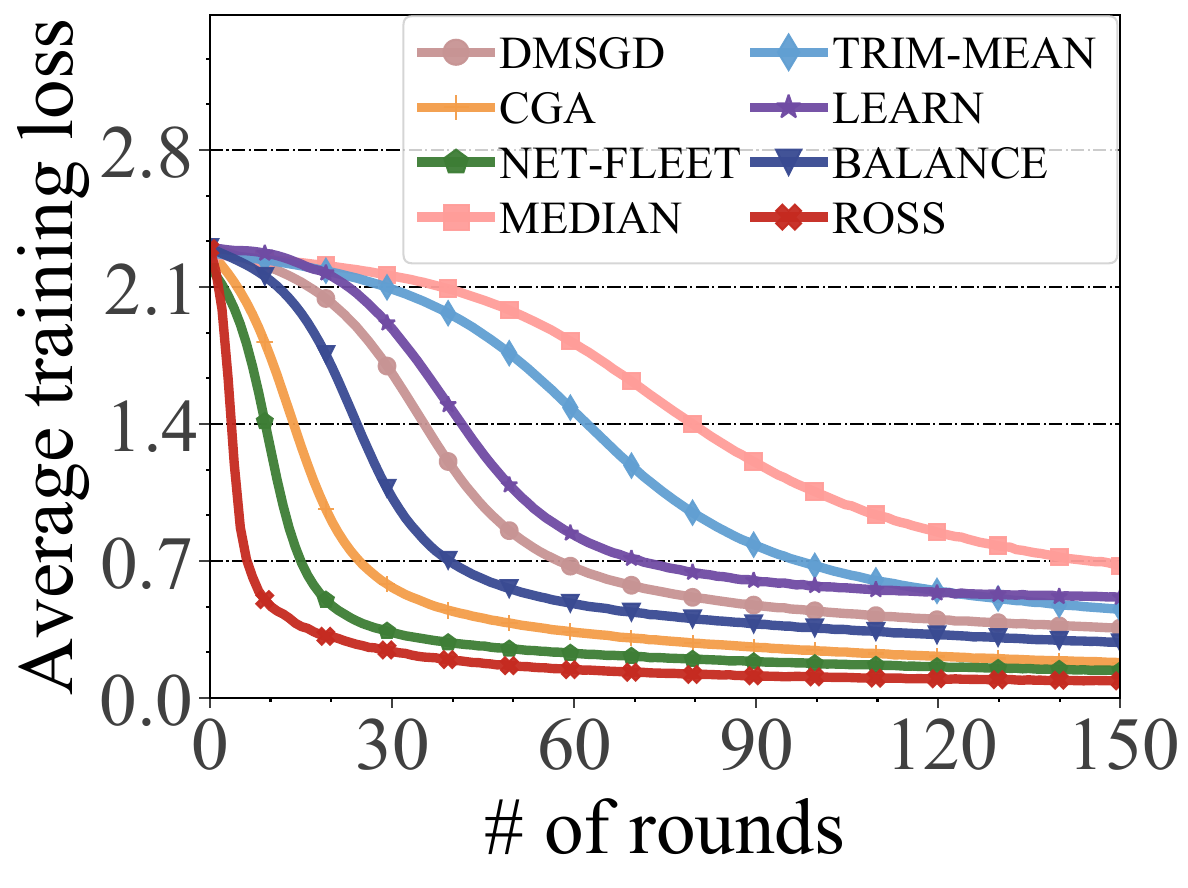}}
        \parbox{.25\textwidth}{\center\scriptsize(a1) Long-tailed ($N=10$)}
        \parbox{.23\textwidth}{\center\scriptsize(a2) Data noise ($N=10$)}
        \parbox{.23\textwidth}{\center\scriptsize(a3) Label noise ($N=10$)}
        \parbox{.25\textwidth}{\center\scriptsize(a4) Gradient poisoning ($N=10$)}
        \parbox{.24\textwidth}{\center\includegraphics[width=.24\textwidth]{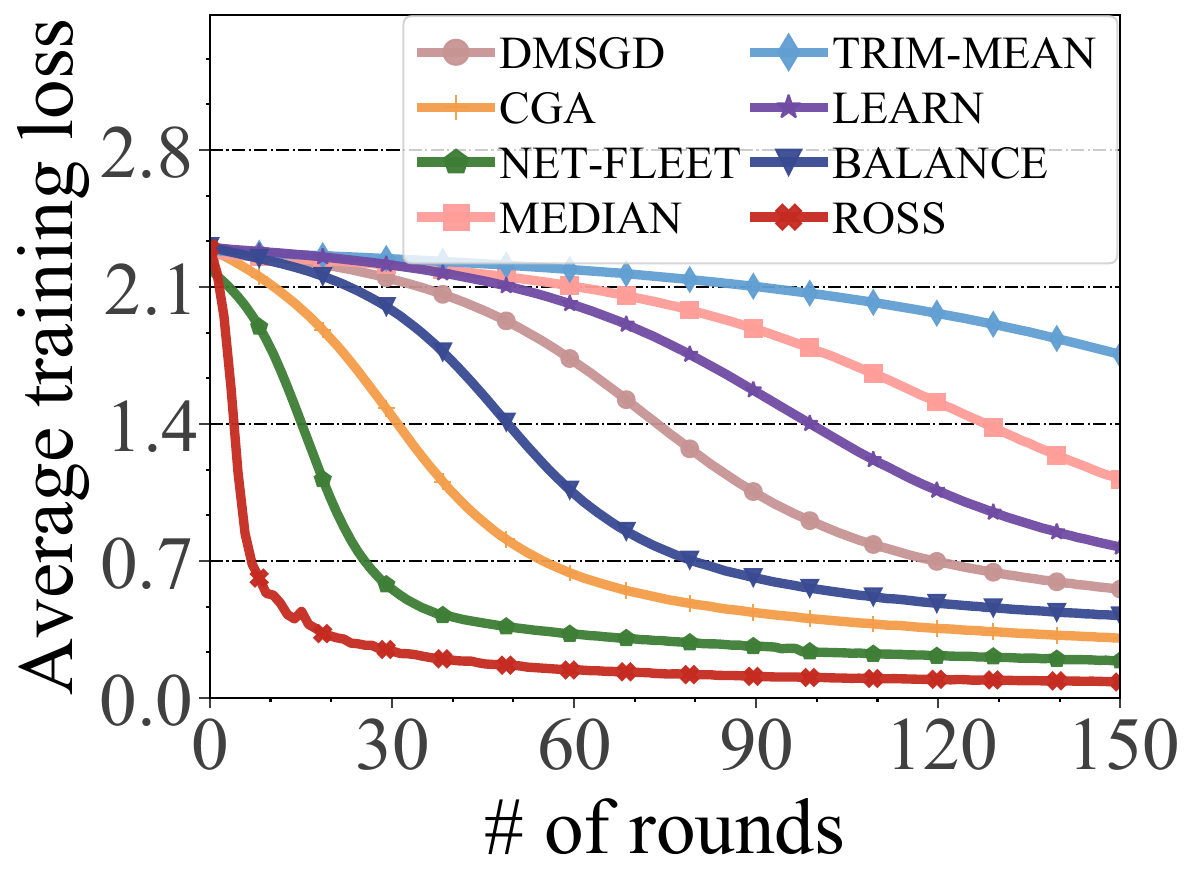}}
        \parbox{.24\textwidth}{\center\includegraphics[width=.24\textwidth]{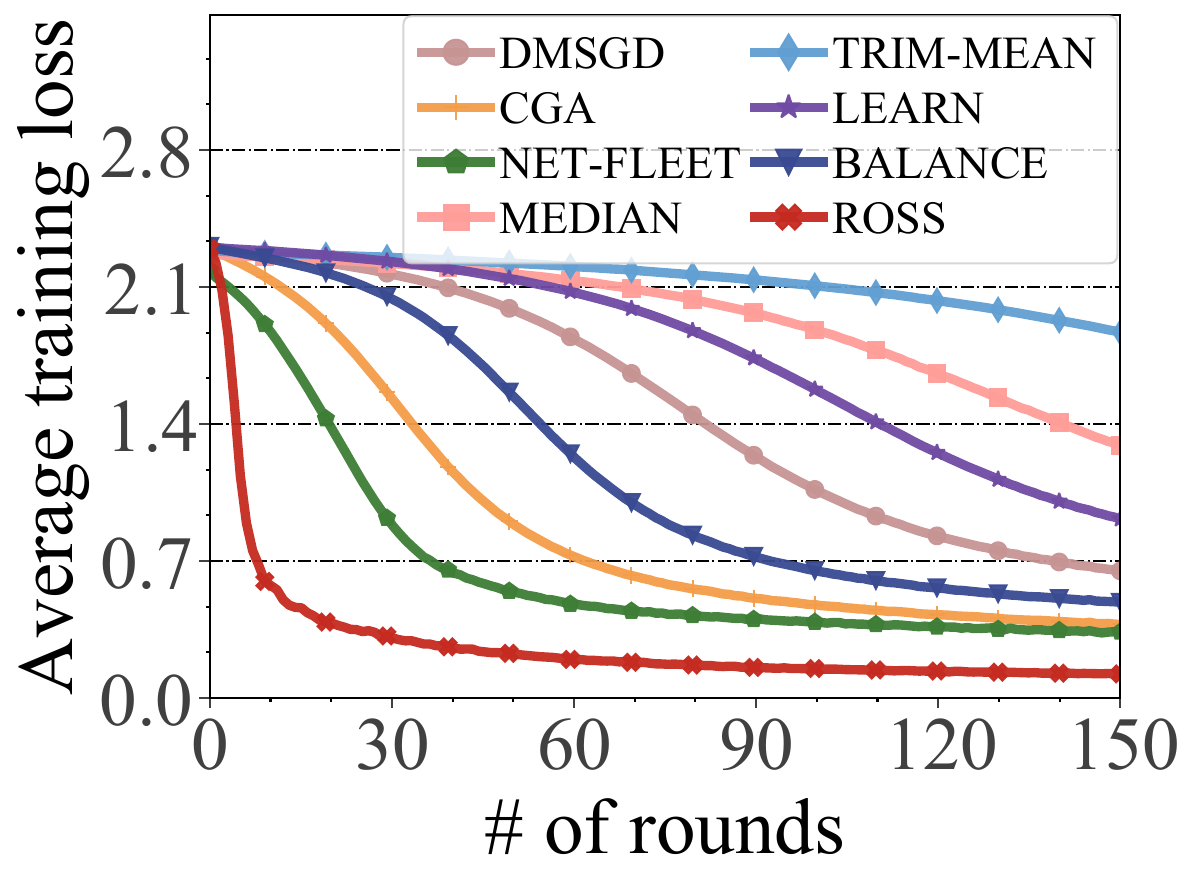}}
        \parbox{.24\textwidth}{\center\includegraphics[width=.24\textwidth]{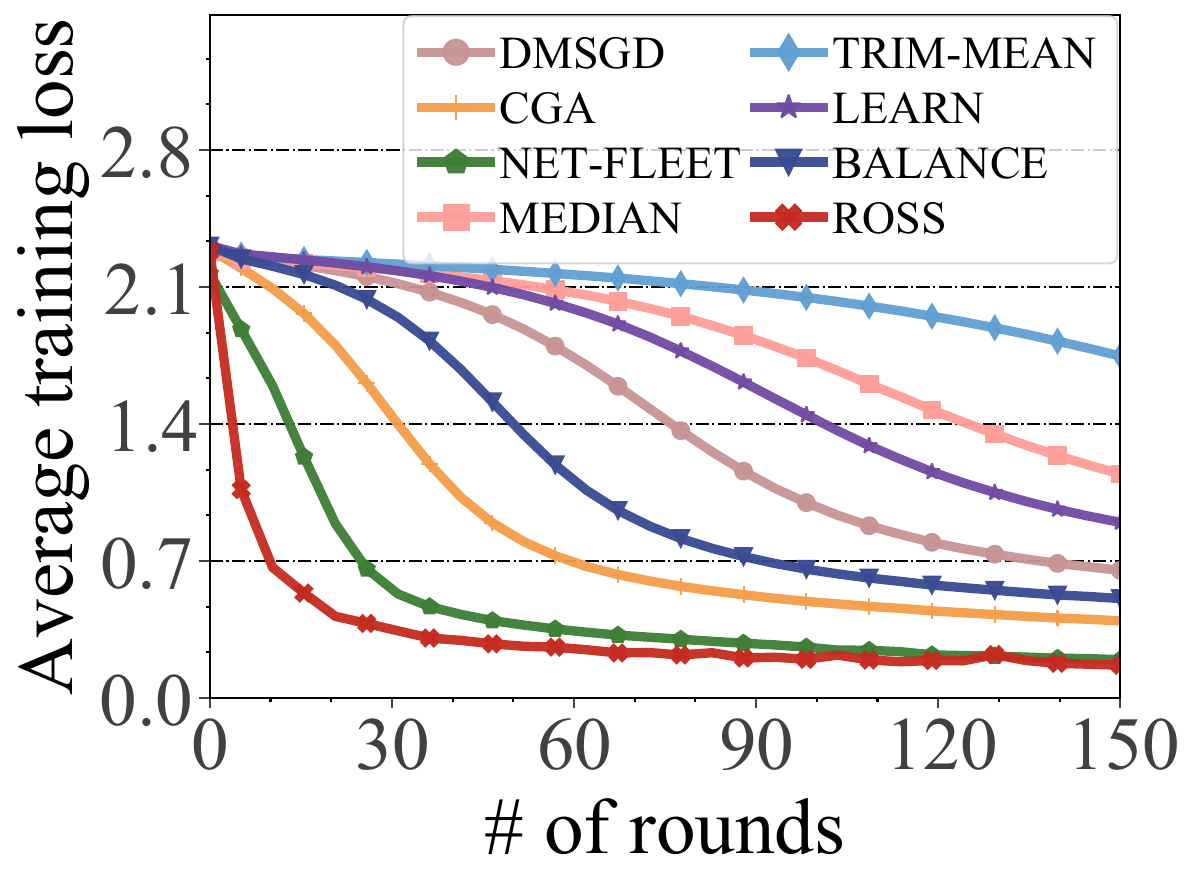}}
        \parbox{.24\textwidth}{\center\includegraphics[width=.24\textwidth]{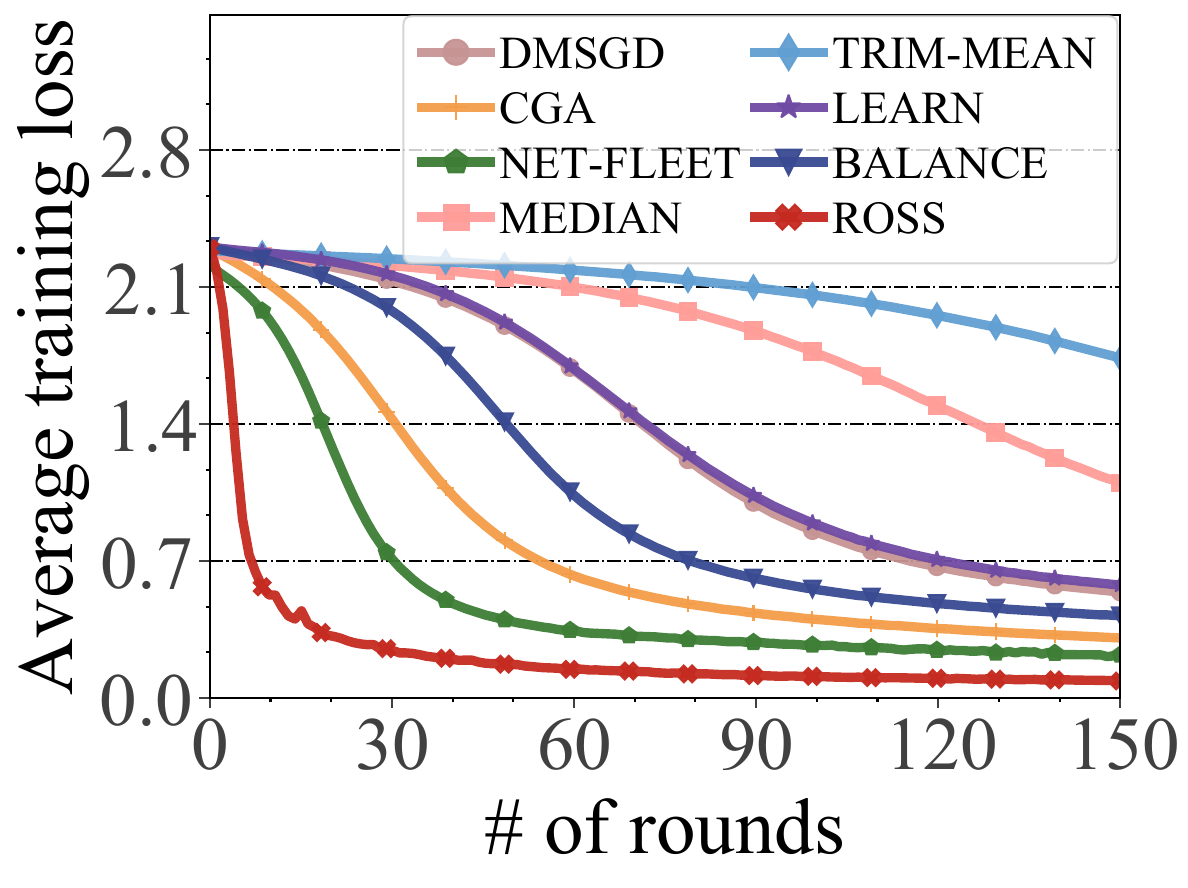}}
        \parbox{.25\textwidth}{\center\scriptsize(b1) Long-tailed ($N=20$)}
        \parbox{.23\textwidth}{\center\scriptsize(b2) Data noise ($N=20$)}
        \parbox{.23\textwidth}{\center\scriptsize(b3) Label noise ($N=20$)}
        \parbox{.25\textwidth}{\center\scriptsize(b4) Gradient poisoning ($N=20$)}
        \parbox{.24\textwidth}{\center\includegraphics[width=.24\textwidth]{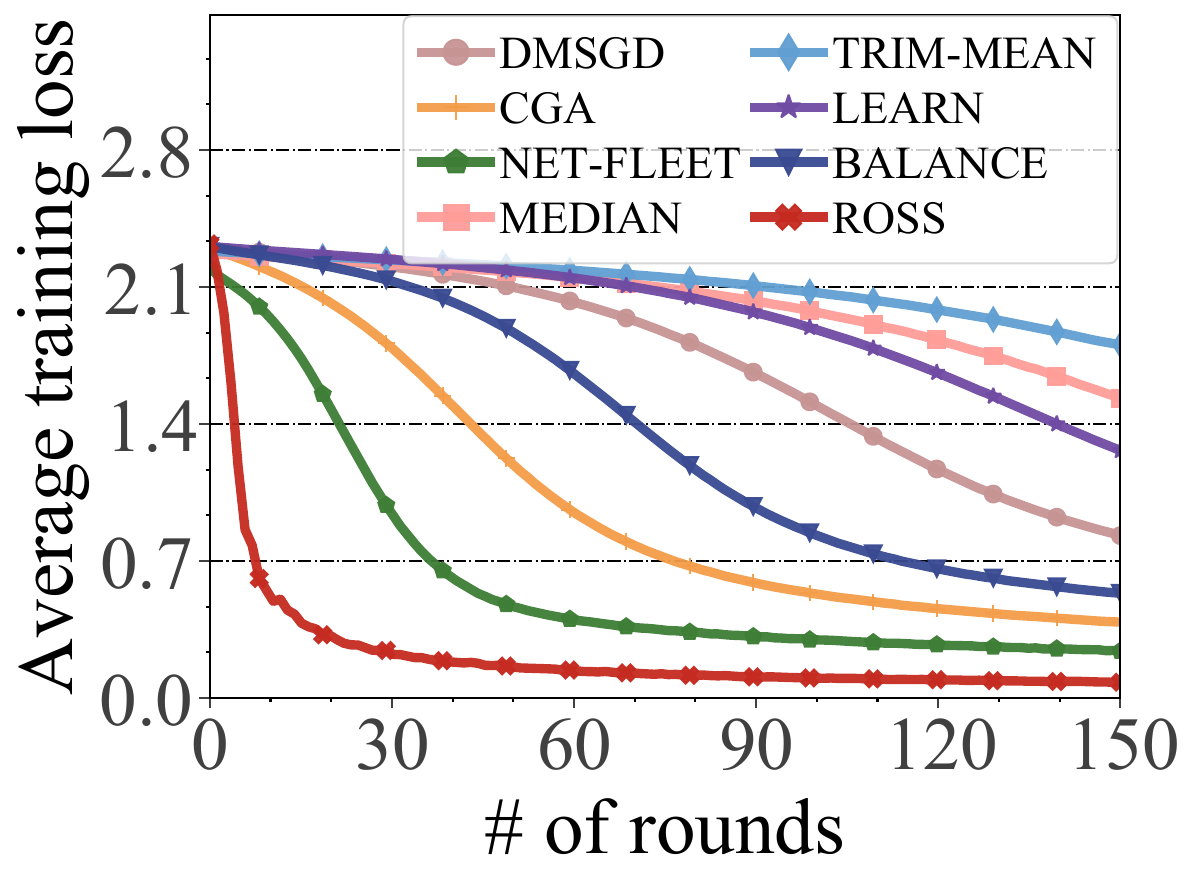}}
        \parbox{.24\textwidth}{\center\includegraphics[width=.24\textwidth]{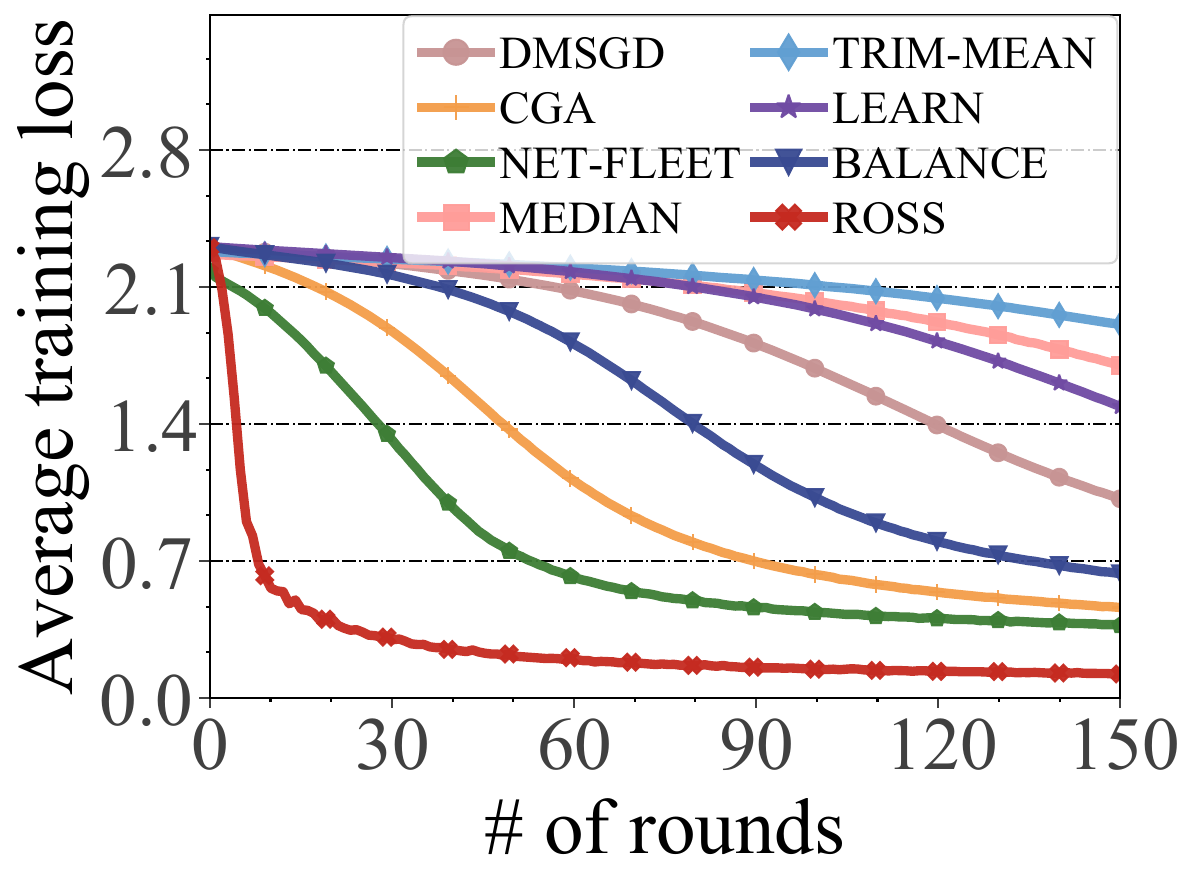}}
        \parbox{.24\textwidth}{\center\includegraphics[width=.24\textwidth]{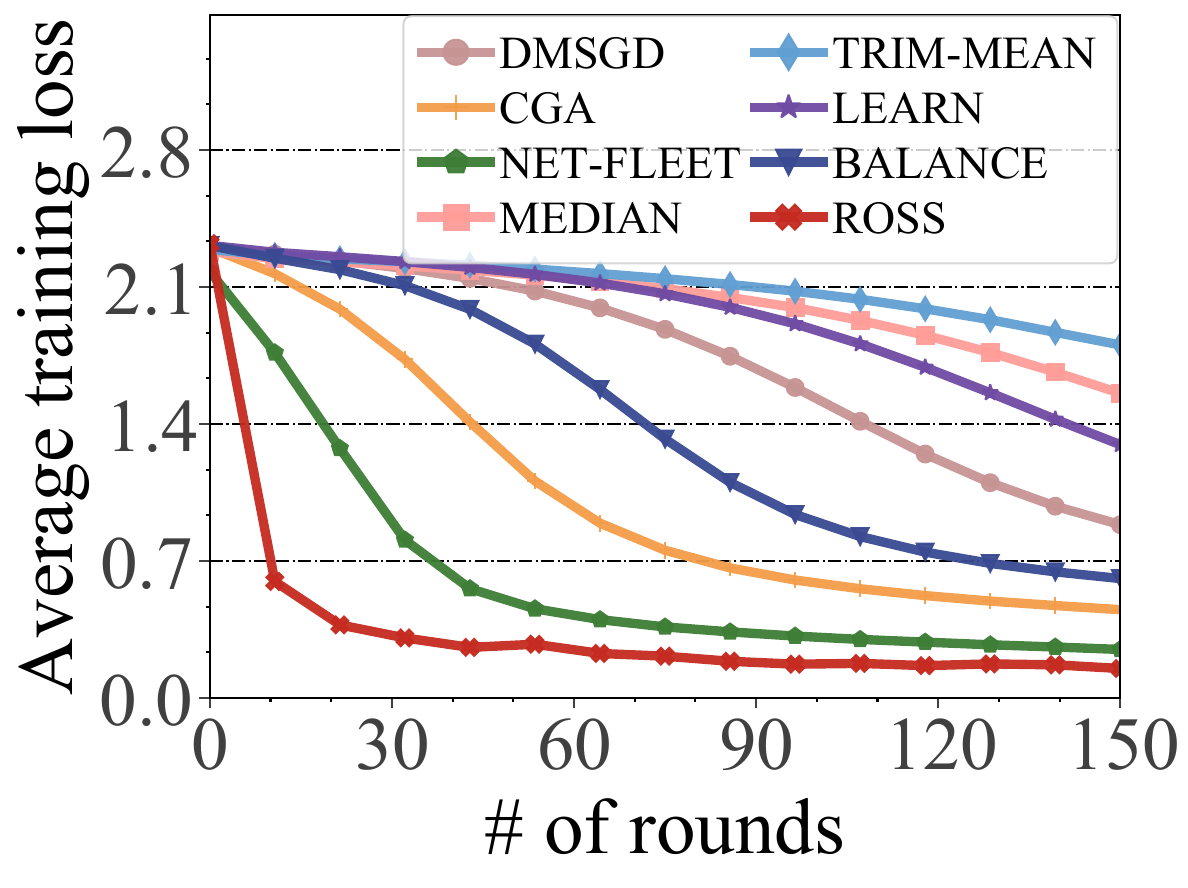}}
        \parbox{.24\textwidth}{\center\includegraphics[width=.24\textwidth]{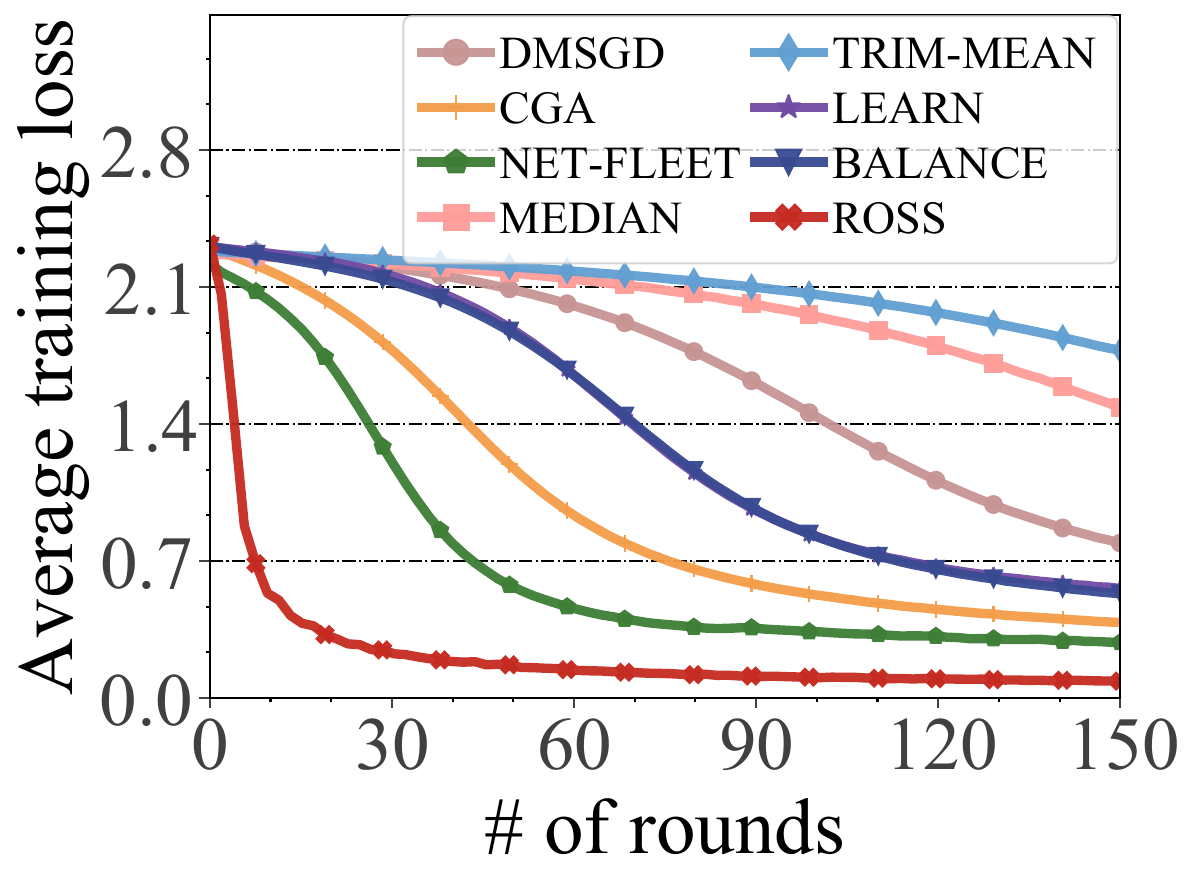}}
        \parbox{.25\textwidth}{\center\scriptsize(c1) Long-tailed ($N=30$)}
        \parbox{.23\textwidth}{\center\scriptsize(c2) Data noise ($N=30$)}
        \parbox{.23\textwidth}{\center\scriptsize(c3) Label noise ($N=30$)}
        \parbox{.25\textwidth}{\center\scriptsize(c4) Gradient poisoning ($N=30$)}
      \caption{Comparison results on MNIST dataset over fully connected graphs.}
      \label{fig:mnist-loss-full}
      \end{center}
      \end{figure*}

      We first let $N=10, 20, 30$ and adopt the fully connected communication graph. The results in Fig.~\ref{fig:mnist-loss-full} show that ROSS achieves a much faster convergence rate and lower average loss than the seven baseline algorithms across all settings. Specifically, the average loss of ROSS approaches its minimum within about $60$ rounds. Even as the number of agents increases, ROSS consistently converges within the same time horizon, whereas the baselines require substantially more communication rounds. For example, with $N=20,30$, CGA converges in roughly $120$ rounds and NET-FLEET in about $90$ rounds, while DMSGD, BALANCE, MEDIAN, TRIM-MEAN, and LEARN fail to converge even after $150$ rounds. Moreover, when $N=10$, the average loss of ROSS is about $0.1$ at convergence; it is $2\times$ smaller than CGA, $3\times$ smaller than BALANCE, $4\times$ smaller than DMSGD, $5\times$ smaller than LEARN and TRIM-MEAN, and $7\times$ smaller than MEDIAN. Another noteworthy observation is that ROSS maintains nearly the same average loss as the number of agents increases (e.g., $N=20,30$), while the baselines experience drastic degradation. For example, under data noise with $N=30$, ROSS achieves a average loss at convergence that is $3–17\times$ smaller than that of the baselines.

\begin{figure*}[htb!]
      \begin{center}
        \parbox{.24\textwidth}{\center\includegraphics[width=.24\textwidth]{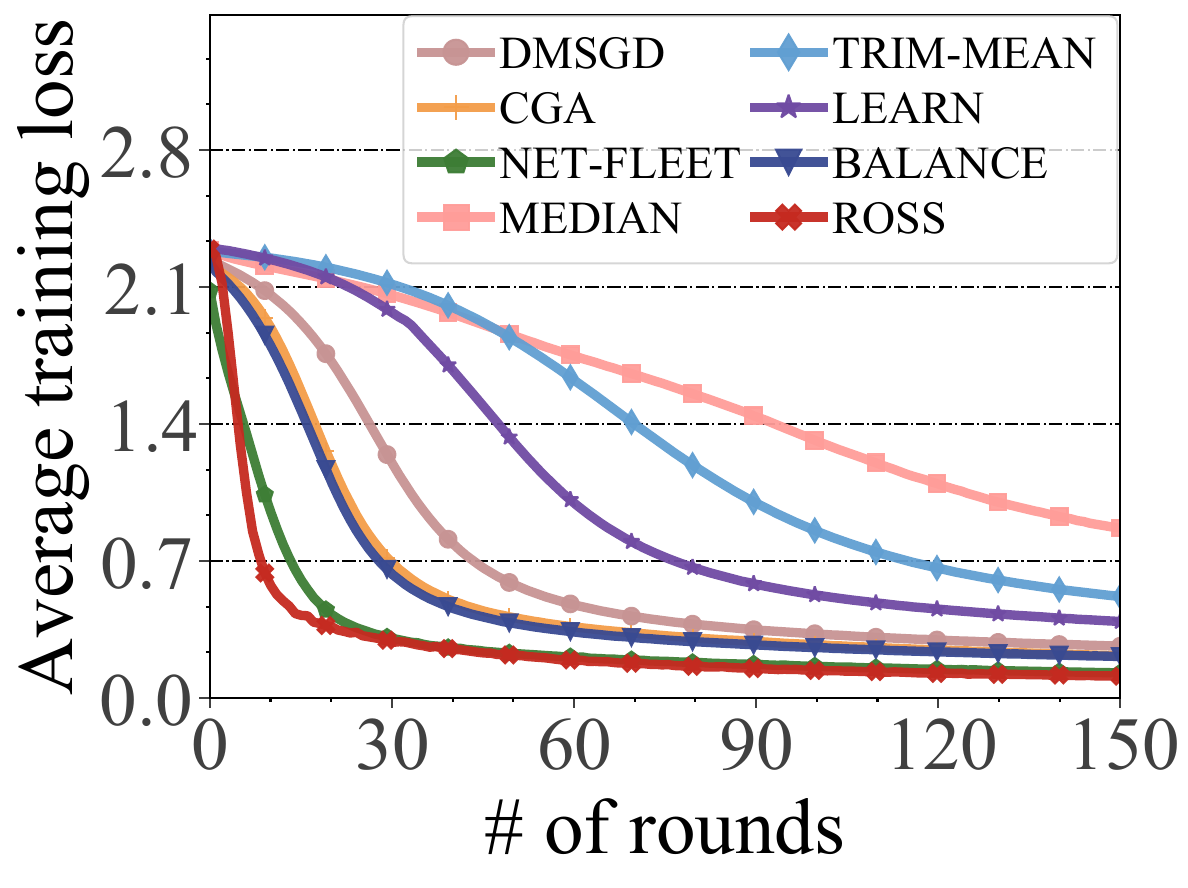}}
        \parbox{.24\textwidth}{\center\includegraphics[width=.24\textwidth]{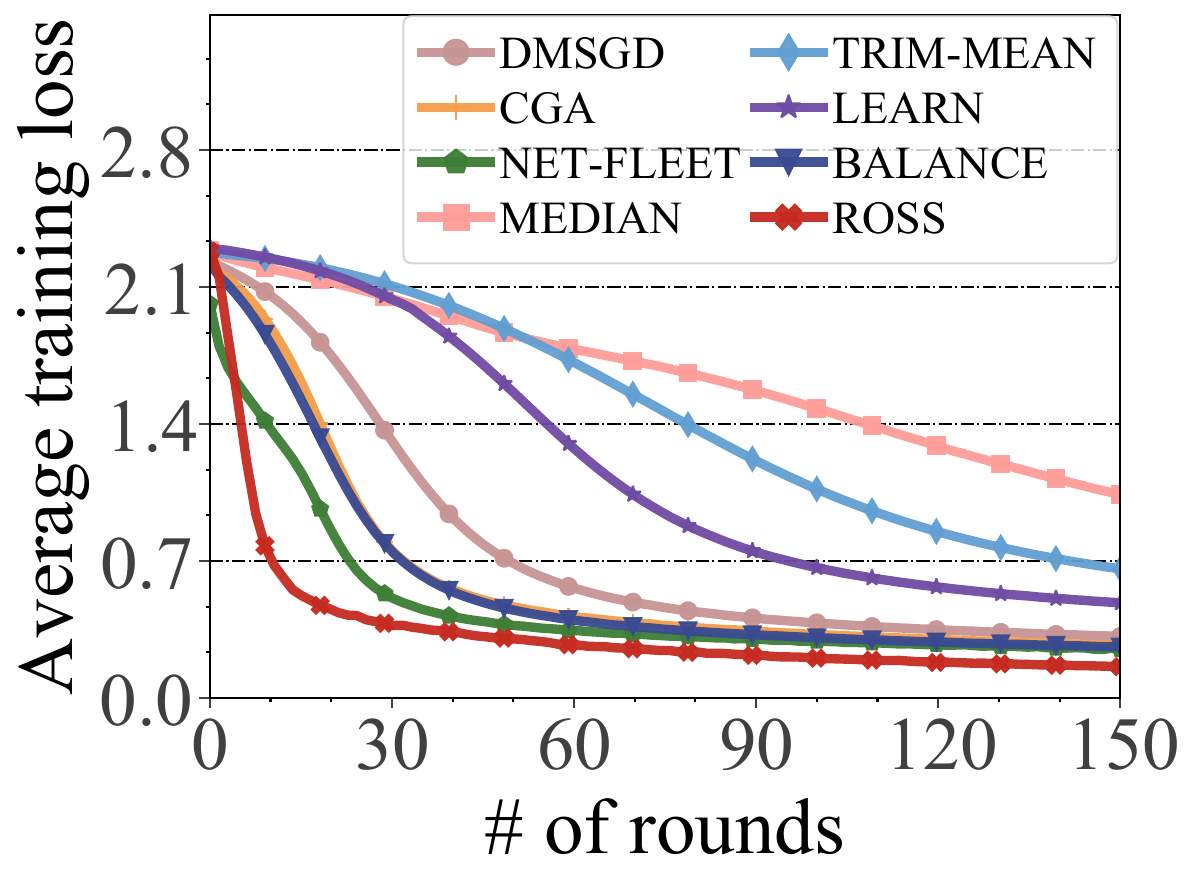}}
        \parbox{.24\textwidth}{\center\includegraphics[width=.24\textwidth]{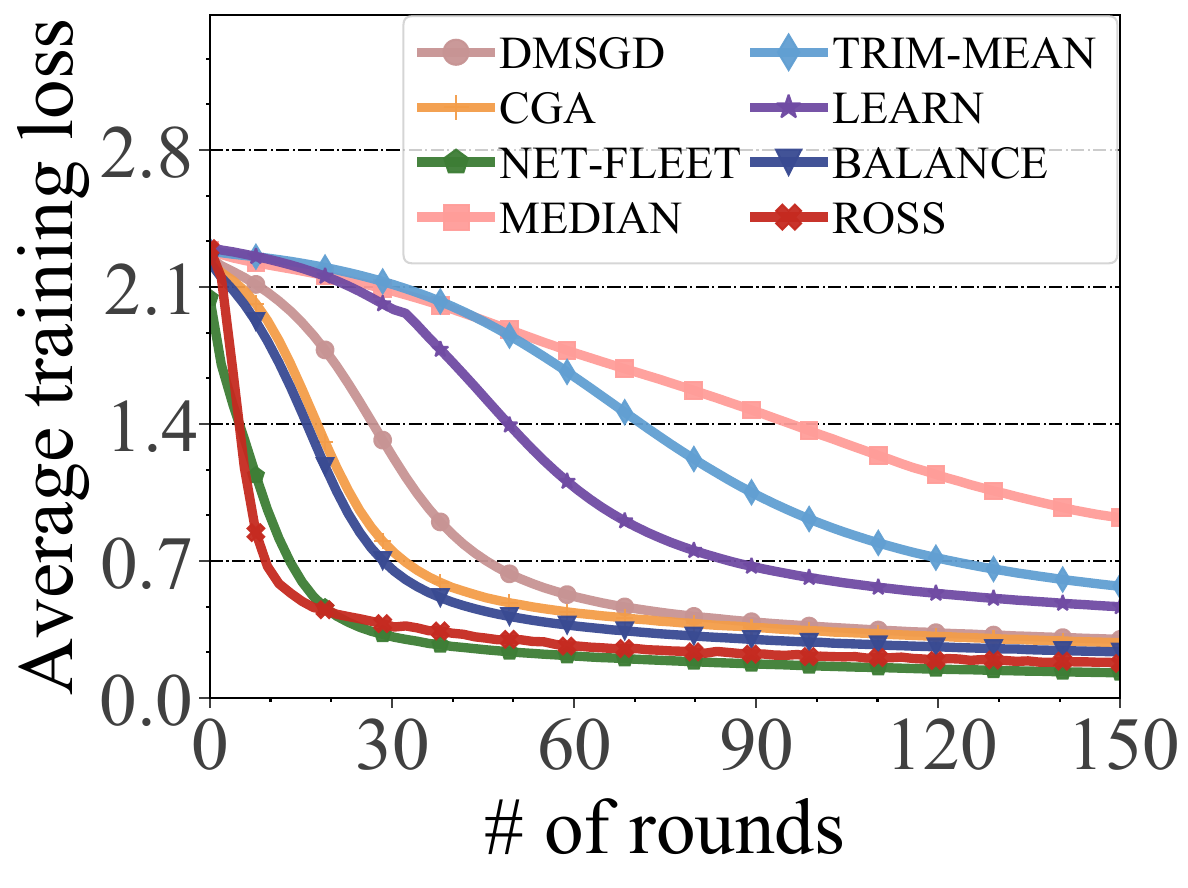}}
        \parbox{.24\textwidth}{\center\includegraphics[width=.24\textwidth]{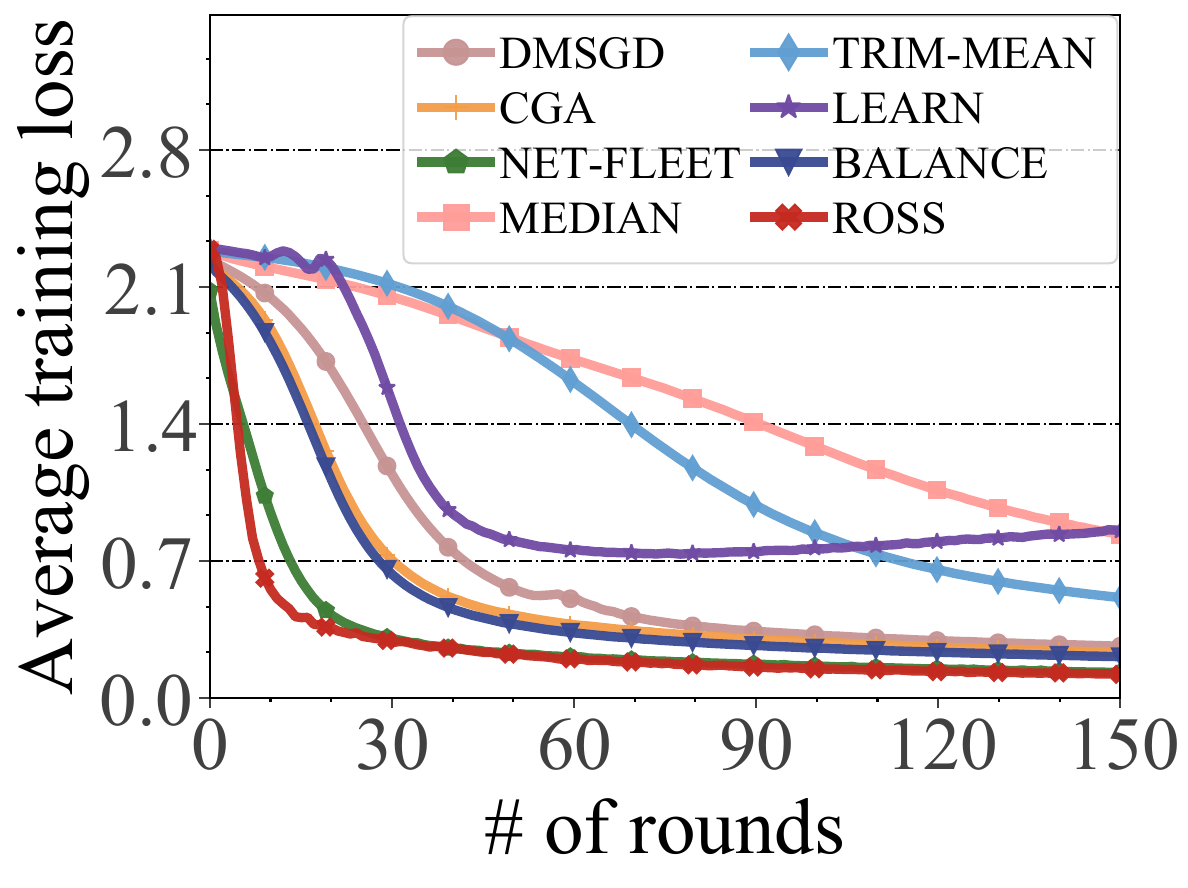}}
        \parbox{.25\textwidth}{\center\scriptsize(a1) Long-tailed ($N=10$)}
        \parbox{.23\textwidth}{\center\scriptsize(a2) Data noise ($N=10$)}
        \parbox{.23\textwidth}{\center\scriptsize(a3) Label noise ($N=10$)}
        \parbox{.25\textwidth}{\center\scriptsize(a4) Gradient poisoning ($N=10$)}
        \parbox{.24\textwidth}{\center\includegraphics[width=.24\textwidth]{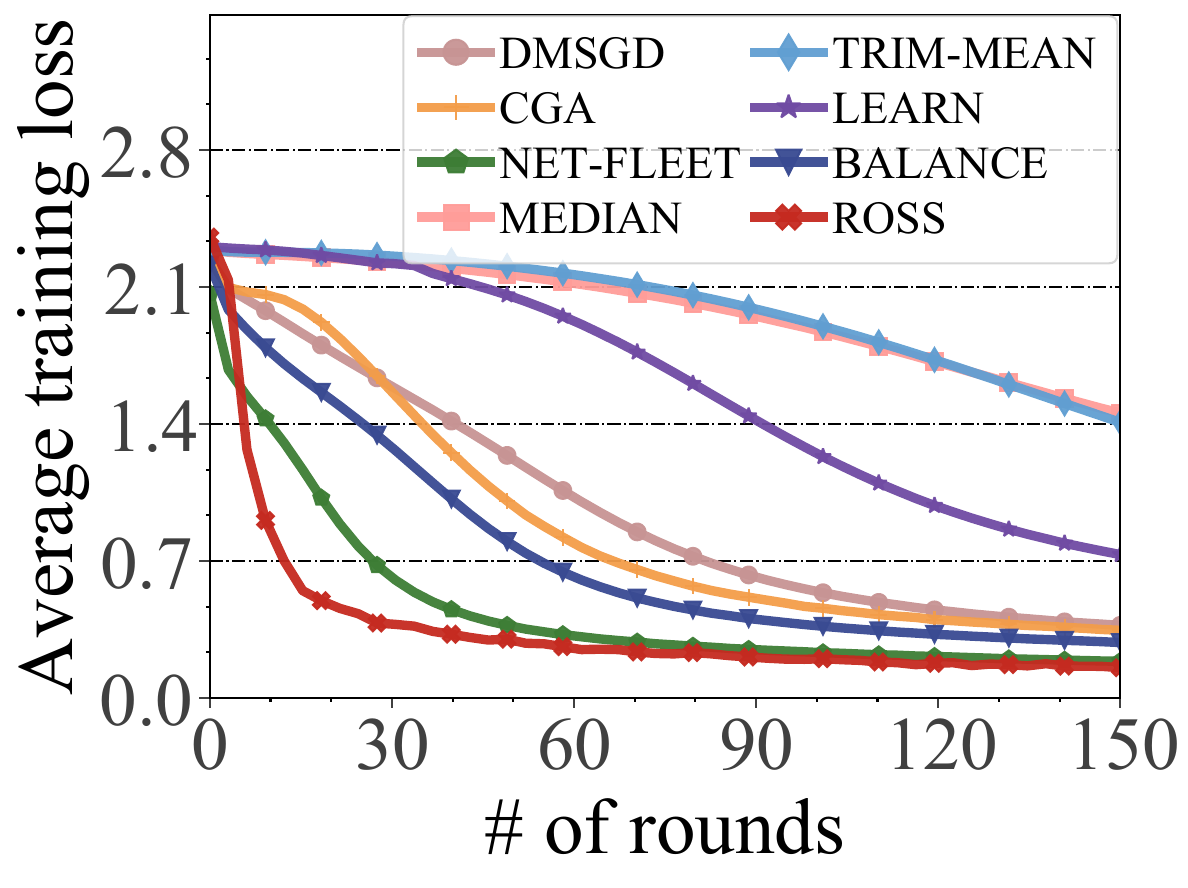}}
        \parbox{.24\textwidth}{\center\includegraphics[width=.24\textwidth]{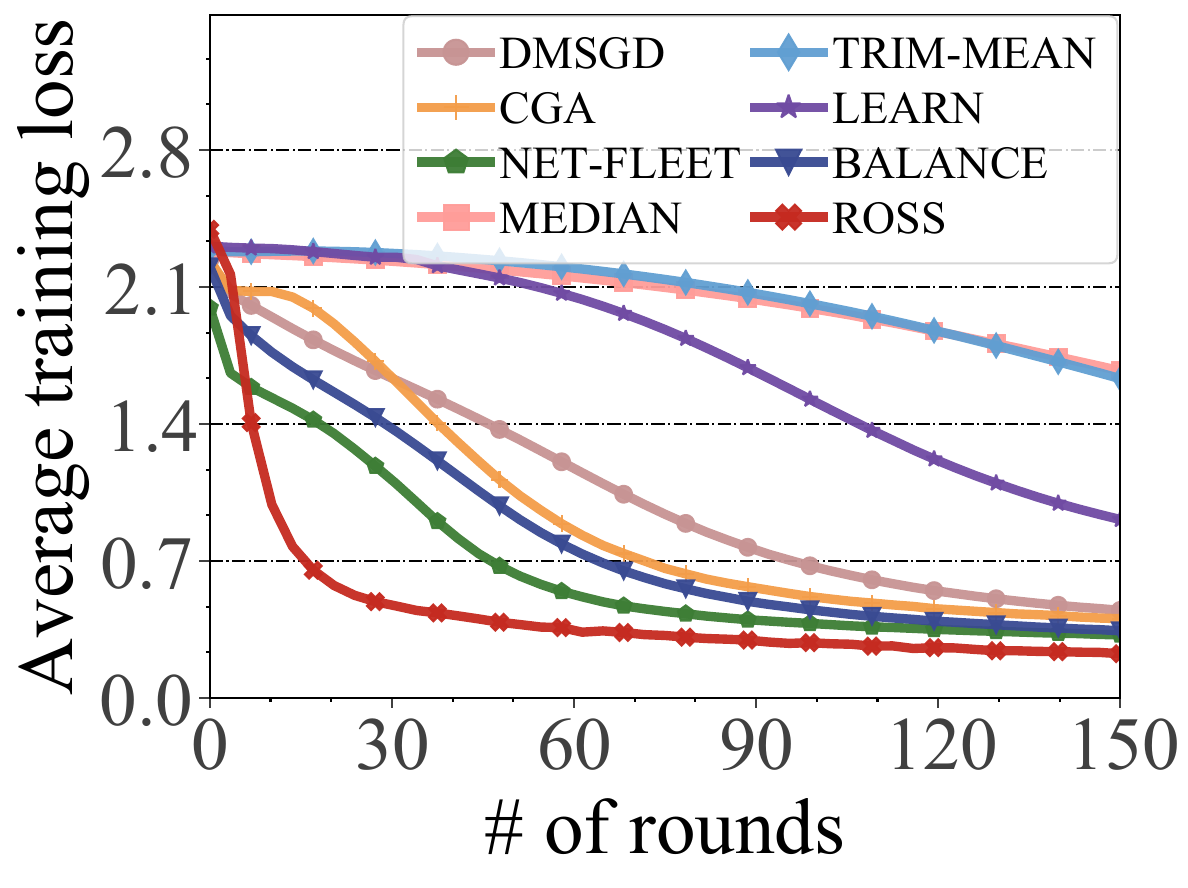}}
        \parbox{.24\textwidth}{\center\includegraphics[width=.24\textwidth]{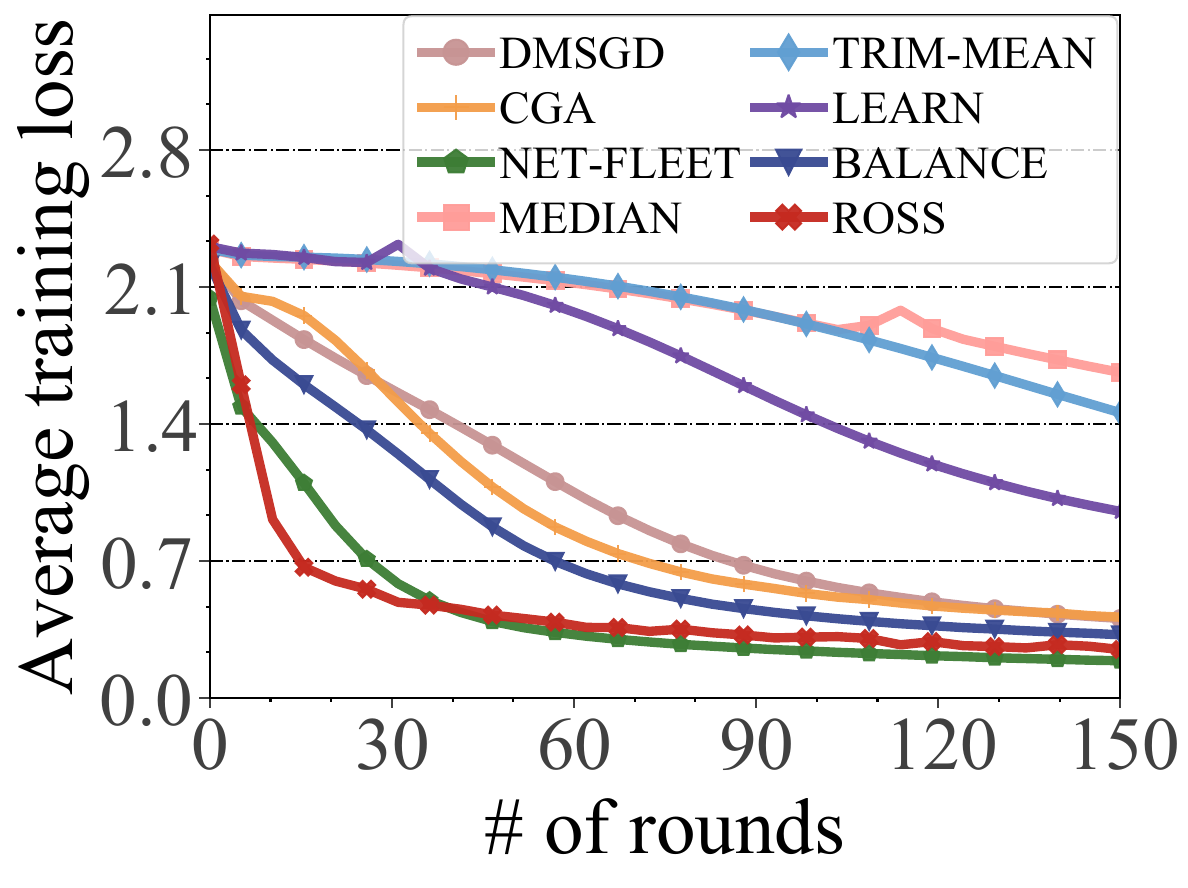}}
        \parbox{.24\textwidth}{\center\includegraphics[width=.24\textwidth]{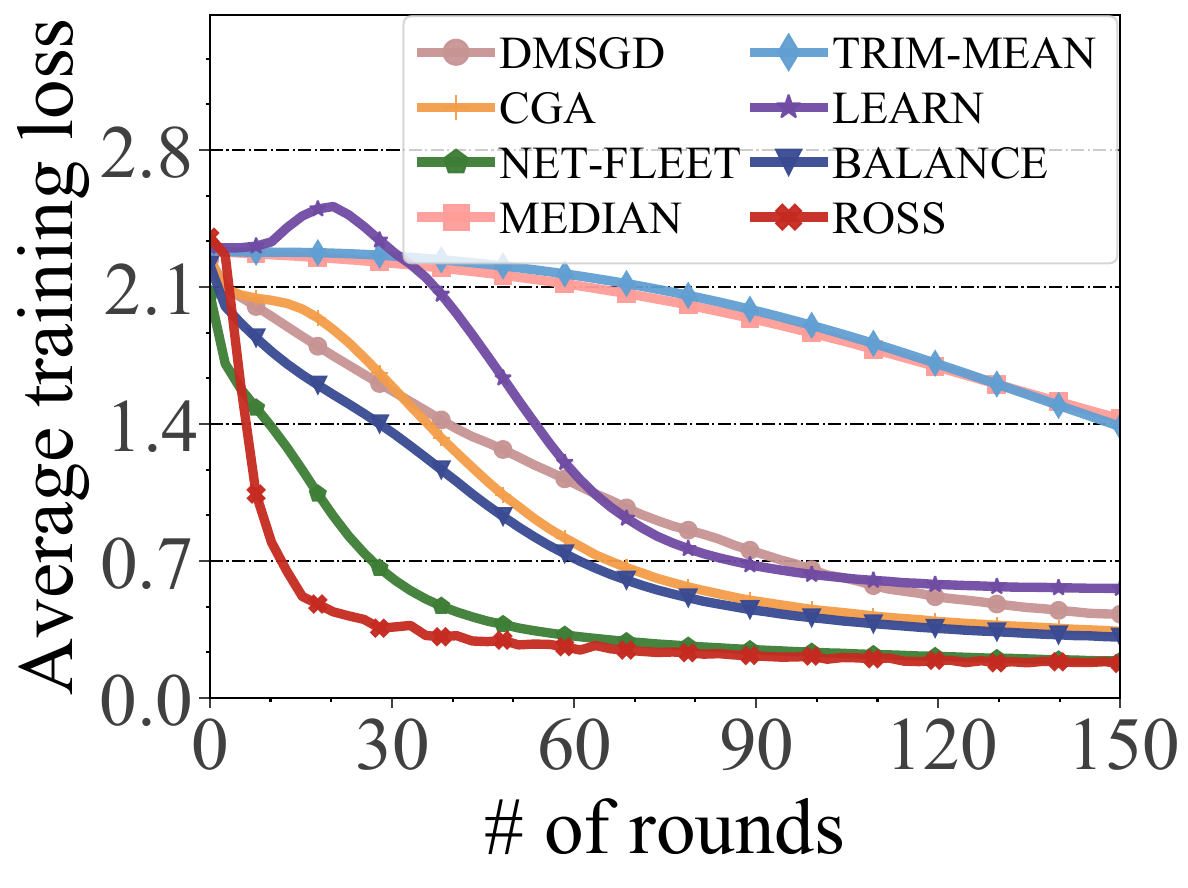}}
        \parbox{.25\textwidth}{\center\scriptsize(b1) Long-tailed ($N=20$)}
        \parbox{.23\textwidth}{\center\scriptsize(b2) Data noise ($N=20$)}
        \parbox{.23\textwidth}{\center\scriptsize(b3) Label noise ($N=20$)}
        \parbox{.25\textwidth}{\center\scriptsize(b4) Gradient poisoning ($N=20$)}
        \parbox{.24\textwidth}{\center\includegraphics[width=.24\textwidth]{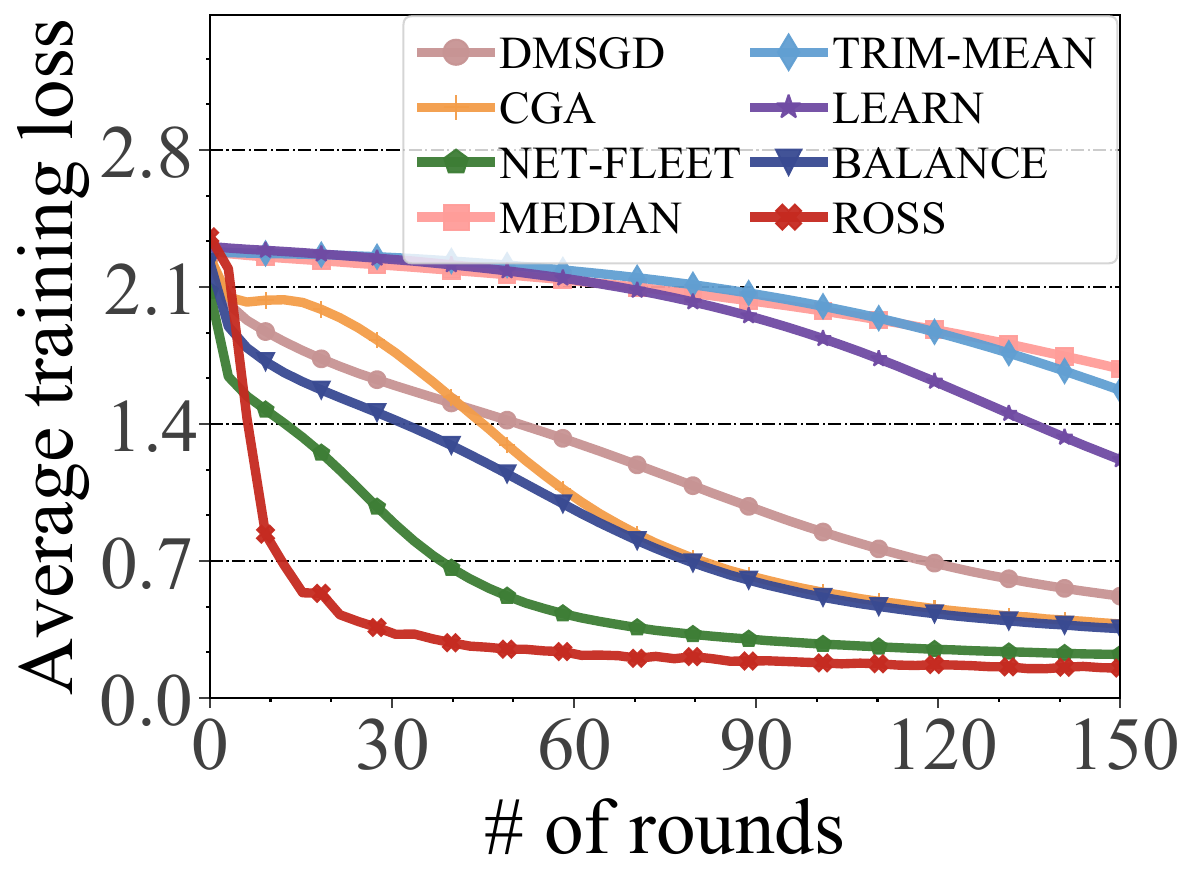}}
        \parbox{.24\textwidth}{\center\includegraphics[width=.24\textwidth]{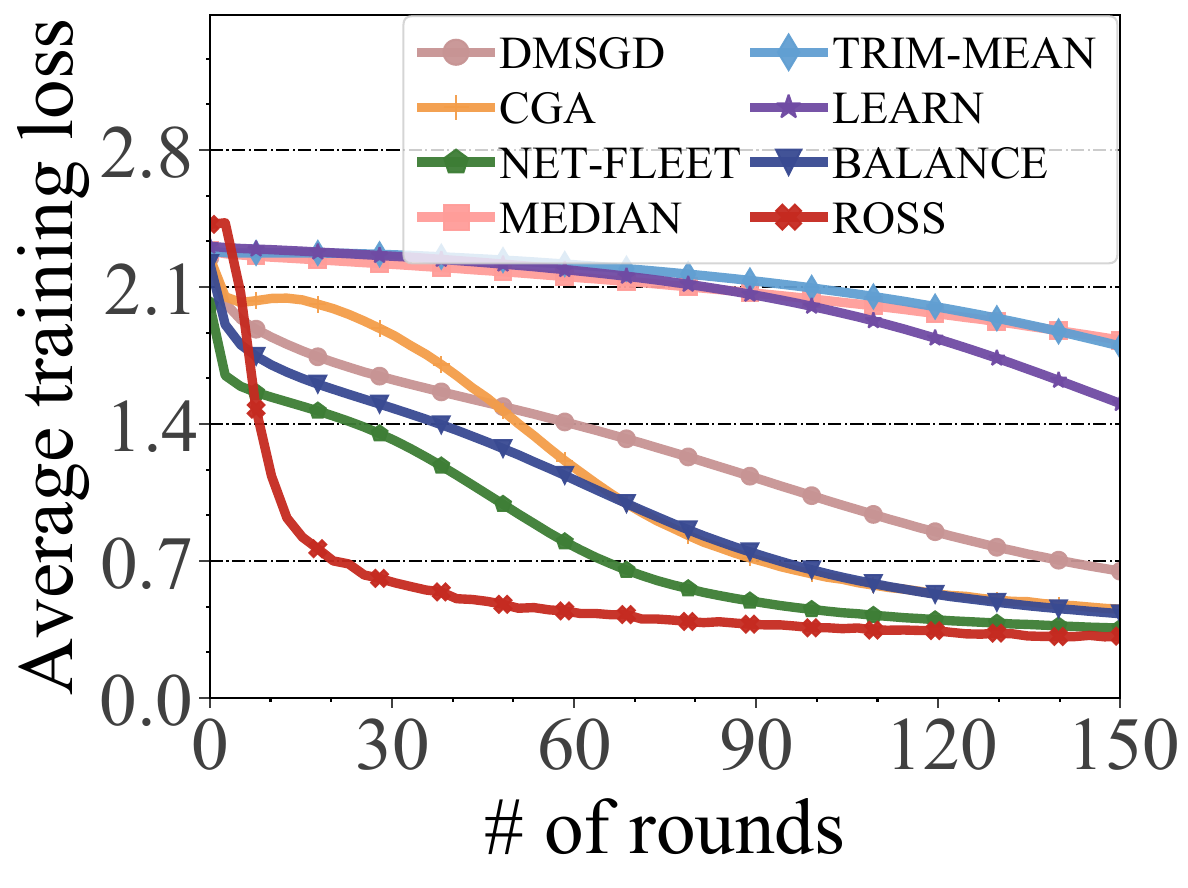}}
        \parbox{.24\textwidth}{\center\includegraphics[width=.24\textwidth]{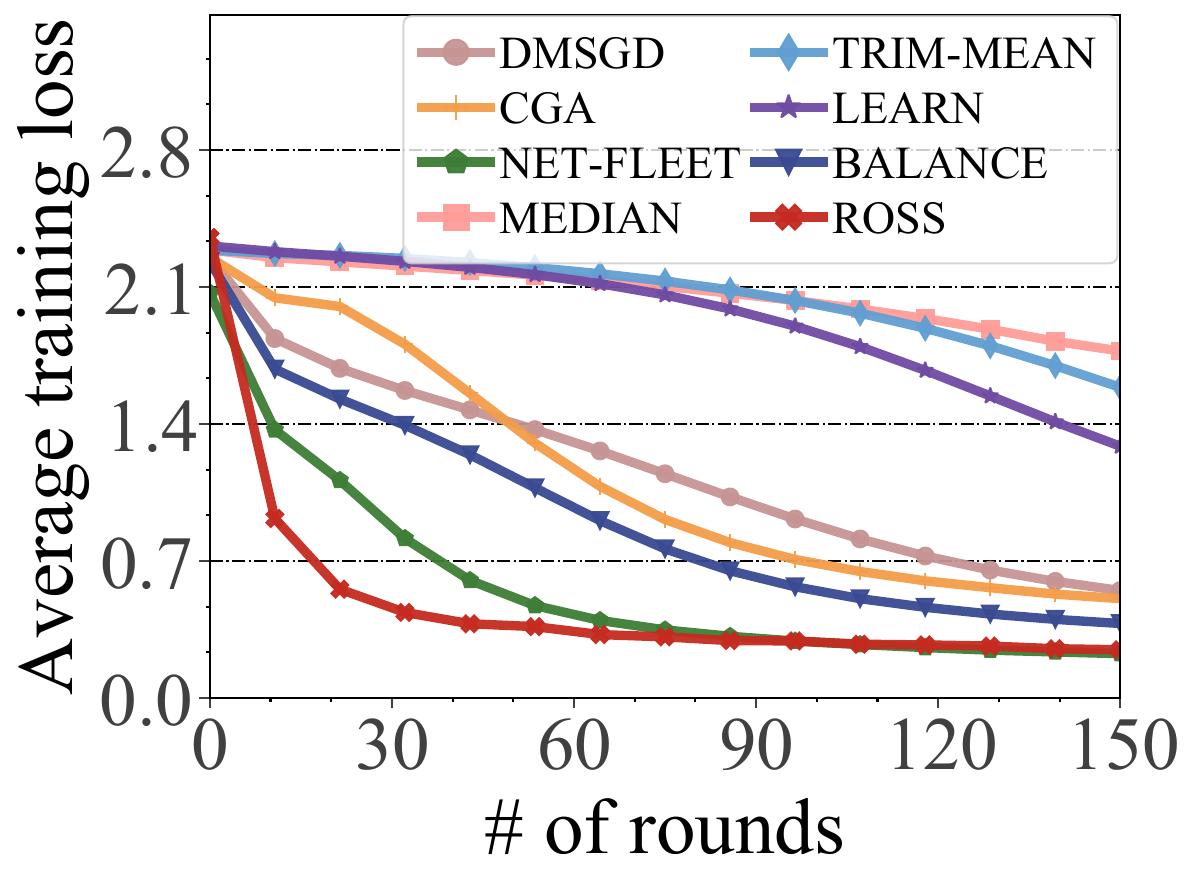}}
        \parbox{.24\textwidth}{\center\includegraphics[width=.24\textwidth]{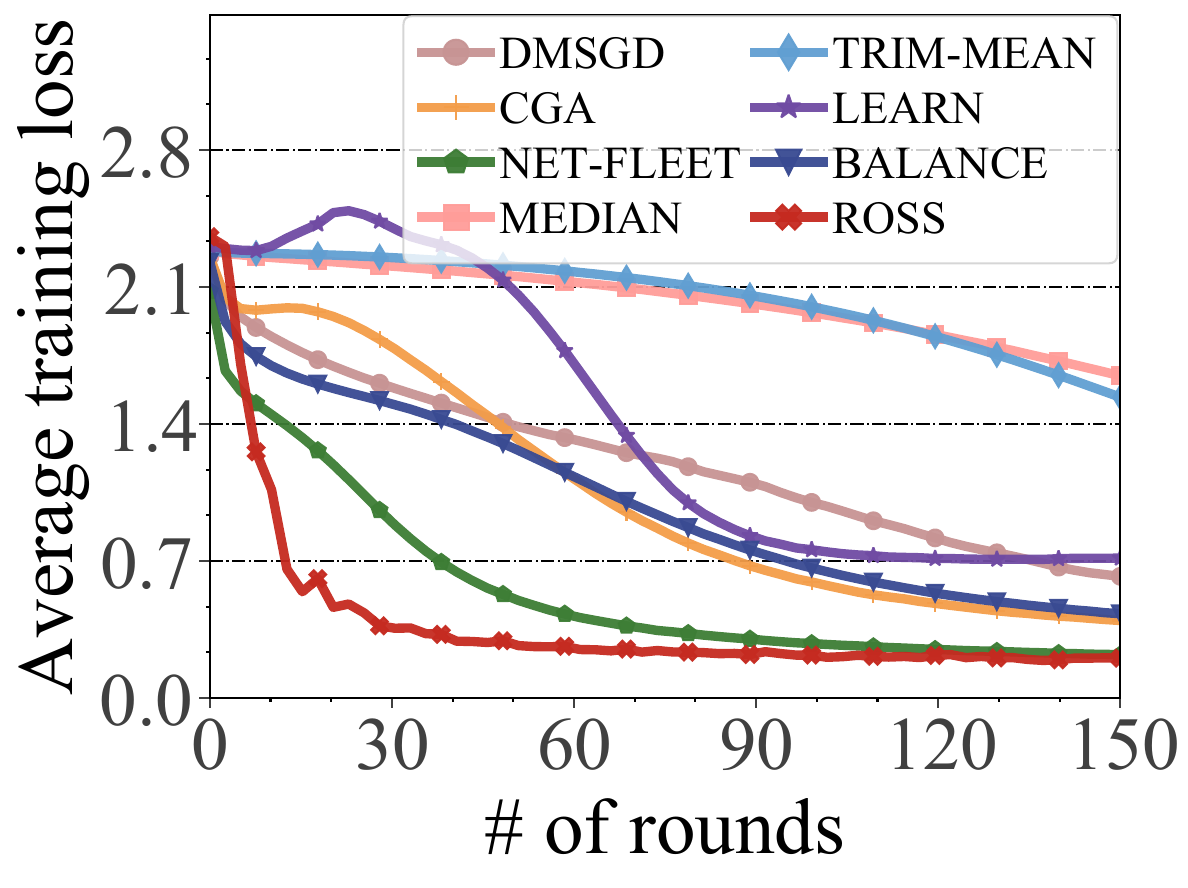}}
        \parbox{.25\textwidth}{\center\scriptsize(c1) Long-tailed ($N=30$)}
        \parbox{.23\textwidth}{\center\scriptsize(c2) Data noise ($N=30$)}
        \parbox{.23\textwidth}{\center\scriptsize(c3) Label noise ($N=30$)}
        \parbox{.25\textwidth}{\center\scriptsize(c4) Gradient poisoning ($N=30$)}
      \caption{Comparison results on MNIST dataset over bipartite graphs.}
      \label{fig:mnist-loss-bipartite}
      \end{center}
      \end{figure*}

      We also report the average loss on bipartite graphs in Fig.~\ref{fig:mnist-loss-bipartite}. The results show that, even over a sparser communication topology, ROSS converges within about $60–70$ rounds across all settings, whereas CGA, BALANCE, MEDIAN, TRIM-MEAN, LEARN, and DMSGD require significantly more rounds to converge. We further observe that ROSS and NET-FLEET exhibit similar convergence behavior; however, ROSS demonstrates a considerable advantage in terms of test accuracy, as will be shown later.

\begin{figure*}[htb!]
      \begin{center}
        \parbox{.32\textwidth}{\center\includegraphics[width=.32\textwidth]{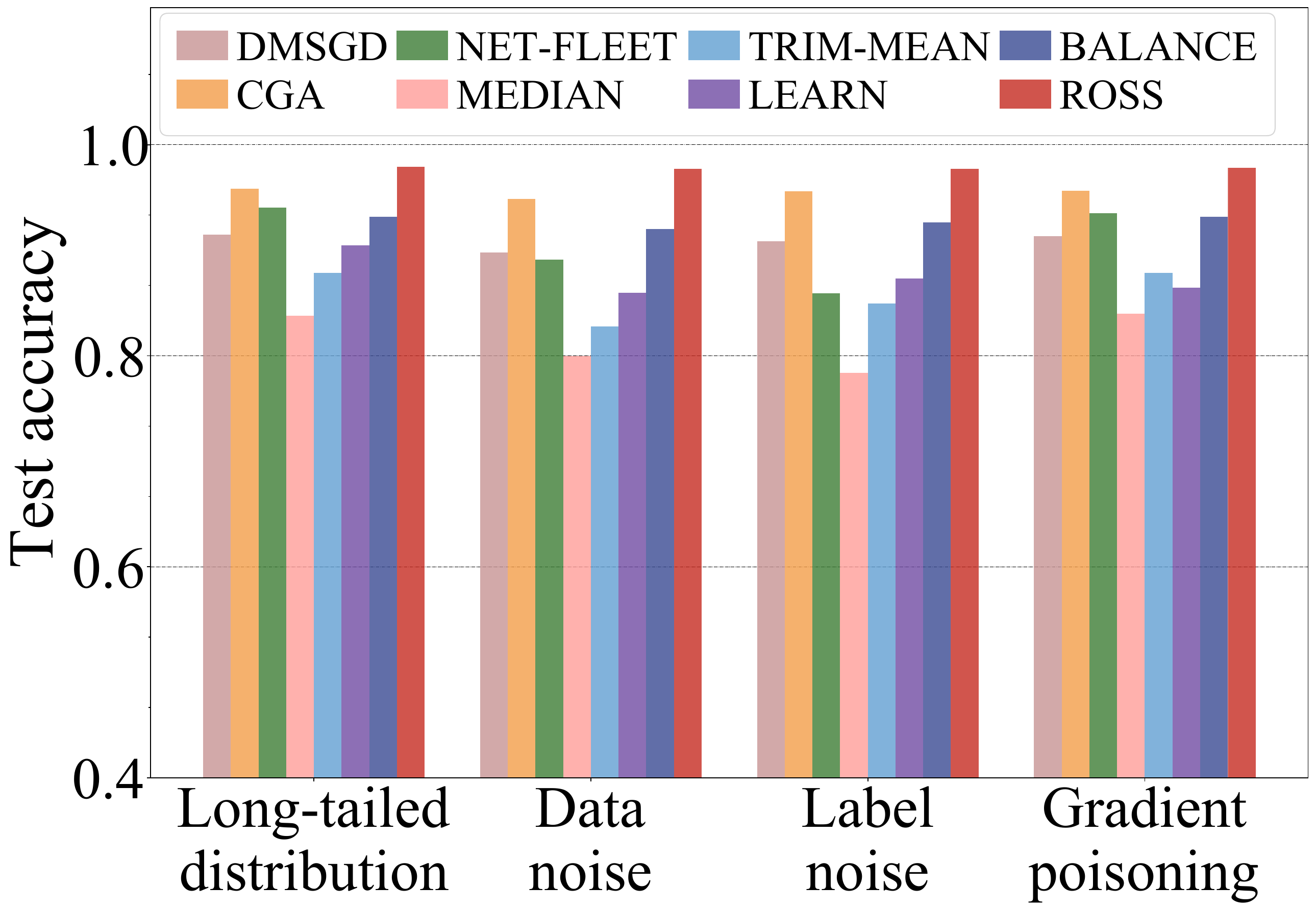}}
        \parbox{.32\textwidth}{\center\includegraphics[width=.32\textwidth]{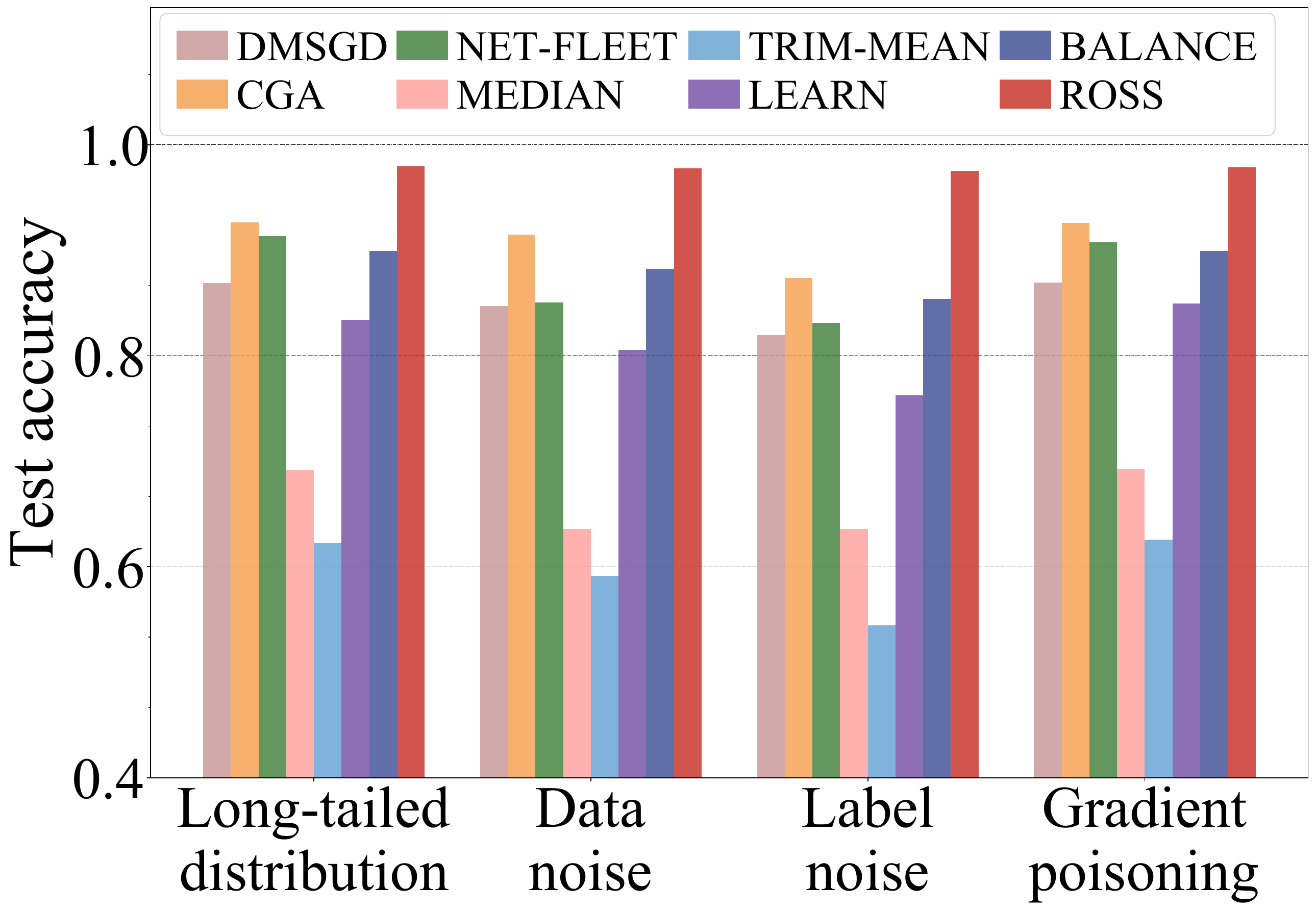}}
        \parbox{.32\textwidth}{\center\includegraphics[width=.32\textwidth]{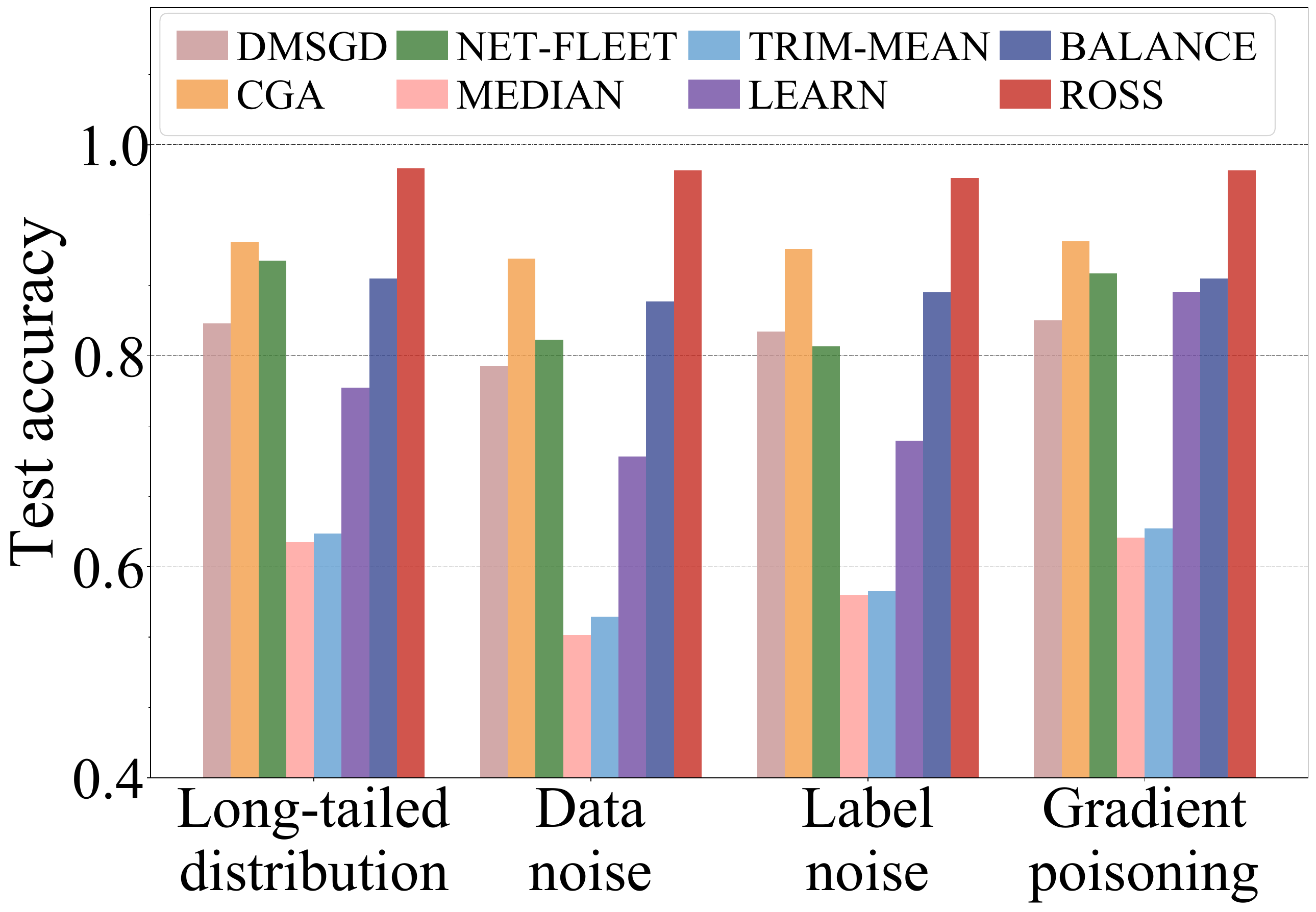}}
        \parbox{.32\textwidth}{\center\scriptsize(a) $N=10$}
        \parbox{.32\textwidth}{\center\scriptsize(b) $N=20$}
        \parbox{.32\textwidth}{\center\scriptsize(c) $N=30$}
      \caption{Comparison results in terms of test accuracy on MNIST dataset over fully connected graphs.}
      \label{fig:mnist-acc-full}
      \end{center}
      \end{figure*}

\begin{figure*}[htb!]
      \begin{center}
        \parbox{.32\textwidth}{\center\includegraphics[width=.32\textwidth]{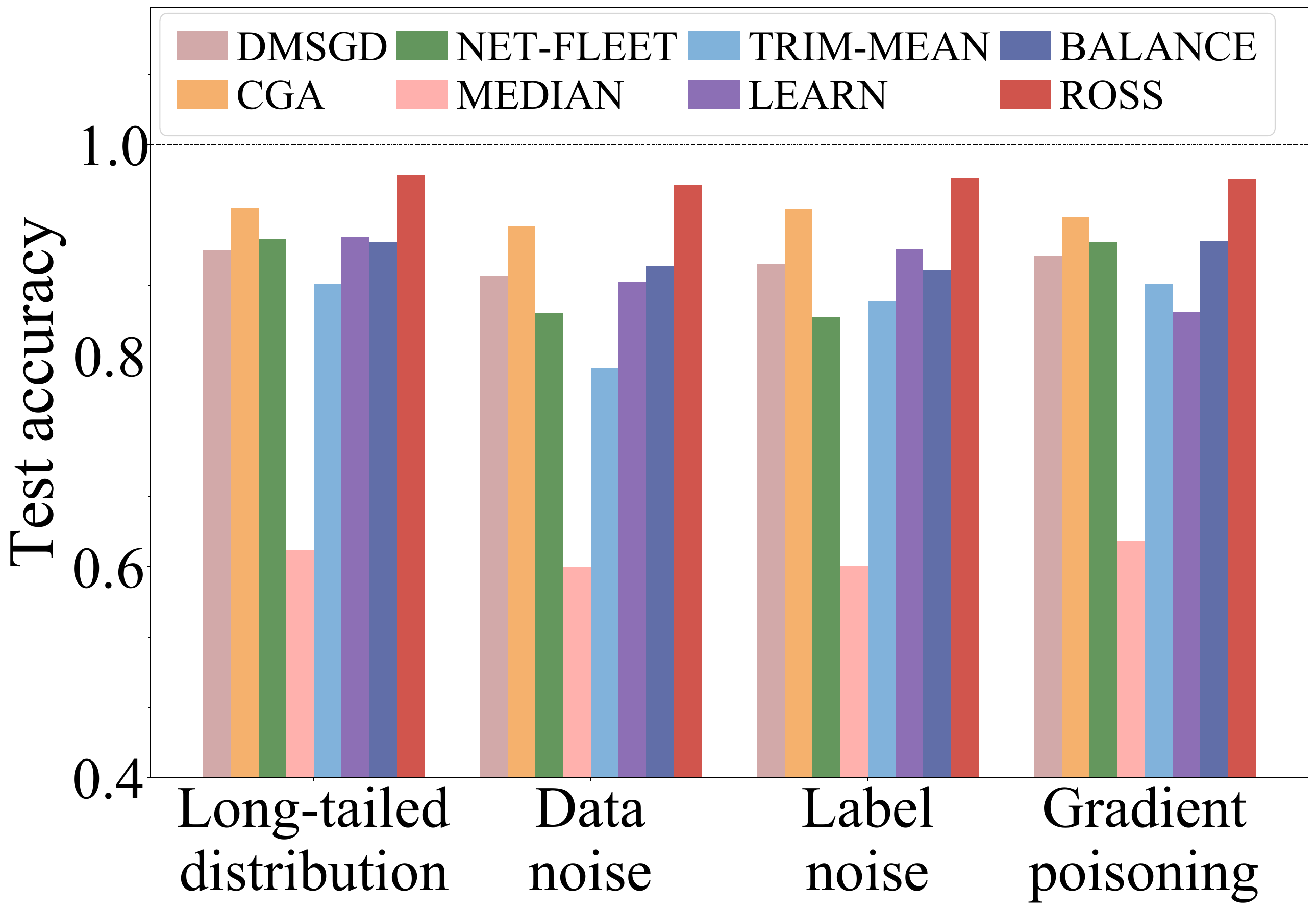}}
        \parbox{.32\textwidth}{\center\includegraphics[width=.32\textwidth]{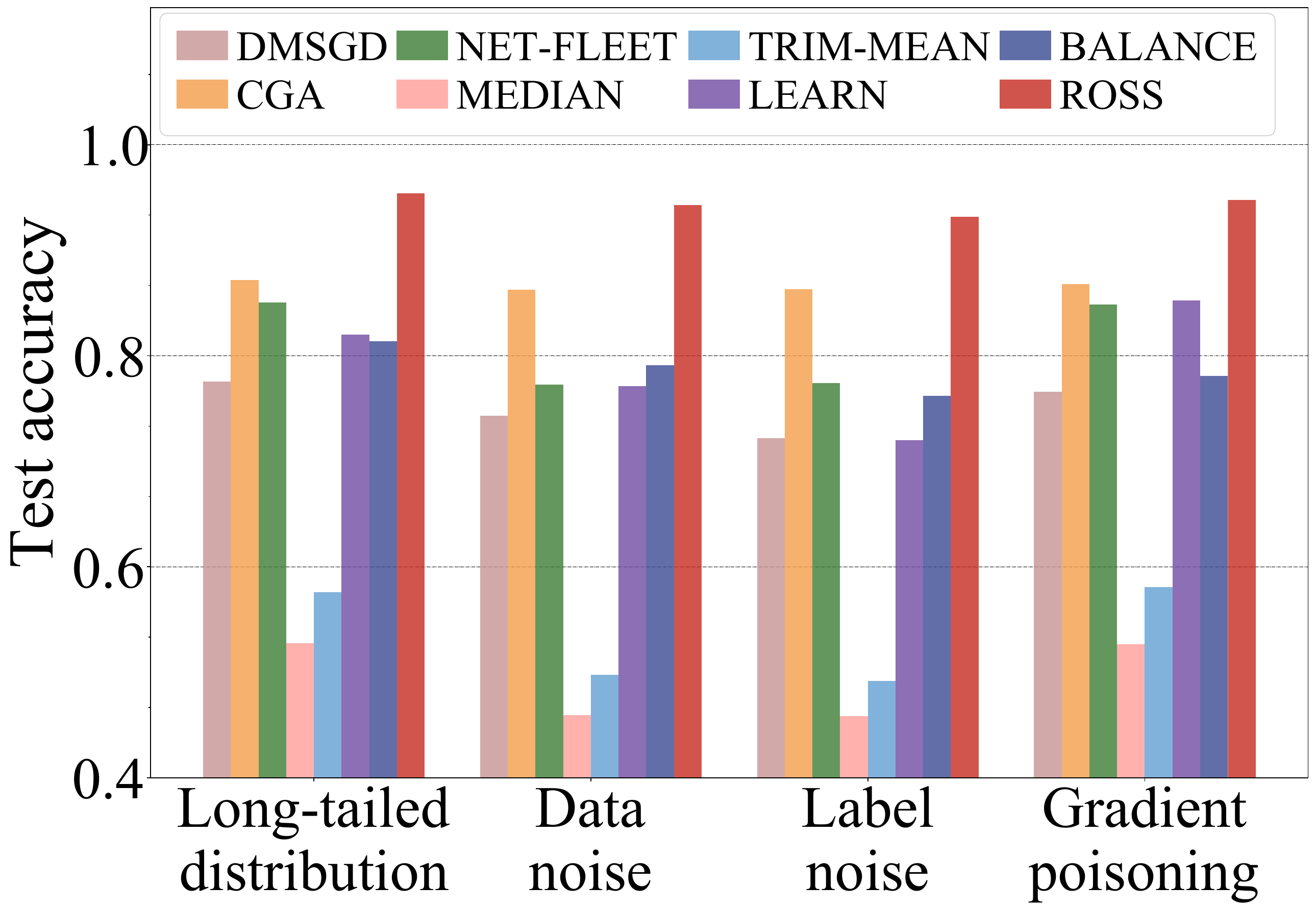}}
        \parbox{.32\textwidth}{\center\includegraphics[width=.32\textwidth]{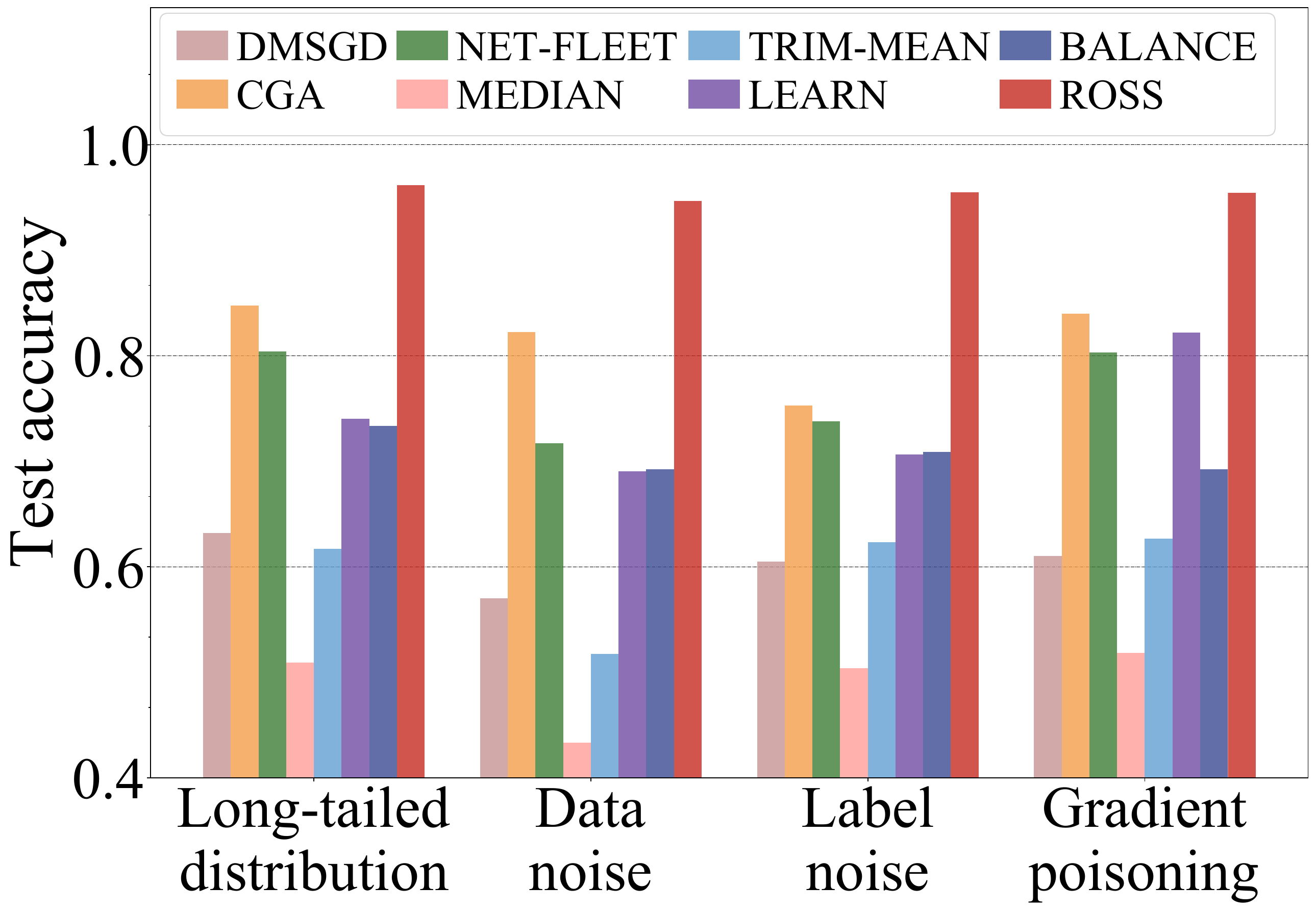}}
        \parbox{.32\textwidth}{\center\scriptsize(a) $N=10$}
        \parbox{.32\textwidth}{\center\scriptsize(b) $N=20$}
        \parbox{.32\textwidth}{\center\scriptsize(c) $N=30$}
      \caption{Comparison results in terms of test accuracy on MNIST dataset over bipartite graphs.}
      \label{fig:mnist-acc-bipartite}
      \end{center}
      \end{figure*}

      We present the prediction accuracy of the different algorithms on the test datasets in Fig.~\ref{fig:mnist-acc-full}-\ref{fig:mnist-acc-bipartite}. As shown in Fig.~\ref{fig:mnist-acc-full} with fully connected communication graphs, our ROSS algorithm achieves a test accuracy of about $0.98$ across all settings, and this performance remains stable even as the number of agents increases. In contrast, the baseline algorithms have lower accuracies. For example, under the long-tailed distribution with $N=10$, the test accuracies of DMSGD, NET-FLEET, CGA, MEDIAN, TRIM-MEAN, LEARN, and BALANCE are $0.92$, $0.94$, $0.96$, $0.84$, $0.88$, $0.90$, and $0.93$, respectively. Moreover, their performance degrades significantly as the number of agents grows. With $N=30$, their accuracies drop to the range of $0.62–0.91$. We also evaluate the performance of the algorithms on bipartite graphs. As shown in Fig.~\ref{fig:mnist-acc-bipartite}, our ROSS algorithm outperforms the baselines under this sparse topology. For instance, when $N=30$ in the label noise setting, the accuracy of ROSS is $1.25-1.92$ times higher than the ones of the baseline algorithms.

    \subsubsection{Experiment Results on CIFAR-10 Dataset} \label{ssec:res-cifar10}
\begin{figure*}[htb!]
      \begin{center}
        \parbox{.24\textwidth}{\center\includegraphics[width=.24\textwidth]{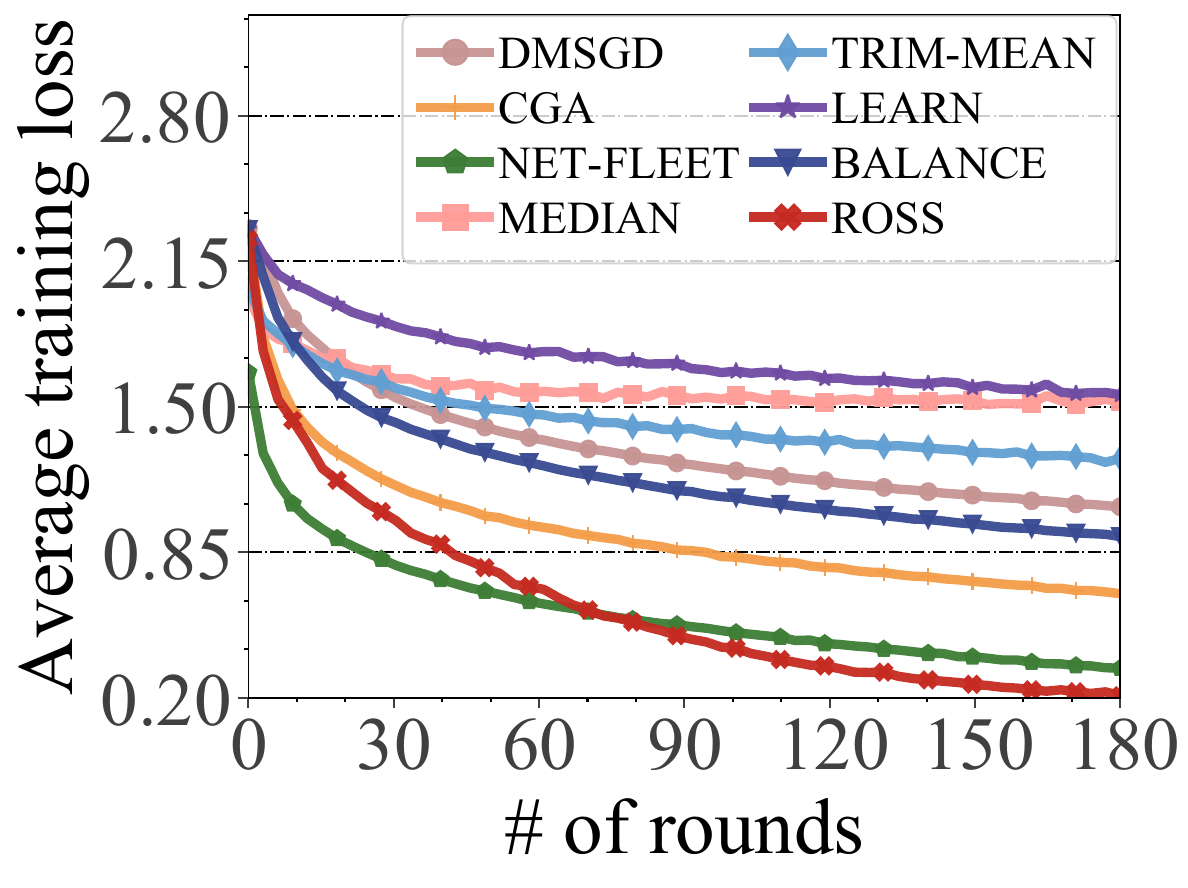}}
        \parbox{.24\textwidth}{\center\includegraphics[width=.24\textwidth]{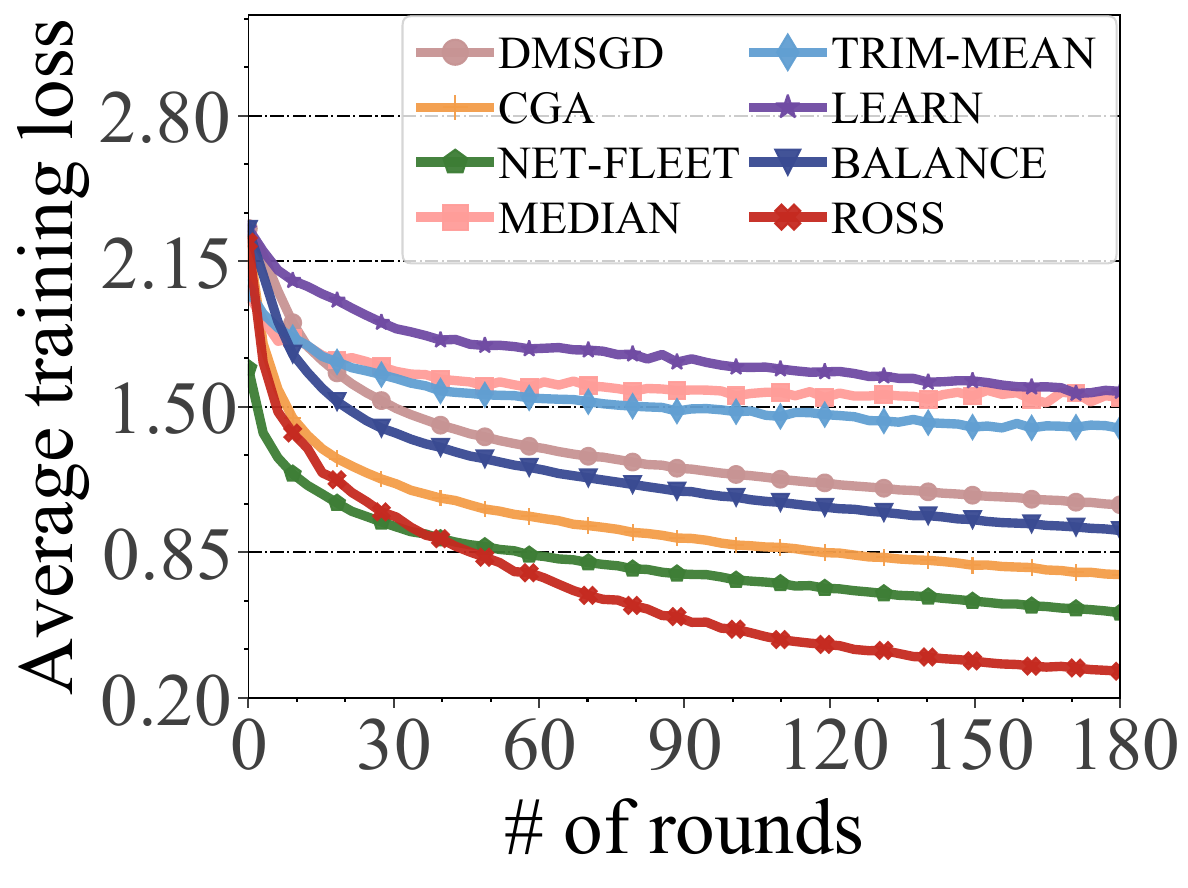}}
        \parbox{.24\textwidth}{\center\includegraphics[width=.24\textwidth]{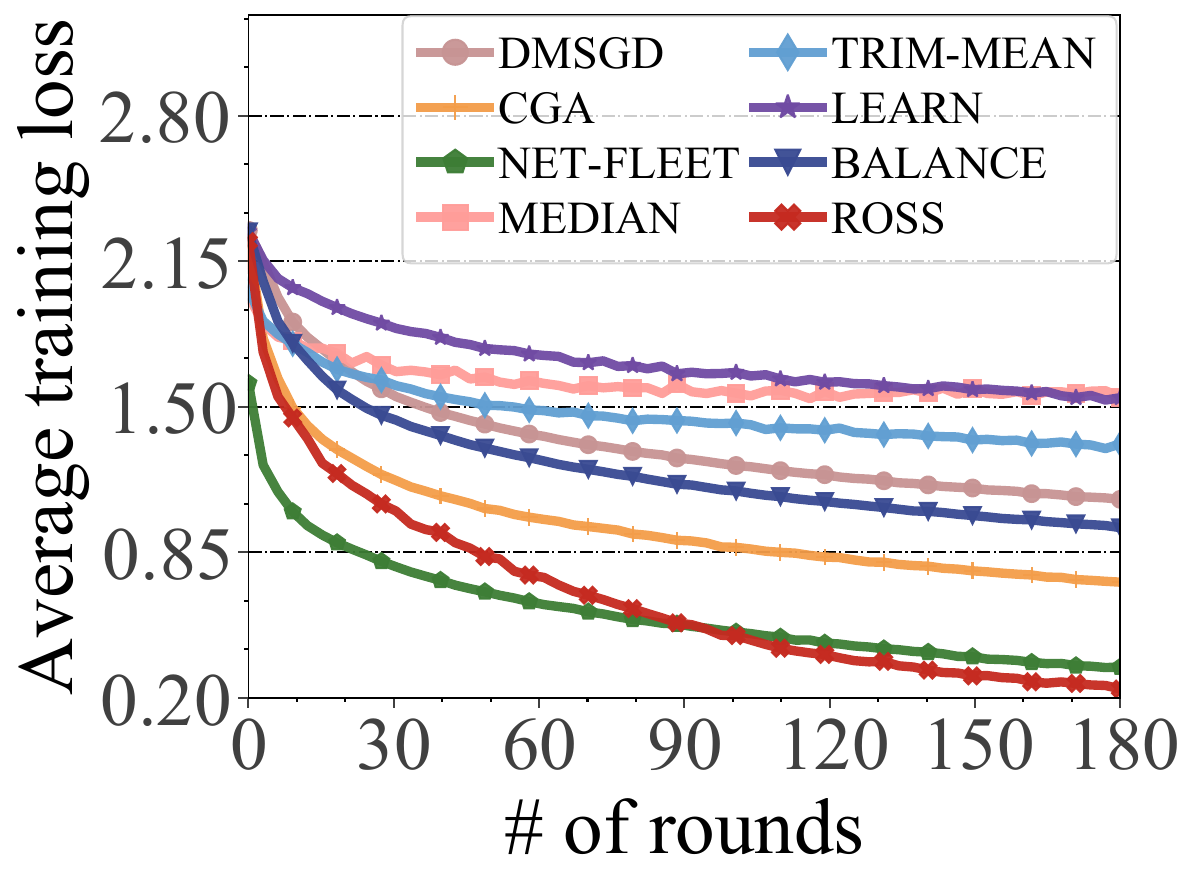}}
        \parbox{.24\textwidth}{\center\includegraphics[width=.24\textwidth]{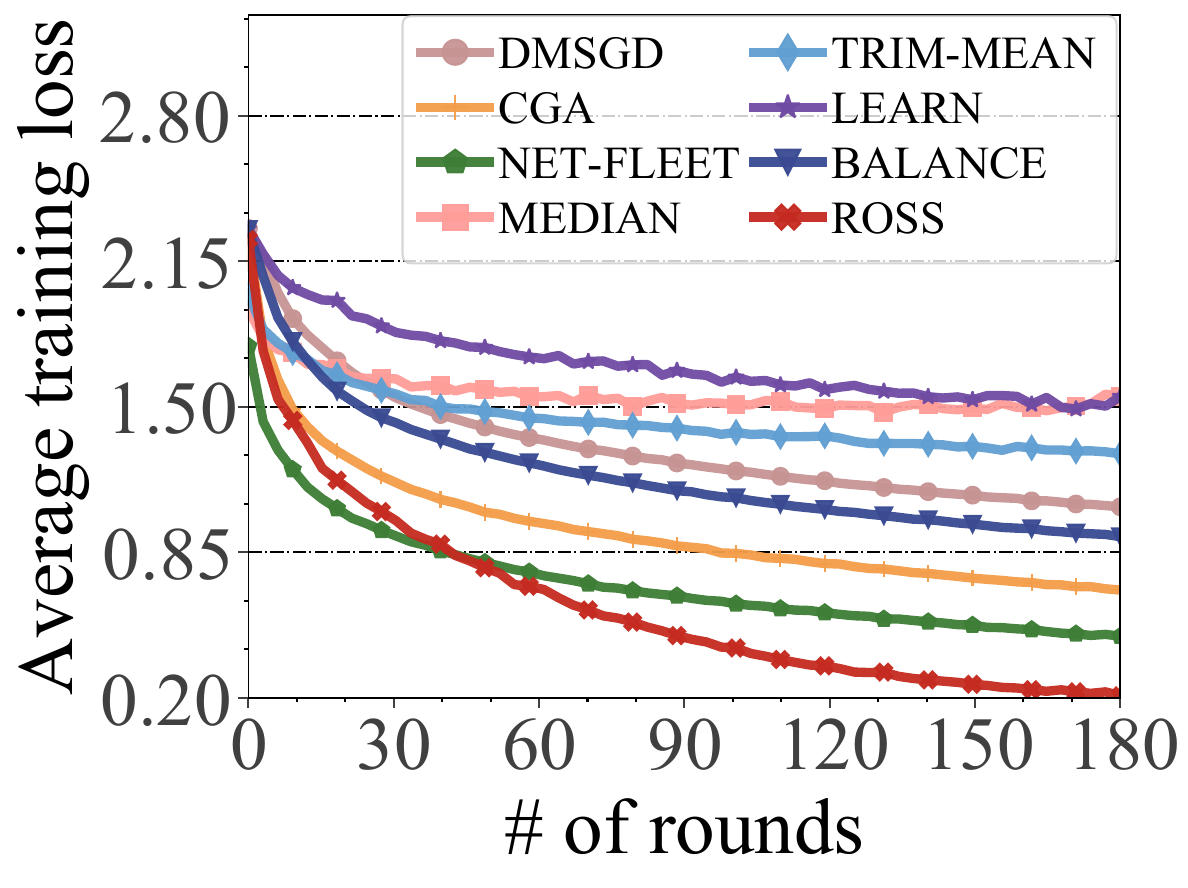}}
        \parbox{.25\textwidth}{\center\scriptsize(a1) Long-tailed ($N=10$)}
        \parbox{.23\textwidth}{\center\scriptsize(a2) Data noise ($N=10$)}
        \parbox{.23\textwidth}{\center\scriptsize(a3) Label noise ($N=10$)}
        \parbox{.25\textwidth}{\center\scriptsize(a4) Gradient poisoning ($N=10$)}
        \parbox{.24\textwidth}{\center\includegraphics[width=.24\textwidth]{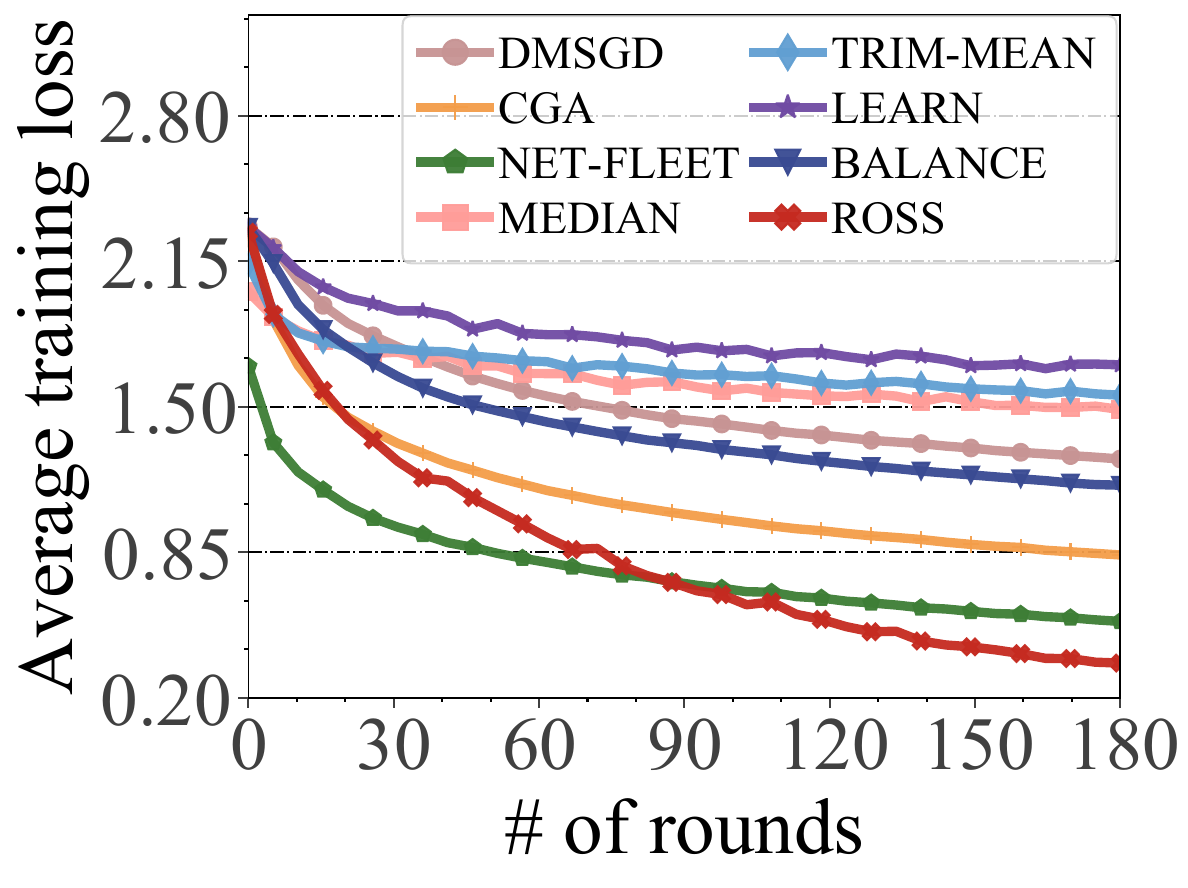}}
        \parbox{.24\textwidth}{\center\includegraphics[width=.24\textwidth]{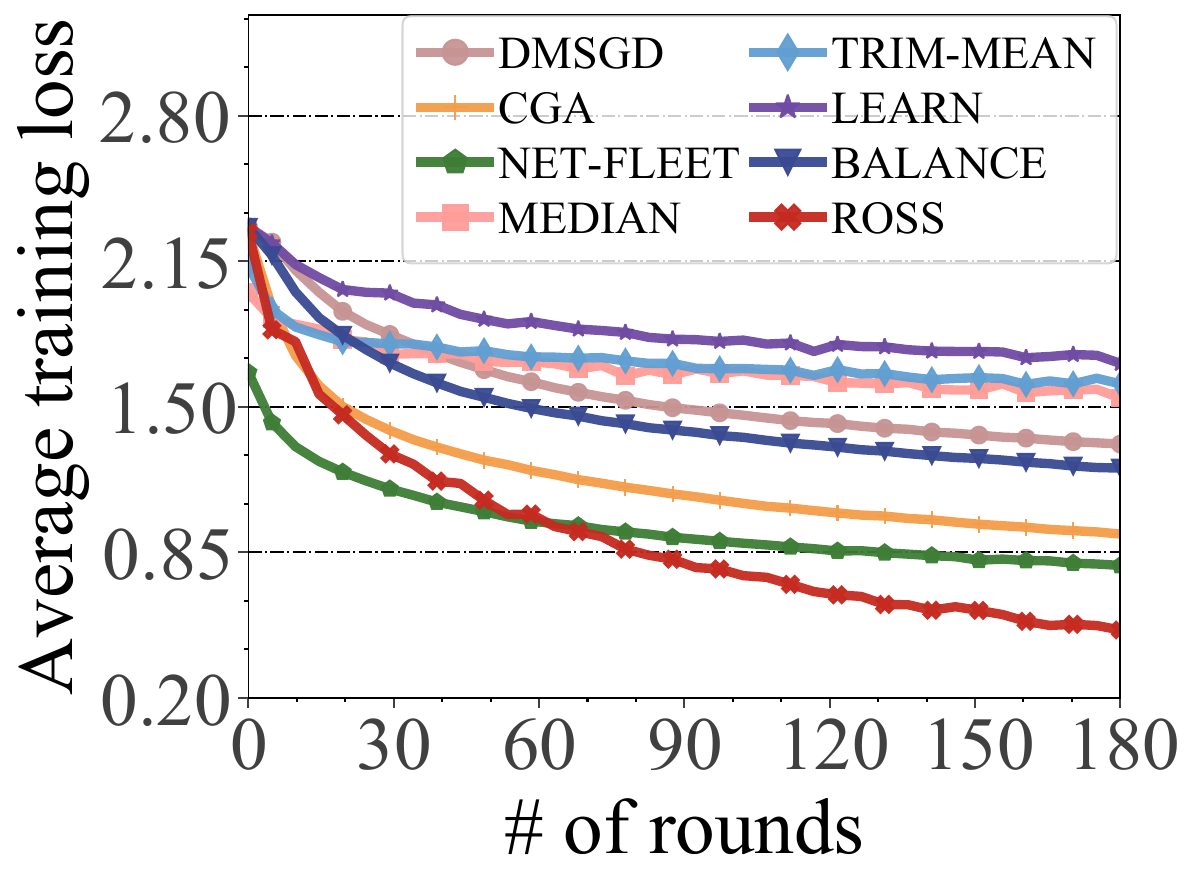}}
        \parbox{.24\textwidth}{\center\includegraphics[width=.24\textwidth]{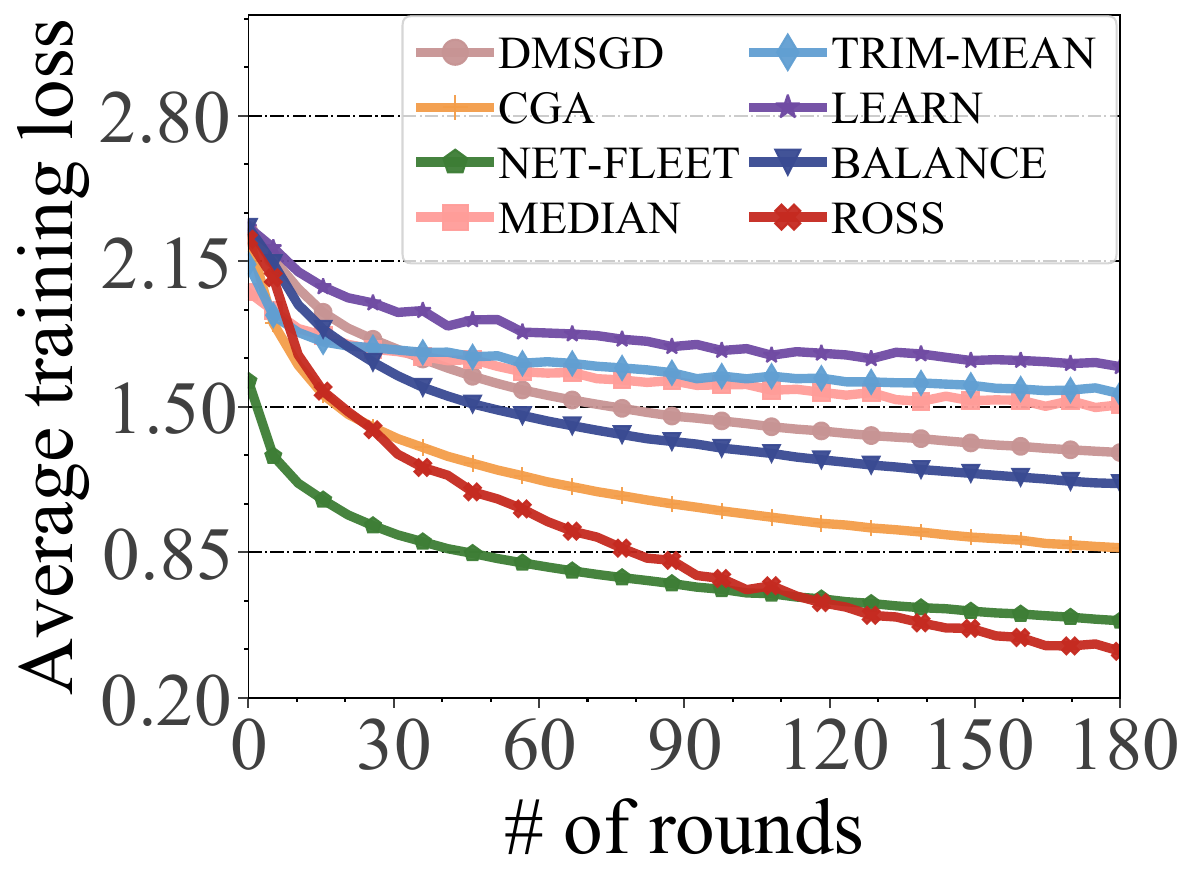}}
        \parbox{.24\textwidth}{\center\includegraphics[width=.24\textwidth]{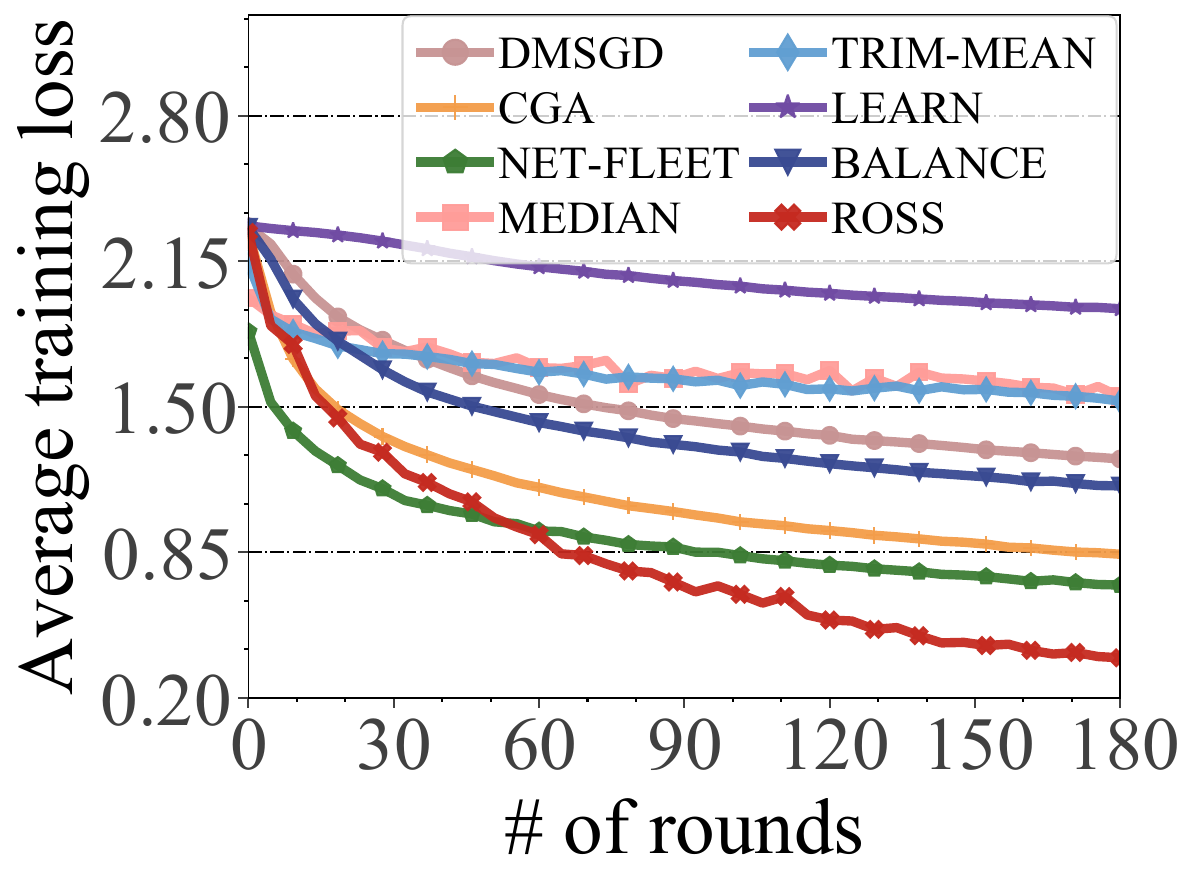}}
        \parbox{.25\textwidth}{\center\scriptsize(b1) Long-tailed ($N=20$)}
        \parbox{.23\textwidth}{\center\scriptsize(b2) Data noise ($N=20$)}
        \parbox{.23\textwidth}{\center\scriptsize(b3) Label noise ($N=20$)}
        \parbox{.25\textwidth}{\center\scriptsize(b4) Gradient poisoning ($N=20$)}
        \parbox{.24\textwidth}{\center\includegraphics[width=.24\textwidth]{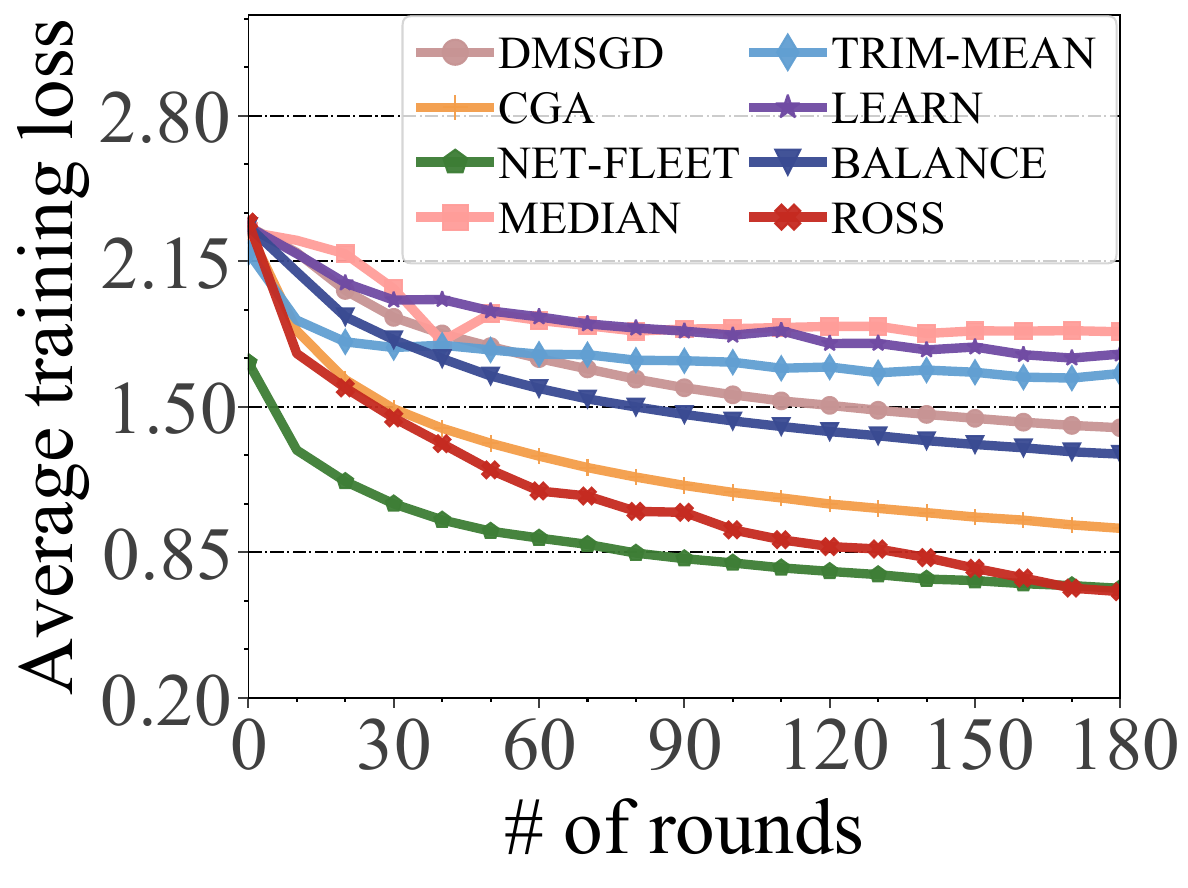}}
        \parbox{.24\textwidth}{\center\includegraphics[width=.24\textwidth]{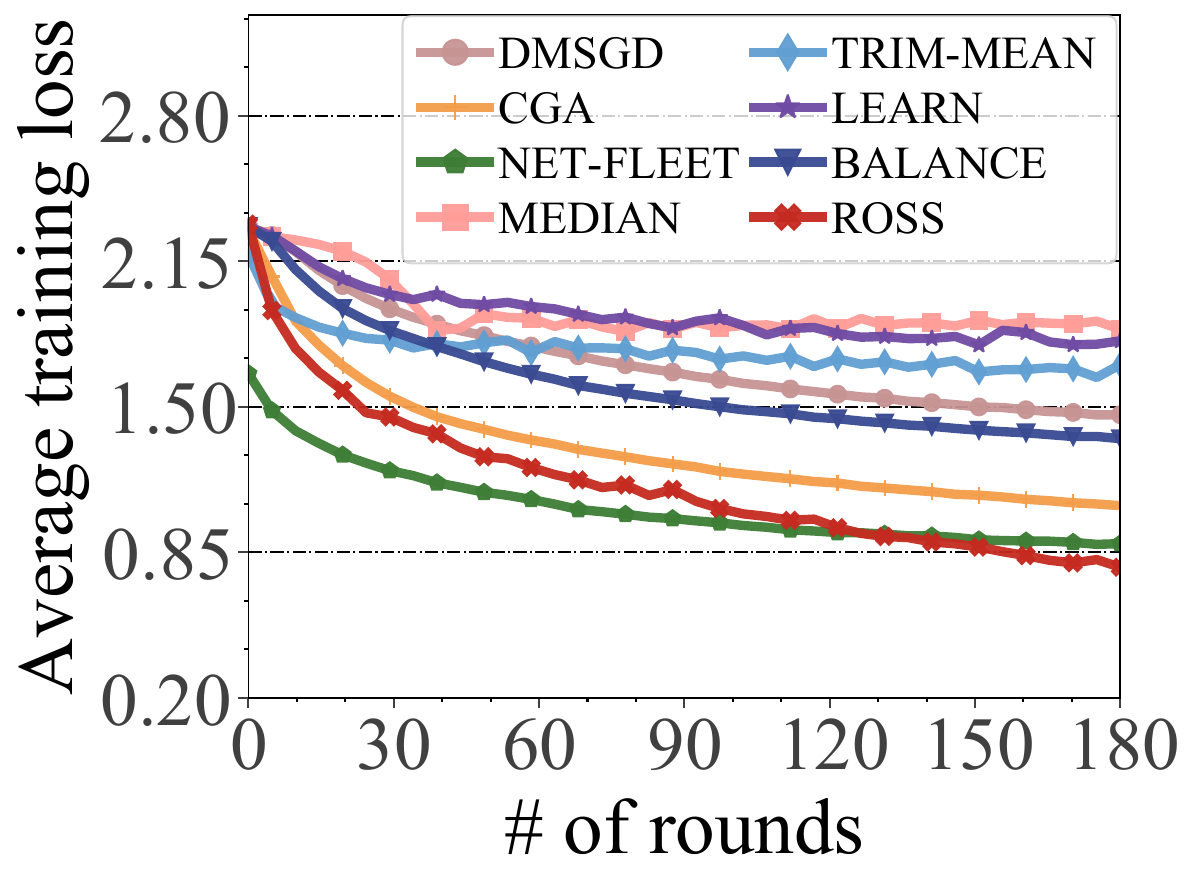}}
        \parbox{.24\textwidth}{\center\includegraphics[width=.24\textwidth]{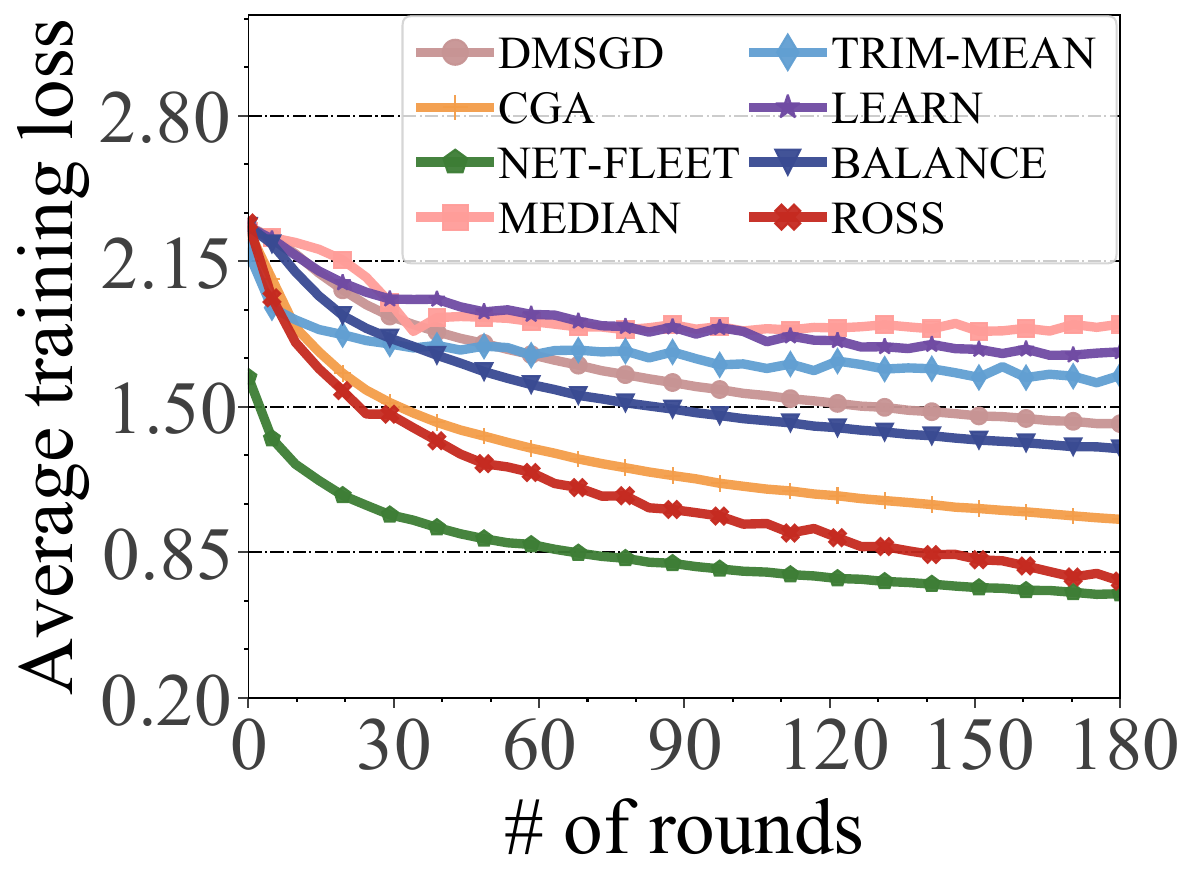}}
        \parbox{.24\textwidth}{\center\includegraphics[width=.24\textwidth]{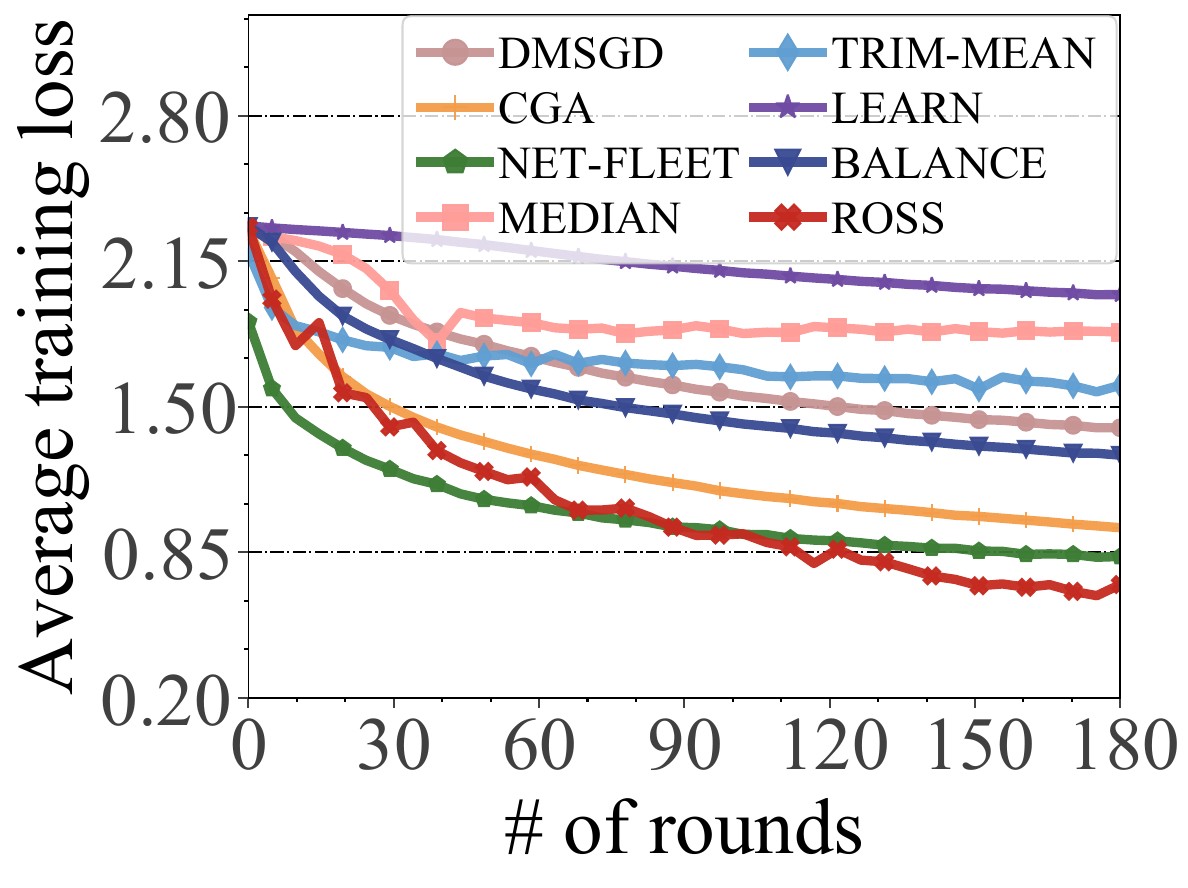}}
        \parbox{.25\textwidth}{\center\scriptsize(c1) Long-tailed ($N=30$)}
        \parbox{.23\textwidth}{\center\scriptsize(c2) Data noise ($N=30$)}
        \parbox{.23\textwidth}{\center\scriptsize(c3) Label noise ($N=30$)}
        \parbox{.25\textwidth}{\center\scriptsize(c4) Gradient poisoning ($N=30$)}
      \caption{Comparison results on CIFAR-10 dataset over fully connected graphs.}
      \label{fig:cifar-loss-full}
      \end{center}
      \end{figure*}

\begin{figure*}[htb!]
      \begin{center}
        \parbox{.24\textwidth}{\center\includegraphics[width=.24\textwidth]{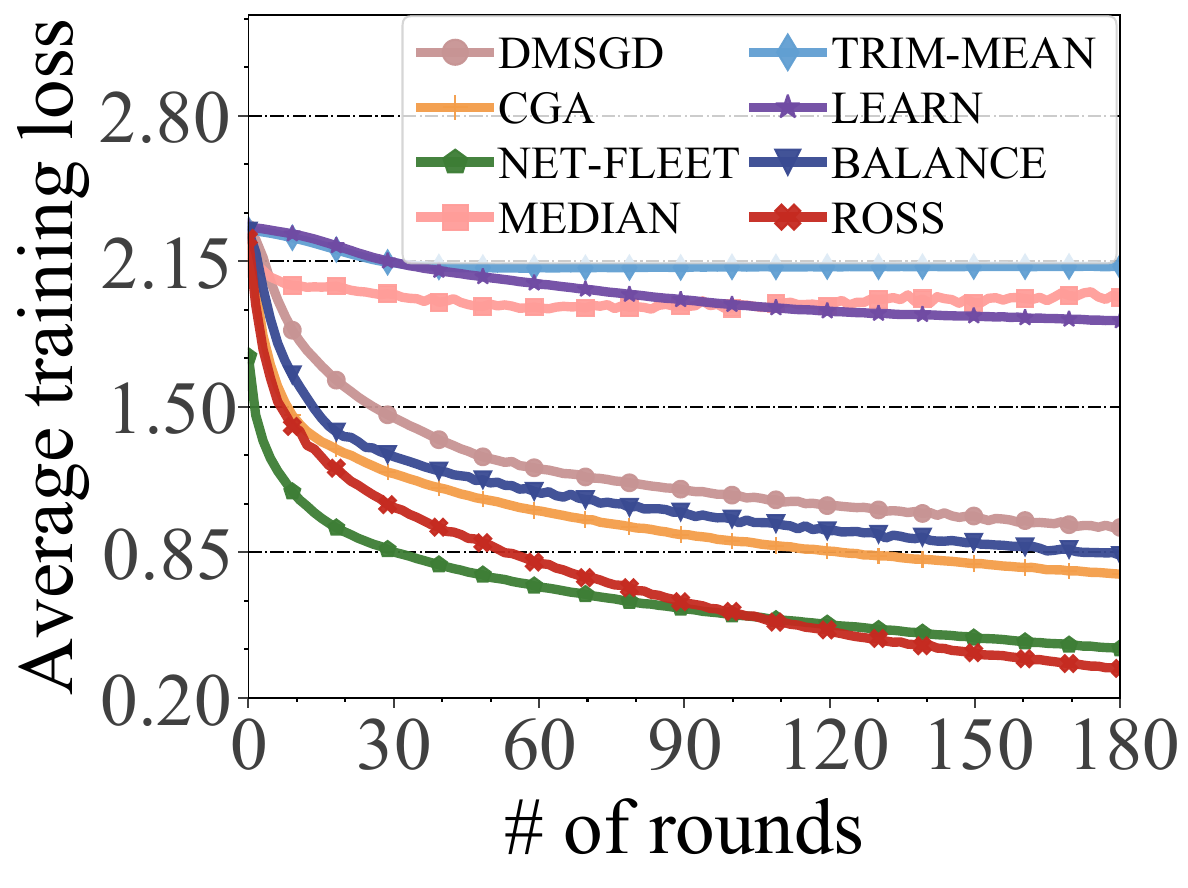}}
        \parbox{.24\textwidth}{\center\includegraphics[width=.24\textwidth]{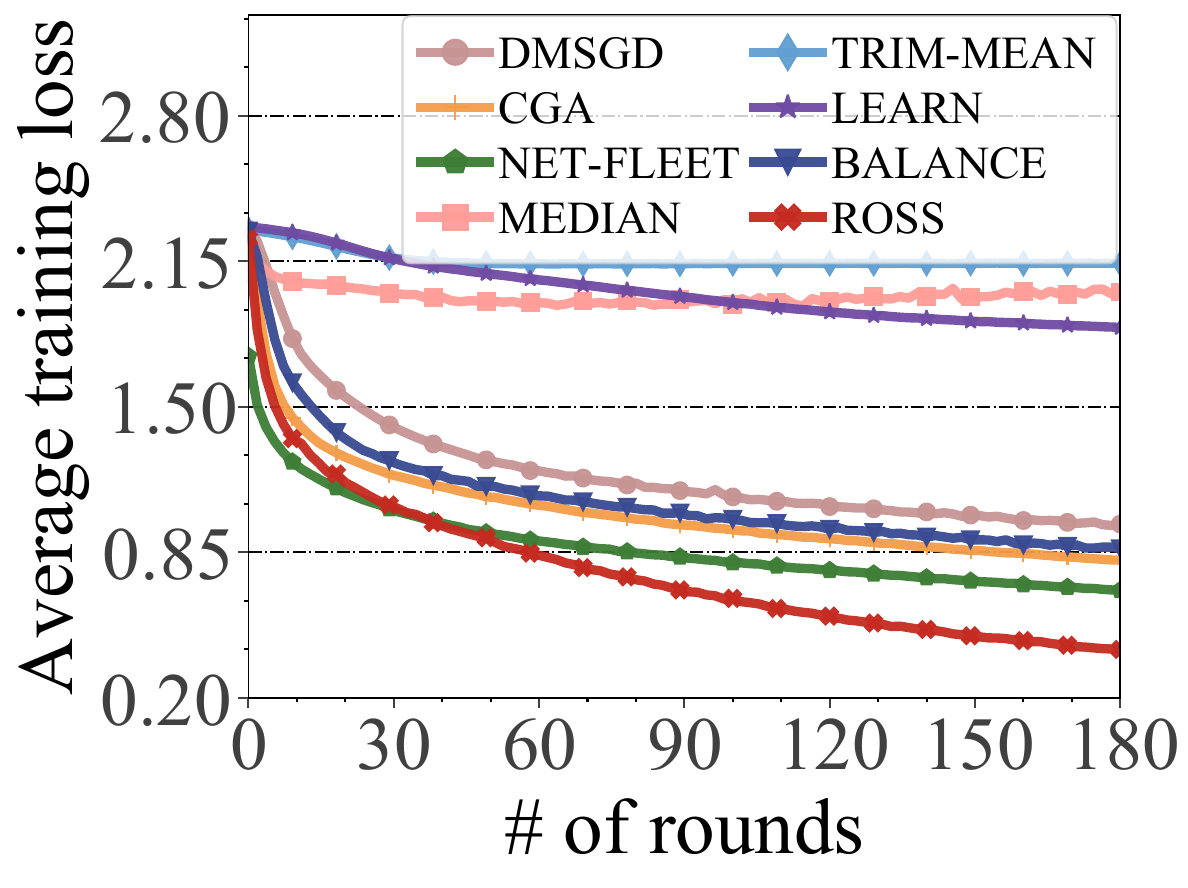}}
        \parbox{.24\textwidth}{\center\includegraphics[width=.24\textwidth]{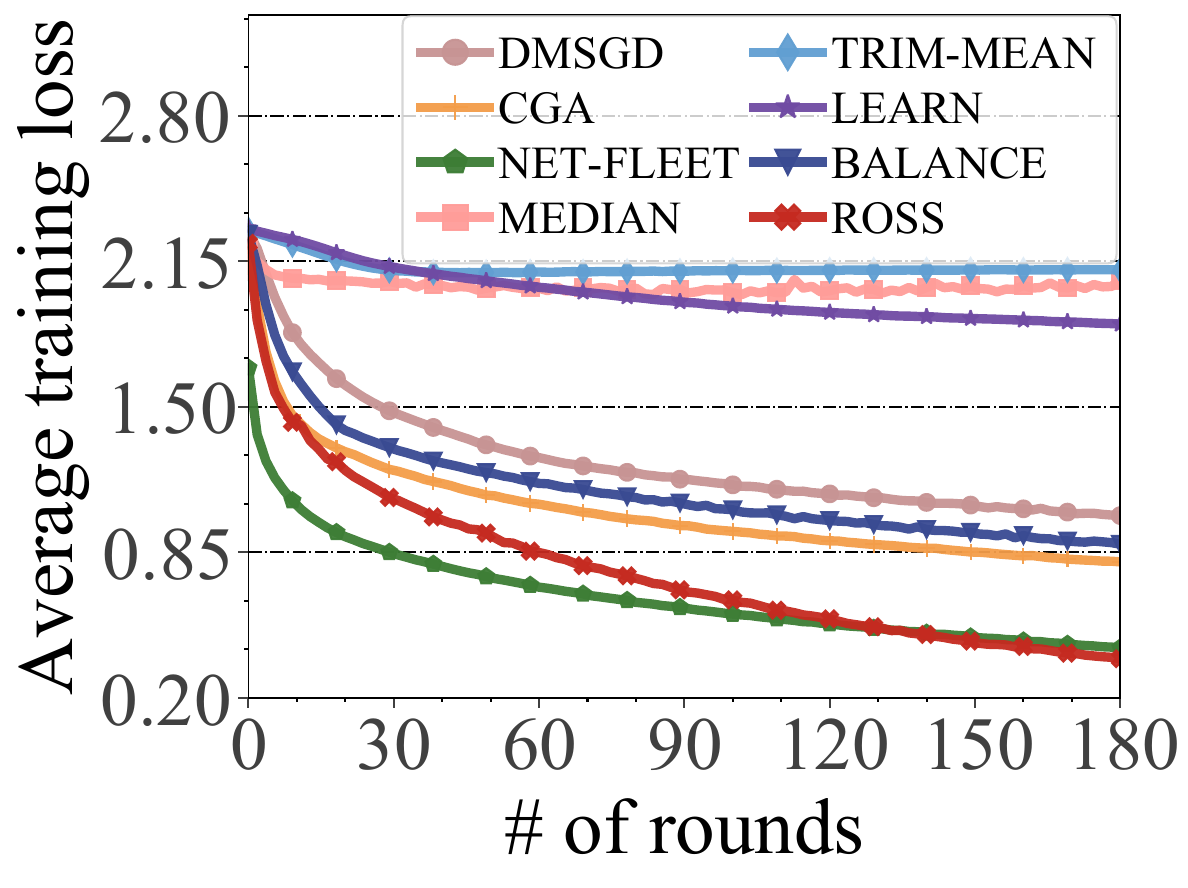}}
        \parbox{.24\textwidth}{\center\includegraphics[width=.24\textwidth]{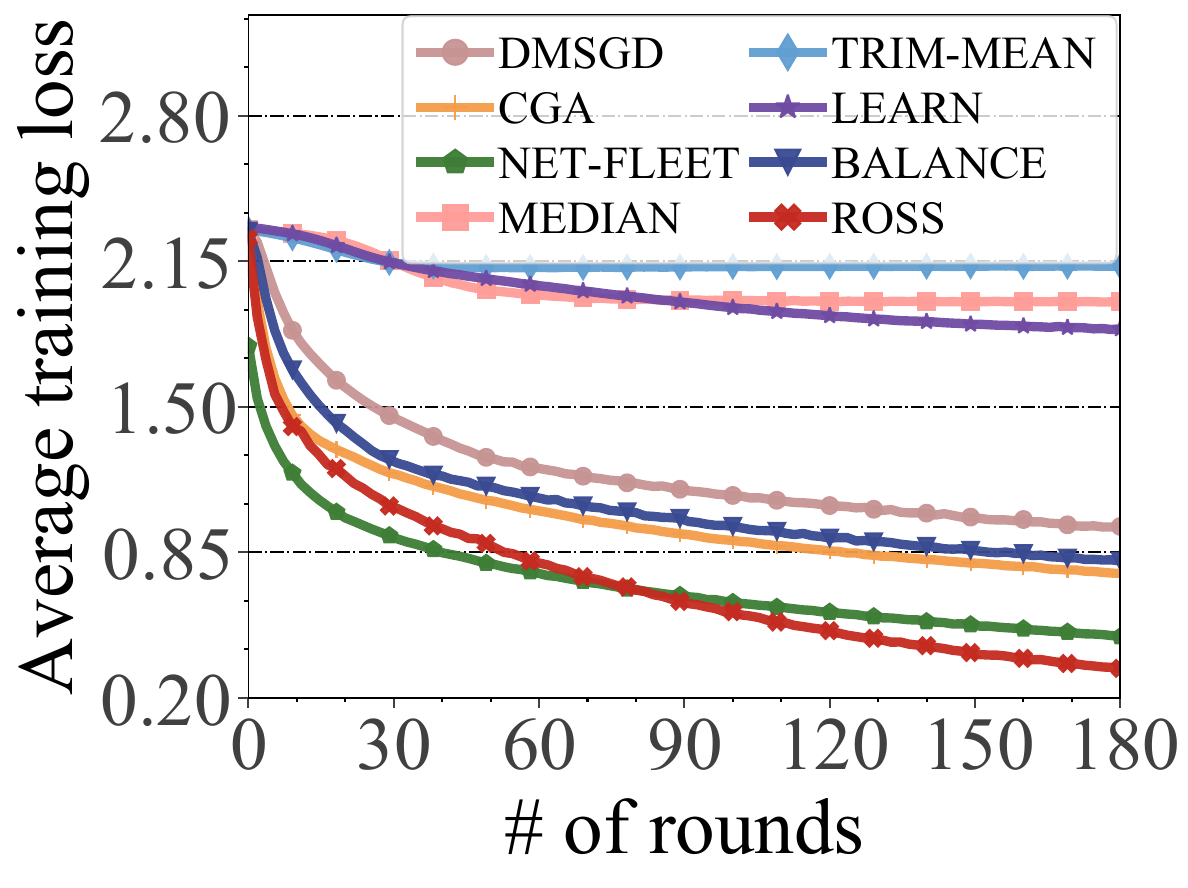}}
        \parbox{.25\textwidth}{\center\scriptsize(a1) Long-tailed ($N=10$)}
        \parbox{.23\textwidth}{\center\scriptsize(a2) Data noise ($N=10$)}
        \parbox{.23\textwidth}{\center\scriptsize(a3) Label noise ($N=10$)}
        \parbox{.25\textwidth}{\center\scriptsize(a4) Gradient poisoning ($N=10$)}
        \parbox{.24\textwidth}{\center\includegraphics[width=.24\textwidth]{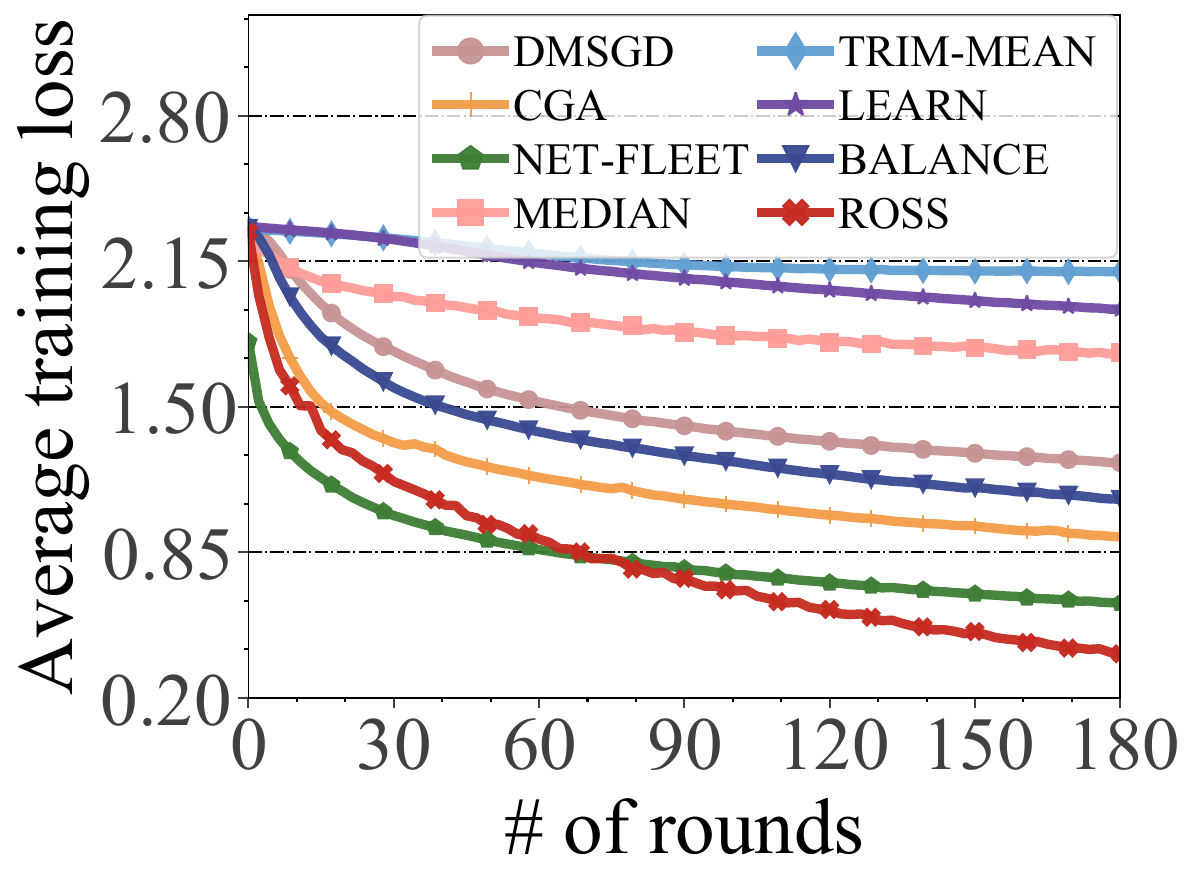}}
        \parbox{.24\textwidth}{\center\includegraphics[width=.24\textwidth]{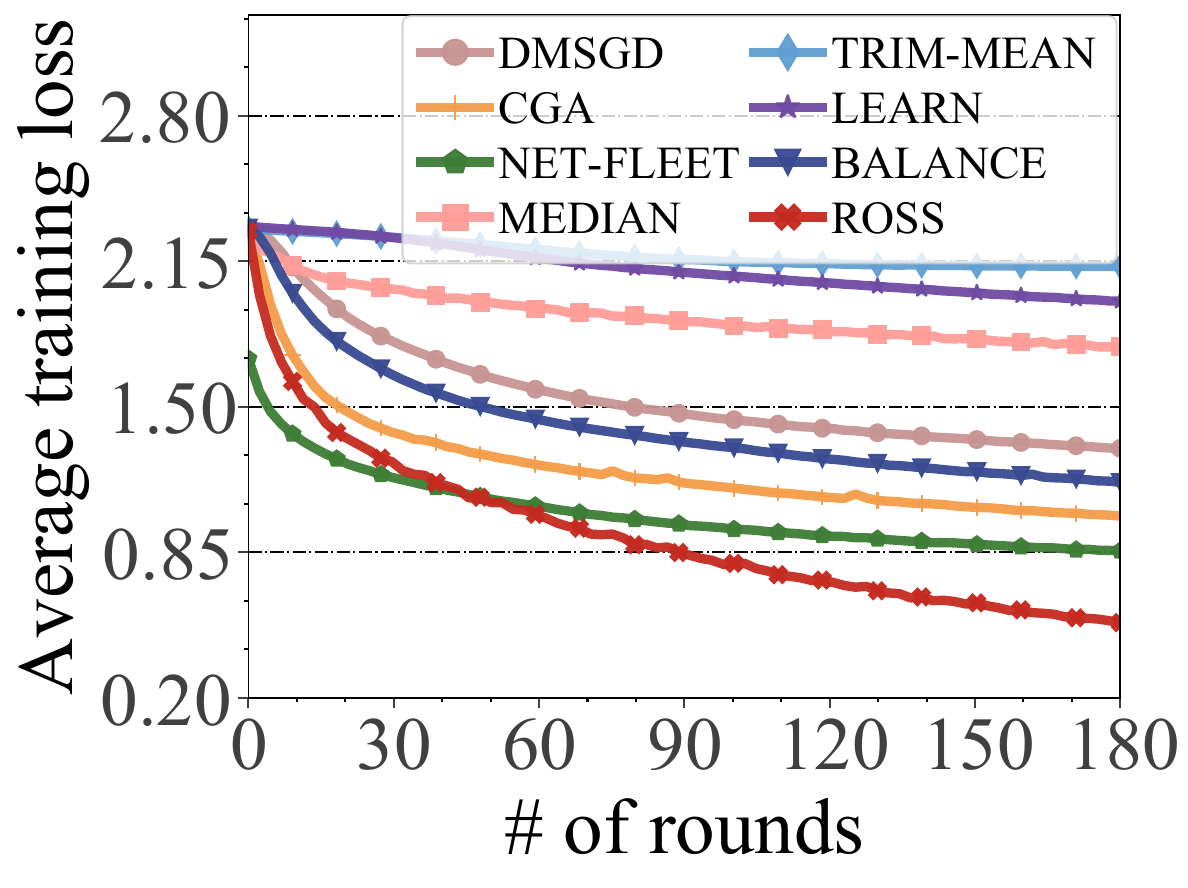}}
        \parbox{.24\textwidth}{\center\includegraphics[width=.24\textwidth]{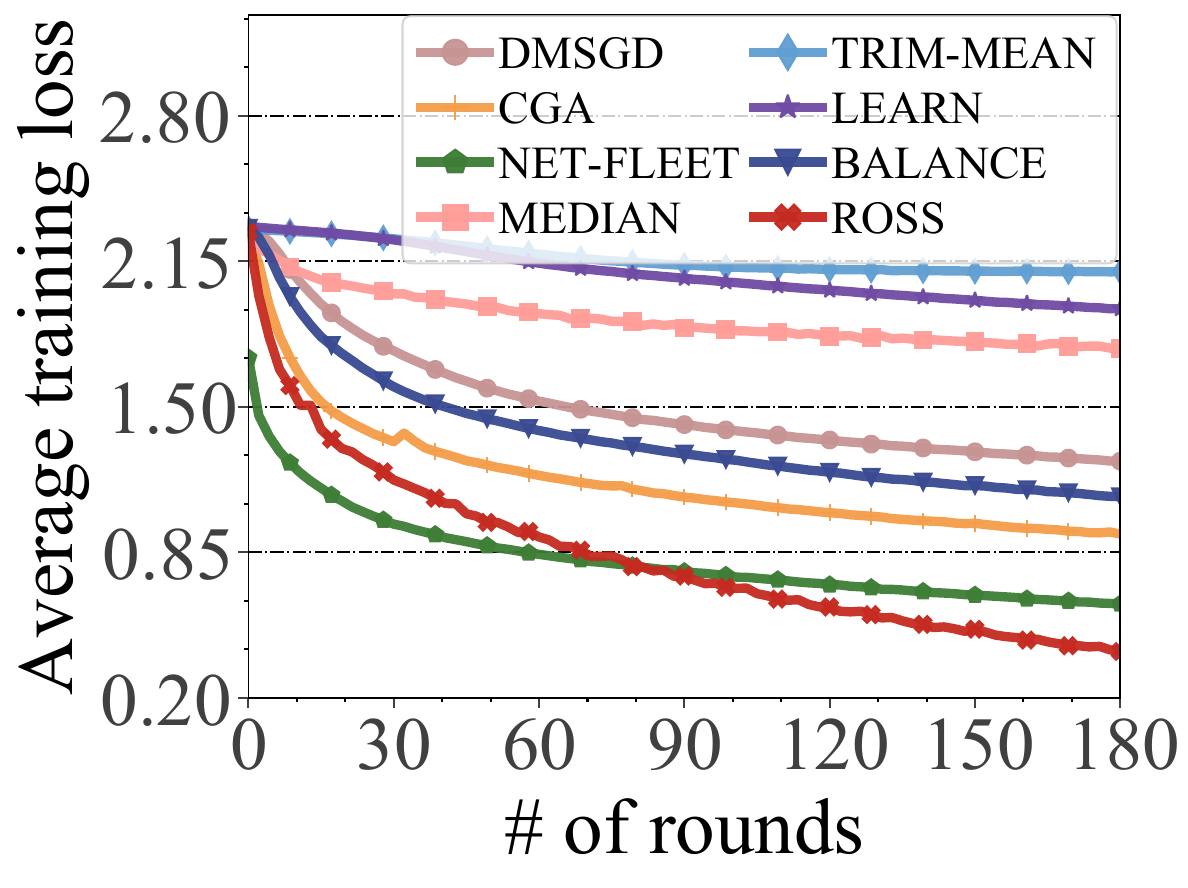}}
        \parbox{.24\textwidth}{\center\includegraphics[width=.24\textwidth]{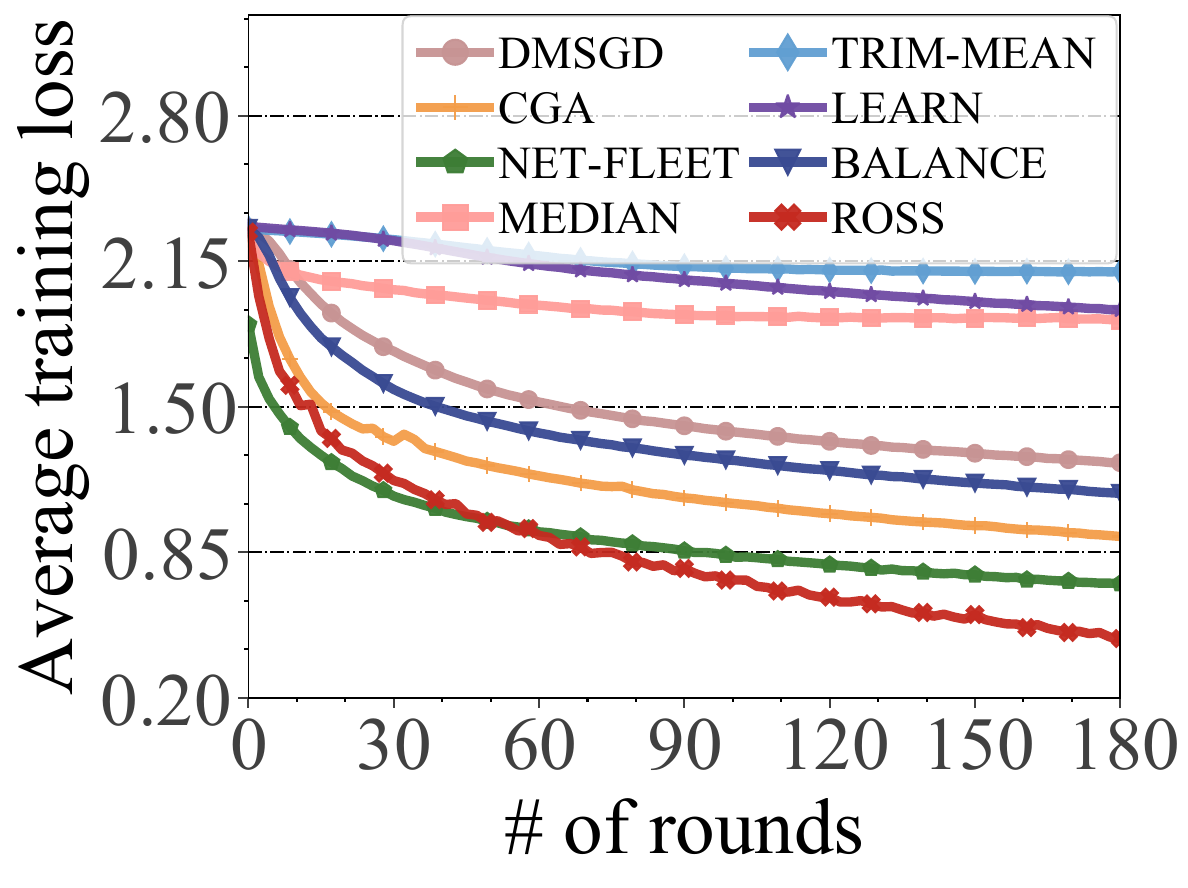}}
        \parbox{.25\textwidth}{\center\scriptsize(b1) Long-tailed ($N=20$)}
        \parbox{.23\textwidth}{\center\scriptsize(b2) Data noise ($N=20$)}
        \parbox{.23\textwidth}{\center\scriptsize(b3) Label noise ($N=20$)}
        \parbox{.25\textwidth}{\center\scriptsize(b4) Gradient poisoning ($N=20$)}
        \parbox{.24\textwidth}{\center\includegraphics[width=.24\textwidth]{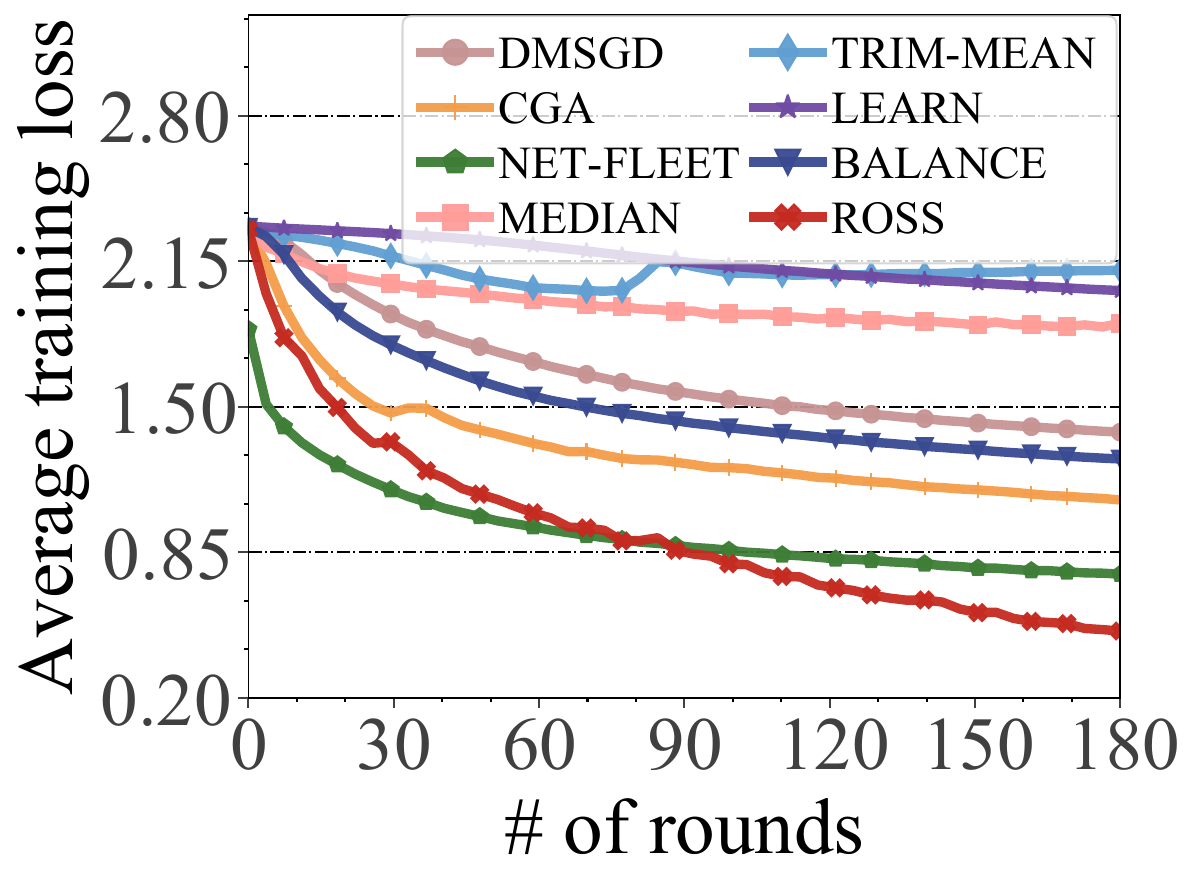}}
        \parbox{.24\textwidth}{\center\includegraphics[width=.24\textwidth]{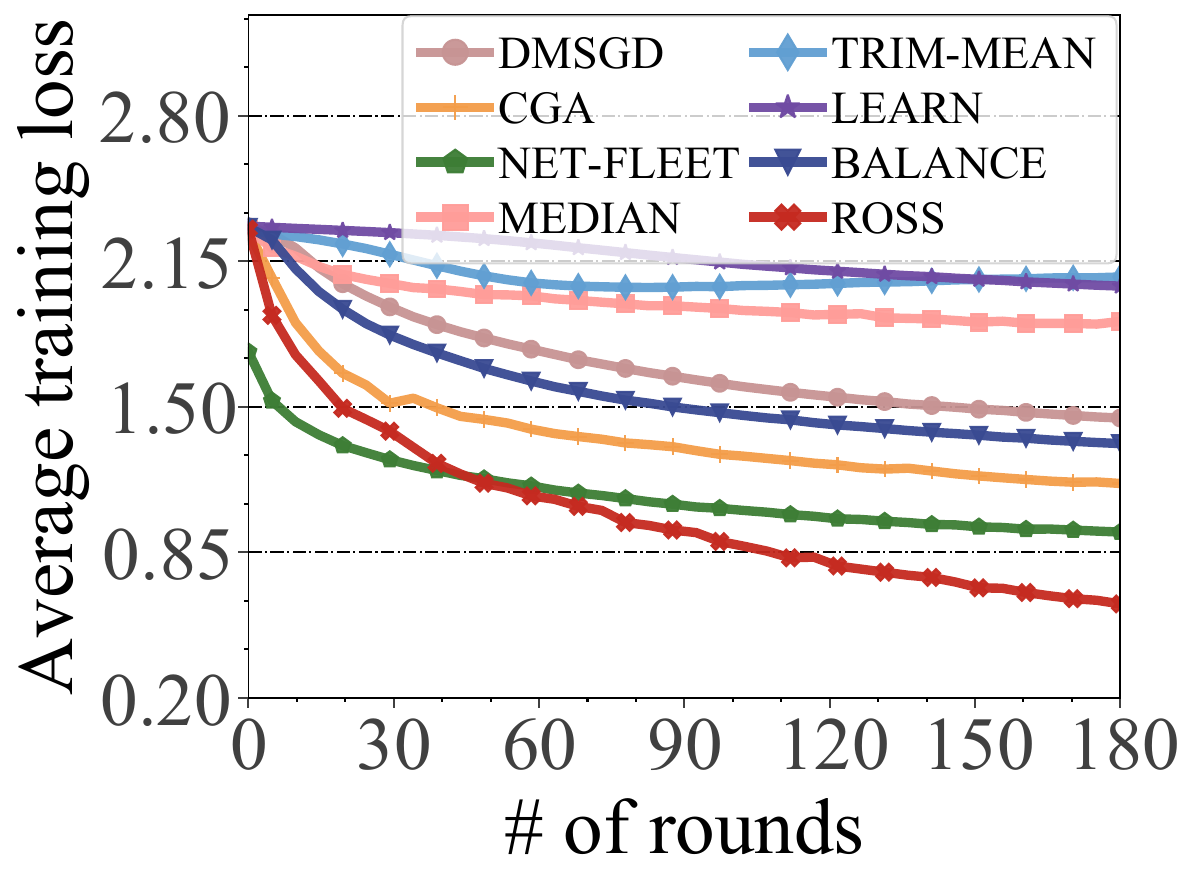}}
        \parbox{.24\textwidth}{\center\includegraphics[width=.24\textwidth]{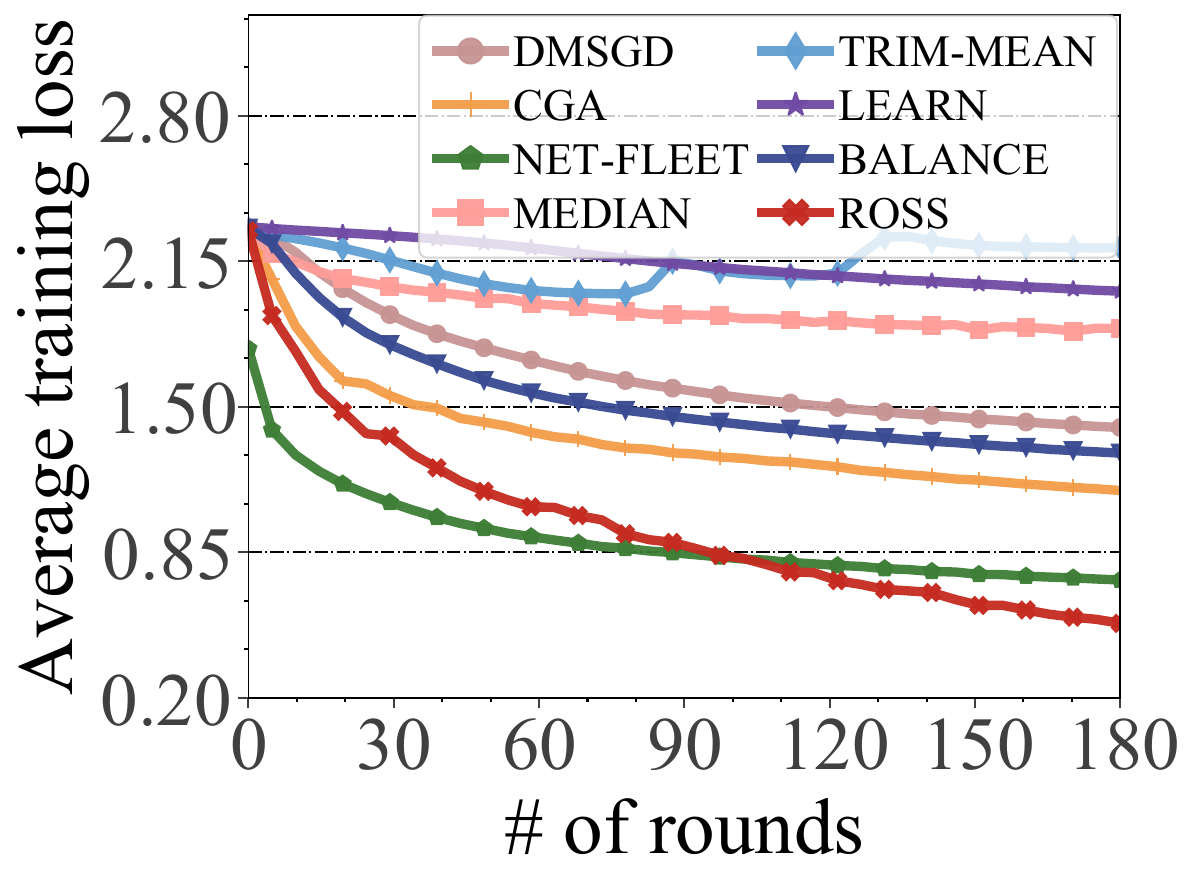}}
        \parbox{.24\textwidth}{\center\includegraphics[width=.24\textwidth]{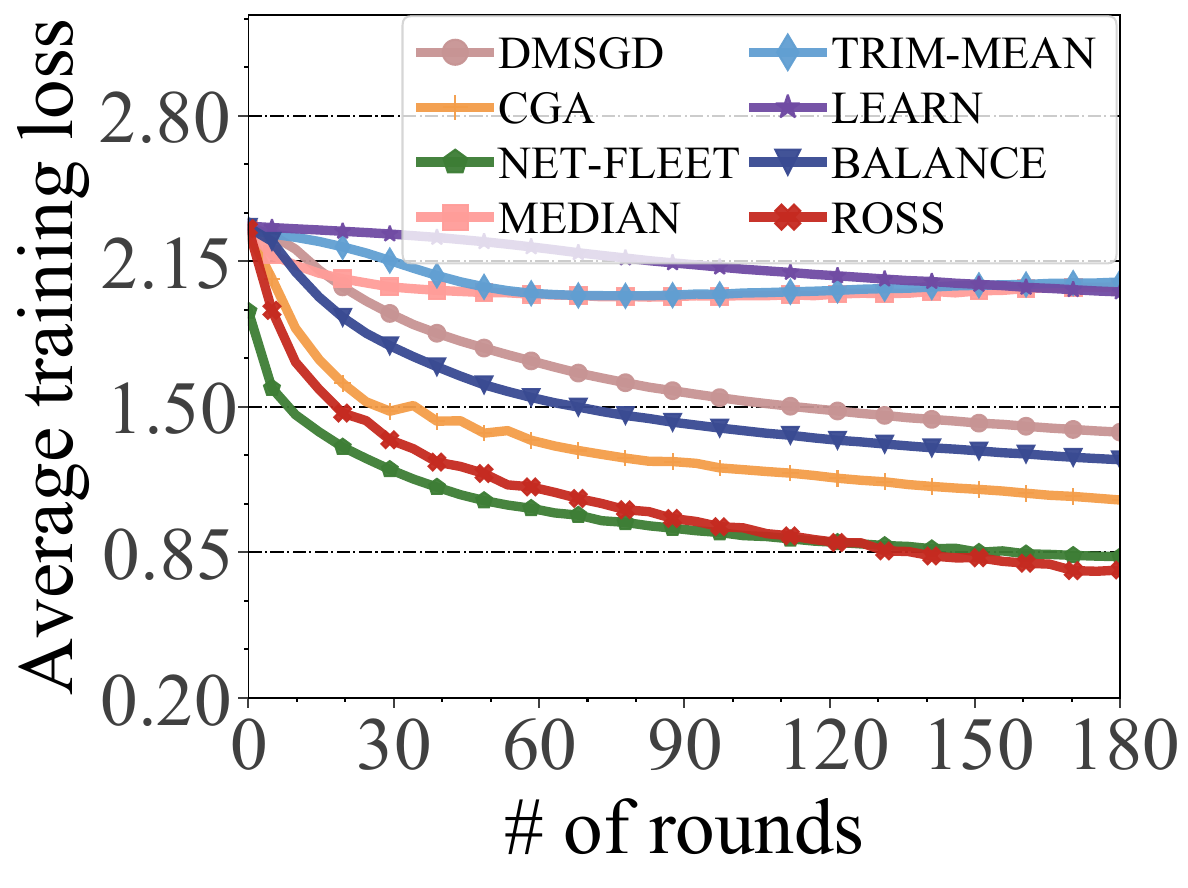}}
        \parbox{.25\textwidth}{\center\scriptsize(c1) Long-tailed ($N=30$)}
        \parbox{.23\textwidth}{\center\scriptsize(c2) Data noise ($N=30$)}
        \parbox{.23\textwidth}{\center\scriptsize(c3) Label noise ($N=30$)}
        \parbox{.25\textwidth}{\center\scriptsize(c4) Gradient poisoning ($N=30$)}
      \caption{Comparison results on CIFAR-10 dataset over bipartite graphs.}
      \label{fig:cifar-loss-bipartite}
      \end{center}
      \end{figure*}

      We next evaluate our ROSS algorithm and the baseline algorithms on the CIFAR-10 dataset. Fig.~\ref{fig:cifar-loss-full} and Fig.~\ref{fig:cifar-loss-bipartite} show the average losses of the algorithms across four settings under different communication topologies. The results indicate that our ROSS algorithm converges faster and achieves lower average loss than DMSGD, CGA, MEDIAN, TRIM-MEAN, LEARN, and BALANCE in all cases. For example, as shown in Fig.~\ref{fig:cifar-loss-full}, with $N=10$ under the long-tailed distribution on a fully connected graph, the average loss of ROSS algorithm is about $0.21$ at convergence; it is $3.1\times$ smaller than CGA algorithm, $4.4\times$ smaller than BALANCE algorithm, $5\times$ smaller than DMSGD algorithm, $6\times$ smaller than TRIM-MEAN, and $7.25\times$ smaller than both MEDIAN and LEARN. Furthermore, when the number of agents increases (e.g., $N=20,30$), our ROSS algorithm maintains a clear advantage over the baselines, especially DMSGD, CGA, MEDIAN, TRIM-MEAN, LEARN, and BALANCE. Similar to the results in Sec.~\ref{ssec:res-mnist}, NET-FLEET algorithm exhibits comparable (and occasionally faster) convergence, but as we show later, its test accuracy is consistently lower than that of ROSS.

\begin{figure*}[htb!]
      \begin{center}
        \parbox{.32\textwidth}{\center\includegraphics[width=.32\textwidth]{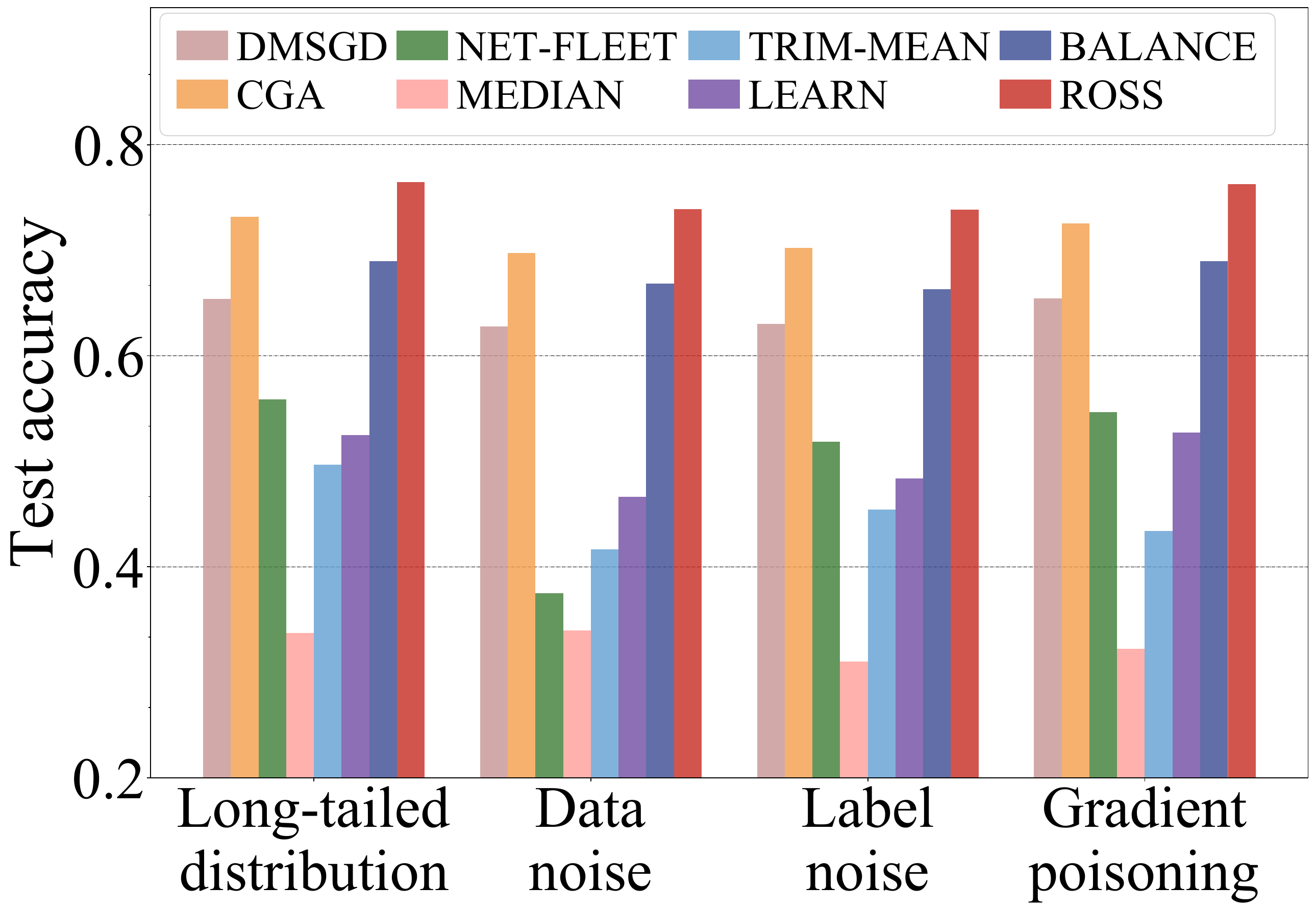}}
        \parbox{.32\textwidth}{\center\includegraphics[width=.32\textwidth]{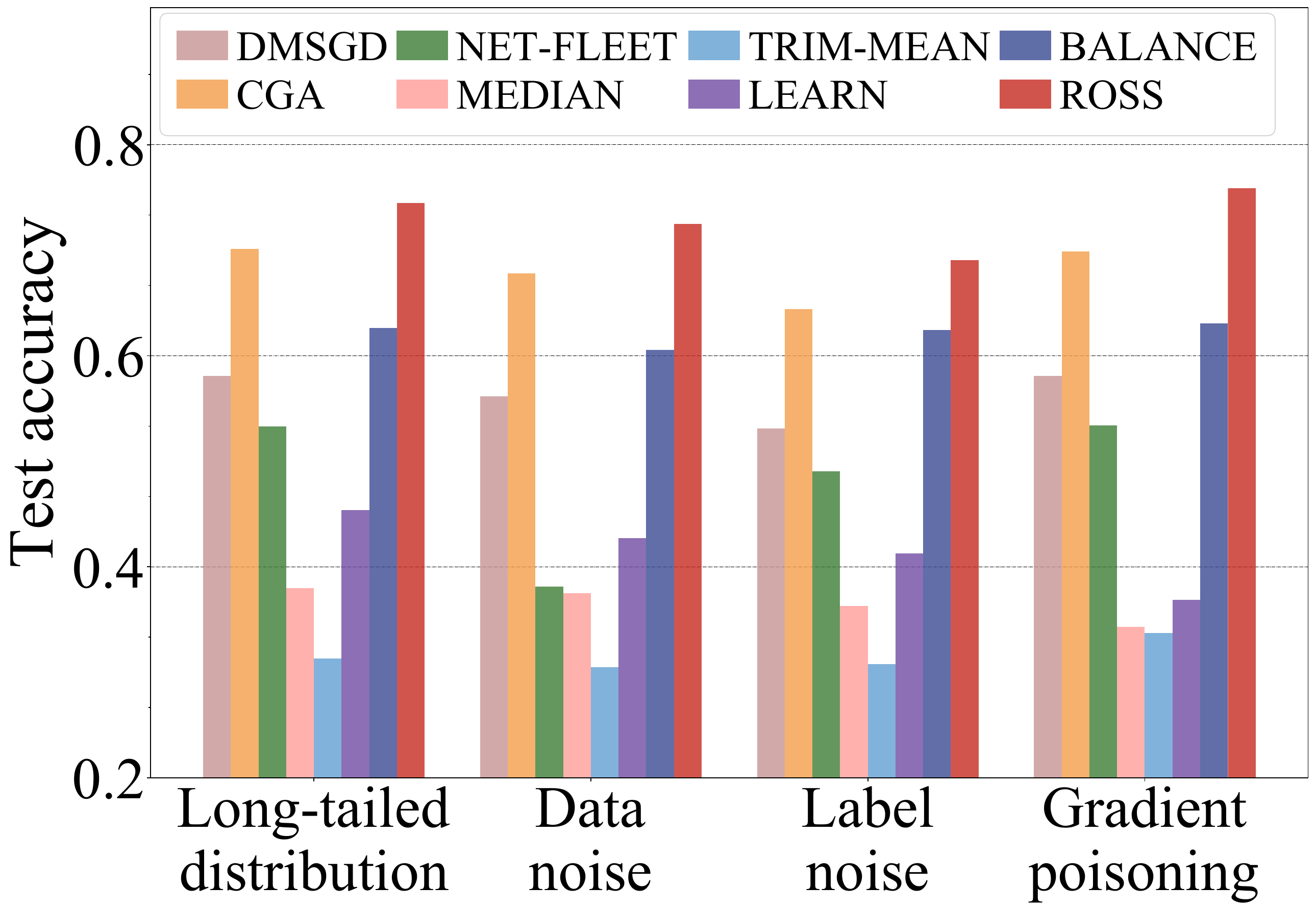}}
        \parbox{.32\textwidth}{\center\includegraphics[width=.32\textwidth]{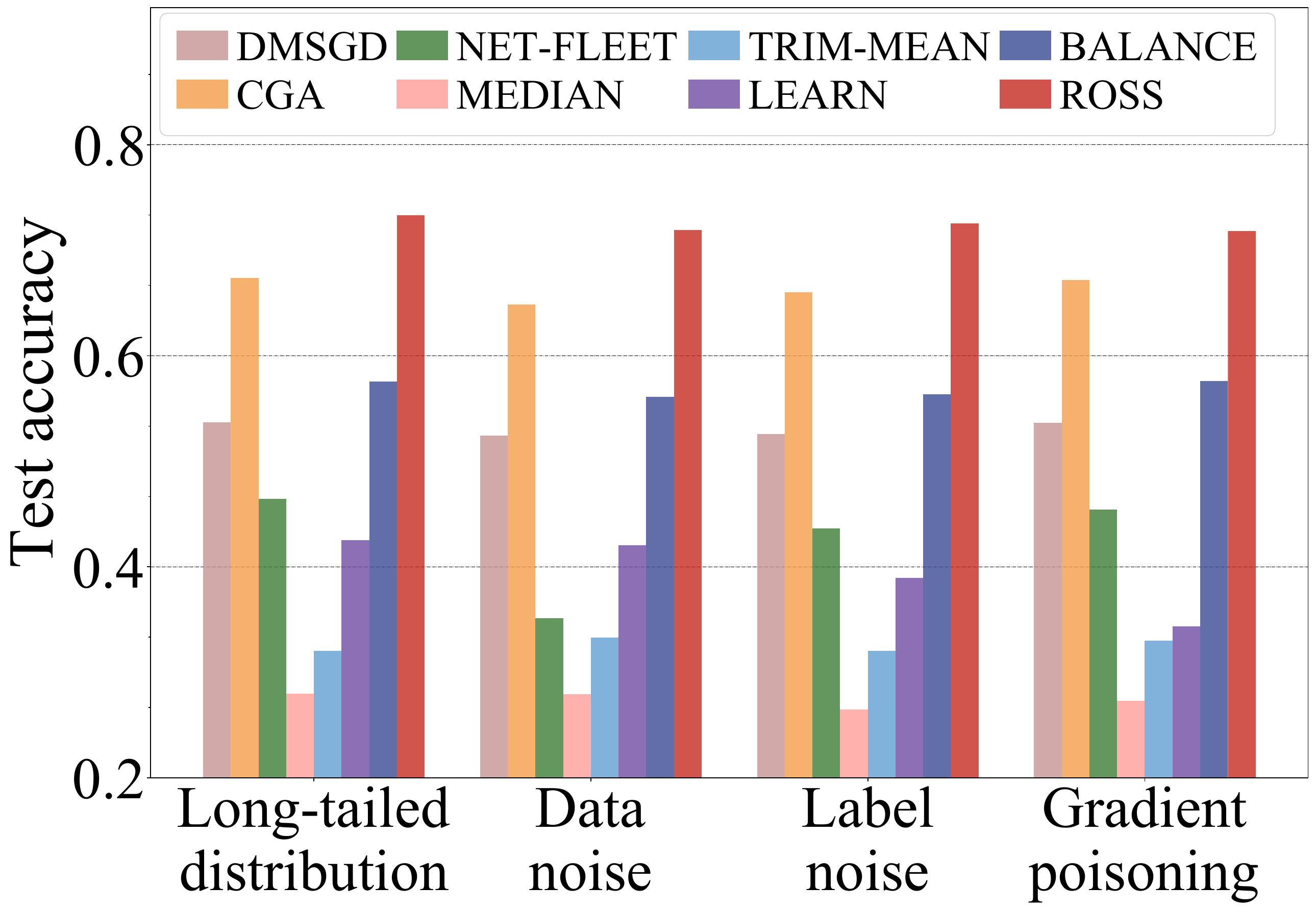}}
        \parbox{.32\textwidth}{\center\scriptsize(a) $N=10$}
        \parbox{.32\textwidth}{\center\scriptsize(b) $N=20$}
        \parbox{.32\textwidth}{\center\scriptsize(c) $N=30$}
      \caption{Comparison results in terms of test accuracy on CIFAR-10 dataset over fully connected graphs.}
      \label{fig:cifar-acc-full}
      \end{center}
      \end{figure*}

\begin{figure*}[htb!]
      \begin{center}
        \parbox{.32\textwidth}{\center\includegraphics[width=.32\textwidth]{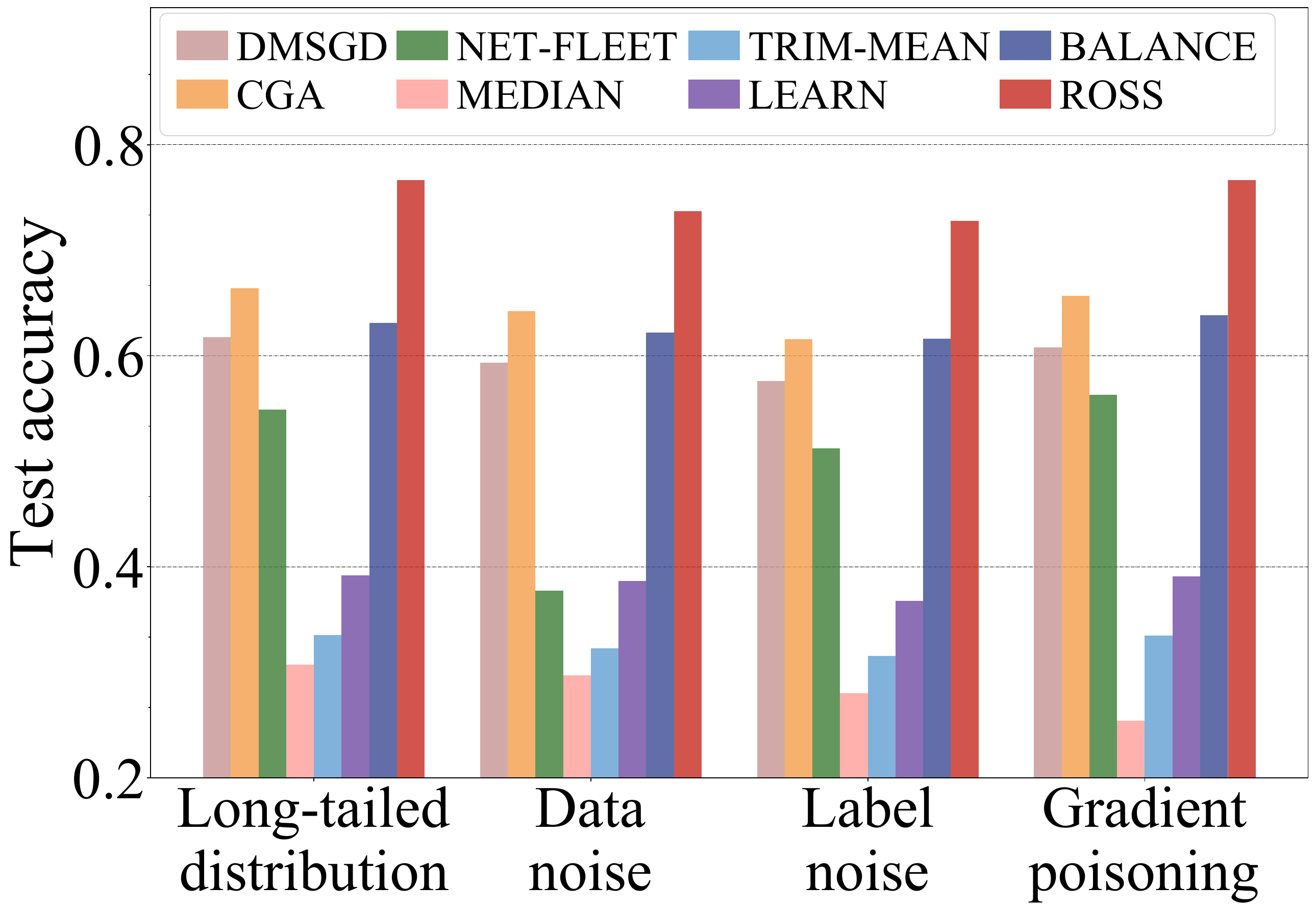}}
        \parbox{.32\textwidth}{\center\includegraphics[width=.32\textwidth]{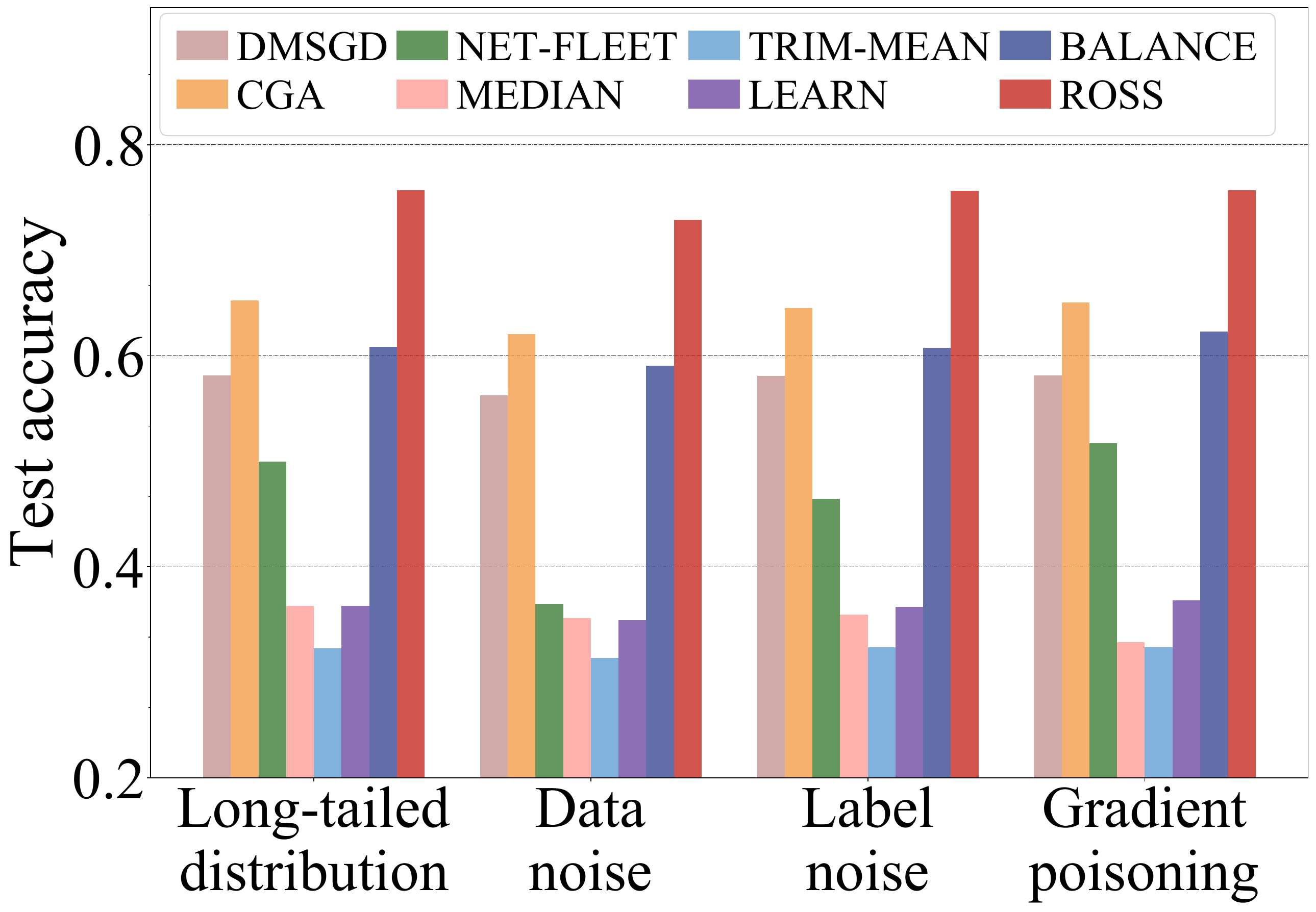}}
        \parbox{.32\textwidth}{\center\includegraphics[width=.32\textwidth]{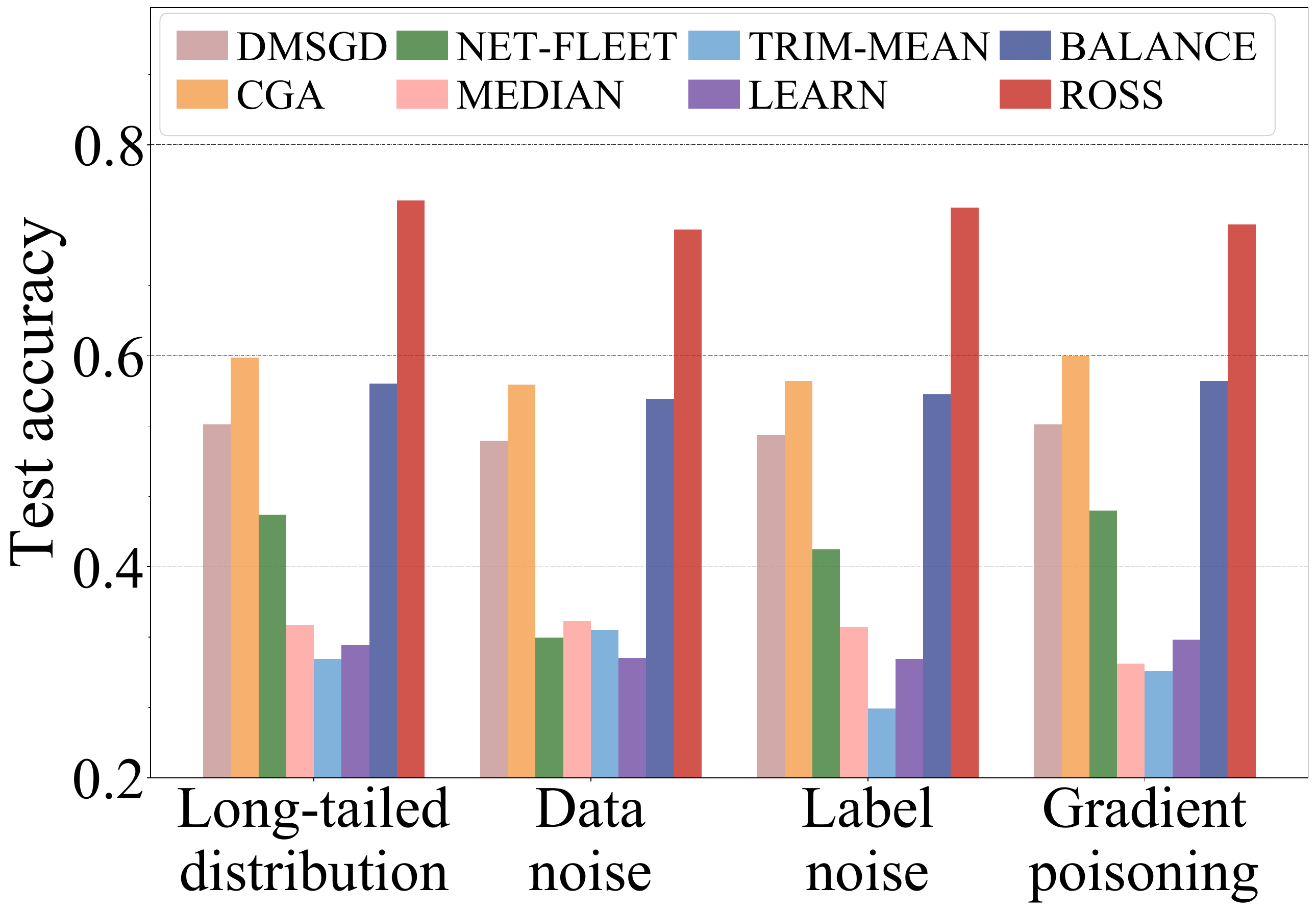}}
        \parbox{.32\textwidth}{\center\scriptsize(a) $N=10$}
        \parbox{.32\textwidth}{\center\scriptsize(b) $N=20$}
        \parbox{.32\textwidth}{\center\scriptsize(c) $N=30$}
      \caption{Comparison results in terms of test accuracy on CIFAR-10 dataset over bipartite graphs.}
      \label{fig:cifar-acc-bipartite}
      \end{center}
      \end{figure*}

      The prediction accuracies of the different algorithms are reported in Fig.~\ref{fig:cifar-acc-full}-\ref{fig:cifar-acc-bipartite}. The results show that ROSS consistently achieves higher test accuracy than all baselines across settings. For example, with $N=10$ on fully connected graphs under data noise, ROSS reaches a test accuracy of $0.74$, which is $1.06-2.18\times$ higher than the baselines. Moreover, the baseline algorithms suffer significant accuracy drops as the number of agents increases. Particularly, when $N=30$, ROSS achieves test accuracies that are $1.1-2.57\times$ higher than those of the baselines. Importantly, ROSS maintains its superiority even on sparse communication graphs. Specifically, with $N=30$ on bipartite graphs under data noise, ROSS attains a test accuracy of $0.71$, while the baselines achieve only $0.31-0.56$.

\section{Conclusion and Future Work} \label{sec:con}
  In this paper, in order to improve the robustness of decentralized learning to data challenges, we propose ROSS, a decentralized stochastic learning algorithm based on Shapley values. Specifically, we innovate in leveraging the notion of Shapley values to measure the contributions of the cross-gradients at each agent for calculating the weighted aggregation of the gradients. We not only provide a solid convergence analysis for our proposed algorithm, but also conduct extensive experiments on two real-world datasets to verify the practical effectiveness of our algorithm.

  As noted above, cross-gradient information is crucial for ensuring the performance of our algorithm under data challenges, but it also introduces additional communication cost. As future work, we plan to incorporate gradient/model compression or pruning~\cite{KarimireddyRSJ-ICML19, CuiSZL-INFOCOM22, JiangXXWLQQ-TMC24} to further reduce this overhead. Furthermore, recent studies have introduced stronger model poisoning attacks~\cite{FangCJG-Security20,BaruchBG-NIPS19}. The model poisoning attacks typically target the stage of local model aggregation, e.g., the aggregation of momentum variables and local models in our case. Since it is inherently difficult for an agent to distinguish a malicious neighbor from a benign one with different data distribution, ensuring robustness against both non-IID data åand adversarial data/model poisoning remains a significant challenge from both theoretical and practical perspectives~\cite{GuoZYXMXL-TCSVT22,FangZHKLLLG-CCS24}. To address this, we are on the way of exploring extensions of ROSS that combine Shapley values with model similarity measures, as suggested in \cite{FangZHKLLLG-CCS24}, in order to enhance robustness particularly at the stage of model aggregation.

\bibliographystyle{unsrt}
\bibliography{ref.bib}

\appendices

\section{Proof of \textbf{Lemma}~\ref{lem:diffhatgrad}}
  This lemma can be proved as follows
  \begin{align*}
    & \mathbb{E} \left[ \left\| \frac{1}{N} \sum^N_{i=1} \left( \bar{g}^{[t]}_i - g^{[t]}_{i,i} \right) \right\|^2 \right]  \\
    = & \mathbb{E} \left[ \left\| \frac{1}{N} \sum^N_{i=1} \sum_{j \in \mathcal{N}_i} \pi^{[t]}_{i,j} \nabla F_j \left( x^{[t-1]}_i; \xi_{j,t} \right) - \frac{1}{N} \sum^N_{i=1} \nabla F_i \left( x^{[t-1]}_i; \xi_{i,t} \right) \right\|^2 \right]  \\
    \stackrel{\scriptsize{\circled{1}}} \leq & 2\mathbb{E} \left[ \left\| \frac{1}{N} \sum^N_{i=1} \sum_{j \in \mathcal{N}_i} \frac{\hat{\varphi}^{[t]}_{i,j}}{\omega_{i,j} \sum_{k\in\mathcal{N}_i} \hat{\varphi}^{[t]}_{i,k}} \nabla F_j \left( x^{[t-1]}_i; \xi_{j,t} \right) \right\|^2 \right] + 2\mathbb{E} \left[ \left\| \frac{1}{N} \sum^N_{i=1} \nabla F_i \left( x^{[t-1]}_i; \xi_{i,t} \right) \right\|^2 \right]  \\
    \stackrel{\scriptsize{\circled{2}}} \leq & \frac{2}{N} \sum^N_{i=1} \sum_{j \in \mathcal{N}_i} \omega_{i,j} \mathbb{E} \left[ \left\| \frac{\hat{\varphi}^{[t]}_{i,j}}{\omega^2_{i,j} \sum_{k\in\mathcal{N}_i} \hat{\varphi}^{[t]}_{i,k}} \right\|^2 \left\| \nabla F_j \left( x^{[t-1]}_i; \xi_{j,t} \right)  \right\|^2 \right] + 2\mathbb{E} \left[ \left\| \frac{1}{N} \sum^N_{i=1} \nabla F_i \left( x^{[t-1]}_i; \xi_{i,t} \right) \right\|^2 \right]   \\
    = & \frac{2}{N} \sum^N_{i=1} \sum_{j \in \mathcal{N}_i} \frac{\omega_{i,j}}{\omega^4_{i,j}} \mathbb{E} \left[ \left\| \frac{\hat{\varphi}^{[t]}_{i,j}}{ \sum_{k\in\mathcal{N}_i} \hat{\varphi}^{[t]}_{i,k}} \right\|^2 \left\| \nabla F_j \left( x^{[t-1]}_i; \xi_{j,t} \right)  \right\|^2 \right] + 2\mathbb{E} \left[ \left\| \frac{1}{N} \sum^N_{i=1} \nabla F_i \left( x^{[t-1]}_i; \xi_{i,t} \right) \right\|^2 \right]   \\
    \stackrel{\scriptsize{\circled{3}}} \leq & \frac{2}{N} \sum^N_{i=1} \sum_{j \in \mathcal{N}_i}  \frac{\omega_{i,j} \cdot  \left(\hat{\varphi}^{[t]}_{\max}\right)^2 }{\omega^4_{i,j}} \mathbb{E} \left[ \left\| \nabla F_j \left(x^{[t-1]}_i; \xi_{j,t} \right)  \right\|^2 \right]  + 2 \mathbb{E} \left[ \left\| \frac{1}{N} \sum^N_{i=1} \nabla f_i \left( x^{[t-1]}_i \right) \right\|^2 \right]  \\
    & + 2 \mathbb{E} \left[ \left\| \frac{1}{N} \sum^N_{i=1} \left( 
    \nabla F_i \left(x^{[t-1]}_i; \xi_{i,t} \right) - \nabla f_i \left(x^{[t-1]}_i \right) \right) \right\|^2 \right]   \\
    \leq & \frac{2}{N} \sum^N_{i=1} \sum_{j \in \mathcal{N}_i}  \frac{\omega_{i,j} \cdot \left(\hat{\varphi}^{[t]}_{\max}\right)^2 }{\omega^4_{i,j}} \mathbb{E} \left[ \left\| \nabla F_j \left(x^{[t-1]}_i; \xi_{j,t} \right) - \nabla f_j \left(x^{[t-1]}_i \right) \right\|^2 \right] + \frac{2}{N} \sum^N_{i=1} \sum_{j \in \mathcal{N}_i}  \frac{\omega_{i,j} \cdot \left(\hat{\varphi}^{[t]}_{\max}\right)^2 }{\omega^4_{i,j}} \mathbb{E} \left[ \left\| \nabla f_j \left(x^{[t-1]}_i \right) \right\|^2 \right]  \\
    & + 2 \mathbb{E} \left[ \left\| \frac{1}{N} \sum^N_{i=1} \left( 
    \nabla F_i \left(x^{[t-1]}_i; \xi_{i,t} \right) - \nabla f_i \left(x^{[t-1]}_i \right) \right) \right\|^2 \right] + 2 \mathbb{E} \left[ \left\| \frac{1}{N} \sum^N_{i=1} \nabla f_i \left(x^{[t-1]}_i \right) \right\|^2 \right]  \\
    = & \frac{2}{N} \sum^N_{i=1} \sum_{j \in \mathcal{N}_i}  \frac{\omega_{i,j} \cdot \left(\hat{\varphi}^{[t]}_{\max}\right)^2 }{\omega^4_{i,j}} \mathbb{E} \left[ \left\| \nabla F_j \left(x^{[t-1]}_i; \xi_{j,t} \right) - \nabla f_j \left(x^{[t-1]}_i \right) \right\|^2 \right]  \\
    & + \frac{2}{N} \sum^N_{i=1} \sum_{j \in \mathcal{N}_i} \frac{\omega_{i,j} \cdot \left(\hat{\varphi}^{[t]}_{\max}\right)^2 }{\omega^4_{i,j}} \mathbb{E} \left[ \left\|
    \begin{aligned} &  \nabla f_j \left(x^{[t-1]}_i \right) - \nabla f_j \left(\bar{x}^{[t-1]} \right) + \nabla f_j \left(\bar{x}^{[t-1]} \right) - \frac{1}{N} \sum^N_{n=1} \nabla f_n \left(\bar{x}^{[t-1]} \right) \\ & + \frac{1}{N} \sum^N_{n=1} \nabla f_n \left(\bar{x}^{[t-1]} \right)  - \frac{1}{N} \sum^N_{n=1} \nabla f_n \left(x^{[t-1]}_n \right) + \frac{1}{N} \sum^N_{n=1} \nabla f_n \left(x^{[t-1]}_n \right)  \end{aligned} \right\|^2 \right] \\
    & + 2 \mathbb{E} \left[ \left\| \frac{1}{N} \sum^N_{i=1} \left( 
    \nabla F_i \left(x^{[t-1]}_i; \xi_{i,t} \right) - \nabla f_i \left(x^{[t-1]}_i \right) \right) \right\|^2 \right] + 2 \mathbb{E} \left[ \left\| \frac{1}{N} \sum^N_{i=1} \nabla f_i \left(x^{[t-1]}_i \right) \right\|^2 \right]  \\
    \stackrel{\scriptsize{\circled{4}}} \leq & \frac{2}{N} \sum^N_{i=1} \sum_{j \in \mathcal{N}_i} \frac{\omega_{i,j} \cdot \left(\hat{\varphi}^{[t]}_{\max}\right)^2 }{\omega^4_{i,j}} \mathbb{E} \left[ \left\| \nabla F_j \left(x^{[t-1]}_i; \xi_{j,t} \right) - \nabla f_j \left(x^{[t-1]}_i \right) \right\|^2 \right]  \\
    & + \frac{8}{N} \sum^N_{i=1} \sum_{j \in \mathcal{N}_i} \frac{\omega_{i,j} \cdot \left(\hat{\varphi}^{[t]}_{\max}\right)^2
    }{\omega^4_{i,j}} \mathbb{E} \left[ \left\| \nabla f_j \left(x^{[t-1]}_i \right) - \nabla f_j \left(\bar{x}^{[t-1]} \right) \right\|^2 \right]    \\
    & + \frac{8}{N} \sum^N_{i=1} \sum_{j \in \mathcal{N}_i} \frac{\omega_{i,j} \cdot \left(\hat{\varphi}^{[t]}_{\max}\right)^2 }{\omega^4_{i,j}} \mathbb{E} \left[ \left\| \nabla f_j \left(\bar{x}^{[t-1]} \right) - \frac{1}{N} \sum^N_{n=1} \nabla f_n \left(\bar{x}^{[t-1]} \right) \right\|^2 \right]  \\
    & + \frac{8}{N} \sum^N_{i=1} \sum_{j \in \mathcal{N}_i} \frac{\omega_{i,j}\cdot \left(\hat{\varphi}^{[t]}_{\max}\right)^2 }{\omega^4_{i,j}} \mathbb{E} \left[ \left\| \frac{1}{N} \sum^N_{n=1} \nabla f_n \left(\bar{x}^{[t-1]} \right) - \frac{1}{N} \sum^N_{n=1} \nabla f_n \left(x^{[t-1]}_n \right)  \right\|^2 \right]  \\
    & + \frac{8}{N} \sum^N_{i=1} \sum_{j \in \mathcal{N}_i} \frac{\omega_{i,j}\cdot \left(\hat{\varphi}^{[t]}_{\max}\right)^2 }{\omega^4_{i,j}} \mathbb{E} \left[ \left\| \frac{1}{N} \sum^N_{n=1}  \nabla f_n \left(x^{[t-1]}_n \right)  \right\|^2 \right]  \\
    & + \frac{2}{N^2} \sum^N_{i=1} \mathbb{E} \left[ \left\| \nabla F_i \left(x^{[t-1]}_i; \xi_{i,t} \right) - \nabla f_i \left(x^{[t-1]}_i \right) \right\|^2 \right] + 2 \mathbb{E} \left[ \left\| \frac{1}{N} \sum^N_{i=1} \nabla f_i \left(x^{[t-1]}_i \right) \right\|^2 \right]  \\
    \stackrel{\scriptsize{\circled{5}}}\leq &  \frac{2}{N} \sum^N_{i=1} \sum_{j \in \mathcal{N}_i} \frac{\omega_{i,j} \cdot \left(\hat{\varphi}^{[t]}_{\max}\right)^2 }{\omega^4_{i,j}} \sigma^2 + \frac{8}{N} \sum^N_{i=1} \sum_{j \in \mathcal{N}_i} \frac{\omega_{i,j} \cdot \left(\hat{\varphi}^{[t]}_{\max}\right)^2
    }{\omega^4_{i,j}} \varsigma^2   \\
    & + \frac{8}{N} \sum^N_{i=1} \sum_{j \in \mathcal{N}_i} \frac{\omega_{i,j} \cdot \left(\hat{\varphi}^{[t]}_{\max}\right)^2 }{\omega^4_{i,j}} L^2 \mathbb{E} \left[ \left\| x^{[t-1]}_i - \bar{x}^{[t-1]} \right\|^2 \right]   \\
    & + \frac{8}{N} \sum^N_{i=1} \sum_{j \in \mathcal{N}_i} \frac{\omega_{i,j} \cdot \left(\hat{\varphi}^{[t]}_{\max}\right)^2 }{\omega^4_{i,j}} \frac{1}{N} \sum^N_{n=1} L^2 \mathbb{E} \left[ \left\| \bar{x}^{[t-1]} - x^{[t-1]}_n \right\|^2 \right] \\
    & + \frac{8}{N} \sum^N_{i=1} \sum_{j \in \mathcal{N}_i} \frac{\omega_{i,j} \cdot \left(\hat{\varphi}^{[t]}_{\max}\right)^2 }{\omega^4_{i,j}} \mathbb{E} \left[ \left\| \frac{1}{N} \sum^N_{n=1}  \nabla f_n \left(x^{[t-1]}_n \right)  \right\|^2 \right]   \\
    & + \frac{2}{N^2} \sum^N_{i=1} \sigma^2 +  2 \mathbb{E} \left[ \left\| \frac{1}{N} \sum^N_{i=1} \nabla f_i \left(x^{[t-1]}_i \right) \right\|^2 \right] \\
    \stackrel{\scriptsize{\circled{6}}} \leq & \frac{2\sigma^2\hat{\varphi}^2_{\max}}{\omega^4_{\min}} + \frac{2\sigma^2}{N} + \frac{8\varsigma^2\hat{\varphi}^2_{\max}}{\omega^4_{\min}} + \frac{16L^2\hat{\varphi}^2_{\max}} {N \omega^4_{\min}} \sum^N_{i=1} \mathbb{E} \left[ \left\| x^{[t-1]}_i - \bar{x}^{[t-1]} \right\|^2 \right]
    + \left( \frac{8\hat{\varphi}^2_{\max}}{\omega^4_{\min}} + 2 \right) \mathbb{E} \left[ \left\| \frac{1}{N} \sum^N_{i=1} \nabla f_i \left(x^{[t-1]}_i \right)  \right\|^2 \right]
  \end{align*}
  where $\scriptsize{\circled{1}}$ follows from the basic inequality $\|a + b\|^2 \leq 2\|a\|^2 + 2\|b\|^2 $ for any vectors $a$ and $b$;  $\scriptsize{\circled{2}}$ follows from the convexity of $\|\cdot\|^2$ and \textit{Jensen's Inequality}; $\scriptsize{\circled{3}}$ follows from $\hat{\varphi}^{[t]}_{\max} =\max_{i\in\mathcal{N},j\in\mathcal{N}_i}\frac{\hat{\varphi}^{[t]}_{i,j}} {\sum_{k\in\mathcal{N}_i} \hat{\varphi}^{[t]}_{i,k}}$ and the facts that $\mathbb{E} \left[ g^{[t]}_{i,i} \right] = \nabla f_i \left( x^{[t-1]}_i \right)$ and $\mathbb{E} \left[ \|a\|^2 \right] = \mathbb{E} \left[ \|a-\mathbb{E}[a]\|^2 \right] + \mathbb{E}\left[ \|\mathbb{E}[a]\|^2 \right]$ holds for any random vector $a$; $\scriptsize{\circled{4}}$ follows from the basic inequality $\|a + b + c + d \|^2 \leq 4\|a\|^2 + 4\|b\|^2 + 4\|c\|^2 + 4\|d\|^2 $ for any vectors $a,b,c,d$ and the facts that $z_i := \nabla F_i \left( x^{[t-1]}_i; \xi_{i,t} \right) - \nabla f_i \left( x^{[t-1]}_i \right)$'s are independent random vectors (with respect to $i$) with zero means and $\mathbb{E} \left[ \left\| \sum^N_{i=1} z_i \right\|^2 \right] = \sum^N_{i=1} \mathbb{E} \left[ \left\| z_i \right\|^2 \right]$; $\scriptsize{\circled{5}}$ follows from \textbf{Assumption}~\ref{ass:bd-varia} and $\left\| \nabla f_j \left( x^{[t-1]}_i \right) - \nabla f_j \left( \bar{x}^{[t-1]} \right) \right\|^2 \leq L^2 \left\| x^{[t-1]}_i - \bar{x}^{[t-1]} \right\|^2 $ by the smoothness of $f_j$ in \textbf{Assumption}~\ref{ass:L-smooth}, and $ \left\| \frac{1}{N} \sum^N_{n=1} \nabla f_n(\bar{x}^{[t-1]}) - \frac{1}{N} \sum^N_{n=1} \nabla f_n({x}^{[t-1]}_n) \right\|^2 \leq \frac{1}{N} \sum^N_{n=1} \left\| \nabla f_n(\bar{x}^{[t-1]}) - \nabla f_n({x}^{[t-1]}_n) \right\|^2 \leq  L^2 \frac{1}{N}\sum^N_{n=1} \left\| \bar{x}^{[t-1]} - {x}^{[t-1]}_n \right\|^2 $ where the first inequality holds due to the convexity of $\|\cdot\|^2$ and \textit{Jensen's Inequality}; $\scriptsize{\circled{6}}$ follows from $ \omega_{\min} = \min_{i \in \mathcal{N}, j \in \mathcal{N}_i}  \omega_{i,j}$ and $\hat{\varphi}_{\max} = \max_{t\in\{1,2,\cdots, T\}} \hat{\varphi}^{[t]}_{\max} $.

\section{Proof of \textbf{Lemma}~\ref{lem:seqdiff}}
  As shown in Line~19 and Line~15 in \textbf{Algorithm}~\ref{alg:ross}, we have
  \begin{align} \label{eq:seqdiff00}
    \bar{u}^{[t]} & = \frac{1}{N} \sum^{N}_{i=1} u^{[t]}_i
    = \frac{1}{N} \sum^{N}_{i=1} \sum_{j \in  \mathcal{N}_i} \omega_{i,j} \hat{u}^{[t]}_j 
    = \frac{1}{N} \sum^{N}_{i=1} \hat{u}^{[t]}_i \left(\sum^{N}_{j=1} w_{j,i} \right) \nonumber\\
    & = \frac{1}{N} \sum^{N}_{i=1} \hat{u}^{[t]}_i
    = \frac{1}{N} \sum^{N}_{i=1} \left( \alpha u^{[t-1]}_i + \bar{g}^{[t]}_i \right)
    = \alpha \bar{u}^{[t-1]} + \frac{1}{N} \sum^{N}_{i=1} \bar{g}^{[t]}_i
  \end{align}
  Moreover, according to Line~20 and Line~16 in \textbf{Algorithm}~\ref{alg:ross}, we have
  \begin{align} \label{eq:seqdiff01}
    \bar{x}^{[t]} & = \frac{1}{N} \sum^{N}_{i=1} x^{[t]}_i
    = \frac{1}{N} \sum^N_{i=1} \sum_{j \in  \mathcal{N}_i} \omega_{i,j} \hat{x}^{[t]}_j
    = \frac{1}{N} \sum^N_{i=1} \hat{x}^{[t]}_i \left(\sum^N_{j=1} w_{j,i} \right) \nonumber\\
    & = \frac{1}{N} \sum^N_{i=1} \hat{x}^{[t]}_i 
    = \frac{1}{N} \sum^N_{i=1} \left(x^{[t-1]}_i - \gamma \hat{u}^{t}_i \right)
    = \bar{x}^{[t-1]} - \gamma \bar{u}^{[t]}
  \end{align}
  We then prove this lemma by induction. When $t=1$, we have
  \begin{align*}
    \bar{S}^{[1]} - \bar{S}^{[0]} =& \frac{1}{1-\alpha} \bar{x}^{[1]} - \frac{\alpha}{1-\alpha} \bar{x}^{[0]} - \bar{x}^{[0]}
    = \frac{1}{1-\alpha} \left( \bar{x}^{[1]} - \bar{x}^{[0]} \right)  \nonumber\\
    =& \frac{-\gamma \bar{u}^{[1]}}{1-\alpha}
    = \frac{-\gamma}{N(1-\alpha)} \sum^N_{i=1} \bar{g}_i^{[1]}  \nonumber
  \end{align*}
  by considering $\gamma \bar{u}^{[1]} = x^{[0]}-x^{[1]}$ due to (\ref{eq:seqdiff01}) and $\bar{u}^{[1]} =  \frac{1}{N} \sum^N_{i=1} \bar{g}_i^{[1]}$ due to (\ref{eq:seqdiff00}). For any $t > 1$, we have
  \begin{align*}
    \bar{S}^{[t]} - \bar{S}^{[t-1]} =& \left( \frac{1}{1-\alpha} \bar{x}^{[t]} - \frac{\alpha}{1-\alpha} \bar{x}^{[t-1]} \right) - \left( \frac{1}{1-\alpha} \bar{x}^{[t-1]} - \frac{\alpha}{1-\alpha} \bar{x}^{[t-2]} \right)  \nonumber\\
    =& \frac{1}{1-\alpha} \left( \bar{x}^{[t]} - \bar{x}^{[t-1]} \right) - \frac{\alpha}{1-\alpha} \left( \bar{x}^{[t-1]} - \bar{x}^{[t-2]} \right)  \nonumber\\
    =& \frac{1}{1-\alpha} \left( -\gamma \bar{u}^{[t]} - \alpha \left(-\gamma \bar{u}^{[t-1]} \right) \right)  \nonumber\\
    =& \frac{-\gamma}{N(1-\alpha)} \sum^N_{i=1} \bar{g}_i^{[t]}  \nonumber
  \end{align*}
  which finally completes the proof.

\section{Proof of \textbf{Lemma}~\ref{lem:sandx}}
  By applying (\ref{eq:seqdiff00}) in \textbf{Lemma}~\ref{lem:seqdiff} recursively, we have
  \begin{equation} \label{eq:up-bar-v}
    \bar{u}^{[t]} = \sum^{t}_{\tau=1} \alpha^{t-\tau} \left( \frac{1}{N}\sum^{N}_{i=1} \bar{g}_i^{[\tau]} \right), ~\forall t \geq 1  
  \end{equation}
  and thus, 
  \begin{align} \label{eq:sandx01}
    \bar{S}^{[t]} - \bar{x}^{[t]} = \frac{-\gamma \alpha}{1-\alpha} \bar{u}^{[t]} = \frac{-\gamma \alpha}{1-\alpha} \sum^{t}_{\tau=1} \alpha^{t-\tau} \left( \frac{1}{N}\sum^{N}_{i=1} \bar{g}_i^{[\tau]} \right)
  \end{align}
  Based on (\ref{eq:sandx01}), we have
  \begin{align*}
    \left\| \bar{S}^{[t]} - \bar{x}^{[t]} \right\|^2 
    =& \left\| \frac{ -\gamma \alpha}{1-\alpha} \sum^{t}_{\tau=1} \alpha^{t-\tau} \left( \frac{1}{N}\sum^{N}_{i=1} \bar{g}_i^{[\tau]} \right) \right\|^2  \\
    \stackrel{\scriptsize{\circled{1}}}=& \frac{ \gamma^2 \alpha^2}{(1-\alpha)^2} \phi^2_t \left\| \sum^{t}_{\tau=1} \frac{\alpha^{t-\tau}}{\phi_t} \left( \frac{1}{N}\sum^{N}_{i=1} \bar{g}_i^{[\tau]} \right) \right\|^2  \\
    \stackrel{\scriptsize{\circled{2}}}\leq& \frac{ \gamma^2 \alpha^2}{(1-\alpha)^2} \phi^2_t \sum^{t}_{\tau=1} \frac{\alpha^{t-\tau}}{\phi_t} \left\| \frac{1}{N}\sum^{N}_{i=1} \bar{g}_i^{[\tau]} \right\|^2  \\
    \stackrel{\scriptsize{\circled{3}}}\leq& \frac{\gamma^2 \alpha^2}{(1-\alpha)^3} \sum^{t}_{\tau=1} \alpha^{t-\tau} \left\| \frac{1}{N}\sum^{N}_{i=1} \bar{g}_i^{[\tau]} \right\|^2
  \end{align*}
  where we have $\scriptsize{\circled{1}}$ by defining  $\phi_t = \sum^{t}_{\tau=1} \alpha^{t-\tau} = \frac{1-\alpha^t}{1-\alpha}$; $\scriptsize{\circled{2}}$ due to the convexity of $\|\cdot\|^2$ and Jensen's Inequality; and $\scriptsize{\circled{3}}$ by considering $ 1-\alpha^t \leq 1$.

  Furthermore, by summing $\left\| \bar{S}^{[t]} - \bar{x}^{[t]} \right\|^2$ over $ t = 1,2, \cdots, T$, we obtain
  \begin{align*}
    \sum^{T}_{t=1} \left\| \bar{S}^{[t]} - \bar{x}^{[t]} \right\|^2
    \leq & \frac{\gamma^2 \alpha^2}{(1-\alpha)^3} \sum^{T}_{t=1} \sum^{t}_{\tau=1} \alpha^{t-\tau} \left\| \frac{1}{N}\sum^{N}_{i=1} \bar{g}_i^{[\tau]} \right\|^2  \nonumber\\
    = & \frac{\gamma^2 \alpha^2}{(1-\alpha)^3} \sum^{T}_{t=1} \left( \left\| \frac{1}{N}\sum^{N}_{i=1} \bar{g}_i^{[t]} \right\|^2 \sum^{T}_{\tau=t} \alpha^{\tau-t} \right)  \nonumber\\
    \stackrel{\scriptsize{\circled{1}}}\leq & \frac{\gamma^2 \alpha^2}{(1-\alpha)^4} \sum^{T}_{t=1} \left\| \frac{1}{N}\sum^{N}_{i=1} \bar{g}_i^{[t]} \right\|^2  
  \end{align*}
  where we have $\scriptsize{\circled{1}}$ by considering $\sum^{T}_{\tau=t} \alpha^{\tau-t} \leq \frac{1}{1-\alpha}$ .

\section{Proof of \textbf{Lemma}~\ref{lem:bdavggrad}}
  Recall that $g^{[t]}_{i,i} = \nabla F_i \left( x^{[t-1]}_i; \xi_{i,t} \right)$ is mutually independent unbiased stochastic gradient at $x^{[t-1]}_i$ such that $\mathbb{E} \left[ g^{[t]}_{i,i} \right] = \nabla f_i \left( x^{[t-1]}_i \right)$ for $\forall i \in \mathcal{N}$. We have
  \begin{align*}
    \mathbb{E} \left[ \left\| \frac{1}{N}\sum^{N}_{i=1} \bar{g}^{[t]}_i \right\|^2 \right]
    = & \mathbb{E} \left[ \left\| \frac{1}{N}\sum^{N}_{i=1} \left( \bar{g}^{[t]}_i - g^{[t]}_{i,i} \right) + \frac{1}{N}\sum^{N}_{i=1}  g^{[t]}_{i,i} \right\|^2 \right] \\
    \stackrel{\scriptsize{\circled{1}}}\leq & 2\mathbb{E} \left[ \left\| \frac{1}{N}\sum^{N}_{i=1} \left( \bar{g}^{[t]}_i - g^{[t]}_{i,i} \right) \right\|^2 \right] + 2\mathbb{E} \left[ \left\| \frac{1}{N}\sum^{N}_{i=1} g^{[t]}_{i,i} \right\|^2 \right]  \\
    \stackrel{\scriptsize{\circled{2}}} \leq & \frac{4\sigma^2\hat{\varphi}^2_{\max}}{\omega^4_{\min}} + \frac{4\sigma^2}{N} + \frac{16\varsigma^2\hat{\varphi}^2_{\max}}{\omega^4_{\min}} + \frac{32L^2\hat{\varphi}^2_{\max}} {N \omega^4_{\min}} \sum^N_{i=1} \mathbb{E} \left[ \left\| x^{[t-1]}_i - \bar{x}^{[t-1]} \right\|^2 \right]   \\ 
    & + \left( \frac{16\hat{\varphi}^2_{\max}}{\omega^4_{\min}} + 4 \right) \mathbb{E} \left[ \left\| \frac{1}{N} \sum^N_{i=1} \nabla f_i \left(x^{[t-1]}_i \right)  \right\|^2 \right] + 2\mathbb{E} \left[ \left\| \frac{1}{N}\sum^{N}_{i=1} g^{[t]}_{i,i} \right\|^2 \right]  \\
    \stackrel{\scriptsize{\circled{3}}} \leq & \frac{4\sigma^2\hat{\varphi}^2_{\max}}{\omega^4_{\min}} + \frac{4\sigma^2}{N} + \frac{16\varsigma^2\hat{\varphi}^2_{\max}}{\omega^4_{\min}} + \frac{32L^2\hat{\varphi}^2_{\max}} {N \omega^4_{\min}} \sum^N_{i=1} \mathbb{E} \left[ \left\| x^{[t-1]}_i - \bar{x}^{[t-1]} \right\|^2 \right] \\
    & + \left( \frac{16\hat{\varphi}^2_{\max}}{\omega^4_{\min}} + 4 \right) \mathbb{E} \left[ \left\| \frac{1}{N} \sum^N_{i=1} \nabla f_i \left(x^{[t-1]}_i \right)  \right\|^2 \right] 
    + 2 \mathbb{E} \left[ \left\| \frac{1}{N}\sum^{N}_{i=1} \nabla f_i \left( x^{[t-1]}_i \right) \right\|^2 \right]  \\
    & + 2 \mathbb{E} \left[ \left\| \frac{1}{N}\sum^{N}_{i=1} \left( \nabla F_i \left( x^{[t-1]}_i; \xi_{i,t} \right)- \nabla f_i \left(x^{[t-1]}_i \right) \right) \right\|^2 \right]   \\
    \stackrel{\scriptsize{\circled{4}}} \leq & \frac{4\sigma^2\hat{\varphi}^2_{\max}}{\omega^4_{\min}} + \frac{4\sigma^2}{N} + \frac{16\varsigma^2\hat{\varphi}^2_{\max}}{\omega^4_{\min}} + \frac{32L^2\hat{\varphi}^2_{\max}} {N \omega^4_{\min}} \sum^N_{i=1} \mathbb{E} \left[ \left\| x^{[t-1]}_i - \bar{x}^{[t-1]} \right\|^2 \right]  \\
    & + \left( \frac{16\hat{\varphi}^2_{\max}}{\omega^4_{\min}} + 4 \right) \mathbb{E} \left[ \left\| \frac{1}{N} \sum^N_{i=1} \nabla f_i \left(x^{[t-1]}_i \right)  \right\|^2 \right]  
    + 2 \mathbb{E} \left[ \left\| \frac{1}{N} \sum^N_{i=1} \nabla f_i \left(x^{[t-1]}_i \right) \right\|^2 \right]  \\
    & + \frac{2}{N^2} \sum^{N}_{i=1} \mathbb{E} \left[ \left\|  \nabla F_i \left( x^{[t-1]}_i; \xi_{i,t} \right) - \nabla f_i \left( x^{[t-1]}_i \right)  \right\|^2 \right]  \\
    \stackrel{\scriptsize{\circled{5}}} \leq & \frac{4\sigma^2\hat{\varphi}^2_{\max}}{\omega^4_{\min}} + \frac{6\sigma^2}{N} + \frac{16\varsigma^2\hat{\varphi}^2_{\max}}{\omega^4_{\min}} + \frac{32L^2\hat{\varphi}^2_{\max}} {N \omega^4_{\min}} \sum^N_{i=1} \mathbb{E} \left[ \left\| x^{[t-1]}_i - \bar{x}^{[t-1]} \right\|^2 \right] \\
    & + \left( \frac{16\hat{\varphi}^2_{\max}}{\omega^4_{\min}} + 6 \right) \mathbb{E} \left[ \left\| \frac{1}{N} \sum^N_{i=1} \nabla f_i \left(x^{[t-1]}_i \right)  \right\|^2 \right]  
  \end{align*}
  where we have $\scriptsize{\circled{1}}$ by considering the inequality $\|a+b\|^2 \leq 2\|a\|^2 + 2\|b\|^2$; $\scriptsize{\circled{2}}$ according to \textbf{Lemma}~\ref{lem:diffhatgrad}; $\scriptsize{\circled{3}}$ based on the facts that $\mathbb{E} \left[ g^{[t]}_{i,i} \right] = \nabla f_i(x^{[t-1]}_i)$ and $\mathbb{E} \left[ \|a\|^2 \right] = \mathbb{E} \left[ \|a-\mathbb{E}[a]\|^2 \right] + \mathbb{E} \left[ \left\| \mathbb{E}[a] \right\|^2 \right]$ holds for any random vector $a$; $\scriptsize{\circled{4}}$ according to the facts that $z_i := \nabla F_i \left( x^{[t-1]}_i; \xi_{i,t} \right) - \nabla f_i \left( x^{[t-1]}_i \right)$'s are independent random vectors (with respect to $i$) with zero means and $\mathbb{E} \left[ \left\| \sum^N_{i=1} z_i \right\|^2 \right] = \sum^N_{i=1} \mathbb{E} \left[ \left\| z_i \right\|^2 \right]$; and $\scriptsize{\circled{5}}$ due to \textbf{Assumption}~\ref{ass:bd-varia}.

\section{Proof of \textbf{Lemma}~\ref{lem:barG-G}}
  According to \eqref{eq:notation} and \textbf{Fact}~\ref{fact:fact-2}, we have
  \begin{align*}
    & \mathbb{E} \left[ \left\| \bar{\mathbf{G}}^{[t]} - \mathbf{G}^{[t]} \right\|^2_{\mathfrak{F}} \right]  \\
    = & \sum^N_{i=1} \mathbb{E} \left[ \left\| \bar{g}^{[t]}_i - g^{[t]}_{i,i} \right\|^2 \right]  \\
    = & \sum^N_{i=1} \mathbb{E} \left[ \left\| \sum_{j \in \mathcal{N}_i} \pi^{[t]}_{i,j} \nabla F_j \left(x^{[t-1]}_i; \xi_{j,t} \right) - \nabla F_i \left(x^{[t-1]}_i; \xi_{i,t} \right) \right\|^2 \right] \\
    \leq & 2\sum^N_{i=1} \mathbb{E} \left[ \left\| \sum_{j \in \mathcal{N}_i} \pi^{[t]}_{i,j} \nabla F_j \left( x^{[t-1]}_i; \xi_{j,t} \right) \right\|^2 \right] + 2\sum^N_{i=1} \mathbb{E} \left[ \left\| \nabla F_i \left( x^{[t-1]}_i; \xi_{i,t} \right) \right\|^2 \right] \\
    \stackrel{\scriptsize{\circled{1}}} \leq & 2\sum^N_{i=1} \sum_{j \in \mathcal{N}_i} \omega_{i,j} \mathbb{E} \left[ \bigg\| \frac{\hat{\varphi}^{[t]}_{i,j}}{\omega^2_{i,j} \sum_{k\in\mathcal{N}_i} \hat{\varphi}^{[t]}_{i,k}} \bigg\|^2 \bigg\| \nabla F_j \left( x^{[t-1]}_i; \xi_{j,t} \right) \bigg\|^2 \right] + 2\sum^N_{i=1} \mathbb{E} \left[ \left\| \nabla F_i \left( x^{[t-1]}_i; \xi_{i,t} \right) \right\|^2 \right] \\
    = & 2\sum^N_{i=1} \sum_{j \in \mathcal{N}_i} \frac{\omega_{i,j}}{\omega^4_{i,j}} \mathbb{E} \left[ \bigg\| \frac{\hat{\varphi}^{[t]}_{i,j}}{ \sum_{k\in\mathcal{N}_i} \hat{\varphi}^{[t]}_{i,k}} \bigg\|^2 \bigg\| \nabla F_j \left( x^{[t-1]}_i; \xi_{j,t} \right) \bigg\|^2 \right] + 2\sum^N_{i=1} \mathbb{E} \left[ \left\| \nabla F_i \left( x^{[t-1]}_i; \xi_{i,t} \right) \right\|^2 \right]  \\
    \stackrel{\scriptsize{\circled{2}}} \leq & 2\sum^N_{i=1} \sum_{j \in \mathcal{N}_i} \frac{\omega_{i,j} \cdot \left(\hat{\varphi}^{[t]}_{\max}\right)^2}{\omega^4_{i,j}} \mathbb{E} \left[ \left\| \nabla F_j \left( x^{[t-1]}_i; \xi_{j,t} \right) \right\|^2 \right] + 2\sum^N_{i=1} \mathbb{E} \left[ \left\| \nabla F_i \left( x^{[t-1]}_i; \xi_{i,t} \right) \right\|^2 \right]  \\
    \stackrel{\scriptsize{\circled{3}}} = & 2\sum^N_{i=1} \sum_{j \in \mathcal{N}_i} \frac{\omega_{i,j} \cdot \left(\hat{\varphi}^{[t]}_{\max}\right)^2}{\omega^4_{i,j}} \mathbb{E} \left[ \left\| \nabla F_j \left( x^{[t-1]}_i; \xi_{j,t} \right) - \nabla f_j \left( x^{[t-1]}_i \right) \right\|^2 \right] + 2\sum^N_{i=1} \sum_{j \in \mathcal{N}_i} \frac{\omega_{i,j} \cdot \left(\hat{\varphi}^{[t]}_{\max}\right)^2}{\omega^4_{i,j}} \mathbb{E} \left[ \left\| \nabla f_j \left( x^{[t-1]}_i \right) \right\|^2 \right]   \\
    & + 2\sum^N_{i=1} \mathbb{E} \left[ \left\| \nabla F_i \left( x^{[t-1]}_i; \xi_{i,t} \right) - \nabla f_i \left( x^{[t-1]}_i \right) \right\|^2 \right] + 2\sum^N_{i=1} \mathbb{E} \left[ \left\| \nabla f_i \left( x^{[t-1]}_i \right) \right\|^2 \right]    \\
    = & 2\sum^N_{i=1} \sum_{j \in \mathcal{N}_i} \frac{\omega_{i,j} \cdot \left(\hat{\varphi}^{[t]}_{\max}\right)^2}{\omega^4_{i,j}} \mathbb{E} \left[ \left\| \nabla F_j \left( x^{[t-1]}_i; \xi_{j,t} \right) - \nabla f_j \left( x^{[t-1]}_i \right) \right\|^2 \right]  \\
    & + 2\sum^N_{i=1} \sum_{j \in \mathcal{N}_i} \frac{\omega_{i,j} \cdot \left(\hat{\varphi}^{[t]}_{\max}\right)^2}{\omega^4_{i,j}} \mathbb{E} \left[ \left\| \begin{aligned} & \nabla f_j \left( x^{[t-1]}_i \right) - \nabla f_j \left( \bar{x}^{[t-1]} \right) + \nabla f_j \left( \bar{x}^{[t-1]} \right) - \frac{1}{N} \sum^N_{n=1} \nabla f_n \left( \bar{x}^{[t-1]} \right) \\ & + \frac{1}{N} \sum^N_{n=1} \nabla f_n \left( \bar{x}^{[t-1]} \right) - \frac{1}{N} \sum^N_{n=1} \nabla f_n \left( x^{[t-1]}_n \right) + \frac{1}{N} \sum^N_{n=1} \nabla f_n \left( x^{[t-1]}_n \right) \end{aligned} \right\|^2 \right]      \\
    & + 2\sum^N_{i=1} \mathbb{E} \left[ \left\| \nabla F_i \left( x^{[t-1]}_i, \xi_{i,t} \right) - \nabla f_i \left( x^{[t-1]}_i \right) \right\|^2 \right] \\
    & + 2\sum^N_{i=1} \mathbb{E} \left[ \left\| \begin{aligned} & \nabla f_i \left( x^{[t-1]}_i \right) - \nabla f_i \left( \bar{x}^{[t-1]} \right) + \nabla f_i \left( \bar{x}^{[t-1]} \right) - \frac{1}{N} \sum^N_{n=1} \nabla f_n \left( \bar{x}^{[t-1]} \right)  \\  & + \frac{1}{N} \sum^N_{n=1} \nabla f_n \left( \bar{x}^{[t-1]} \right) - \frac{1}{N} \sum^N_{n=1} \nabla f_n \left( x^{[t-1]}_n \right) + \frac{1}{N} \sum^N_{n=1} \nabla f_n \left( x^{[t-1]}_n \right) \end{aligned} \right\|^2 \right]    \\
    \stackrel{\scriptsize{\circled{4}}} \leq & 2\sum^N_{i=1} \sum_{j \in \mathcal{N}_i} \frac{\omega_{i,j} \cdot \left(\hat{\varphi}^{[t]}_{\max}\right)^2}{\omega^4_{i,j}} \mathbb{E} \left[ \left\| \nabla F_j \left( x^{[t-1]}_i; \xi_{j,t} \right) - \nabla f_j \left( x^{[t-1]}_i \right) \right\|^2 \right]  \\
    & + 8\sum^N_{i=1} \sum_{j \in \mathcal{N}_i} \frac{\omega_{i,j} \cdot \left(\hat{\varphi}^{[t]}_{\max}\right)^2}{\omega^4_{i,j}} \mathbb{E} \left[ \left\| \nabla f_j \left( x^{[t-1]}_i \right) - \nabla f_j \left( \bar{x}^{[t-1]} \right)  \right\|^2 \right]  \\
    & + 8\sum^N_{i=1} \sum_{j \in \mathcal{N}_i} \frac{\omega_{i,j} \cdot \left(\hat{\varphi}^{[t]}_{\max}\right)^2}{\omega^4_{i,j}} \mathbb{E} \left[ \left\| \nabla f_j \left( \bar{x}^{[t-1]} \right) - \frac{1}{N} \sum^N_{n=1} \nabla f_n \left( \bar{x}^{[t-1]} \right) \right\|^2 \right]  \\
    & + 8\sum^N_{i=1} \sum_{j \in \mathcal{N}_i} \frac{\omega_{i,j} \cdot \left(\hat{\varphi}^{[t]}_{\max}\right)^2}{\omega^4_{i,j}} \mathbb{E} \left[ \left\| \frac{1}{N} \sum^N_{n=1} \nabla f_n \left( \bar{x}^{[t-1]} \right) - \frac{1}{N} \sum^N_{n=1} \nabla f_n \left( x^{[t-1]}_n \right)  \right\|^2 \right]  \\
    & + 8\sum^N_{i=1} \sum_{j \in \mathcal{N}_i} \frac{\omega_{i,j} \cdot \left(\hat{\varphi}^{[t]}_{\max}\right)^2}{\omega^4_{i,j}} \mathbb{E} \left[ \left\| \frac{1}{N} \sum^N_{n=1} \nabla f_n \left( x^{[t-1]}_n \right)  \right\|^2 \right]  \\
    & +  2\sum^N_{i=1} \mathbb{E} \left[ \left\| \nabla F_i \left( x^{[t-1]}_i; \xi_{i,t} \right) - \nabla f_i \left( x^{[t-1]}_i \right) \right\|^2 \right] \\
    & + 8\sum^N_{i=1} \mathbb{E} \left[ \left\| \nabla f_i \left( x^{[t-1]}_i \right) - \nabla f_i \left( \bar{x}^{[t-1]} \right) \right\|^2 \right] \\
    & + 8\sum^N_{i=1} \mathbb{E} \left[ \left\| \nabla f_i \left( \bar{x}^{[t-1]} \right) - \frac{1}{N} \sum^N_{n=1} \nabla f_n \left( \bar{x}^{[t-1]} \right) \right\|^2 \right] \\
    & + 8\sum^N_{i=1} \mathbb{E} \left[ \left\| \frac{1}{N} \sum^N_{n=1} \nabla f_n \left( \bar{x}^{[t-1]} \right) - \frac{1}{N} \sum^N_{n=1} \nabla f_n \left( x^{[t-1]}_n \right) \right\|^2 \right] \\
    & + 8\sum^N_{i=1} \mathbb{E} \left[ \left\| \frac{1}{N} \sum^N_{n=1} \nabla f_n \left( x^{[t-1]}_n \right) \right\|^2 \right] \\
    \stackrel{\scriptsize{\circled{5}}} \leq & 2\sum^N_{i=1} \sum_{j \in \mathcal{N}_i} \frac{\omega_{i,j} \cdot \left(\hat{\varphi}^{[t]}_{\max}\right)^2}{\omega^4_{i,j}} \sigma^2 + 8\sum^N_{i=1} \sum_{j \in \mathcal{N}_i} \frac{\omega_{i,j} \cdot \left(\hat{\varphi}^{[t]}_{\max}\right)^2}{\omega^4_{i,j}} \varsigma^2  \\
    & + 8\sum^N_{i=1} \sum_{j \in \mathcal{N}_i} \frac{\omega_{i,j} \cdot \left(\hat{\varphi}^{[t]}_{\max}\right)^2}{\omega^4_{i,j}} L^2 \mathbb{E} \left[ \left\| x^{[t-1]}_i - \bar{x}^{[t-1]} \right\|^2 \right]  \\
    & + 8\sum^N_{i=1} \sum_{j \in \mathcal{N}_i} \frac{\omega_{i,j} \cdot \left(\hat{\varphi}^{[t]}_{\max}\right)^2}{\omega^4_{i,j}} \mathbb{E} \left[ \left\| \frac{1}{N} \sum^N_{n=1} \nabla f_n \left( x^{[t-1]}_n \right)  \right\|^2 \right]  \\ 
    & + 8\sum^N_{i=1} \sum_{j \in \mathcal{N}_i} \frac{\omega_{i,j} \cdot \left(\hat{\varphi}^{[t]}_{\max}\right)^2}{\omega^4_{i,j}} \frac{1}{N} \sum^N_{n=1} L^2 \mathbb{E} \left[ \left\| \bar{x}^{[t-1]} - x^{[t-1]}_n \right\|^2 \right]  \\
    & + 2\sum^N_{i=1} \sigma^2  + 8\sum^N_{i=1} L^2 \mathbb{E} \left[ \left\| x^{[t-1]}_i - \bar{x}^{[t-1]} \right\|^2 \right]    \\ 
    & + 8\sum^N_{i=1} \varsigma^2 + 8\sum^N_{i=1} \mathbb{E} \left[ \left\| \frac{1}{N} \sum^N_{n=1} \nabla f_n \left( x^{[t-1]}_n \right) \right\|^2 \right] \\
    & + 8\sum^N_{i=1} \frac{1}{N} \sum^N_{n=1} L^2 \mathbb{E} \left[ \left\| \bar{x}^{[t-1]} - x^{[t-1]}_n \right\|^2 \right]  \\
    \stackrel{\scriptsize{\circled{6}}} \leq & \frac{2N\sigma^2\hat{\varphi}^2_{\max}}{\omega^4_{\min}} + 2N\sigma^2 + \frac{8N\varsigma^2\hat{\varphi}^2_{\max}}{\omega^4_{\min}} +  8N\varsigma^2   \\
    & + \left( \frac{16L^2\hat{\varphi}^2_{\max}}{\omega^4_{\min}} + 16L^2 \right) \sum^N_{i=1} \mathbb{E} \left[ \left\| x^{[t-1]}_i - \bar{x}^{[t-1]} \right\|^2 \right]  \\
    & +  \left( \frac{8N\hat{\varphi}^2_{\max}}{\omega^4_{\min}} + 8N \right) \mathbb{E} \left[ \left\| \frac{1}{N} \sum^N_{i=1} \nabla f_i \left(x^{[t-1]}_i \right) \right\|^2 \right]
  \end{align*}
  where $\scriptsize{\circled{1}}$ follows from the convexity of $\|\cdot\|^2$ and \textit{Jensen's Inequality}; $\scriptsize{\circled{2}}$ follows from $\hat{\varphi}^{[t]}_{\max} =\max_{i\in \mathcal{N}, j\in \mathcal{N}_i}\frac{\hat{\varphi}^{[t]}_{i,j}} {\sum_{k\in\mathcal{N}_i} \hat{\varphi}^{[t]}_{i,k}}$; $\scriptsize{\circled{3}}$ follows from the facts that $\mathbb{E} \left[ g^{[t]}_{i,i} \right] = \nabla f_i \left( x^{[t-1]}_i \right)$ and $\mathbb{E} \left[ \|a\|^2 \right] = \mathbb{E} \left[ \|a-\mathbb{E}[a]\|^2 \right] + \mathbb{E}\left[ \|\mathbb{E}[a]\|^2 \right]$ holds for any random vector $a$; $\scriptsize{\circled{4}}$ follows from the basic inequality $\|a + b + c + d \|^2 \leq 4\|a\|^2 + 4\|b\|^2 + 4\|c\|^2 + 4\|d\|^2 $ for any vectors $a,b,c,d$; $\scriptsize{\circled{5}}$ follows from \textbf{Assumption}~\ref{ass:bd-varia}, and $\left\| \nabla f_j \left( x^{[t-1]}_i \right) - \nabla f_j \left( \bar{x}^{[t-1]} \right) \right\|^2 \leq L^2 \left\| x^{[t-1]}_i - \bar{x}^{[t-1]} \right\|^2 $ by the smoothness of $f_j$ in \textbf{Assumption}~\ref{ass:L-smooth}, and $ \left\| \frac{1}{N} \sum^N_{n=1} \nabla f_n(\bar{x}^{[t-1]}) - \frac{1}{N} \sum^N_{n=1} \nabla f_n({x}^{[t-1]}_n) \right\|^2 \leq \frac{1}{N} \sum^N_{n=1} \left\| \nabla f_n(\bar{x}^{[t-1]}) - \nabla f_n({x}^{[t-1]}_n) \right\|^2 \leq  L^2 \frac{1}{N}\sum^N_{n=1} \left\| \bar{x}^{[t-1]} - {x}^{[t-1]}_n \right\|^2 $ where the first inequality holds due to the convexity of $\|\cdot\|^2$ and \textit{Jensen's Inequality}; $\scriptsize{\circled{6}}$ follows from $ \omega_{\min} = \min_{i \in \mathcal{N}, j \in \mathcal{N}_i} \omega_{i,j}$ and $\hat{\varphi}_{\max} = \max_{t\in\{1,2,\cdots, T\}} \hat{\varphi}^{[t]}_{\max}$.

\section{Proof of \textbf{Lemma}~\ref{lem:barx-x}}
  Recalling the definition of $\mathbf{U}^{[t]}$ and $\hat{\mathbf{U}}^{[t]}$ in \eqref{eq:notation}, according to Line~19 and Line~15 in \textbf{Algorithm}~\ref{alg:ross}, for all $t \geq 1$, we have
  \begin{align} \label{eq:seqdiff00mat}
    \mathbf{U}^{[t]} & = \hat{\mathbf{U}}^{[t]} \mathbf{W}
    = \left( \alpha \mathbf{U}^{[t-1]} + \bar{\mathbf{G}}^{[t]} \right) \mathbf{W}
  \end{align}
  By applying (\ref{eq:seqdiff00mat}) recursively, we obtain
  \begin{align}
    \mathbf{U}^{[t]} & = \alpha^t \mathbf{U}^{[0]} \mathbf{W}^t + \sum^{t}_{\tau=1} \bar{\mathbf{G}}^{[\tau]} \alpha^{t-\tau} \mathbf{W}^{t-\tau+1} 
    \stackrel{\mathbf{U}^{[0]} = 0} = \sum^{t}_{\tau=1} \bar{\mathbf{G}}^{[\tau]} \alpha^{t-\tau} \mathbf{W}^{t-\tau+1} 
  \end{align}
  Recalling the definition of $\mathbf{X}^{[t]}$ in \eqref{eq:notation}, according to Line~20 and Line~16 in \textbf{Algorithm}~\ref{alg:ross}, for all $t \geq 1$, we have
  \begin{align} \label{eq:seqdiff01mat}
    \mathbf{X}^{[t]} & = \left( \mathbf{X}^{[t-1]} - \gamma \hat{\mathbf{U}}^{[t]} \right) \mathbf{W}
    = \mathbf{X}^{[t-1]} \mathbf{W} - \gamma \hat{\mathbf{U}}^{[t]} \mathbf{W} 
    = \mathbf{X}^{[t-1]} \mathbf{W} - \gamma \mathbf{U}^{[t]}  
  \end{align}
  By applying (\ref{eq:seqdiff01mat}) recursively, we can obtain 
  \begin{align}
    \mathbf{X}^{[t]} = \mathbf{X}^{[0]} \mathbf{W}^{t} - \gamma \sum^t_{\tau=1} \mathbf{U}^{[\tau]} \mathbf{W}^{t-\tau}   
  \end{align}
  Multiplying both sides by $(\mathbf{I} - \mathbf{Q})$ with $\mathbf{Q}$ defined in \textbf{Fact}~\ref{fact:fact-1}, we have
  \begin{align}
    \mathbf{X}^{[t]} (\mathbf{I} - \mathbf{Q}) &= \mathbf{X}^{[0]} \mathbf{W}^t (\mathbf{I} - \mathbf{Q}) - \gamma \sum^t_{\tau=1} \mathbf{U}^{[\tau]} \mathbf{W}^{t-\tau} (\mathbf{I} - \mathbf{Q})  \nonumber\\
    & \stackrel{\scriptsize{\circled{1}}} = \mathbf{X}^{[0]} (\mathbf{I} - \mathbf{Q}) \mathbf{W}^t - \gamma \sum^t_{\tau=1} \mathbf{U}^{[\tau]} (\mathbf{I} - \mathbf{Q}) \mathbf{W}^{t-\tau}
  \end{align}
  where $\scriptsize{\circled{1}}$ follows from $\mathbf{W} (\mathbf{I} - \mathbf{Q}) = (\mathbf{I} - \mathbf{Q}) \mathbf{W} $ in \textbf{Fact}~\ref{fact:fact-1}.

  By considering $\mathbf{X}^{[0]} (\mathbf{I} - \mathbf{Q}) = 0$ due to all columns of $\mathbf{X}^{[0]}$ are identical, we have
  \begin{align} \label{eq:XI-Q-00}
    & \mathbb{E} \left[ \left\| \mathbf{X}^{[t]} (\mathbf{I} - \mathbf{Q}) \right\|^2_{\mathfrak{F}} \right] \nonumber\\
    = & \mathbb{E} \left[ \left\| -\gamma \sum^t_{\tau=1} \mathbf{U}^{[\tau]} (\mathbf{I} - \mathbf{Q}) \mathbf{W}^{t-\tau} \right\|^2_{\mathfrak{F}} \right] \nonumber\\
    = & \mathbb{E} \left[ \left\| -\gamma \sum^t_{\tau=1} \sum^{\tau}_{\tau'=1} \bar{\mathbf{G}}^{[\tau']} \alpha^{\tau-\tau'} \mathbf{W}^{\tau-\tau'+1} (\mathbf{I} - \mathbf{Q}) \mathbf{W}^{t-\tau} \right\|^2_{\mathfrak{F}} \right] \nonumber\\
    = & \mathbb{E} \left[ \left\| -\gamma \sum^t_{\tau=1} \sum^{\tau}_{\tau'=1} \bar{\mathbf{G}}^{[\tau']} \alpha^{\tau-\tau'} \mathbf{W}^{t-\tau'+1} (\mathbf{I} - \mathbf{Q}) \right\|^2_{\mathfrak{F}} \right] \nonumber\\
    = & \mathbb{E} \left[ \left\| -\gamma \sum^{t}_{\tau=1} \bar{\mathbf{G}}^{[\tau]} ( \sum^{t}_{\tau'= \tau} \alpha^{\tau'-\tau} \mathbf{W}^{t-\tau+1} ) (\mathbf{I} - \mathbf{Q}) \right\|^2_{\mathfrak{F}} \right]  \nonumber\\
    = & \mathbb{E} \left[ \left\| -\gamma \sum^{t}_{\tau=1} \bar{\mathbf{G}}^{[\tau]} ( \sum^{t}_{\tau'= \tau} \alpha^{\tau'-\tau} ) \mathbf{W}^{t-\tau+1} (\mathbf{I} - \mathbf{Q}) \right\|^2_{\mathfrak{F}} \right]  \nonumber\\
    = & \mathbb{E} \left[ \left\| -\gamma \sum^{t}_{\tau=1} \frac{1-\alpha^{t-\tau}}{1-\alpha} \bar{\mathbf{G}}^{[\tau]} (\mathbf{I} - \mathbf{Q}) \mathbf{W}^{t-\tau+1} \right\|^2_{\mathfrak{F}} \right]  \nonumber\\
    = & \gamma^2 \mathbb{E} \left[ \left\| \sum^{t}_{\tau=1} \frac{1-\alpha^{t-\tau}}{1-\alpha} \left( ( \bar{\mathbf{G}}^{[\tau]} - \mathbf{G}^{[\tau]}) (\mathbf{I} - \mathbf{Q}) \mathbf{W}^{t-\tau+1} + \mathbf{G}^{[\tau]} (\mathbf{I} - \mathbf{Q}) \mathbf{W}^{t-\tau+1} \right) \right\|^2_{\mathfrak{F}} \right]  \nonumber\\
    \stackrel{\scriptsize{\circled{1}}} \leq & 2\gamma^2 \underbrace{ \mathbb{E} \left[ \left\| \sum^{t}_{\tau=1} \frac{1-\alpha^{t-\tau}}{1-\alpha} \left( \bar{\mathbf{G}}^{[\tau]} - \mathbf{G}^{[\tau]} \right) \left( \mathbf{I} - \mathbf{Q} \right) \mathbf{W}^{t-\tau+1} \right\|^2_{\mathfrak{F}} \right] }_{\mathrm{\MakeTextUppercase{\romannumeral 1}}} \nonumber\\
    & + 2\gamma^2 \underbrace{ \mathbb{E} \left[ \left\| \sum^{t}_{\tau=1} \frac{1-\alpha^{t-\tau}}{1-\alpha} \mathbf{G}^{[\tau]} (\mathbf{I} - \mathbf{Q}) \mathbf{W}^{t-\tau+1} \right\|^2_{\mathfrak{F}} \right] }_{\mathrm{\MakeTextUppercase{\romannumeral 2}}}
  \end{align}
  where we have $\scriptsize{\circled{1}}$ by considering the inequality $\| \mathbf{A} + \mathbf{B} \|^2_{\mathfrak{F}} \leq 2 \| \mathbf{A} \|^2_{\mathfrak{F}} + 2 \| \mathbf{B} \|^2_{\mathfrak{F}}$.

  In the following, we first give the upper bound of term $\mathrm{\MakeTextUppercase{\romannumeral 1}}$
  \begin{align} \label{eq:bdtermI}
    & \mathbb{E} \left[ \left\| \sum^{t}_{\tau=1} \frac{1-\alpha^{t-\tau}}{1-\alpha} \left( \bar{\mathbf{G}}^{[\tau]} - \mathbf{G}^{[\tau]} \right) \left( \mathbf{I} - \mathbf{Q} \right) \mathbf{W}^{t-\tau+1} \right\|^2_{\mathfrak{F}} \right]  \nonumber\\
    \stackrel{\scriptsize{\circled{1}}} \leq & \sum^{t}_{\tau=1} \sum^{t}_{\tau'=1} \mathbb{E} \left[ \left\| \frac{1-\alpha^{t-\tau}}{1-\alpha} ( \bar{\mathbf{G}}^{[\tau]} - \mathbf{G}^{[\tau]} ) ( \mathbf{I} - \mathbf{Q} ) \mathbf{W}^{t-\tau+1} \right\|_{\mathfrak{F}}  \left\| \frac{1-\alpha^{t-\tau'}}{1-\alpha}  ( \bar{\mathbf{G}}^{[\tau']} - \mathbf{G}^{[\tau']} ) ( \mathbf{I} - \mathbf{Q} ) \mathbf{W}^{t-\tau'+1} \right\|_{\mathfrak{F}} \right]   \nonumber\\
    \stackrel{\scriptsize{\circled{2}}} \leq & \frac{1}{(1-\alpha)^2} \sum^{t}_{\tau=1} \sum^{t}_{\tau'=1} \mathbb{E} \left[ \left\| \bar{\mathbf{G}}^{[\tau]} - \mathbf{G}^{[\tau]} \right\|_{\mathfrak{F}} \left\| (\mathbf{I} - \mathbf{Q}) \mathbf{W}^{t-\tau+1} \right\|_{\mathfrak{s}} \left\|  \bar{\mathbf{G}}^{[\tau']} - \mathbf{G}^{[\tau']} \right\|_{\mathfrak{F}} \left\| (\mathbf{I} - \mathbf{Q}) \mathbf{W}^{t-\tau'+1} \right\|_{\mathfrak{s}} \right]   \nonumber\\
    \stackrel{\scriptsize{\circled{3}}} \leq & \frac{1}{(1-\alpha)^2} \sum^{t}_{\tau=1} \sum^{t}_{\tau'=1} \rho^{(t+1-\frac{\tau+\tau'}{2})} \mathbb{E} \left[ \left\| \bar{\mathbf{G}}^{[\tau]} - \mathbf{G}^{[\tau]} \right\|_{\mathfrak{F}} \left\|  \bar{\mathbf{G}}^{[\tau']} - \mathbf{G}^{[\tau']} \right\|_{\mathfrak{F}} \right]    \nonumber\\
    \stackrel{\scriptsize{\circled{4}}} \leq & \frac{1}{(1-\alpha)^2} \sum^{t}_{\tau=1} \sum^{t}_{\tau'=1} \rho^{(t+1-\frac{\tau+\tau'}{2})} \left( \frac{1}{2} \mathbb{E} \left[  \left\| \bar{\mathbf{G}}^{[\tau]} - \mathbf{G}^{[\tau]} \right\|^2_{\mathfrak{F}} \right] + \frac{1}{2} \mathbb{E} \left[ \left\|  \bar{\mathbf{G}}^{[\tau']} - \mathbf{G}^{[\tau']} \right\|^2_{\mathfrak{F}} \right] \right)    \nonumber\\
    \stackrel{\scriptsize{\circled{5}}} \leq & \frac{1}{(1-\alpha)^2 (1-\sqrt{\rho})} \sum^{t}_{\tau=1} \rho^{(\frac{t+1-\tau}{2})} \mathbb{E} \left[ \left\| \bar{\mathbf{G}}^{[\tau]} - \mathbf{G}^{[\tau]} \right\|^2_{\mathfrak{F}} \right]     \nonumber\\
    \stackrel{\scriptsize{\circled{6}}} \leq & \frac{1}{(1-\alpha)^2 (1-\sqrt{\rho})} \sum^{t}_{\tau=1} \rho^{(\frac{t+1-\tau}{2})} \left( \begin{aligned} & \frac{2N\sigma^2\hat{\varphi}^2_{\max}}{\omega^4_{\min}} + 2N\sigma^2 + \frac{8N\varsigma^2\hat{\varphi}^2_{\max}}{\omega^4_{\min}} +  8N\varsigma^2   \\
    & + \left( \frac{16L^2\hat{\varphi}^2_{\max}}{\omega^4_{\min}} + 16L^2 \right) \sum^N_{i=1} \mathbb{E} \left[ \left\| x^{[\tau-1]}_i - \bar{x}^{[\tau-1]} \right\|^2 \right] \\
    & +  \left( \frac{8N\hat{\varphi}^2_{\max}}{\omega^4_{\min}} + 8N \right) \mathbb{E} \left[ \left\| \frac{1}{N} \sum^N_{i=1} \nabla f_i \left(x^{[\tau-1]}_i \right) \right\|^2 \right] \end{aligned} \right)  \nonumber\\
    \leq & \frac{2N\sigma^2 + 8N\varsigma^2}{(1-\alpha)^2 (1-\sqrt{\rho})^2} + \frac{ \left(2N\sigma^2 + 8N\varsigma^2\right)\hat{\varphi}^2_{\max}}{\omega^4_{\min}(1-\alpha)^2 (1-\sqrt{\rho})^2}  \nonumber\\
    & + \frac{16L^2\hat{\varphi}^2_{\max}}{\omega^4_{\min}(1-\alpha)^2 (1-\sqrt{\rho})} \sum^{t}_{\tau=1} \rho^{(\frac{t+1-\tau}{2})}  \sum^N_{i=1} \mathbb{E} \left[ \left\|  x^{[\tau-1]}_i - \bar{x}^{[\tau-1]} \right\|^2 \right]      \nonumber\\
    & + \frac{16L^2}{(1-\alpha)^2 (1-\sqrt{\rho})} \sum^{t}_{\tau=1} \rho^{(\frac{t+1-\tau}{2})} \sum^N_{i=1} \mathbb{E} \left[ \left\|  x^{[\tau-1]}_i - \bar{x}^{[\tau-1]} \right\|^2 \right]  \nonumber\\
    & + \frac{8N\hat{\varphi}^2_{\max}}{\omega^4_{\min}(1-\alpha)^2 (1-\sqrt{\rho})} \sum^{t}_{\tau=1} \rho^{(\frac{t+1-\tau}{2})} \mathbb{E} \left[ \left\| \frac{1}{N} \sum^N_{i=1} \nabla f_i \left( x^{[\tau-1]}_i \right) \right\|^2 \right]  \nonumber\\
    & + \frac{8N}{(1-\alpha)^2 (1-\sqrt{\rho})} \sum^{t}_{\tau=1} \rho^{(\frac{t+1-\tau}{2})} \mathbb{E} \left[ \left\| \frac{1}{N} \sum^N_{i=1} \nabla f_i \left( x^{[\tau-1]}_i \right) \right\|^2 \right]
  \end{align}
  where $\scriptsize{\circled{1}}$ follows from \textbf{Fact}~\ref{fact:fact-2}; $\scriptsize{\circled{2}}$ follows from the inequality $\left| \frac{1-\alpha^{t-\tau}}{1-\alpha}\right| \leq \frac{1}{1-\alpha}$ and the inequality $\|\mathbf{AB}\|_{\mathfrak{F}} \leq \|\mathbf{B}\|_{\mathfrak{s}} \|\mathbf{A}\|_{\mathfrak{F}}$; $\scriptsize{\circled{3}}$ follows from \textbf{Fact}~\ref{fact:fact-1}; $\scriptsize{\circled{4}}$ follows from the inequality $\|\mathbf{A}\|_{\mathfrak{F}} \|\mathbf{B}\|_{\mathfrak{F}} \leq \frac{1}{2}(\|\mathbf{A}\|^2_{\mathfrak{F}} + \|\mathbf{B}\|^2_{\mathfrak{F}})$; $\scriptsize{\circled{5}}$ follows from the inequality $\sum^{t}_{\tau'=1} \rho^{\left( t+1-\frac{\tau + \tau'}{2} \right)} \leq \rho^{\frac{t+1-\tau}{2}} (1-\sqrt{\rho})^{-1}$; $\scriptsize{\circled{6}}$ follows from \textbf{Lemma}~\ref{lem:barG-G}.
  We then bound term $\mathrm{\MakeTextUppercase{\romannumeral 2}}$ as follows
  \begin{align} \label{eq:bdtermII-00}
    & \mathbb{E} \left[ \left\| \sum^{t}_{\tau=1} \frac{1-\alpha^{t-\tau}}{1-\alpha} \left( \mathbf{G}^{[\tau]} (\mathbf{I} - \mathbf{Q}) \mathbf{W}^{t-\tau+1} \right) \right\|^2_{\mathfrak{F}} \right] \nonumber\\
    \leq & \sum^{t}_{\tau=1} \sum^{t}_{\tau'=1} \mathbb{E} \left[ \left\| \frac{1-\alpha^{t-\tau}}{1-\alpha} \left( \mathbf{G}^{[\tau]} (\mathbf{I} - \mathbf{Q}) \mathbf{W}^{t-\tau+1} \right) \right\|_{\mathfrak{F}} \left\| \frac{1-\alpha^{t-\tau'}}{1-\alpha} \left( \mathbf{G}^{[\tau']} (\mathbf{I} - \mathbf{Q}) \mathbf{W}^{t-\tau'+1} \right) \right\|_{\mathfrak{F}} \right]    \nonumber\\
    \leq & \frac{1}{(1-\alpha)^2} \sum^{t}_{\tau=1} \sum^{t}_{\tau'=1} \rho^{(t+1-\frac{\tau + \tau'}{2})} \mathbb{E} \left[ \left\| \mathbf{G}^{[\tau]} \right\|_{\mathfrak{F}} \left\| \mathbf{G}^{[\tau']} \right\|_{\mathfrak{F}} \right]    \nonumber\\
    \leq & \frac{1}{(1-\alpha)^2} \sum^{t}_{\tau=1} \sum^{t}_{\tau'=1} \rho^{(t+1-\frac{\tau + \tau'}{2})} \left( \frac{1}{2} \mathbb{E} \left[ \left\| \mathbf{G}^{[\tau]} \right\|^2_{\mathfrak{F}} \right] +     \frac{1}{2} \mathbb{E} \left[ \left\| \mathbf{G}^{[\tau']} \right\|^2_{\mathfrak{F}} \right] \right)    \nonumber\\
    = & \frac{1}{(1-\alpha)^2} \sum^{t}_{\tau=1} \sum^{t}_{\tau'=1} \rho^{(t+1-\frac{\tau + \tau'}{2})} \mathbb{E} \left[ \left\| \mathbf{G}^{[\tau]} \right\|^2_{\mathfrak{F}} \right]    \nonumber\\
    \leq & \frac{1}{(1-\alpha)^2 (1-\sqrt{\rho})} \sum^{t}_{\tau=1} \rho^{(\frac{t+1-\tau}{2})} \mathbb{E} \left[ \left\| \mathbf{G}^{[\tau]} \right\|^2_{\mathfrak{F}} \right]
  \end{align}
  It is apparent that the bound of $\mathrm{\MakeTextUppercase{\romannumeral 2}}$ is dependent on $\mathbb{E} \left[ \left\| \mathbf{G}^{[\tau]} \right\|^2_{\mathfrak{F}} \right]$ .
  \begin{align} \label{eq:bdtermII-01}
    \mathbb{E} \left[ \left\| \mathbf{G}^{[\tau]} \right\|^2_{\mathfrak{F}} \right]   = & \mathbb{E} \left[ \left\| \mathbf{G}^{[\tau]} - \mathbf{H}^{[\tau]} + \mathbf{H}^{[\tau]} \right\|^2_{\mathfrak{F}} \right]    \nonumber\\
    \stackrel{\scriptsize{\circled{1}}} \leq & 2 \mathbb{E} \left[ \left\| \mathbf{G}^{[\tau]} - \mathbf{H}^{[\tau]} \right\|^2_{\mathfrak{F}} \right] + 2 \mathbb{E} \left[ \left\| \mathbf{H}^{[\tau]} \right\|^2_{\mathfrak{F}} \right]    \nonumber\\
    \leq & 2 \mathbb{E} \left[\sum^N_{i=1} \left\| \nabla F_i \left( x^{[\tau-1]}_i; \xi_{i,\tau} \right) - \nabla f_i \left( x^{[\tau-1]}_i \right) \right\|^2 \right] \nonumber\\
    & + 2\mathbb{E} \left[ \left\| \mathbf{H}^{[\tau]} -   \mathbf{J}^{[\tau]} + \mathbf{J}^{[\tau]} - \mathbf{J}^{[\tau]}\mathbf{Q} + \mathbf{J}^{[\tau]}\mathbf{Q} - \mathbf{H}^{[\tau]}\mathbf{Q} + \mathbf{H}^{[\tau]}\mathbf{Q} \right\|^2_{\mathfrak{F}} \right]    \nonumber\\
    \leq & 2 \mathbb{E} \left[\sum^N_{i=1} \left\| \nabla F_i \left( x^{[\tau-1]}_i; \xi_{i,\tau} \right) - \nabla f_i \left( x^{[\tau-1]}_i \right) \right\|^2 \right]   \nonumber\\ 
    & +8\mathbb{E} \left[ \left\| \mathbf{H}^{[\tau]} - \mathbf{J}^{[\tau]} \right\|^2_{\mathfrak{F}} \right] + 8\mathbb{E} \left[ \left\| \mathbf{J}^{[\tau]} - \mathbf{J}^{[\tau]}\mathbf{Q} \right\|^2_{\mathfrak{F}} \right]     \nonumber\\
    & + 8\mathbb{E} \left[ \left\| \mathbf{J}^{[\tau]}\mathbf{Q} - \mathbf{H}^{[\tau]}\mathbf{Q}  \right\|^2_{\mathfrak{F}} \right] + 8\mathbb{E} \left[ \left\| \mathbf{H}^{[\tau]}\mathbf{Q}  \right\|^2_{\mathfrak{F}} \right]    \nonumber\\
    \leq & 2 \mathbb{E} \left[\sum^N_{i=1} \left\| \nabla F_i \left( x^{[\tau-1]}_i; \xi_{i,\tau} \right) - \nabla f_i \left( x^{[\tau-1]}_i \right) \right\|^2 \right]   \nonumber\\
    & + 8 \mathbb{E} \left[\sum^N_{i=1} \left\| \nabla f_i \left( x^{[\tau-1]}_i \right) - \nabla f_i \left( \bar{x}^{[\tau-1]} \right) \right\|^2 \right]   \nonumber\\
    & + 8 \mathbb{E} \left[\sum^N_{i=1} \left\| \nabla f_i \left(\bar{x}^{[\tau-1]} \right) - \frac{1}{N}\sum^N_{n=1} \nabla f_n \left( \bar{x}^{[\tau-1]} \right) \right\|^2 \right]  \nonumber\\
    & + 8 \mathbb{E} \left[\sum^N_{i=1} \left\| \frac{1}{N} \sum^N_{n=1} \nabla f_n \left( \bar{x}^{[\tau-1]} \right) - \frac{1}{N} \sum^N_{n=1} \nabla f_n \left( {x}^{[\tau-1]}_n \right) \right\|^2 \right]  \nonumber\\
    & + 8 \mathbb{E} \left[\sum^N_{i=1} \left\| \frac{1}{N} \sum^N_{n=1} \nabla f_n \left( {x}^{[\tau-1]}_n \right) \right\|^2 \right]  \nonumber\\
    \stackrel{\scriptsize{\circled{2}}} \leq & 2N\sigma^2 + 16L^2\sum^N_{i=1} \mathbb{E} \left[ \left\|  x^{[\tau-1]}_i - \bar{x}^{[\tau-1]} \right\|^2 \right]  + 8N\varsigma^2 + 8N\mathbb{E} \left[ \left\| \frac{1}{N} \sum^N_{i=1} \nabla f_i \left( x^{[\tau-1]}_i \right) \right\|^2 \right]
  \end{align} 
  where we have $\scriptsize{\circled{1}}$ by considering the basic inequality $\| \mathbf{A}+\mathbf{B} \|^2_{\mathfrak{F}} \leq 2\|\mathbf{A}\|^2_{\mathfrak{F}} + 2\|\mathbf{B}\|^2_{\mathfrak{F}}$; and $\scriptsize{\circled{2}}$ since 
  \begin{align*}
    \sum^N_{i=1} \left\| \nabla f_i(x^{[\tau-1]}_i) - \nabla f_i(\bar{x}^{[\tau-1]}) \right\|^2 \leq L^2 \sum^N_{i=1} \left\| x^{[\tau-1]}_i - \bar{x}^{[\tau-1]} \right\|^2
  \end{align*}
  due to the smoothness of each $f_i$ by \textbf{Assumption}~\ref{ass:L-smooth}, and 
  \begin{align*}
    \sum^N_{i=1} \left\| \frac{1}{N} \sum^N_{n=1} \nabla f_n(\bar{x}^{[\tau-1]}) - \frac{1}{N} \sum^N_{n=1} \nabla f_n({x}^{[\tau-1]}_n) \right\|^2 \leq \sum^N_{i=1} \frac{1}{N} \sum^N_{n=1} \left\| \nabla f_n(\bar{x}^{[\tau-1]}) - \nabla f_n({x}^{[\tau-1]}_n) \right\|^2 \leq L^2 \sum^N_{n=1} \left\| \bar{x}^{[\tau-1]} - {x}^{[\tau-1]}_n \right\|^2
  \end{align*}
  due to the convexity of $\|\cdot\|^2$ and \textit{Jensen's Inequality}.
  By substituting (\ref{eq:bdtermII-01}) into (\ref{eq:bdtermII-00}), we continue to derive the bound of $\mathbf{\MakeTextUppercase{\romannumeral 2}}$ as follows.
  \begin{align} \label{eq:bdtermII-02}
    & \mathbb{E} \left[ \left\| \sum^{t}_{\tau=1} \frac{1-\alpha^{t-\tau}}{1-\alpha} \left( \mathbf{G}^{[\tau]} (\mathbf{I} - \mathbf{Q}) \mathbf{W}^{t-\tau+1} \right) \right\|^2_{\mathfrak{F}} \right]    \nonumber\\
    \leq & \frac{1}{(1-\alpha)^2 (1-\sqrt{\rho})} \sum^{t}_{\tau=1} \rho^{(\frac{t+1-\tau}{2})} \left( \begin{aligned}
      & 2N\sigma^2  + 8N\varsigma^2 + 16L^2\sum^N_{i=1} \mathbb{E} \left[ \left\|  x^{[\tau-1]}_i - \bar{x}^{[\tau-1]} \right\|^2 \right] \\
      & + 8N\mathbb{E} \left[ \left\| \frac{1}{N} \sum^N_{i=1} \nabla f_i \left(x^{[\tau-1]}_i \right) \right\|^2 \right]
    \end{aligned}  \right)   \nonumber\\
    \leq & \frac{2N\sigma^2 + 8N\varsigma^2}{(1-\alpha)^2 (1-\sqrt{\rho})^2} 
    + \frac{16L^2}{(1-\alpha)^2 (1-\sqrt{\rho})} \sum^{t}_{\tau=1} \rho^{(\frac{t+1-\tau}{2})} \sum^N_{i=1} \mathbb{E} \left[ \left\|  x^{[\tau-1]}_i - \bar{x}^{[\tau-1]} \right\|^2 \right] \nonumber\\
    & + \frac{8N}{(1-\alpha)^2 (1-\sqrt{\rho})} \sum^{t}_{\tau=1} \rho^{(\frac{t+1-\tau}{2})} \mathbb{E} \left[ \left\| \frac{1}{N} \sum^N_{i=1} \nabla f_i \left(x^{[\tau-1]}_i \right) \right\|^2 \right]
  \end{align}
  By substituting (\ref{eq:bdtermI}) and (\ref{eq:bdtermII-02}) into (\ref{eq:XI-Q-00}), we have
  \begin{align} \label{eq:XI-Q-01}
    \mathbb{E} \left[ \left\| \mathbf{X}^{[t]} (\mathbf{I} - \mathbf{Q}) \right\|^2_{\mathfrak{F}} \right]
    \leq & \frac{4\gamma^2 N \sigma^2 + 16\gamma^2 N \varsigma^2}{(1-\alpha)^2 (1-\sqrt{\rho})^2} + \frac{\left(4\gamma^2 N \sigma^2 + 16\gamma^2 N \varsigma^2\right)\hat{\varphi}^2_{\max}}{\omega^4_{\min}(1-\alpha)^2 (1-\sqrt{\rho})^2} \nonumber\\
    & + \frac{32\gamma^2 L^2\hat{\varphi}^2_{\max}}{\omega^4_{\min}(1-\alpha)^2 (1-\sqrt{\rho})} \sum^{t}_{\tau=1} \rho^{(\frac{t+1-\tau}{2})} \sum^N_{i=1} \mathbb{E} \left[ \left\|  x^{[\tau-1]}_i - \bar{x}^{[\tau-1]} \right\|^2 \right]      \nonumber\\
    & + \frac{32\gamma^2  L^2}{(1-\alpha)^2 (1-\sqrt{\rho})} \sum^{t}_{\tau=1} \rho^{(\frac{t+1-\tau}{2})} \sum^N_{i=1} \mathbb{E} \left[ \left\|  x^{[\tau-1]}_i - \bar{x}^{[\tau-1]} \right\|^2 \right]  \nonumber\\
    & + \frac{16\gamma^2 N\hat{\varphi}^2_{\max}}{\omega^4_{\min}(1-\alpha)^2 (1-\sqrt{\rho})} \sum^{t}_{\tau=1} \rho^{(\frac{t+1-\tau}{2})} \mathbb{E} \left[ \left\| \frac{1}{N} \sum^N_{i=1} \nabla f_i \left( x^{[\tau-1]}_i \right) \right\|^2 \right]  \nonumber\\
    & + \frac{16\gamma^2 N}{(1-\alpha)^2 (1-\sqrt{\rho})} \sum^{t}_{\tau=1} \rho^{(\frac{t+1-\tau}{2})} \mathbb{E} \left[ \left\| \frac{1}{N} \sum^N_{i=1} \nabla f_i \left( x^{[\tau-1]}_i \right) \right\|^2 \right]  \nonumber\\
    & + \frac{4\gamma^2 N \sigma^2 + 16\gamma^2 N \varsigma^2}{(1-\alpha)^2 (1-\sqrt{\rho})^2}   \nonumber\\
    & + \frac{32\gamma^2 L^2}{(1-\alpha)^2 (1-\sqrt{\rho})} \sum^{t}_{\tau=1} \rho^{(\frac{t+1-\tau}{2})} \sum^N_{i=1} \mathbb{E} \left[ \left\|  x^{[\tau-1]}_i - \bar{x}^{[\tau-1]} \right\|^2 \right]    \nonumber\\
    & + \frac{16\gamma^2 N}{(1-\alpha)^2 (1-\sqrt{\rho})} \sum^{t}_{\tau=1} \rho^{\left( \frac{t+1-\tau}{2} \right)} \mathbb{E} \left[ \left\| \frac{1}{N} \sum^N_{i=1} \nabla f_i \left( x^{[\tau-1]}_i \right) \right\|^2 \right]    \nonumber\\
    \leq &  \frac{8\gamma^2 N \sigma^2 + 32\gamma^2 N \varsigma^2}{(1-\alpha)^2 (1-\sqrt{\rho})^2} + \frac{\left(4\gamma^2 N \sigma^2 + 16\gamma^2 N \varsigma^2\right)\hat{\varphi}^2_{\max}}{\omega^4_{\min}(1-\alpha)^2 (1-\sqrt{\rho})^2}  \nonumber\\
    & + \frac{32\gamma^2 L^2\hat{\varphi}^2_{\max}}{\omega^4_{\min}(1-\alpha)^2 (1-\sqrt{\rho})} \sum^{t}_{\tau=1} \rho^{(\frac{t+1-\tau}{2})} \sum^N_{i=1} \mathbb{E} \left[ \left\|  x^{[\tau-1]}_i - \bar{x}^{[\tau-1]} \right\|^2 \right]      \nonumber\\
    & + \frac{64\gamma^2 L^2}{(1-\alpha)^2 (1-\sqrt{\rho})} \sum^{t}_{\tau=1} \rho^{(\frac{t+1-\tau}{2})} \sum^N_{i=1} \mathbb{E} \left[ \left\|  x^{[\tau-1]}_i - \bar{x}^{[\tau-1]} \right\|^2 \right]    \nonumber\\
    & + \frac{16\gamma^2 N\hat{\varphi}^2_{\max}}{\omega^4_{\min}(1-\alpha)^2 (1-\sqrt{\rho})} \sum^{t}_{\tau=1} \rho^{(\frac{t+1-\tau}{2})} \mathbb{E} \left[ \left\| \frac{1}{N} \sum^N_{i=1} \nabla f_i \left( x^{[\tau-1]}_i \right) \right\|^2 \right]  \nonumber\\
    & + \frac{32\gamma^2 N}{(1-\alpha)^2 (1-\sqrt{\rho})} \sum^{t}_{\tau=1} \rho^{\left( \frac{t+1-\tau}{2} \right)} \mathbb{E} \left[ \left\| \frac{1}{N} \sum^N_{i=1} \nabla f_i \left( x^{[\tau-1]}_i \right) \right\|^2 \right]
  \end{align}
  Summing over $t = 1, 2, \cdots, T$ and noting that $\mathbb{E} \left[ \left\| \mathbf{X}^{[0]} (\mathbf{I} - \mathbf{Q}) \right\|^2_{\mathfrak{F}} \right] = 0$, we have
  \begin{align} \label{eq:XI-Q-03}
    \sum^{T}_{t=1} \mathbb{E} \left[ \left\| \mathbf{X}^{[t]} (\mathbf{I} - \mathbf{Q}) \right\|^2_{\mathfrak{F}} \right]
    \leq & T \cdot \frac{8\gamma^2 N \sigma^2 + 32\gamma^2 N \varsigma^2}{(1-\alpha)^2 (1-\sqrt{\rho})^2} + T \cdot \frac{\left(4\gamma^2 N \sigma^2 + 16\gamma^2 N \varsigma^2\right)\hat{\varphi}^2_{\max}}{\omega^4_{\min}(1-\alpha)^2 (1-\sqrt{\rho})^2} \nonumber\\
    & + \frac{32\gamma^2 L^2\hat{\varphi}^2_{\max}}{\omega^4_{\min}(1-\alpha)^2 (1-\sqrt{\rho})} \sum^{T}_{t=1} \sum^{t}_{\tau=1} \rho^{(\frac{t+1-\tau}{2})} \sum^N_{i=1} \mathbb{E} \left[ \left\|  x^{[\tau-1]}_i - \bar{x}^{[\tau-1]} \right\|^2 \right]      \nonumber\\
    & + \frac{64\gamma^2 L^2}{(1-\alpha)^2 (1-\sqrt{\rho})} \sum^{T}_{t=1} \sum^{t}_{\tau=1} \rho^{(\frac{t+1-\tau}{2})} \sum^N_{i=1} \mathbb{E} \left[ \left\|  x^{[\tau-1]}_i - \bar{x}^{[\tau-1]} \right\|^2 \right]    \nonumber\\
    & + \frac{16\gamma^2 N\hat{\varphi}^2_{\max}}{\omega^4_{\min}(1-\alpha)^2 (1-\sqrt{\rho})} \sum^{T}_{t=1} \sum^{t}_{\tau=1} \rho^{(\frac{t+1-\tau}{2})} \mathbb{E} \left[ \left\| \frac{1}{N} \sum^N_{i=1} \nabla f_i \left( x^{[\tau-1]}_i \right) \right\|^2 \right]  \nonumber\\
    & + \frac{32\gamma^2 N}{(1-\alpha)^2 (1-\sqrt{\rho})} \sum^{T}_{t=1} \sum^{t}_{\tau=1} \rho^{\left( \frac{t+1-\tau}{2} \right)} \mathbb{E} \left[ \left\| \frac{1}{N} \sum^N_{i=1} \nabla f_i \left( x^{[\tau-1]}_i \right) \right\|^2 \right]  \nonumber\\
    \leq & 4T \gamma^2 N \frac{\left(\frac{\hat{\varphi}^2_{\max}}{\omega^4_{\min}} +2 \right) \left(\sigma^2 + 4\varsigma^2 \right)}{(1-\alpha)^2(1-\sqrt{\rho})^2}  \nonumber\\
    & + \frac{32\gamma^2 L^2 \left( \frac{\hat{\varphi}^2_{\max}}{\omega^4_{\min}} + 2 \right)}{(1-\alpha)^2 (1-\sqrt{\rho})^2} \sum^{T}_{t=1} \sum^N_{i=1} \mathbb{E} \left[ \left\|  x^{[t]}_i - \bar{x}^{[t]} \right\|^2 \right]     \nonumber\\
    & + \frac{16\gamma^2 N \left( \frac{\hat{\varphi}^2_{\max}}{\omega^4_{\min}} + 2 \right)}{(1-\alpha)^2 (1-\sqrt{\rho})^2} \sum^{T}_{t=1} \mathbb{E} \left[ \left\| \frac{1}{N} \sum^N_{i=1} \nabla f_i \left( x^{[t-1]}_i \right) \right\|^2 \right]
  \end{align}
  Dividing the both sides of (\ref{eq:XI-Q-03}) by $N$, we have
  \begin{align} \label{eq:XI-Q-04}
    \sum^{T}_{t=1} \frac{1}{N} \mathbb{E} \left[ \left\| \mathbf{X}^{[t]} (\mathbf{I} - \mathbf{Q}) \right\|^2_{\mathfrak{F}} \right]
    \leq & 4T \gamma^2 \frac{\left(\frac{\hat{\varphi}^2_{\max}}{\omega^4_{\min}} +2 \right) \left(\sigma^2 + 4\varsigma^2 \right)}{(1-\alpha)^2(1-\sqrt{\rho})^2}   \nonumber\\
    & + \frac{32\gamma^2 L^2 \left( \frac{\hat{\varphi}^2_{\max}}{\omega^4_{\min}} + 2 \right)}{(1-\alpha)^2 (1-\sqrt{\rho})^2} \sum^{T}_{t=1} \frac{1}{N} \mathbb{E} \left[ \left\| \mathbf{X}^{[t]} (\mathbf{I} - \mathbf{Q}) \right\|^2_{\mathfrak{F}} \right]   \nonumber\\
    & + \frac{16\gamma^2 \left( \frac{\hat{\varphi}^2_{\max}}{\omega^4_{\min}} + 2 \right)}{(1-\alpha)^2 (1-\sqrt{\rho})^2} \sum^{T}_{t=1} \mathbb{E} \left[ \left\| \frac{1}{N} \sum^N_{i=1} \nabla f_i(x^{[t-1]}_i) \right\|^2 \right]   
  \end{align}
  Rearranging terms and dividing the both sides of (\ref{eq:XI-Q-04}) by $1- \frac{32\gamma^2 L^2 \left( \frac{\hat{\varphi}^2_{\max}}{\omega^4_{\min}} + 2 \right)}{(1-\alpha)^2 (1-\sqrt{\rho})^2}$, we complete the proof as follows
  \begin{align} \label{eq:lem6final}
    & \sum^{T}_{t=1} \frac{1}{N} \sum^N_{i=1} \mathbb{E} \left[ \left\| \bar{x}^{[t]} - x^{[t]}_i \right\|^2 \right]  \nonumber\\
    \leq & \frac{1}{1- \frac{32\gamma^2 L^2 \left( \frac{\hat{\varphi}^2_{\max}}{\omega^4_{\min}} + 2 \right)}{(1-\alpha)^2 (1-\sqrt{\rho})^2}} \left(4 T \gamma^2 \frac{\left( \frac{\hat{\varphi}^2_{\max}}{\omega^4_{\min}} + 2 \right) \left(\sigma^2 + 4\varsigma^2 \right)}{(1-\alpha)^2(1-\sqrt{\rho})^2} \right)  \nonumber\\
    & + \frac{1}{1- \frac{32\gamma^2 L^2 \left( \frac{\hat{\varphi}^2_{\max}}{\omega^4_{\min}} + 2 \right)}{(1-\alpha)^2 (1-\sqrt{\rho})^2}} \left( \frac{16\gamma^2 \left(\frac{\hat{\varphi}^2_{\max}}{\omega^4_{\min}} + 2 \right) }{(1-\alpha)^2 (1-\sqrt{\rho})^2} \sum^{T}_{t=1} \mathbb{E} \left[ \left\| \frac{1}{N} \sum^N_{i=1} \nabla f_i \left(x^{[t-1]}_i \right) \right\|^2 \right] \right)  \nonumber\\
    \stackrel{\scriptsize{\circled{1}}} \leq & 8 T \gamma^2 \frac{\left( \frac{\hat{\varphi}^2_{\max}}{\omega^4_{\min}} +2 \right) \left( \sigma^2 + 4\varsigma^2 \right)}{(1-\alpha)^2 (1-\sqrt{\rho})^2} + \frac{1}{2L^2} \sum^{T}_{t=1} \mathbb{E} \left[ \left\| \frac{1}{N} \sum^N_{i=1} \nabla f_i \left( x^{[t-1]}_i \right) \right\|^2 \right]
  \end{align}
  where $\scriptsize{\circled{1}}$ follows because $\gamma \leq \frac{(1-\alpha)(1-\sqrt{\rho})}{8L\sqrt{\frac{\hat{\varphi}^2_{\max}}{\omega^4_{\min}} + 2}}$ is chosen to ensure $\left( 1 - \frac{32\gamma^2 L^2 \left(\frac{\hat{\varphi}^2_{\max}}{\omega^4_{\min}} + 2 \right)}{(1-\alpha)^2 (1-\sqrt{\rho})^2} \right)^{-1} \leq 2$ and $\frac{16\gamma^2 \left( \frac{\hat{\varphi}^2_{\max}}{\omega^4_{\min}} + 2 \right) }{(1-\alpha)^2 (1-\sqrt{\rho})^2} \leq \frac{1}{4L^2}$.

\section{More Details of Experiment Setup} \label{sec:refalgo}

  To simulate non-IID data across agents, we use the Dirichlet distribution $\mathsf{Dir}(\mu\mathbf{1}_Y)$ produce the probability vector of the data labels, where $\mu$ controls the degree of heterogeneity, $\mathbf{1}_Y$ is a $Y$-dimensional all-ones vector, and $Y$ denotes the number of labels~\cite{LinKSJ-ICML21,YurochkinAGGHK-ICML19}. With $\mu \to 0$, we have the data distribution more heterogeneous across the agents. In our experiments, we let $\mu = 0.25$ for both MNIST and CIFAR-10 datasets. For a more practical evaluation, we consider the following representative vulnerable scenarios based on the non-IID data setting.
  \begin{itemize}
    \item \textbf{Long-tailed distribution}~\cite{CaoWGAM-NIPS19}. We construct long-tailed versions of MNIST and CIFAR-10 datasets by reducing the number of training samples in each class of the original datasets. Specifically, we control the class-wise sample distribution using the \textit{Imbalance Ratios} (IR), which is defined as the ratio between the number of samples in the most frequent and least frequent class.
    \item \textbf{Data noise}~\cite{SunLZXLLQR-KDD23}. In this case, we assume that malicious agents may inject random noise into their local data. Specifically, the noise obeys Gaussian distribution $\mathcal{N}(0,I)$.
    \item \textbf{Label noise}~\cite{TolpeginTGL-ESORICS20}. We suppose that a malicious agent may perturb the labels of its local data to destroy the convergence of decentralized learning. For example, the label of each local data in the Byzantine agents is flipped from $y$ to $(y+1) \% Y$ where $y \in \{0,1,\cdots,Y-1\}$.
    \item \textbf{Gradient poisoning}~\cite{BaruchBG-NIPS19}. We assume that malicious agents exploit the distributional characteristics of gradients across different dimensions and carefully design perturbation magnitudes to perform slight poisoning on the gradients, thereby disrupting the convergence of the global model.
  \end{itemize}

  In our experiments, we compare ROSS with the following state-of-the-art algorithms. 
  \begin{itemize}
    \item \textbf{DMSGD}~\cite{YuJY-ICML19}: \textit{Decentralized Momentum SGD} (DMSGD) is a decentralized stochastic optimization algorithm based on the notion of momentum. Through this algorithm, multiple agents collaborate to learn from a distributed dataset. Specifically, during the iterative optimization process, each agent shares with its neighbors its local model parameters and gradient-momentum values over a (fixed) communication topology. Although the heavy-ball acceleration method adopted in gradient-based optimization is utilized in DMSGD to improve its efficiency, the abnormal data distribution across the agents is not taken into account.
    \item \textbf{CGA}~\cite{EsfandiariTJBHHS-ICML21}: \textit{Cross-Gradient Aggregation} (CGA) algorithm is a decentralized learning algorithm enabling effective learning from non-IID data distributions. In particular, each agent collects the gradients with respect to its neighbors' datasets and updates its model using a projected gradient based on quadratic programming.
    \item \textbf{NET-FLEET}~\cite{ZhangFLYLZ-MobiHoc22}: In the \textit{Decentralized Networked Federated Learning with Recursive Gradient Correction} (NET-FLEET) algorithm, a recursive gradient correction technique is used to deal with heterogeneous data distribution across the agents, such that each agent runs multiple local updates between two consecutive communication rounds with their neighbors.
    \item \textbf{MEDIAN}~\cite{YinCKB-ICML18}: \textit{Median-based Gradient Descent} (MEDIAN) is originally designed for federated learning with a central server. In our experiments, we adapt it to the decentralized setting. Specifically, in each round, each agent exchanges its local model with its neighbors and then computes the coordinate-wise median of the received models.
    \item \textbf{TRIM-MEAN}~\cite{YinCKB-ICML18}: The \textit{Trimmed-mean-based Gradient Descent} (TRIM-MEAN) algorithm is also originally designed for federated learning. In our decentralized setting, each agent collects local models from its neighbors, discards a fixed fraction of the largest and smallest coordinate values, and then averages the remaining ones.
    \item \textbf{LEARN}~\cite{El-MhamdiFGGHR-NIPS21}: 
    In this algorithm, the agents exchange both local model updates and local models with their neighbor, and adopt Trimmean aggregation rule to combine the received local model updates and local models. Specifically, in each round $t$, each agent aggregates local model updates received from its neighbors for $\lceil \log_2 t \rceil$ times, and then exchanges local models with its neighbors once.
    \item \textbf{BALANCE}~\cite{FangZHKLLLG-CCS24}: \textit{Byzantine-robust Averaging through Local Similarity in Decentralization} (BALANCE) is designed to defend against model poisoning attacks in decentralized learning. Specifically, each agent utilizes its local model as a reference to evaluate the similarity of a received model and determine whether it is benign or malicious. In our experiments, we adapt this novel similarity-based aggregation to the cross-gradient aggregation step, so as to address our data challenges.
  \end{itemize}


\section{Experiment Results on Ring Graphs} \label{sec:ring-loss-acc}
  In this section, we report our experiment results in terms of average loss and test accuracy on ring graphs. Specifically, the experiment results on MNIST dataset are presented in Fig.~\ref{fig:mnist-loss-ring}-\ref{fig:mnist-acc-ring}, while those on CIFAR-10 dataset are reported in Fig.~\ref{fig:cifar-loss-ring}-\ref{fig:cifar-acc-ring}.

  \subsection{Experiment Results on MNIST Dataset}  \label{ssec:res-mnist-ring}
  \begin{figure*}[htb!]
  \begin{center}
    \parbox{.24\textwidth}{\center\includegraphics[width=.24\textwidth]{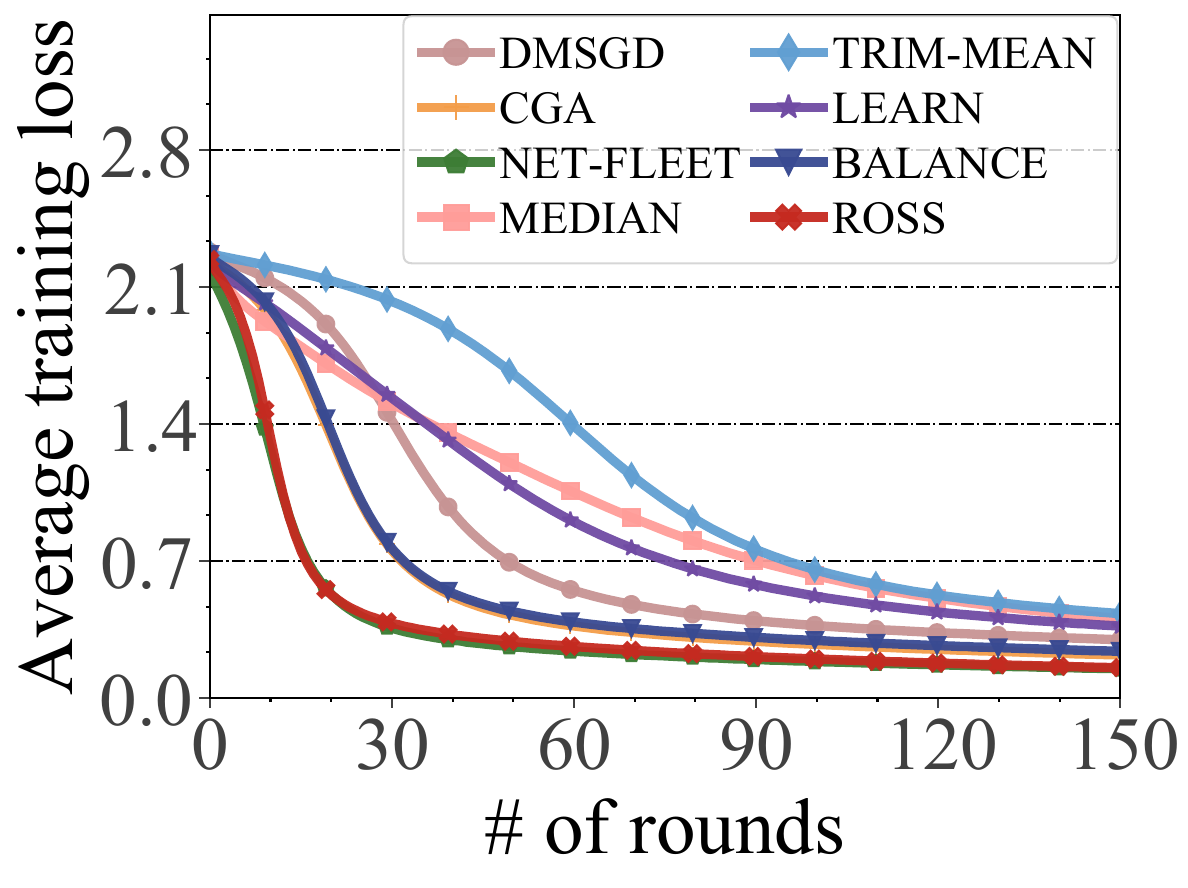}}
    \parbox{.24\textwidth}{\center\includegraphics[width=.24\textwidth]{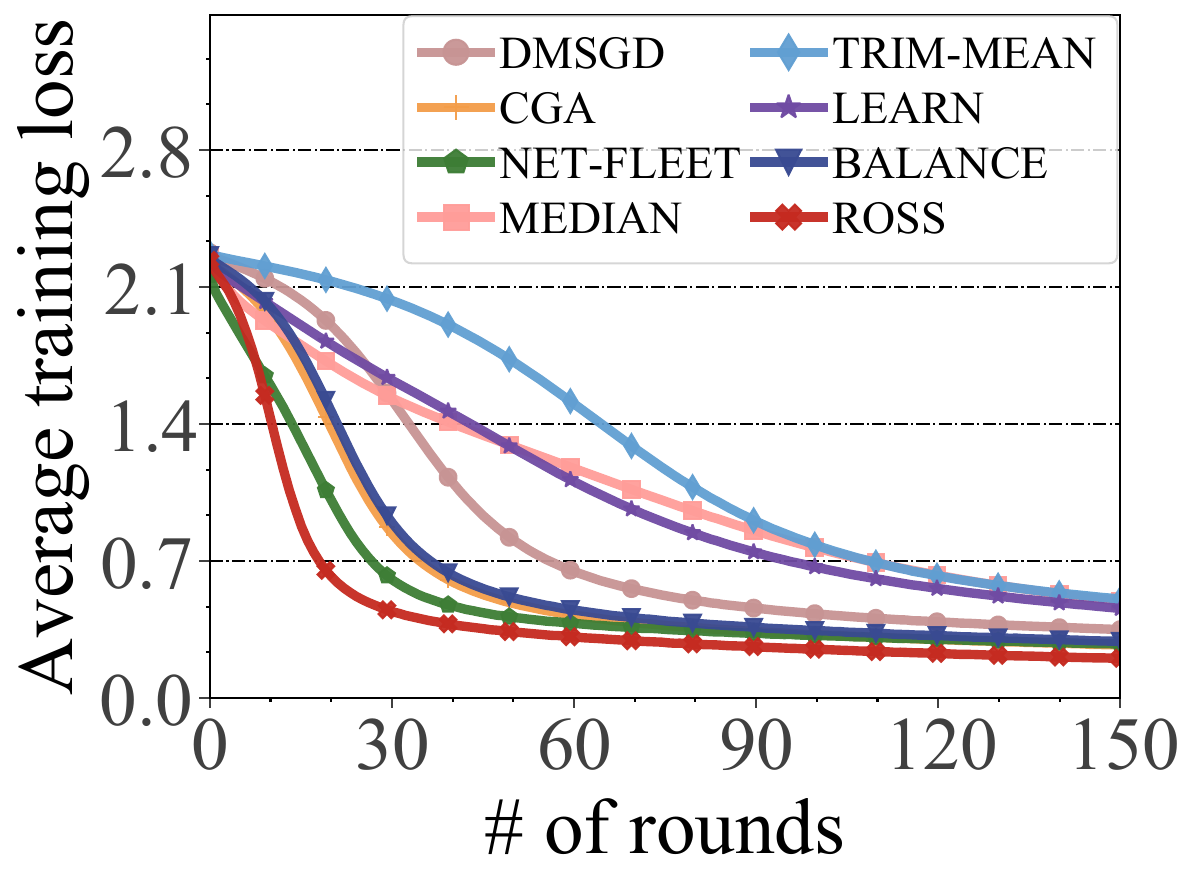}}
    \parbox{.24\textwidth}{\center\includegraphics[width=.24\textwidth]{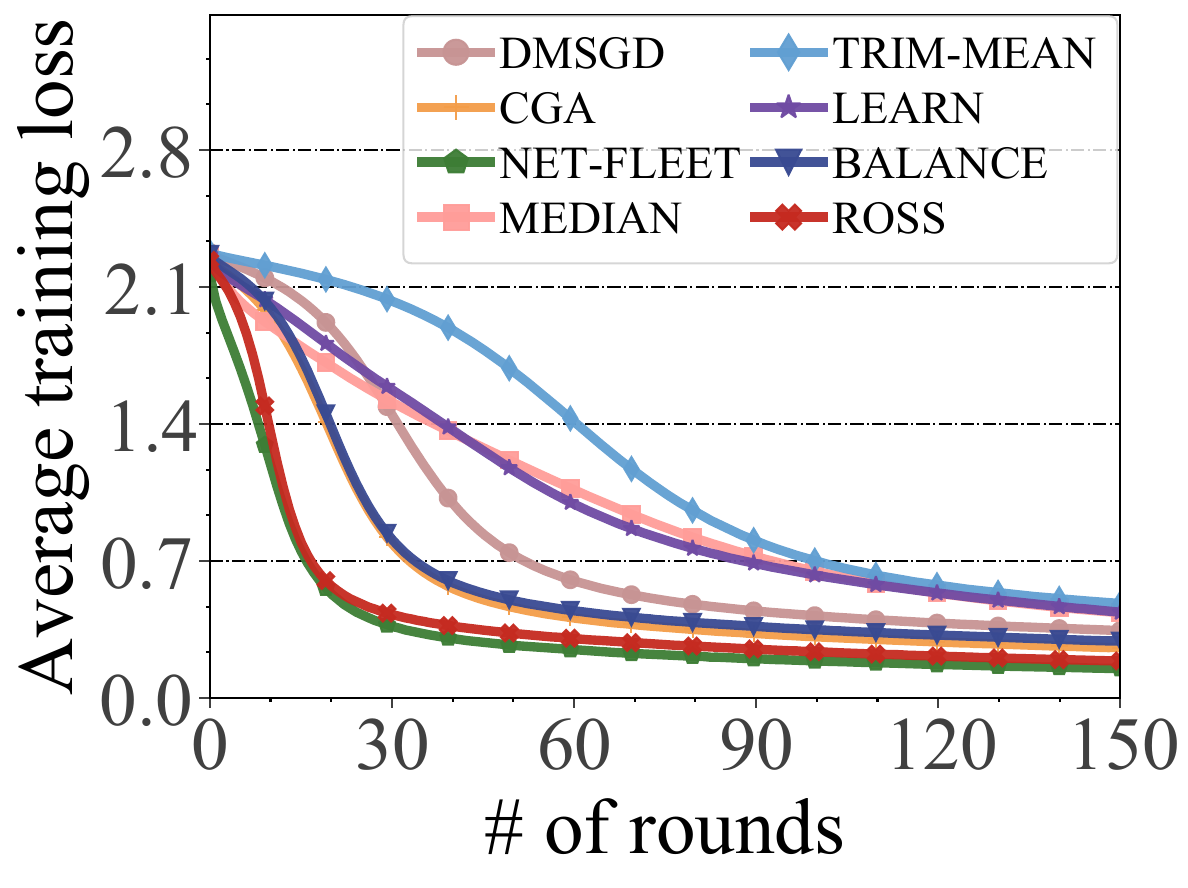}}
    \parbox{.24\textwidth}{\center\includegraphics[width=.24\textwidth]{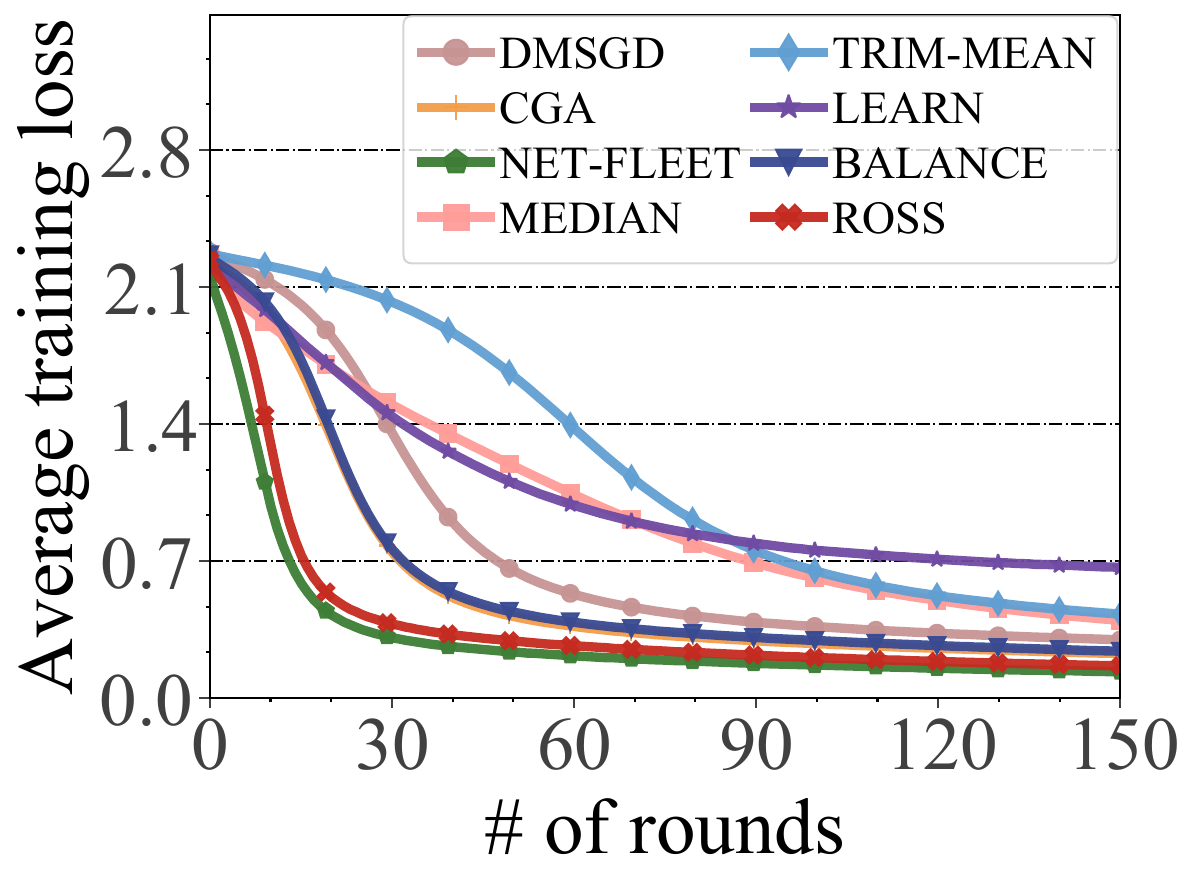}}
    \parbox{.25\textwidth}{\center\scriptsize(a1) Long-tailed ($N=10$)}
    \parbox{.23\textwidth}{\center\scriptsize(a2) Data noise ($N=10$)}
    \parbox{.23\textwidth}{\center\scriptsize(a3) Label noise ($N=10$)}
    \parbox{.25\textwidth}{\center\scriptsize(a4) Gradient poisoning ($N=10$)}
    \parbox{.24\textwidth}{\center\includegraphics[width=.24\textwidth]{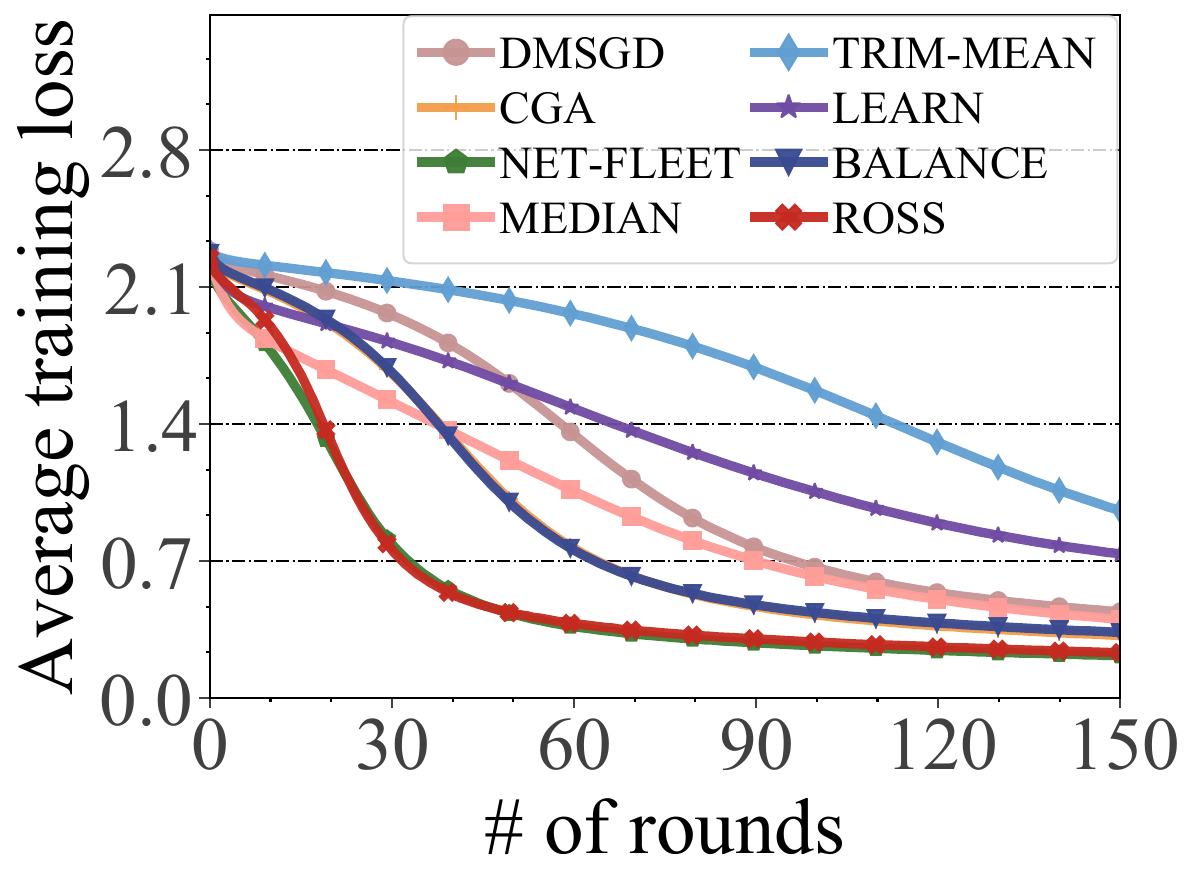}}
    \parbox{.24\textwidth}{\center\includegraphics[width=.24\textwidth]{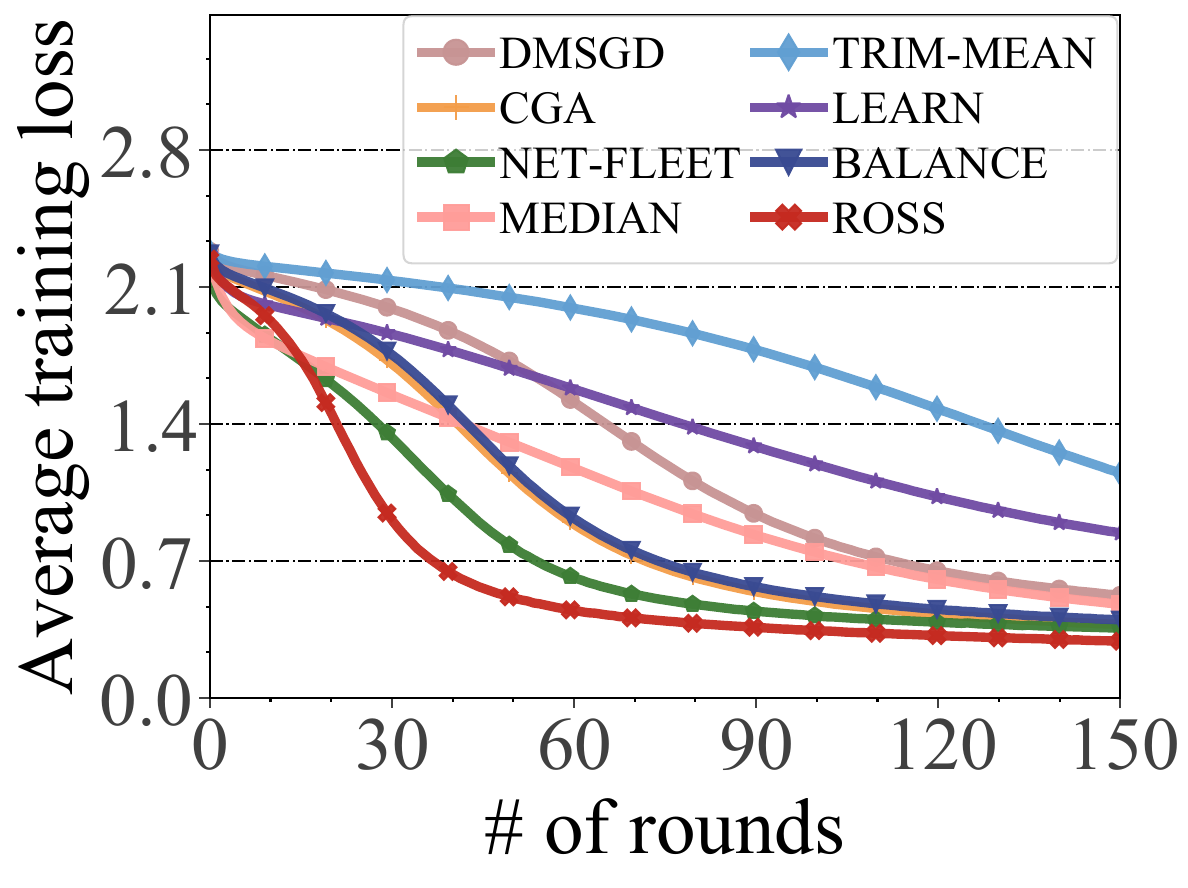}}
    \parbox{.24\textwidth}{\center\includegraphics[width=.24\textwidth]{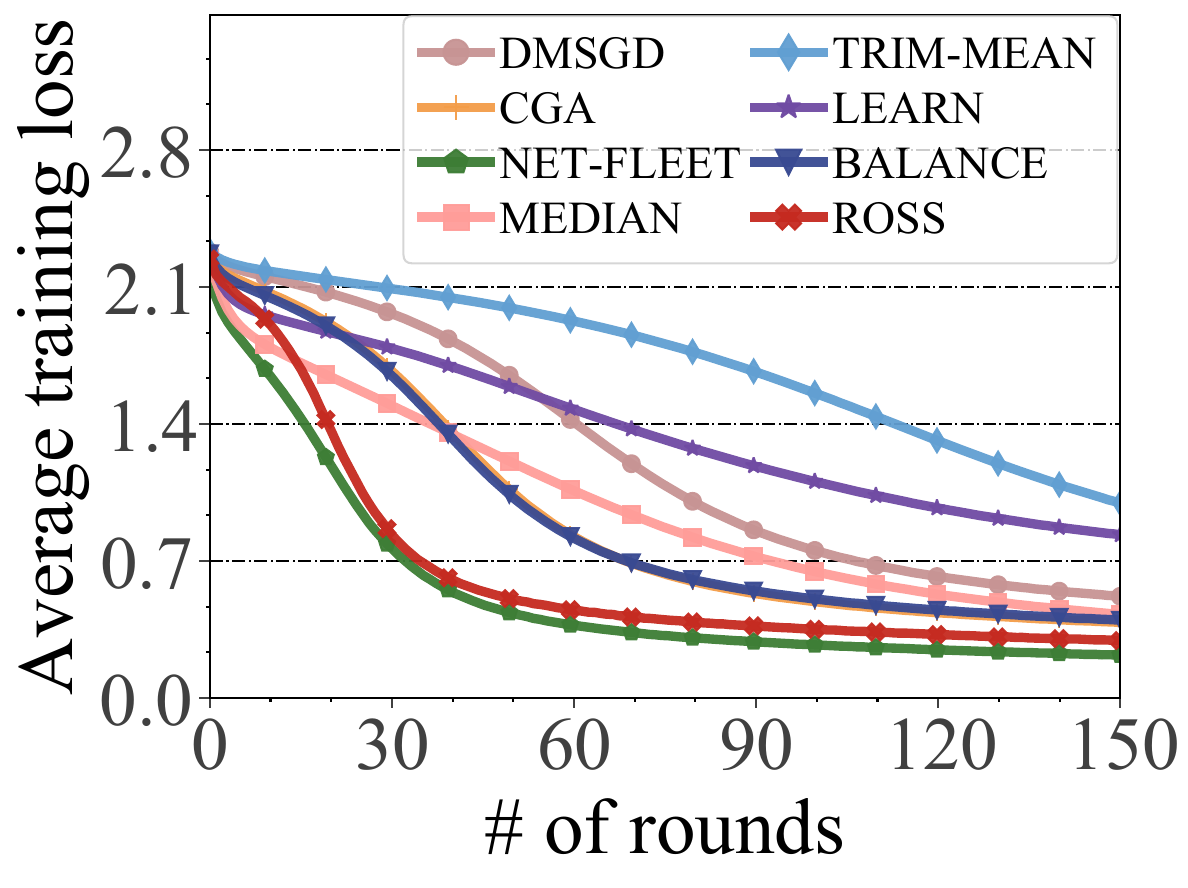}}
    \parbox{.24\textwidth}{\center\includegraphics[width=.24\textwidth]{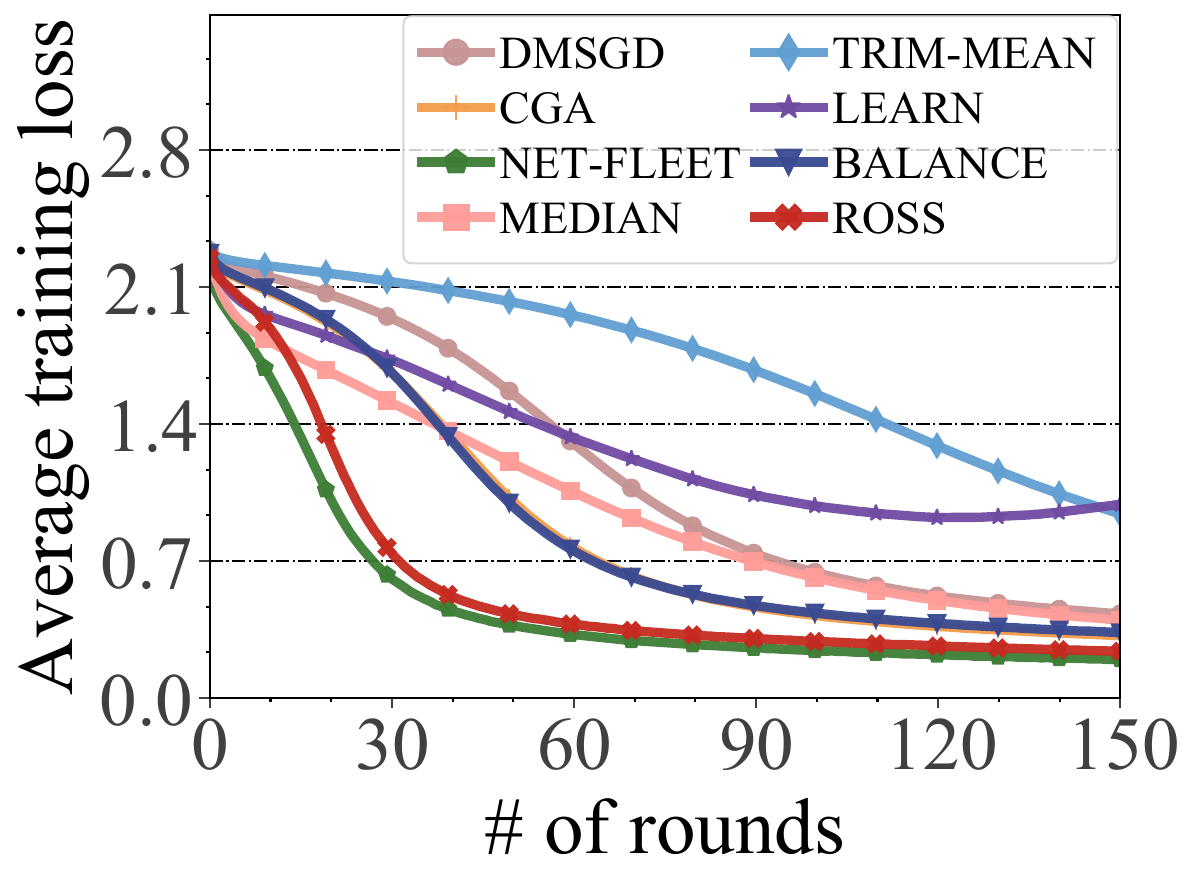}}
    \parbox{.25\textwidth}{\center\scriptsize(b1) Long-tailed ($N=20$)}
    \parbox{.23\textwidth}{\center\scriptsize(b2) Data noise ($N=20$)}
    \parbox{.23\textwidth}{\center\scriptsize(b3) Label noise ($N=20$)}
    \parbox{.25\textwidth}{\center\scriptsize(b4) Gradient poisoning ($N=20$)}
    \parbox{.24\textwidth}{\center\includegraphics[width=.24\textwidth]{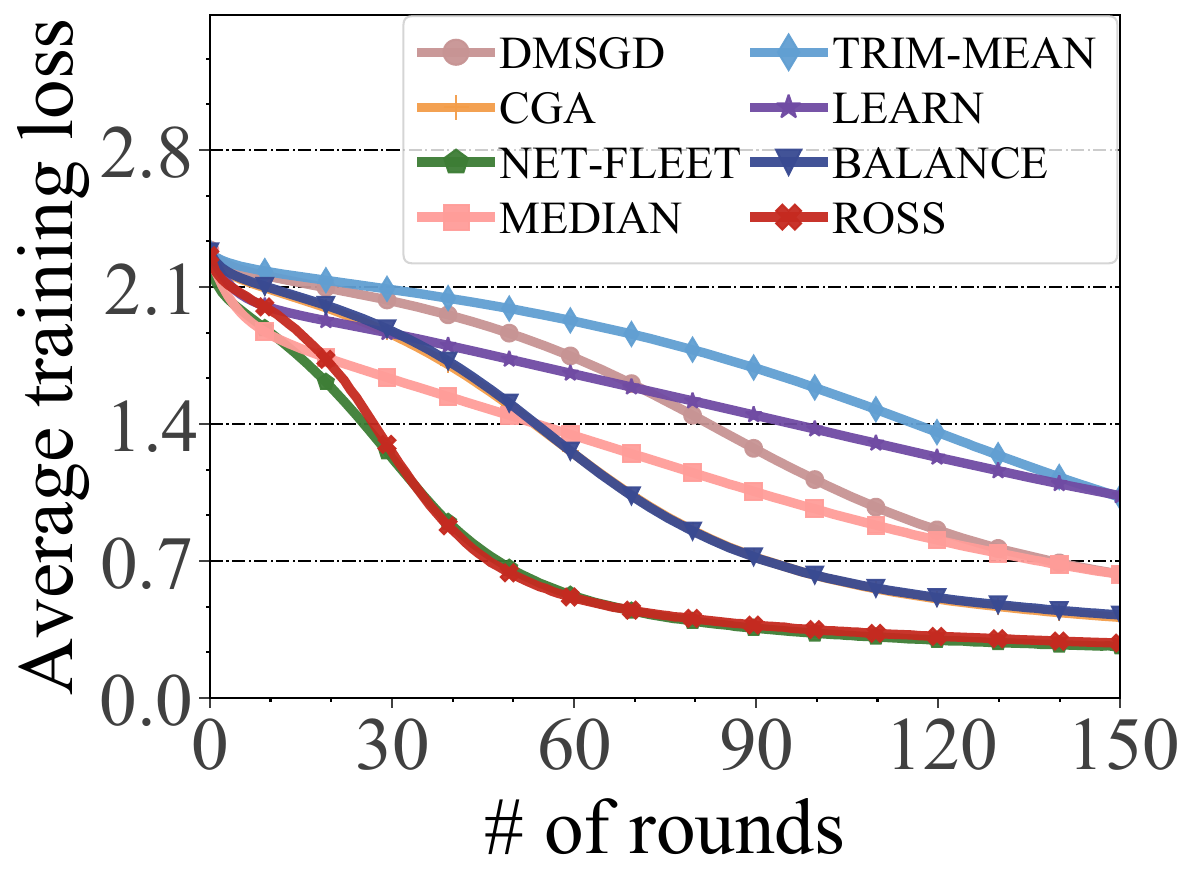}}
    \parbox{.24\textwidth}{\center\includegraphics[width=.24\textwidth]{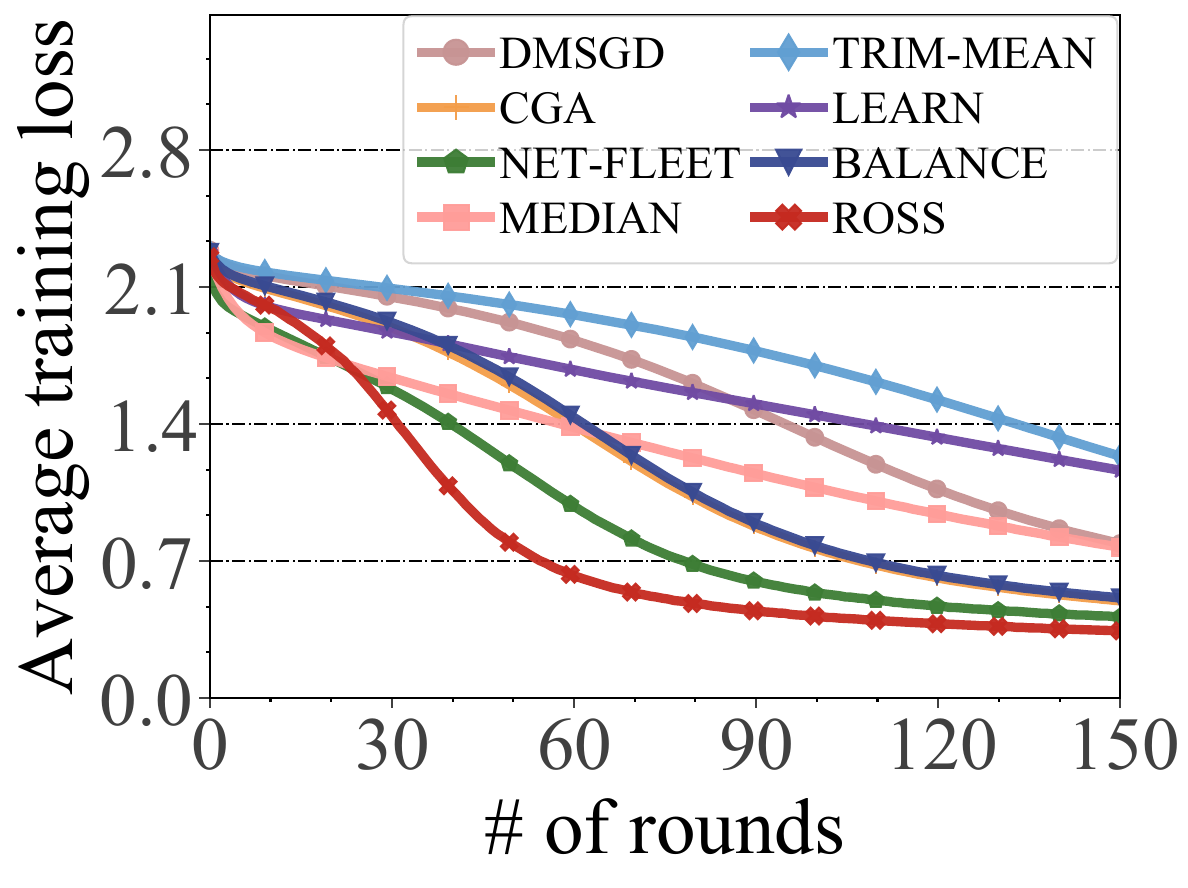}}
    \parbox{.24\textwidth}{\center\includegraphics[width=.24\textwidth]{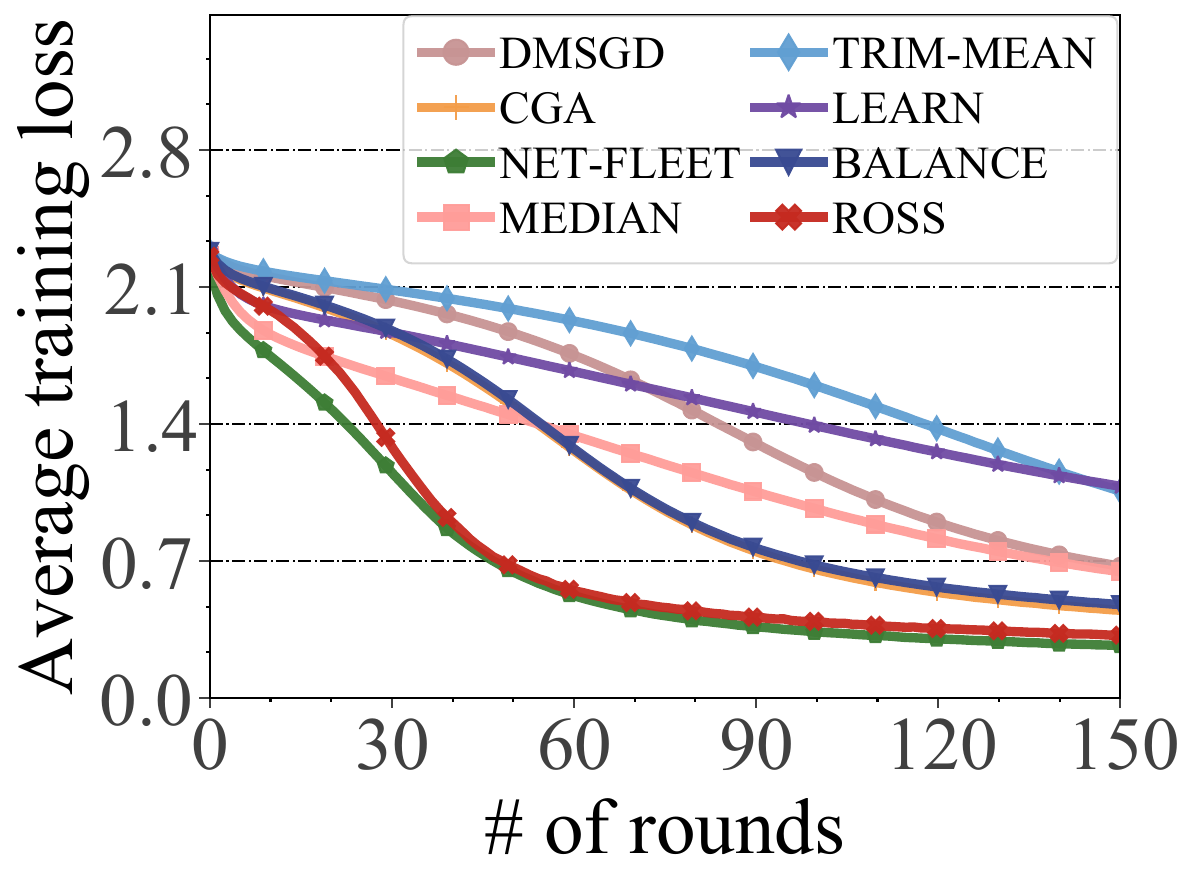}}
    \parbox{.24\textwidth}{\center\includegraphics[width=.24\textwidth]{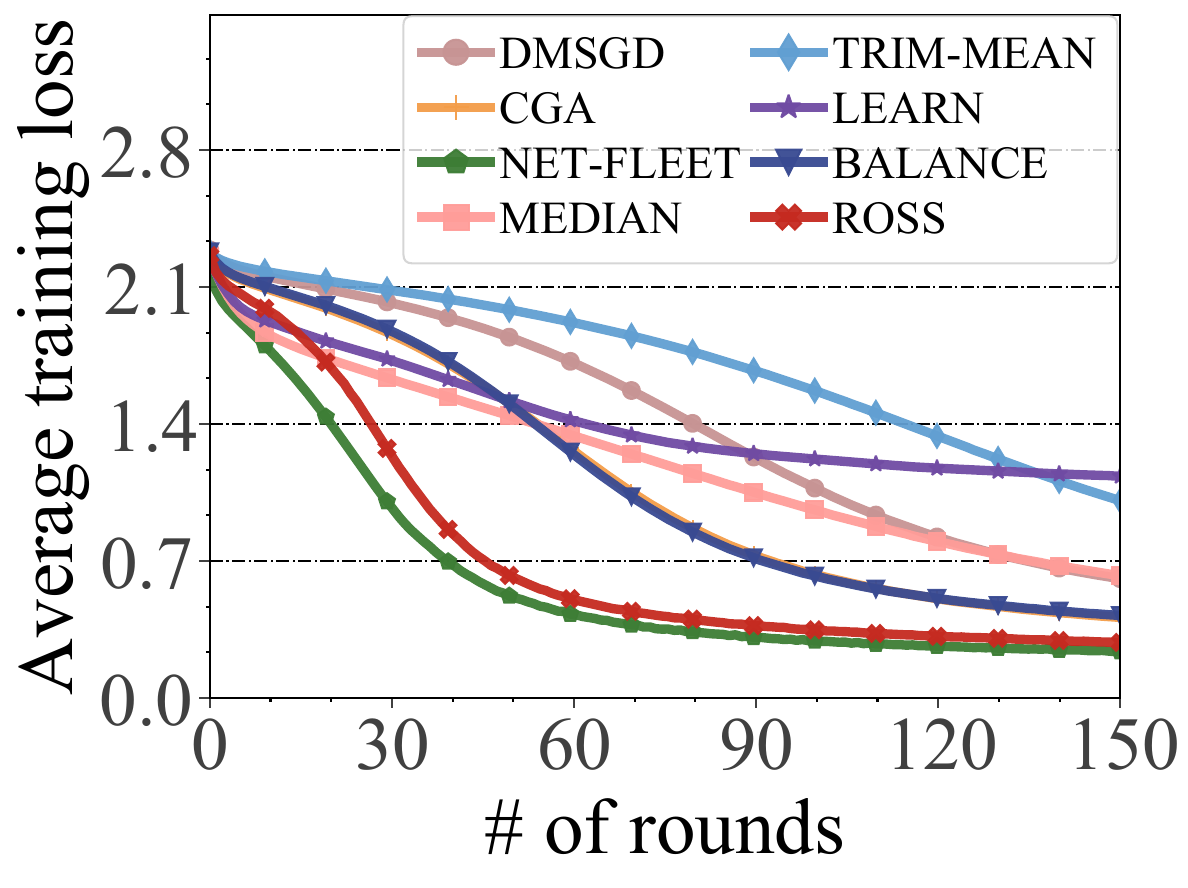}}
    \parbox{.25\textwidth}{\center\scriptsize(c1) Long-tailed ($N=30$)}
    \parbox{.23\textwidth}{\center\scriptsize(c2) Data noise ($N=30$)}
    \parbox{.23\textwidth}{\center\scriptsize(c3) Label noise ($N=30$)}
    \parbox{.25\textwidth}{\center\scriptsize(c4) Gradient poisoning ($N=30$)}
  \caption{Comparison results on MNIST dataset over ring graphs.}
  \label{fig:mnist-loss-ring}
  \end{center}
  \end{figure*}
  \begin{figure*}[htb!]
  \begin{center}
    \parbox{.32\textwidth}{\center\includegraphics[width=.32\textwidth]{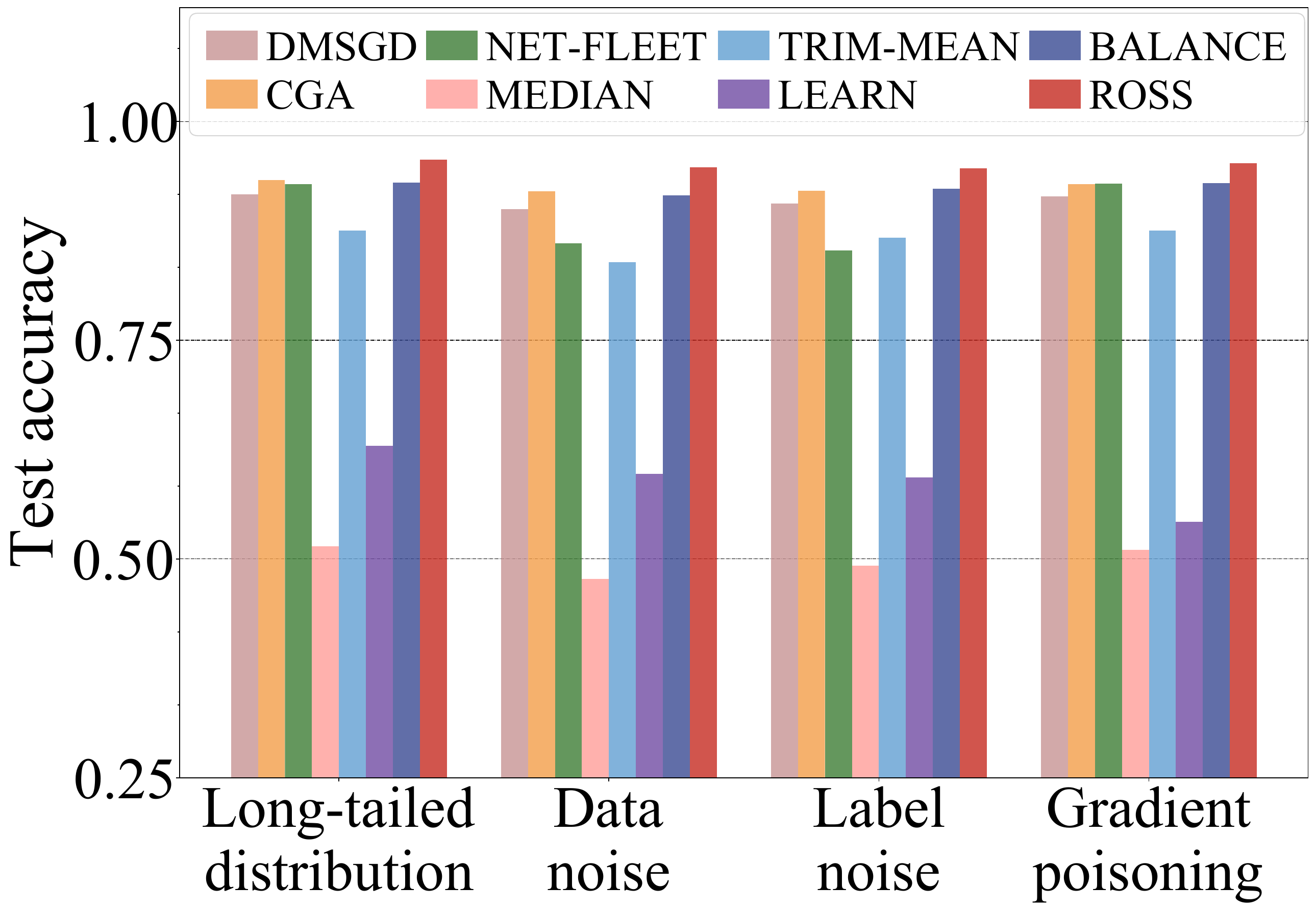}}
    \parbox{.32\textwidth}{\center\includegraphics[width=.32\textwidth]{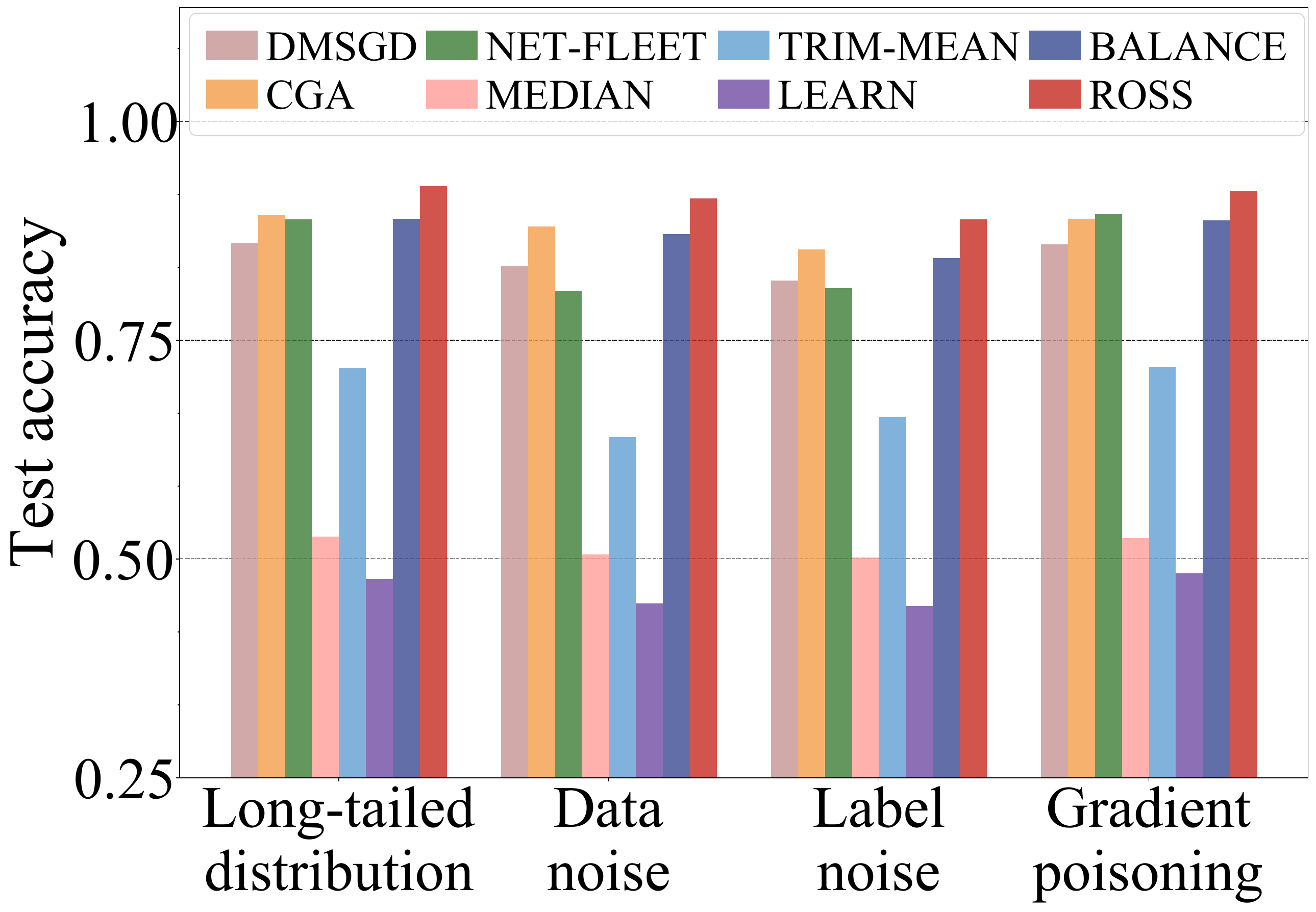}}
    \parbox{.32\textwidth}{\center\includegraphics[width=.32\textwidth]{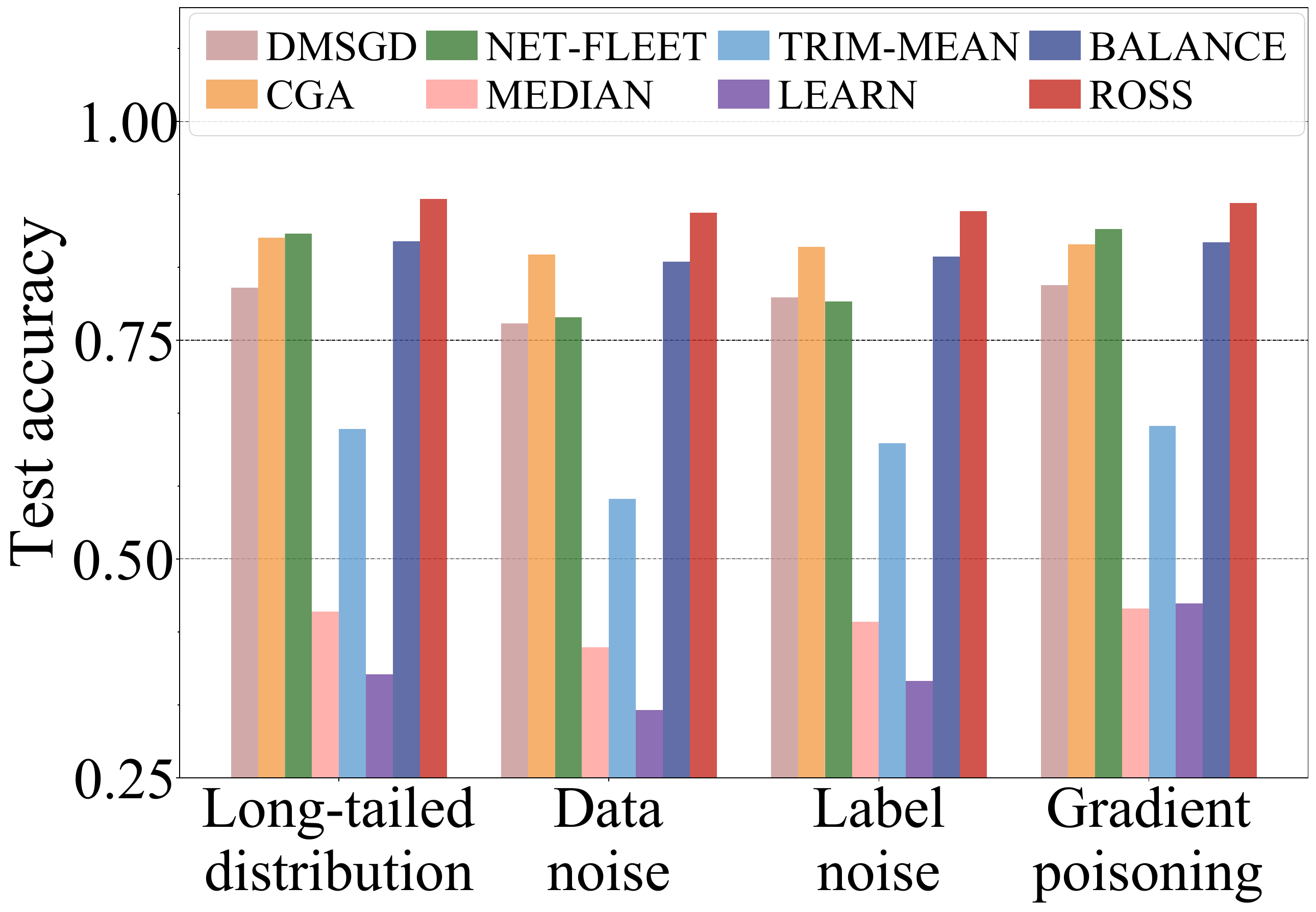}}
    \parbox{.32\textwidth}{\center\scriptsize(a) $N=10$}
    \parbox{.32\textwidth}{\center\scriptsize(b) $N=20$}
    \parbox{.32\textwidth}{\center\scriptsize(c) $N=30$}
  \caption{Comparison results in terms of test accuracy on MNIST dataset over ring graphs.}
  \label{fig:mnist-acc-ring}
  \end{center}
  \end{figure*}
  As shown in Fig.~\ref{fig:mnist-loss-ring}, even over sparser communication graphs, our ROSS algorithm still has much higher convergence rate than the other baseline algorithms across all settings. Specifically, ROSS achieves its convergence in about $70-90$ rounds across all settings, and its convergence rate remains consistent even as the number of agents increases, ensuring the $70-90$ rounds remain sufficient for convergence. However, when the number of agents increases, the other baselines require more communication rounds to achieve convergence. Moreover, compared to other baseline algorithms, ROSS achieves a lower average loss. For example, when $N=30$, the average loss of ROSS algorithm is about $0.25$ in all settings when the convergence is reached; it is $1.5-3.5\times$ smaller than that of the baselines. Furthermore, we observe that ROSS and NET-FLEET may have similar convergence performance, but as we show later, our ROSS algorithm has remarkable advantage in terms of test accuracy.

  We present the experiment results in terms of prediction accuracy obtained on ring graphs in Fig.~\ref{fig:mnist-acc-ring}. It is illustrated that, ROSS has higher test accuracy than the other baseline algorithms across all settings. For example, when $N=10$, ROSS reaches a test accuracy of about $0.95$ across all settings, which is $1.05-2\times$ higher than the baseline algorithms. Moreover, it is observed that, the baseline algorithms, i.e., DMSGD, CGA, NET-FLEET, MEDIAN, TRIM-MEAN, LEARN and BALANCE, have their test accuracies decreased significantly when the number of agents increases. Especially when $N$ is increased to $30$, the test accuracy of ROSS algorithm is $1.06-2.53\times$ higher than those of the baseline algorithms.

  \subsection{Experiment Results on CIFAR-10 Dataset} \label{ssec:res-cifar10-ring}
  \begin{figure*}[htb!]
  \begin{center}
    \parbox{.24\textwidth}{\center\includegraphics[width=.24\textwidth]{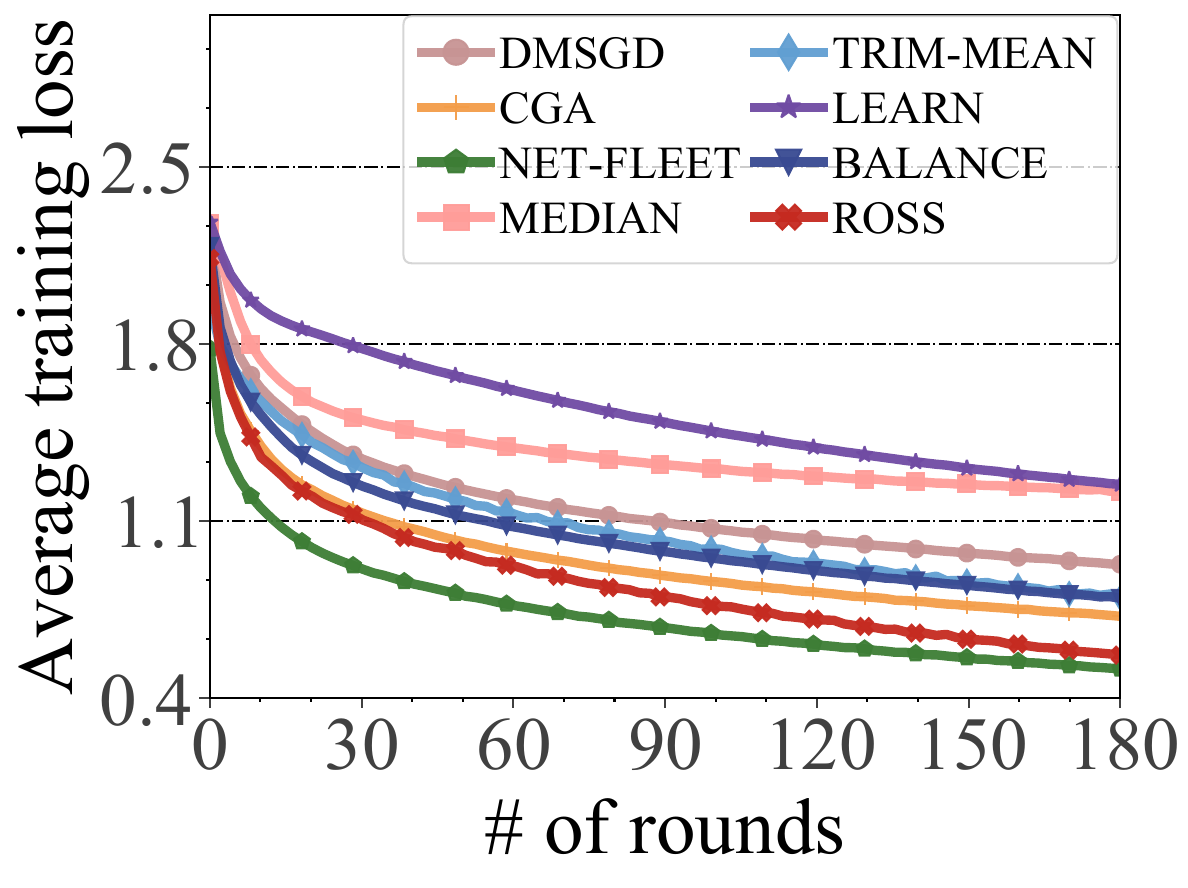}}
    \parbox{.24\textwidth}{\center\includegraphics[width=.24\textwidth]{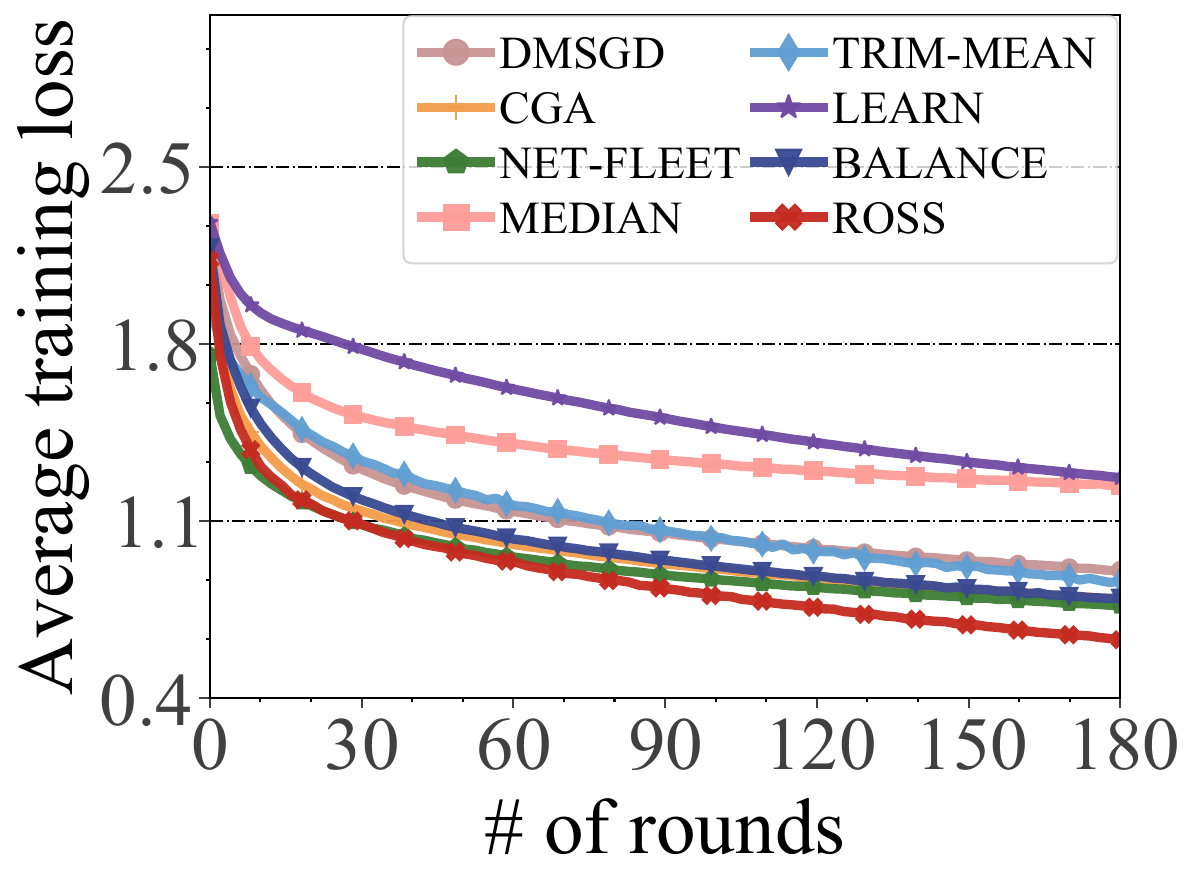}}
    \parbox{.24\textwidth}{\center\includegraphics[width=.24\textwidth]{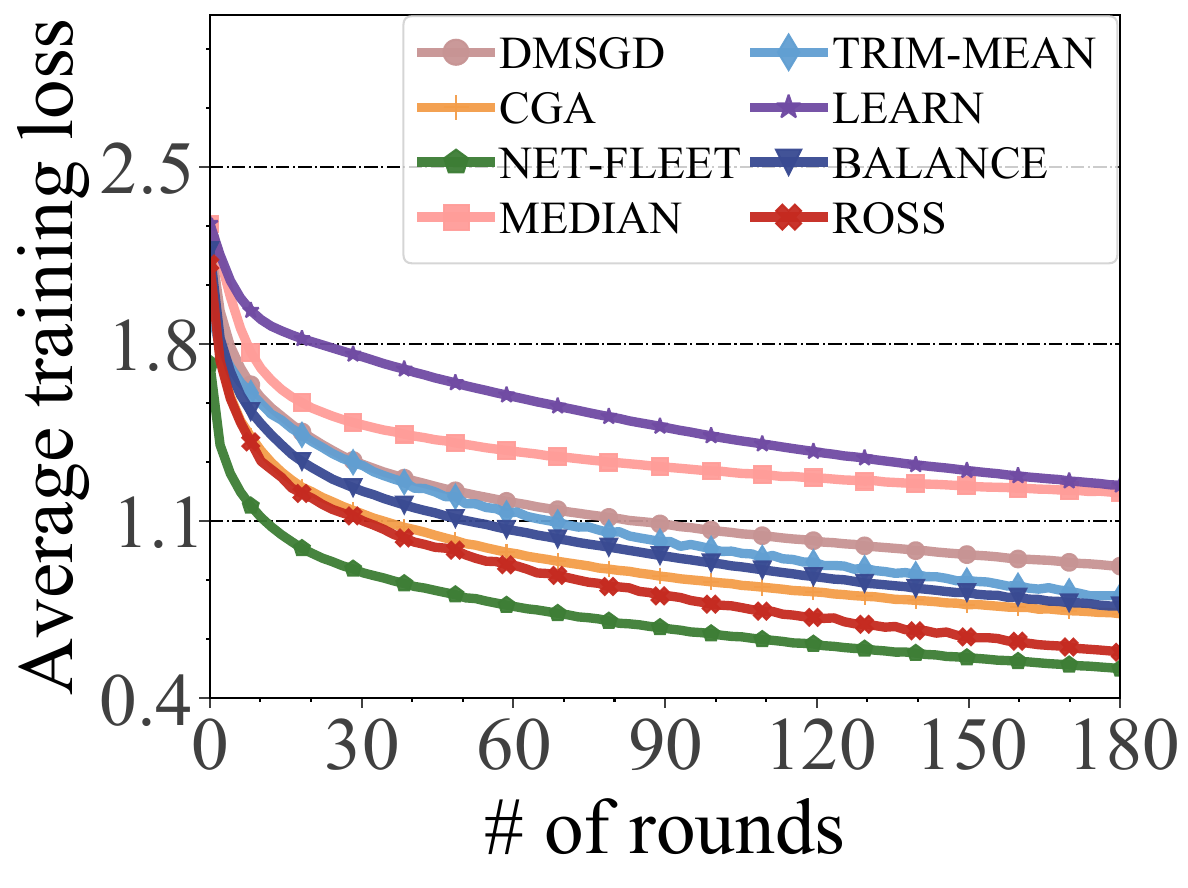}}
    \parbox{.24\textwidth}{\center\includegraphics[width=.24\textwidth]{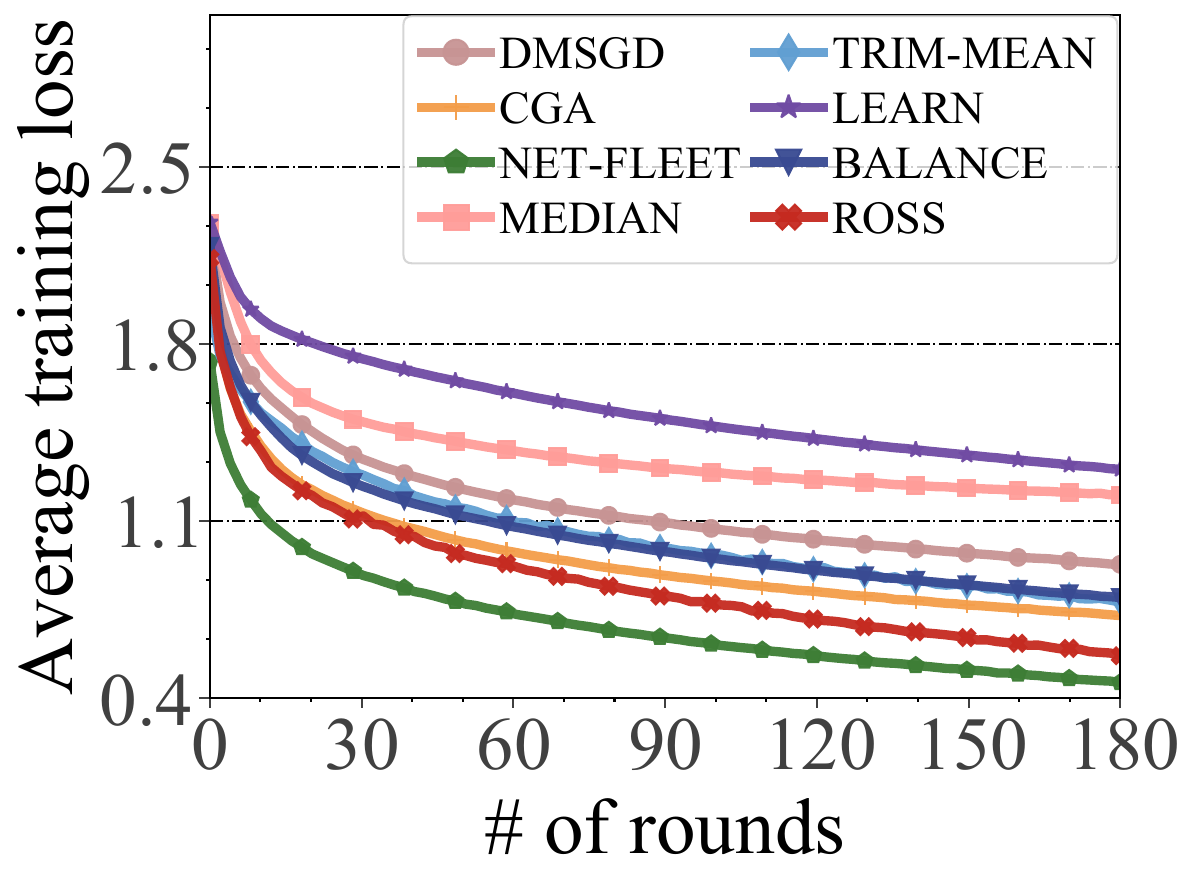}}
    \parbox{.25\textwidth}{\center\scriptsize(a1) Long-tailed ($N=10$)}
    \parbox{.23\textwidth}{\center\scriptsize(a2) Data noise ($N=10$)}
    \parbox{.23\textwidth}{\center\scriptsize(a3) Label noise ($N=10$)}
    \parbox{.25\textwidth}{\center\scriptsize(a4) Gradient poisoning ($N=10$)}
    \parbox{.24\textwidth}{\center\includegraphics[width=.24\textwidth]{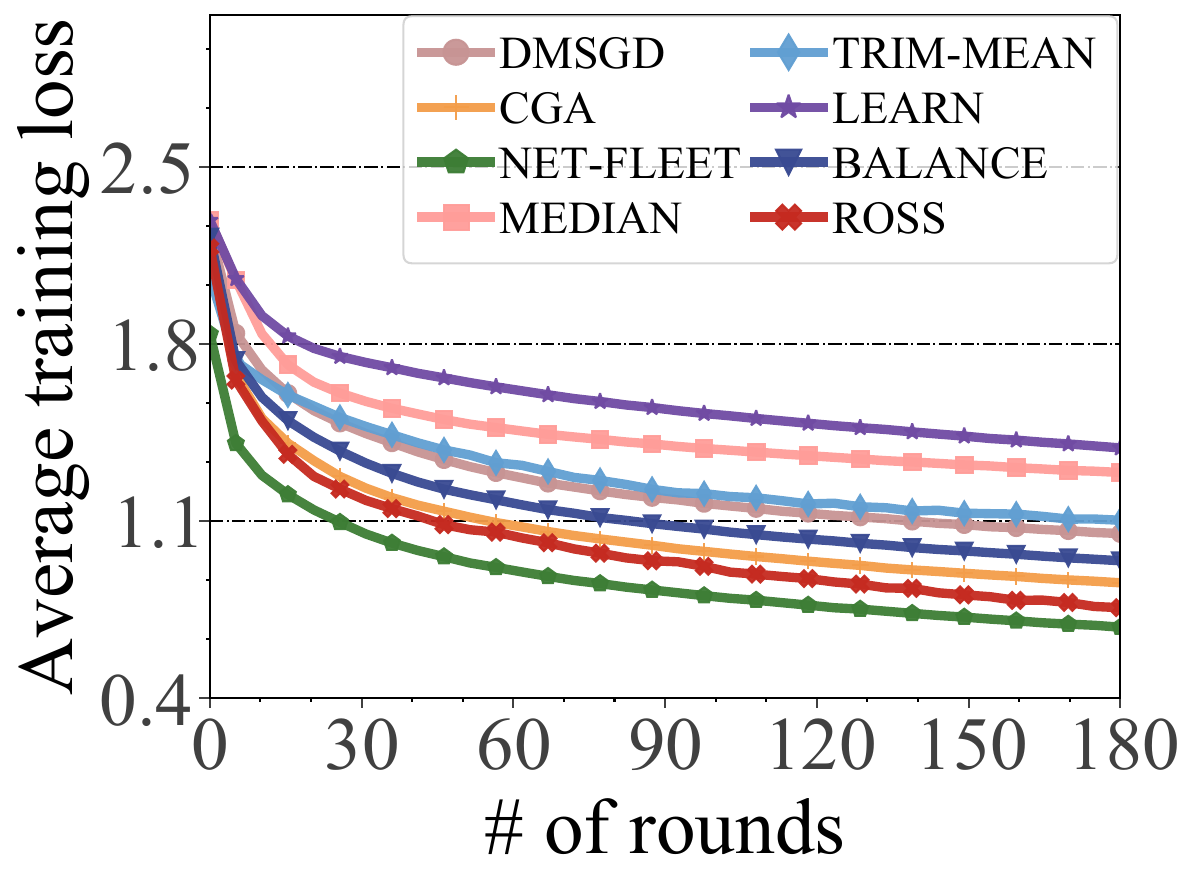}}
    \parbox{.24\textwidth}{\center\includegraphics[width=.24\textwidth]{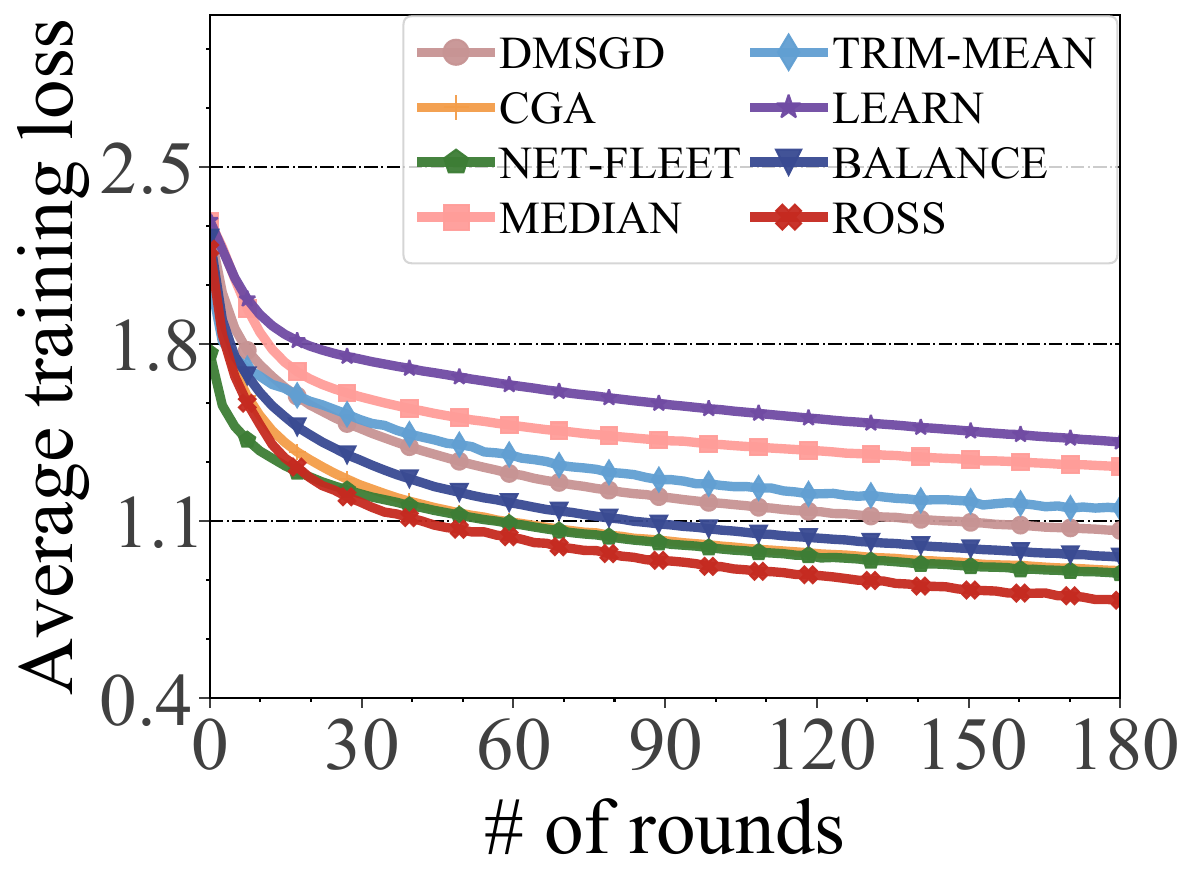}}
    \parbox{.24\textwidth}{\center\includegraphics[width=.24\textwidth]{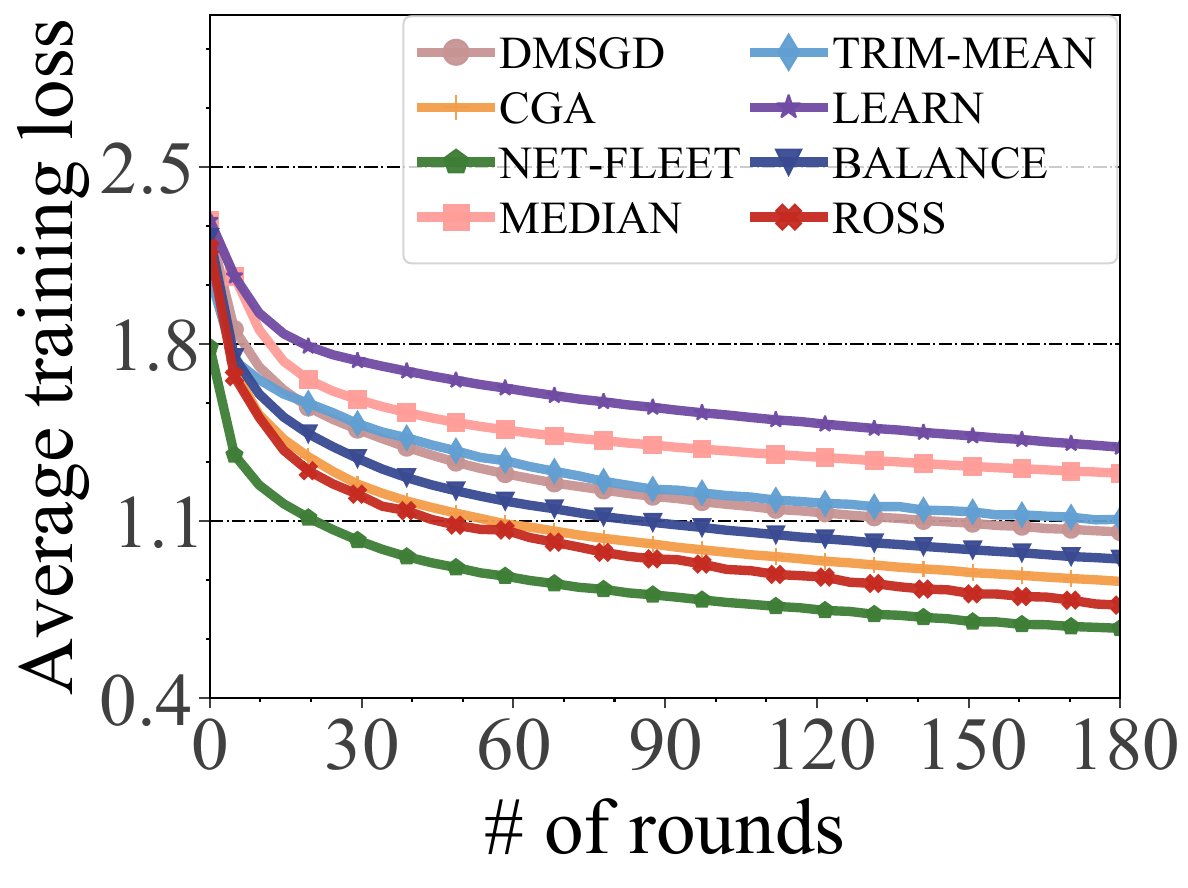}}
    \parbox{.24\textwidth}{\center\includegraphics[width=.24\textwidth]{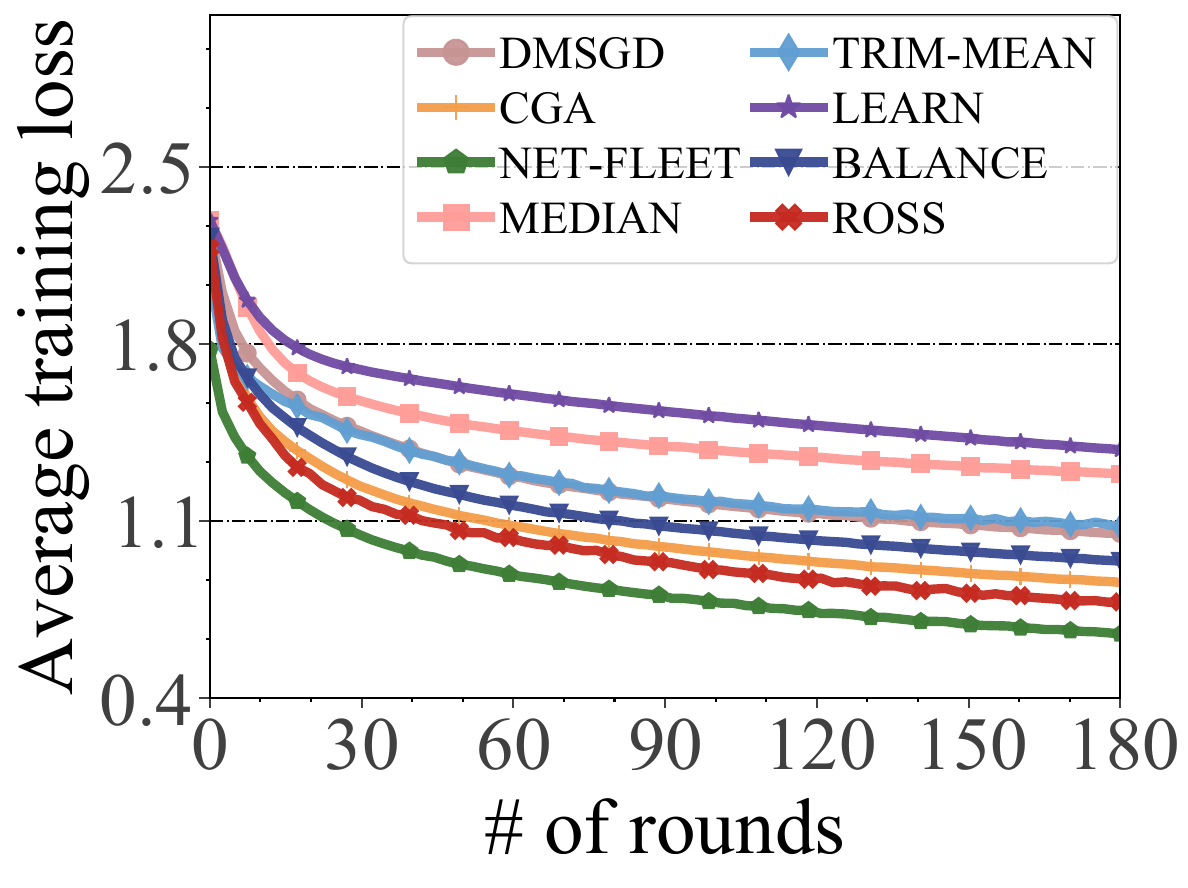}}
    \parbox{.25\textwidth}{\center\scriptsize(b1) Long-tailed ($N=20$)}
    \parbox{.23\textwidth}{\center\scriptsize(b2) Data noise ($N=20$)}
    \parbox{.23\textwidth}{\center\scriptsize(b3) Label noise ($N=20$)}
    \parbox{.25\textwidth}{\center\scriptsize(b4) Gradient poisoning ($N=20$)}
    \parbox{.24\textwidth}{\center\includegraphics[width=.24\textwidth]{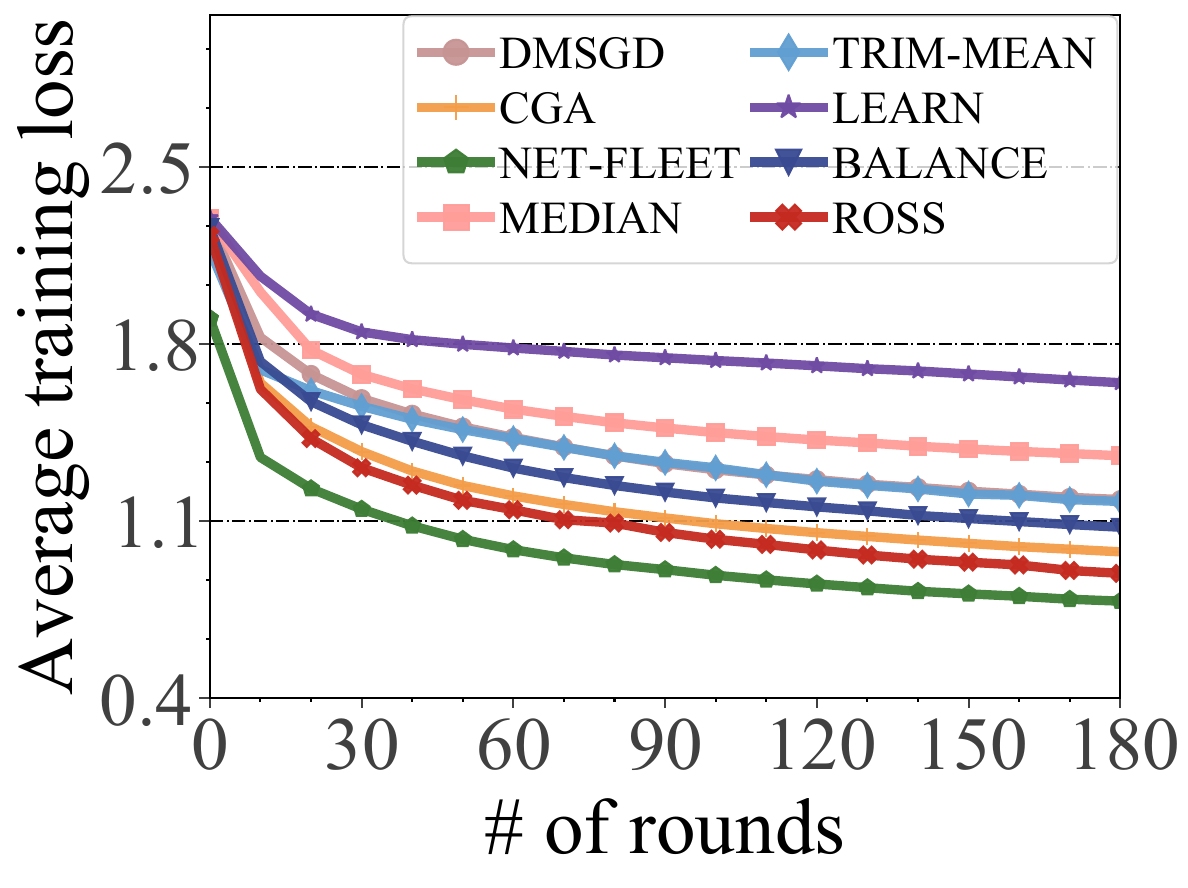}}
    \parbox{.24\textwidth}{\center\includegraphics[width=.24\textwidth]{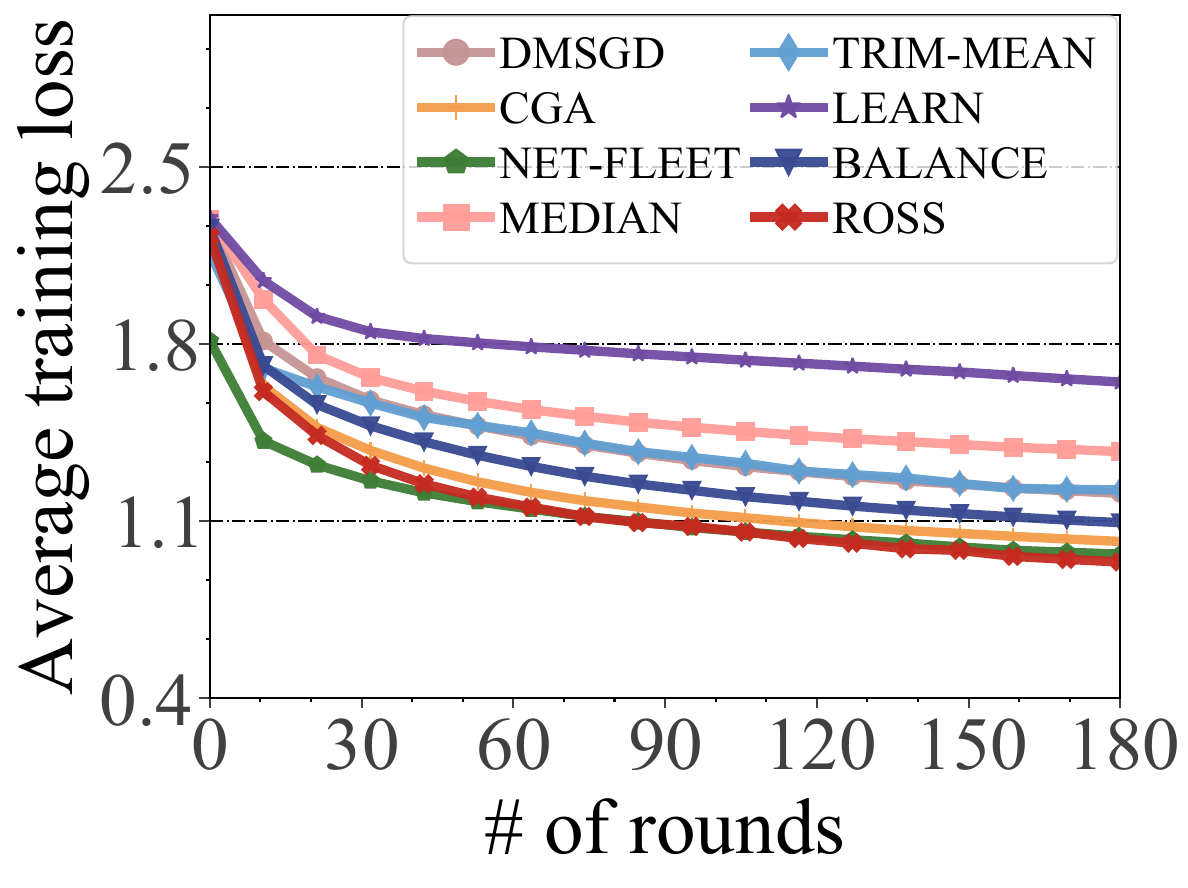}}
    \parbox{.24\textwidth}{\center\includegraphics[width=.24\textwidth]{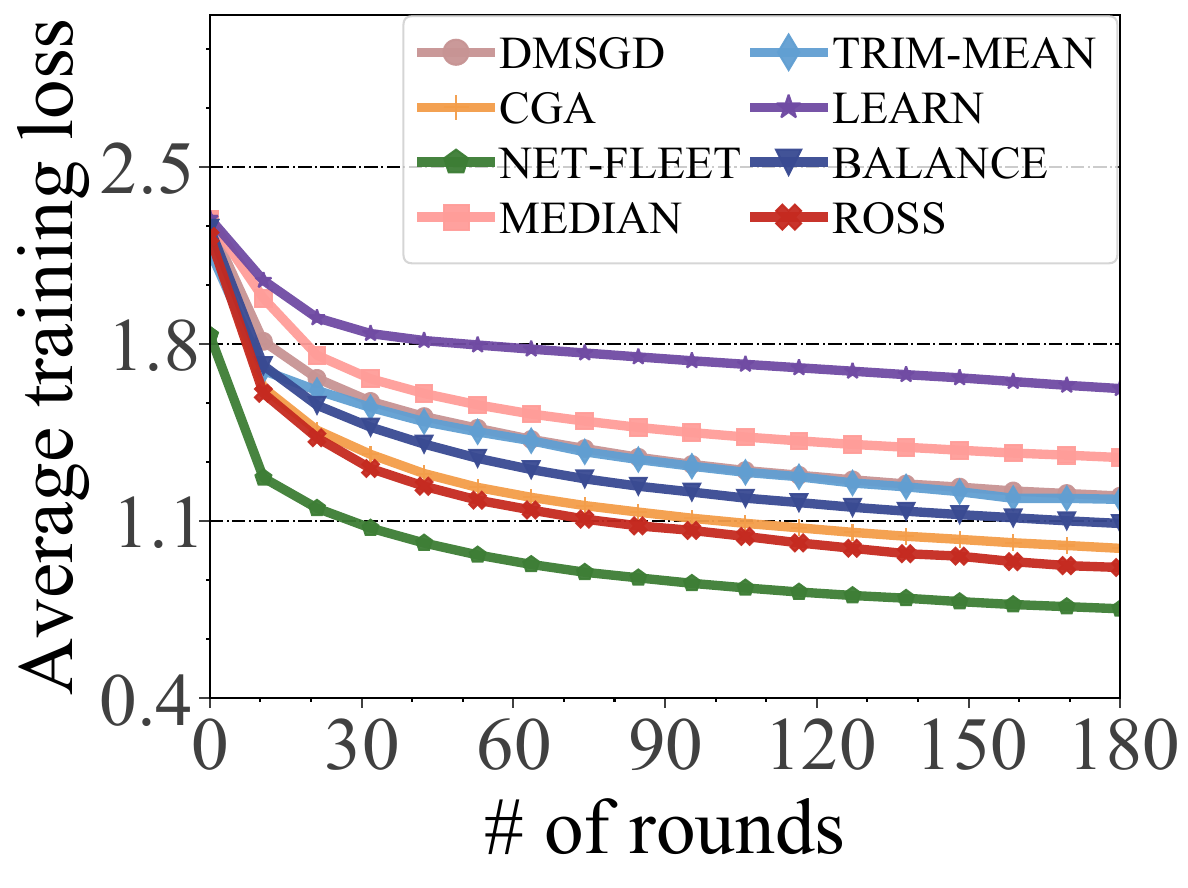}}
    \parbox{.24\textwidth}{\center\includegraphics[width=.24\textwidth]{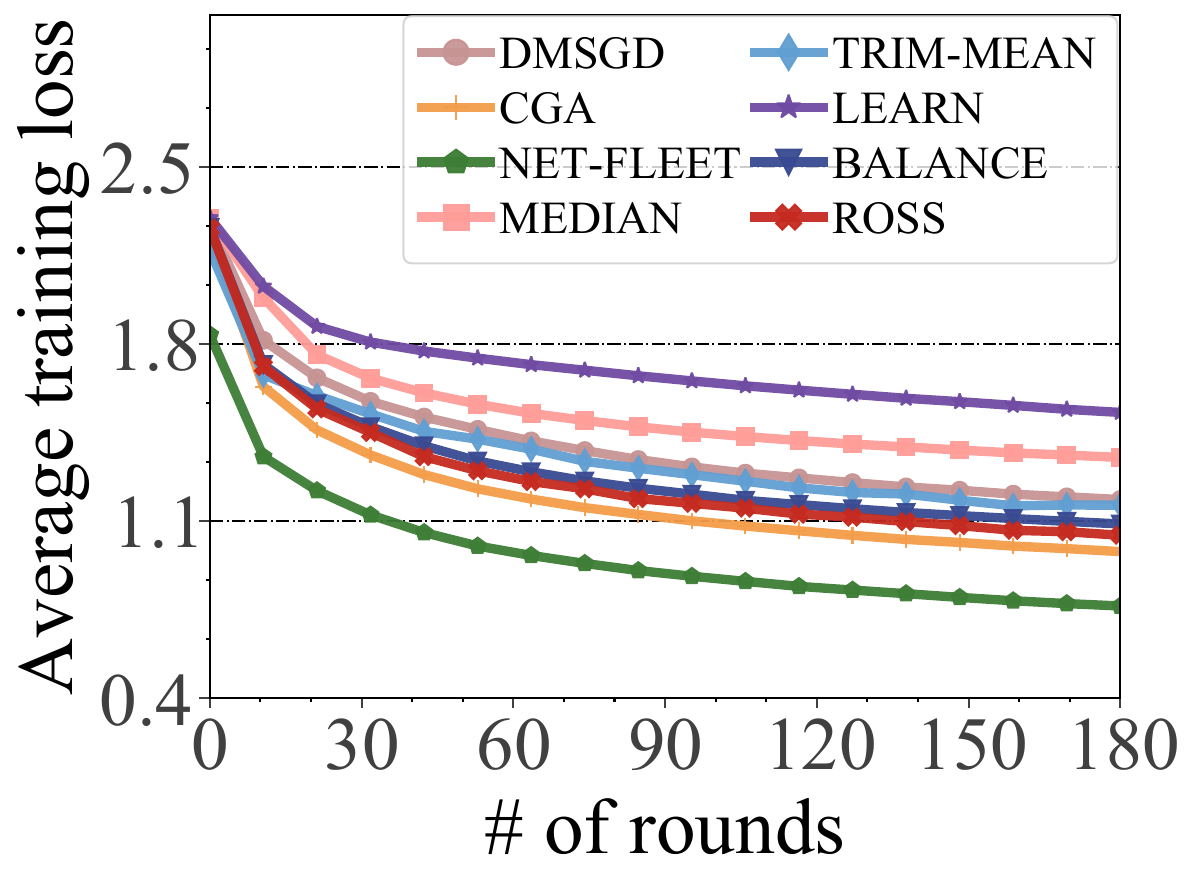}}
    \parbox{.25\textwidth}{\center\scriptsize(c1) Long-tailed ($N=30$)}
    \parbox{.23\textwidth}{\center\scriptsize(c2) Data noise ($N=30$)}
    \parbox{.23\textwidth}{\center\scriptsize(c3) Label noise ($N=30$)}
    \parbox{.25\textwidth}{\center\scriptsize(c4) Gradient poisoning ($N=30$)}
  \caption{Comparison results on CIFAR-10 dataset over ring graphs.}
  \label{fig:cifar-loss-ring}
  \end{center}
  \end{figure*}
  \begin{figure*}[htb!]
  \begin{center}
    \parbox{.32\textwidth}{\center\includegraphics[width=.32\textwidth]{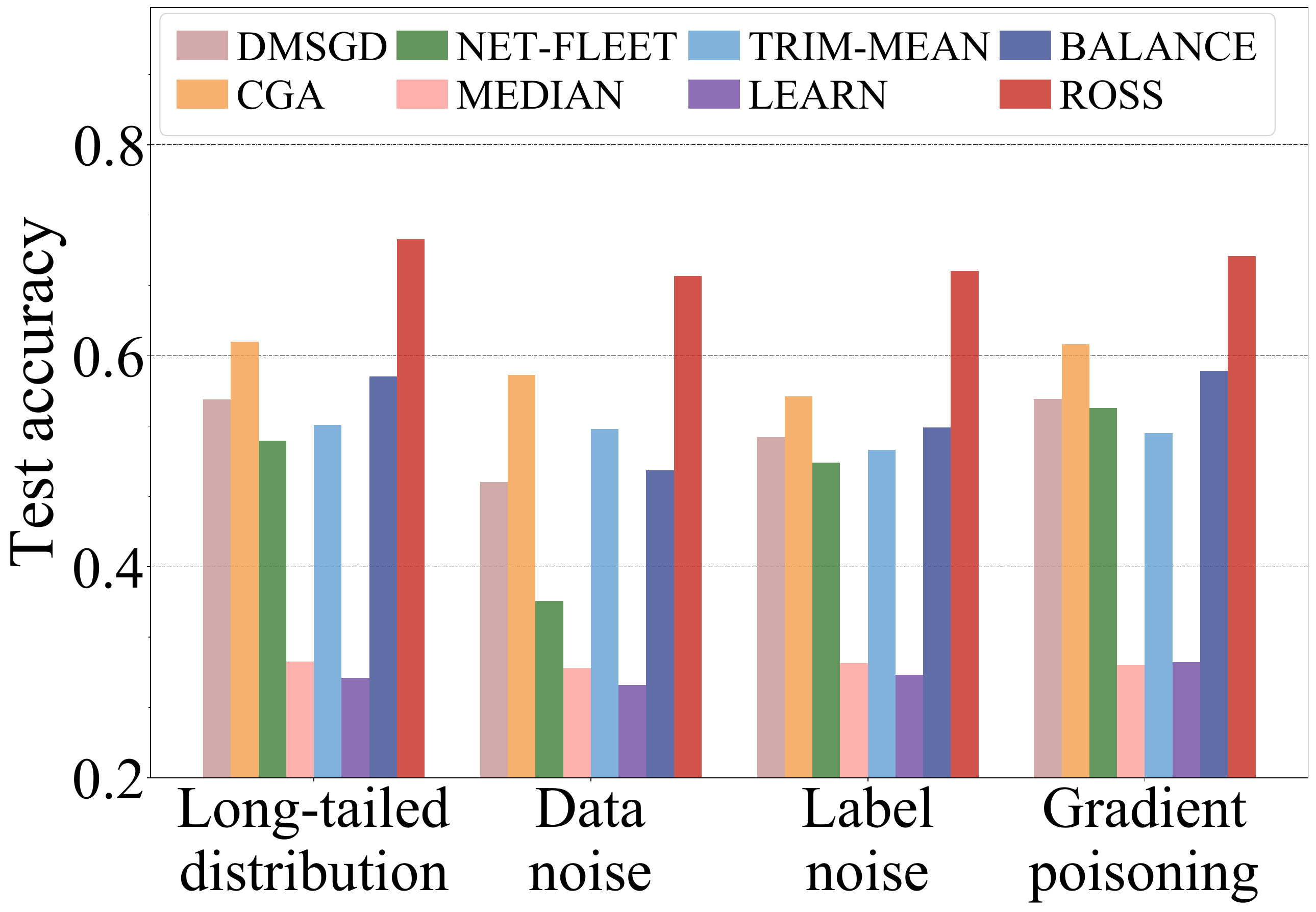}}
    \parbox{.32\textwidth}{\center\includegraphics[width=.32\textwidth]{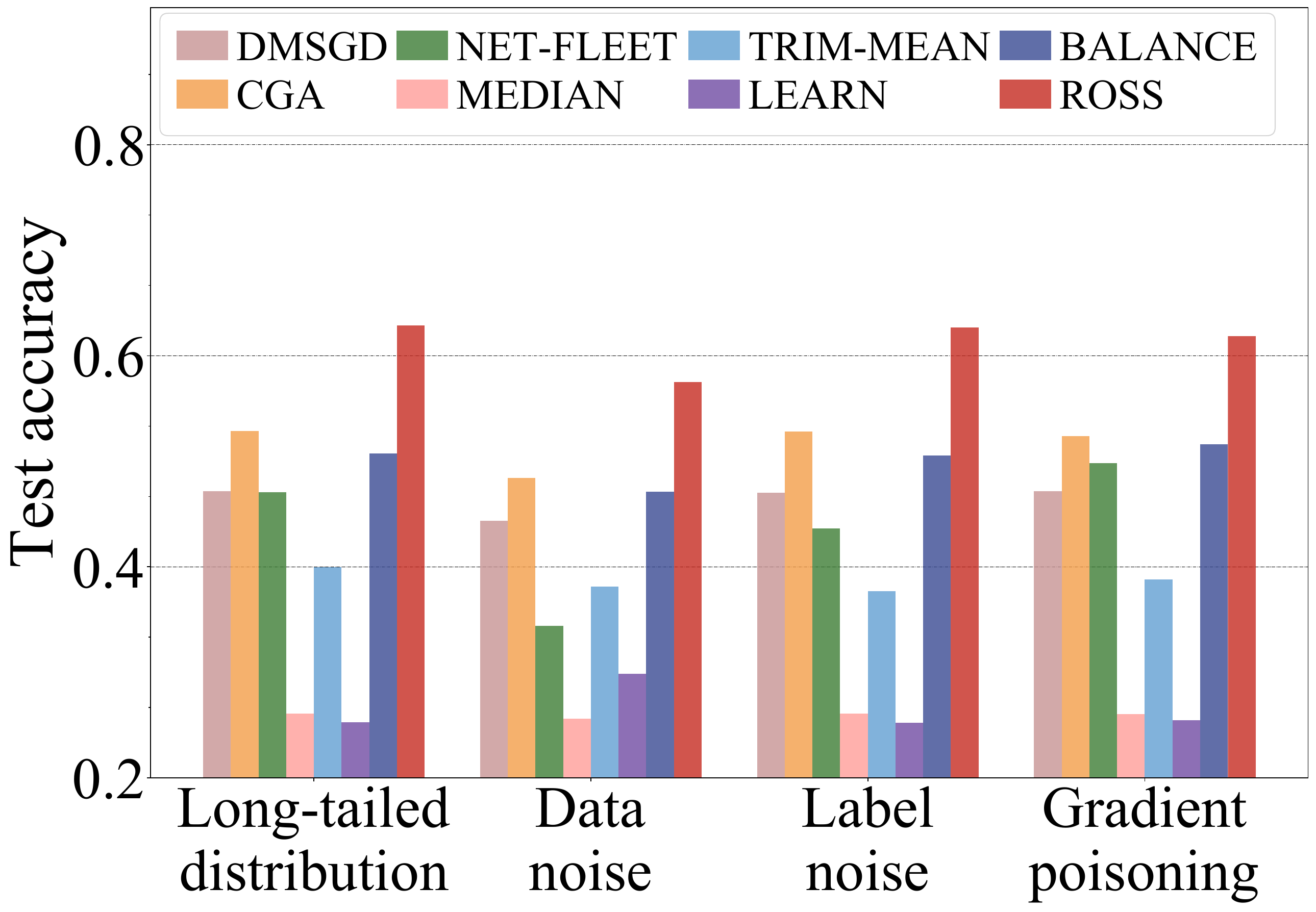}}
    \parbox{.32\textwidth}{\center\includegraphics[width=.32\textwidth]{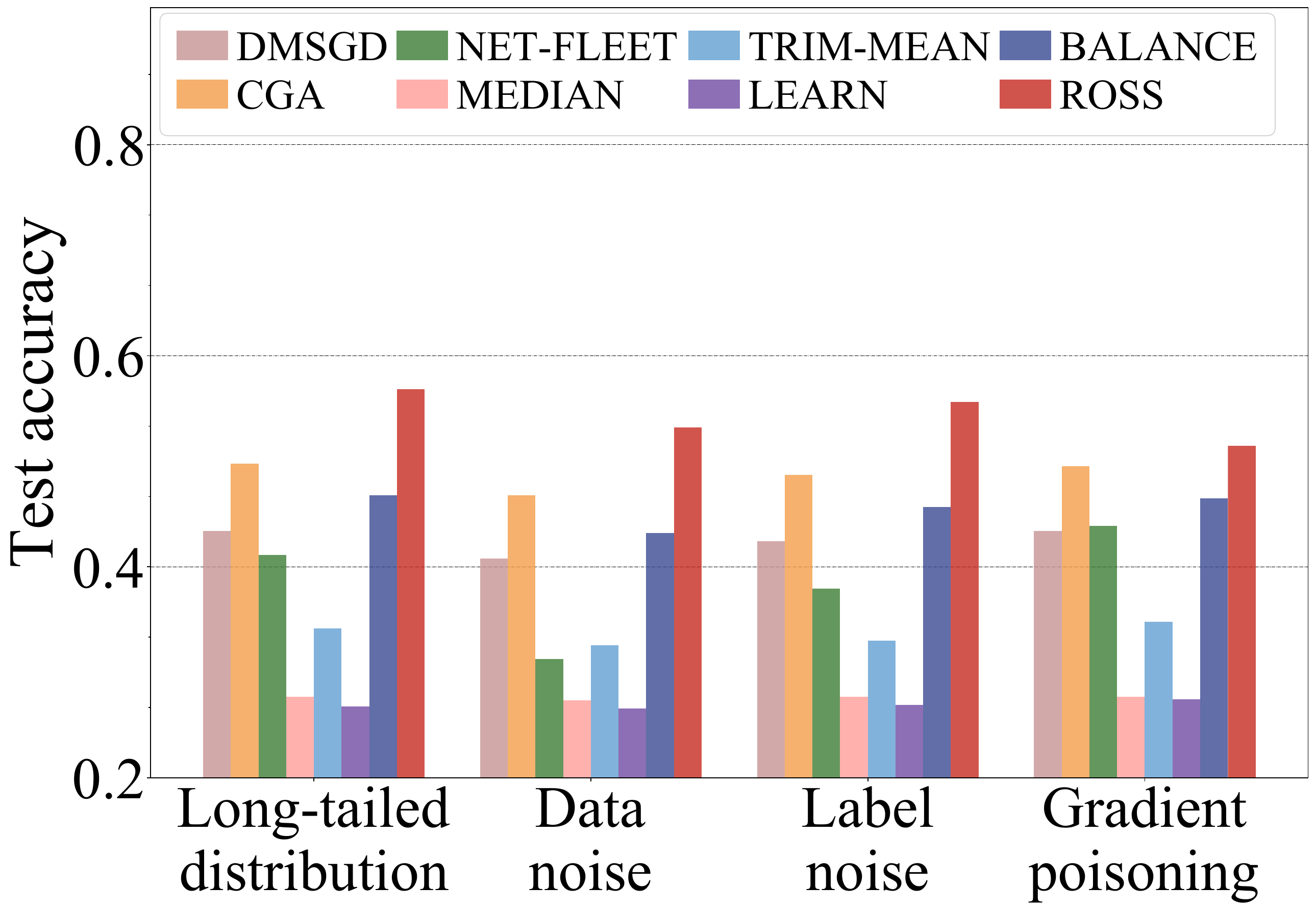}}
    \parbox{.32\textwidth}{\center\scriptsize(a) $N=10$}
    \parbox{.32\textwidth}{\center\scriptsize(b) $N=20$}
    \parbox{.32\textwidth}{\center\scriptsize(c) $N=30$}
  \caption{Comparison results in terms of test accuracy on CIFAR-10 dataset over ring graphs.}
  \label{fig:cifar-acc-ring}
  \end{center}
  \end{figure*}

  We present the convergence results on the CIFAR-10 dataset in Fig.\ref{fig:cifar-loss-ring}. The results show that ROSS consistently converges faster and achieves a lower average loss than DMSGD, CGA, MEDIAN, TRIM-MEAN, LEARN, and BALANCE. For instance, with $N=10$ under the long-tailed distribution, ROSS attains an average loss that is $1.3-2.2\times$ smaller than the baselines. Similar to the observations in Sec.\ref{ssec:res-mnist-ring}, NET-FLEET exhibits convergence performance comparable to (and in some cases better than) ROSS; however, as will be demonstrated later, its test accuracy is significantly lower than that of ROSS.

  The prediction accuracies of different algorithms obtained on ring graphs are shown in Fig.~\ref{fig:cifar-acc-ring}. The results show that our ROSS algorithm achieves higher test accuracy than the other baseline algorithms across all settings. For example, under the data noise with $N=10$, the test accuracy of our ROSS algorithm is around $0.68$, which is $1.2-2.3\times$ higher than the baseline algorithms. Moreover, as the number of agents increases, our ROSS algorithm still have apparent advantage over the baselines. In particularly, when $N=30$, the test accuracy of ROSS remains $1.3-2\times$ higher than those of baselines.

\end{document}